\documentclass[runningheads]{llncs}

\usepackage[final,year=2026]{xxxx}

\usepackage{xxxxabbrv}
\usepackage{graphicx}
\usepackage{booktabs}
\usepackage{multirow}
\usepackage{array}
\usepackage{amsmath}
\usepackage{amssymb}
\usepackage{capt-of}
\usepackage[accsupp]{axessibility}
\usepackage{xcolor}
\usepackage{float}
\usepackage{placeins}
\usepackage{colortbl}
\usepackage{enumitem}
\usepackage{algorithm} %
\usepackage[noend]{algpseudocode}
\usepackage[T1]{fontenc}

\usepackage{listings}
\usepackage{upquote}
\usepackage[most]{tcolorbox}

\lstdefinestyle{promptstyle}{
  basicstyle=\ttfamily\small,
  breaklines=true,
  breakatwhitespace=false,
  columns=fullflexible,
  keepspaces=true,
  showstringspaces=false,
  upquote=true,
  aboveskip=0pt,
  belowskip=0pt,
}

\newtcblisting{promptlisting}[1]{
  enhanced,
  breakable,
  sharp corners,
  colback=black!1,
  colframe=xxxxblue,
  colbacktitle=xxxxblue!10,
  coltitle=black,
  fonttitle=\bfseries,
  title={#1},
  boxrule=0.7pt,
  left=1mm,
  right=1mm,
  top=1mm,
  bottom=1mm,
  listing only,
  listing options={style=promptstyle},
}

\newcommand{\promptmeta}[2]{\noindent\textbf{#1} #2\par}
\newcommand{\placeholder}[1]{\texttt{\textless#1\textgreater}}
\newcommand{\maketitlesupplementary}{%
    \clearpage
    \appendix
    \renewcommand{\thesection}{\Alph{section}}
    \renewcommand{\thesubsection}{\Alph{section}.\arabic{subsection}}
    \setcounter{section}{0}
    \setcounter{figure}{0}
    \setcounter{table}{0}
    \renewcommand{\thefigure}{A\arabic{figure}}
    \renewcommand{\thetable}{A\arabic{table}}
    \begin{center}
        {\Large\bfseries Appendix}
    \end{center}
    \vspace{1em}
}

\usepackage{pgfplots}
\pgfplotsset{compat=1.18}

\usepackage[breaklinks,colorlinks,citecolor=xxxxblue]{hyperref}

\begin{document}

\title{Teaching an Agent to Sketch One Part at a Time}

\author{
Xiaodan Du$^{*}$\inst{1}\and
Ruize Xu$^{*}$\inst{2}\and
David Yunis$^{*}$\inst{1}\and \\
Yael Vinker\inst{3}\and
Greg Shakhnarovich\inst{1}
}

\authorrunning{X. Du et al.}

\institute{
$^1$TTI-Chicago \quad $^2$University of Chicago \quad $^3$MIT CSAIL \\
\email{\{xdu,dyunis,greg\}@ttic.edu \quad richard1xur@uchicago.edu \quad yaelvink@mit.edu}
\\
\small{\url{https://xiaodan.io/teaching-an-agent-to-sketch}}
}

\maketitle

\begin{figure}[H]
\centering
\vspace{-20pt}
\includegraphics[width=1\linewidth]{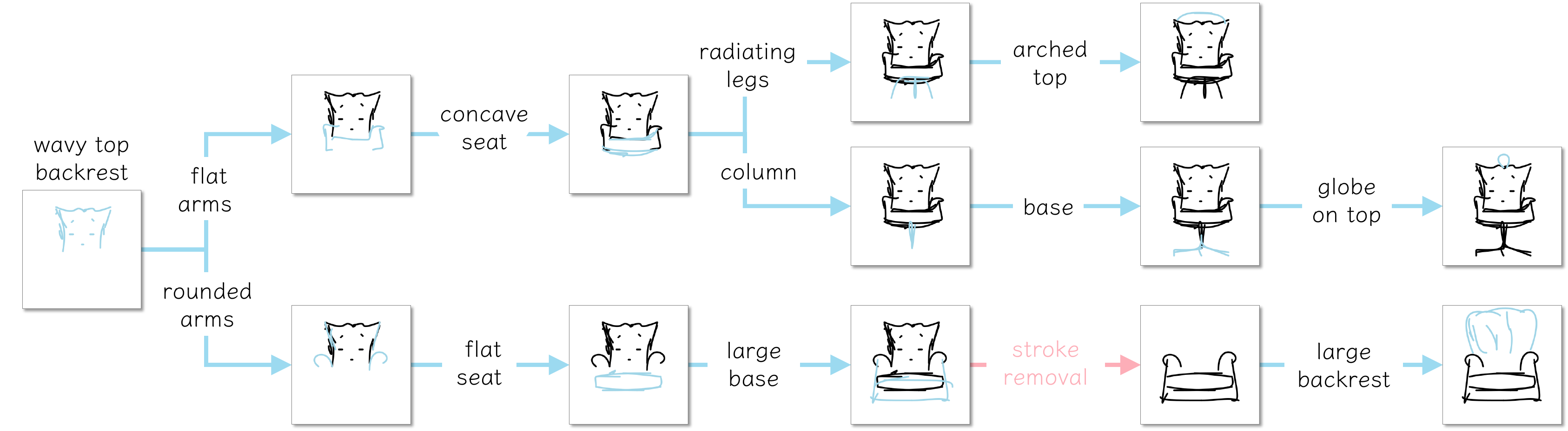}
\caption{
\textbf{Progressive vector sketch generation using our VLM agent.} Trained on our new dataset via SFT + RL training, our agent generates sketches part-by-part, conditioned on text instructions and the evolving canvas. It produces diverse, structurally plausible sketches and supports localized editing via arbitrary stroke removal and replacement.
}
\label{fig:teaser}
\end{figure}

\begin{abstract}

We develop a method for producing vector sketches one part at a time. To do this, we train a multi-modal language model-based agent using a novel multi-turn process-reward reinforcement learning following supervised fine-tuning.  Our approach is enabled by a new dataset we call ControlSketch-Part, containing rich part-level annotations for sketches, obtained using a novel, generic automatic annotation pipeline that segments vector sketches into semantic parts and assigns paths to parts with a structured multi-stage labeling process.  Our results indicate that incorporating structured part-level data and providing agent with the visual feedback through the process enables interpretable, controllable, and locally editable text-to-vector sketch generation.
\end{abstract}
\let\thefootnote\relax\footnote{* Equal contribution.}

\section{Introduction}
\label{sec:intro}

Sketching provides a structured abstraction of visual content and enables rapid ideation and concept exploration in domains from industrial design to digital art. Sketches represented as vector graphics offer many advantages over rasterized canvasses, making them useful in creative workflows: infinite scalability, structured visual elements, support for precise and localized modifications, and more. 
Automatic text-to-vector sketch generation has been widely explored \cite{sketchrnn,beziersketch,sketchode,chirodiff,zhou2025strokefusion,clipdraw,diffsketcher,autosketch,sketchdreamer,sketchagent}. However, the majority of existing works generate the full sketch at once, overlooking the progressive, step-by-step nature of the sketching process.

 Having the strokes in a sketch grouped into meaningful parts makes the sketch more easily \textbf{editable}: parts could be removed, replaced or modified in isolation from the rest of the sketch more efficiently than by modifying individual strokes. 
 It also makes the sketch more \textbf{interpretable} to a human user. Finally, one-shot generation from a long, compositional prompts may lead to failures that are localized but difficult to mitigate. In contrast,  
 incorporating the notion of parts into the generation pipeline provides the designer fine-grained \textbf{control}: if a generated sketch of a part is not right, it can be replaced, and multiple choices can be explored at any intermediate stage before proceeding to other parts.

Most prior work on text-to-vector sketch generation does not allow for part-by-part generation. The one exception is SketchAgent~\cite{sketchagent} which relies on close-sourced vision-language model (VLM) as backend (and thus is not easily adaptable to desired domain or style) and produces simplistic, icon-style outputs.

This leaves a gap: existing methods are unable to achieve free-text guided part-by-part generation and highly detailed vector sketch with a unified model, struggling to support human-friendly workflows which include branching possibilities and creative exploration. In contrast, our method makes these possible, as illustrated  in~\cref{fig:teaser}.

We believe that a necessary element for closing this gap is the right training data.
Work like SketchAgent has demonstrated the potential of VLMs in iteratively generating sketches conditioned on text. However, large language models (LLMs) are known to be extremely data-hungry\cite{villalobos2024position,hoffmann2022chinchilla} and collecting a large amount of high-quality part-annotated datasets for vector sketch created by professionals can be costly and difficult to scale. 
To overcome this obstacle, we propose a scalable pipeline for annotating parts in vector sketches.
Our pipeline relies on a multi-stage labeling process for part decomposition and path assignment. It includes proposal, critique, and revision stages. This pipeline is generic and can be applied to any vector sketch data. 

We apply the aforementioned data collection pipeline on the ControlSketch dataset~\cite{swiftsketch}. We call the resulting part-annotated dataset (which we will release as one of our contributions) \textit{ControlSketch-Part}, and use it to train a VLM on the text-guided part-by-part generation task.
The training uses a two-stage supervised fine-tuning (SFT) and reinforcement learning (RL) framework, where the SFT stage teaches format and initializes the sketching policy for a single turn, and an innovative \textbf{multi-turn process-reward GRPO} RL training stage aligns multi-turn rollouts using intermediate-state rewards \cite{deepseekmath,ppo}. 

Our automatic data pipeline and the proposed training strategy enable free-text guided multi-turn interactive sketch generation. We show, quantitatively using automated metric and user studies, and qualitatively in the paper and the supplementary, that our results significantly improve on prior work. 

\noindent In summary, our contributions are:

\begin{itemize}[topsep=0pt]
    \item A generic scalable pipeline for automated VLM-based part annotation of vector sketches, yielding a short overall caption, a set of semantic part descriptions, and a complete path-to-part assignment for any vector sketch.
        \item A high-quality part-annotated sketch dataset, ControlSketch-Part, and an associated new benchmark for multi-turn text-to-vector sketch generation.
        \item A novel multi-turn process-reward GRPO algorithm for training, enabling us to train a sketching agent with novel capabilities: multi-turn vector sketch generation and progressive editing of sketches with text guidance.
\end{itemize}
The qualitative and quantitative experiment results show the potential of our data pipeline combined with VLM in the field of text-to-vector sketch synthesis.

\section{Related Works}

\subsection{Text-to-Vector Sketch Synthesis}

Previous works on text-to-vector sketch generation fall into two main categories: learning-based approaches and test-time optimization-based approaches.

\noindent\textbf{Learning-based approaches} Sketch-RNN~\cite{sketchrnn} is one of the first works to explore this task by learning to generate polylines autoregressively. BézierSketch~\cite{beziersketch} improves upon Sketch-RNN by replacing polylines with Bézier curves for better smoothness. SketchODE~\cite{sketchode} further extends autoregressive stroke generation by modeling sketches as continuous-time functions via Neural ODEs. More recently, inspired by the success of score-based generative modeling~\cite{ho2020denoising,song2020score}, methods including ChiroDiff~\cite{chirodiff} and StrokeFusion~\cite{zhou2025strokefusion} apply diffusion models to sketch synthesis and denoise all strokes simultaneously, offering no progressive, part-level control over the generation process. These methods are conditioned on pre-defined discrete class/attribute labels rather than free-form natural language, greatly limiting their real-world applicability. They also fall short of producing complex, high-fidelity sketches.

\noindent\textbf{Test-time optimization-based approaches} These methods take a longer time to produce an output but offer greater flexibility and higher visual quality. CLIPDraw~\cite{clipdraw} pioneers text-guided vector sketch synthesis by optimizing SVG paths with a CLIP-based~\cite{clip} objective. Later works utilize CLIP-based optimization for versatile image-to-sketch generation \cite{vinker2022clipasso, Vinker_2023_ICCV}.
DiffSketcher~\cite{diffsketcher}, AutoSketch~\cite{autosketch}, and SketchDreamer~\cite{sketchdreamer} leverage a wider range of supervision, such as LPIPS~\cite{lpips} loss and score distillation sampling (SDS)~\cite{poole2022dreamfusion} loss, to achieve higher visual quality. However, these methods optimize all strokes jointly for a single text input, producing sketches without meaningful stroke ordering or semantic part structure.

The most directly relevant work to part-aware, text-guided sketch synthesis is SketchAgent~\cite{sketchagent}, which uses a closed-source Claude Sonnet model in a zero-shot prompting framework to perform text-guided sequential sketching. SketchAgent's zero-shot nature constrains it to doodle-style outputs that cannot be adapted to higher visual fidelity or specific domains. It also exhibits low spatial grounding accuracy. 

\subsection{Reinforcement Learning for Large Language Models}
Reinforcement learning (RL) has long provided a principled framework for optimizing sequential decision-makers in Markov decision processes (MDPs).  
This MDP perspective is increasingly relevant for modern LLMs, since autoregressive token generation itself can be interpreted as a sequential decision process and, more broadly, many agentic applications expose the model to multi-step environments where errors can accumulate over time. Recently, DeepSeekMath introduced Group Relative Policy Optimization (GRPO), which has been efficient and successful in various reasoning tasks~\cite{deepseekmath}.

\noindent\textbf{Multimodal RL and dense credit assignment}  Extensions of GRPO training go beyond text-only reasoning. In the domain of vector graphics, for example, \emph{Reason-SVG}~\cite{xing2025reasonsvghybridrewardrl} proposes a two-stage scheme (SFT followed by GRPO) to generate SVG sketches via a hybrid reward combining programmatic correctness and visual similarity. A complementary line, \emph{Rendering-Aware Reinforcement Learning}~\cite{rodriguez2025renderingawarereinforcementlearningvector}, uses rendering feedback to compute a visual reward: the similarity between a rendered SVG output and a target image guides policy improvement through GRPO. However, these methods do not use intermediate states of the generation process, whereas we leverage intermediate state representations (i.e., partial sketch) to provide dense credit assignment.

\section{Automated Part Annotation}
\label{sec:dataset}

\begin{figure}[t]
  \centering
  \includegraphics[width=1\linewidth]{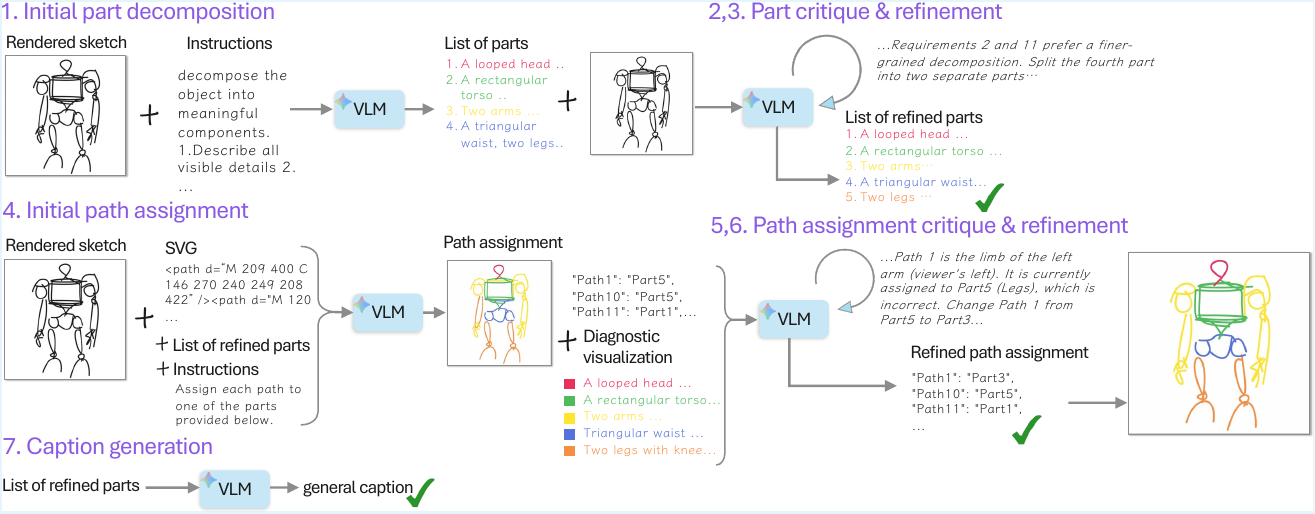}
  \caption{An illustration of our automated part annotation pipeline. The same VLM is used to produce part designations and assignments in some stages and to critique these assignments and suggest improvements in other stages. Green check marks indicate outputs retained in the final dataset.}
  \label{fig:data_pipeline}
\end{figure}

We start with an existing dataset of sketches. Our goal is to enrich each sketch, given as a vector graphics (e.g., SVG) file, with detailed part information: 
\begin{itemize}[topsep=0pt]
\item A short caption describing the entire sketch;
\item A set of part descriptions, on a semantic level related to the sketch content;
\item A path-to-part assignment for each path.
\end{itemize}

\subsection{Data Collection Pipeline}\label{sec:datapipe}
We design a multi-step automatic data annotation pipeline that progressively derives semantic structure from the raw SVG input. An overview of the pipeline is presented in~\cref{fig:data_pipeline}.

\noindent\textbf{(1) Initial part decomposition} The input sketch is rendered into a raster image. Based on this rendering, a VLM proposes a semantic decomposition as a small set of parts. Each part is written as a concise textual description of a distinct object component. The VLM prompt (see Supplementary for all prompt details) instructs it to output non-overlapping yet collectively exhaustive parts.

\noindent\textbf{(2) Part critique} Like others~\cite{liu2024lost,harada2025curse}, we find that even current state-of-the-art VLMs struggle to follow all rules in a complicated task. Therefore, we run an improvement step:  the VLM (acting now as a critic) audits the current set of parts against all the instructions from Step 1 (and the rendered sketch) and returns a structured list of issues, enforced by a schema~\cite{geng2025jsonschemabench}. Each issue contains ``type of violation'', ``severity'', ``reasoning'' and ``suggested fix''. The critique also contains an overall ``summary'' of the issues and a boolean ``should revise'' flag. 

\noindent\textbf{(3) Part refinement} If the flag is ``should revise'', the VLM is instructed to revise the provided previous part decomposition using the critique from (2) and the sketch rendering. The output format is the same as that in (1). 

\noindent\textbf{(4) Initial path assignment} Based on the refined parts, the sketch's SVG text, and the sketch rendering, we instruct the VLM to assign every path to one part. The output is schema-constrained so that: 
\begin{itemize}[topsep=0pt]
    \item parts are assigned part labels as ``Part1'', ``Part2'',\ldots;
    \item each path index (``Path1'', ``Path2'',\ldots) is assigned to exactly one part;
    \item each part contains at least one path.
\end{itemize}

\noindent\textbf{(5) Path assignment critique with diagnostic visualization.} We critique the path assignment similarly to (2), with the addition of a diagnostic visualization (shown in Fig.~\ref{fig:data_pipeline}), as input to the VLM critic. First, we assign each part label a unique color from a pre-defined color palette, and build two panel. In the left panel of the diagnostic visualization, we render a color marker, the part label, and the part description text for each part, in the corresponding color. Thus each part description has an unambiguous visual identity. In the right panel, we recolor the sketch by rendering each path in the color of its assigned part. The two color-coded panels (descriptions and sketch) are concatenated side-by-side, making it easier for the VLM to capture the correspondence between the part description and the path assignment. The VLM receives the original sketch image, the diagnostic image, the previous path assignment, and the task instructions for (4), and is asked to identify incorrect path assignments and provide concrete correction suggestions. The output schema is exactly the same as that of (2).

\noindent\textbf{(6) Path assignment refinement} During this step, a refinement pass receives the sketch rendering, the sketch paths, the refined parts from (3), and initial part assignments from (4) along with step (4) instruction, and the path assignment critique from (5). It updates the path assignment with necessary edits under the same schema constraints as (4). 

\noindent\textbf{(7) Caption generation} Finally, we use the VLM to generate a short general caption that summarizes the object based solely on the refined parts. This ensures the overall text caption remains consistent with part-level semantics.

\subsection{Our Dataset: ControlSketch-Part}\label{controlsketchpart}

The procedure described above is designed to generalize to \textit{any} sketch dataset with SVG (or vector-convertible) sketches. To reduce the data gap discussed in \cref{sec:intro}, we apply it to a complex, realistic-looking sketch dataset: ControlSketch.

ControlSketch is a professional-quality dataset that consists of 35,000 image-sketch pairs\cite{swiftsketch} generated by SDXL \cite{sdxl} and the SDS~\cite{poole2022dreamfusion} loss-based optimization algorithm. It contains sketches for 15 object categories; we do not use or refer to the category labels in any way in training, and only mention them for reference when organizing examples in this paper.

\begin{figure}[t]
\centering
\includegraphics[width=1\textwidth]{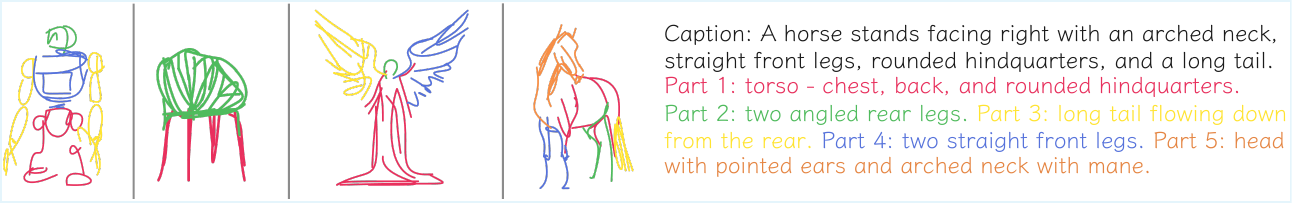}
\vspace{-0.5cm}
\caption{Examples from the ControlSketch-Part dataset. We show part decomposition for 4 sketches with various objects and number of parts. The actual caption and part descriptions are shown for the rightmost sketch. The black text is the overall caption. The color-coded part descriptions and stroke groups demonstrate the part-level semantic annotations. }
\label{fig:data_viz}
\end{figure}

We construct a schema so that the number of parts of a sketch is between 2 and 5, and apply our pipeline using Gemini 3.0 Pro as the VLM. We call the resulting dataset with the newly added captions, part descriptions, and path-to-part assignment \textbf{ControlSketch-Part}. An illustration of examples of ControlSketch-Part data can be found in \cref{fig:data_viz}.

\section{Method}
\label{sec:method_overview}

\begin{figure*}[t]
  \centering
   \includegraphics[width=1\textwidth]{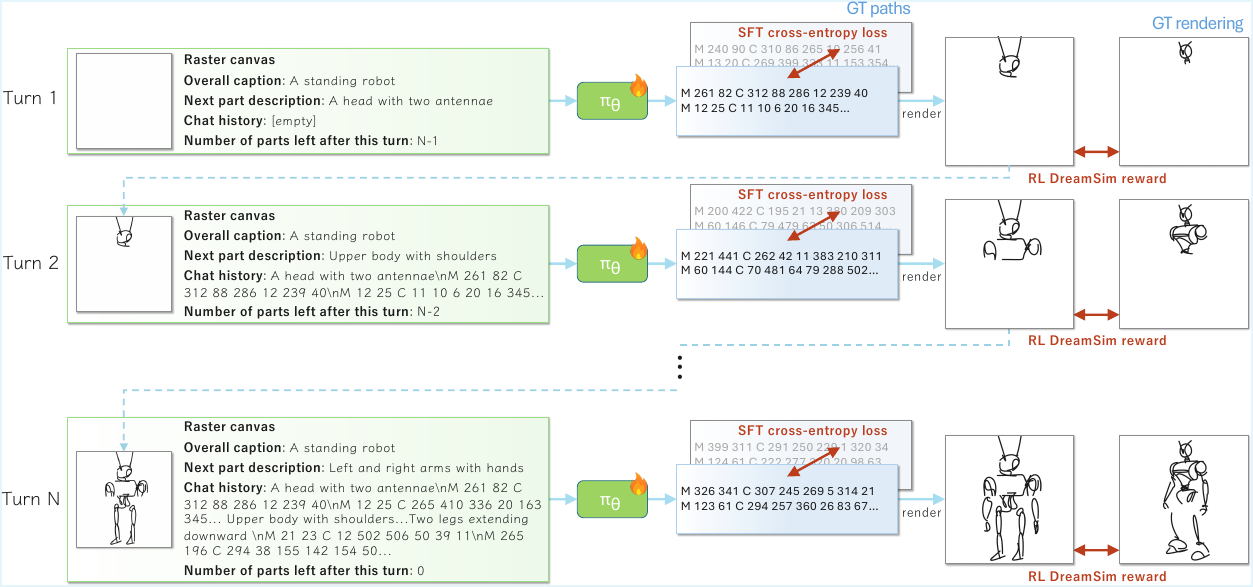}

   \caption{The visualization of the training pipeline. The task of generating vector sketches based on text prompts is split into multiple turns. Blue arrows: sequential computation; red arrows: loss. Cross-entropy loss and DreamSim reward are used at training signal at SFT and RL stages, respectively. $\pi_\theta$ is the policy model, i.e., our VLM.}
   \label{fig:pipeline}
\end{figure*}

We aim to have a VLM agent generate a vector sketch iteratively: draw a part $\rightarrow$ look and reason $\rightarrow$ draw the next part. An overview of our method's pipeline can be found in \cref{fig:pipeline}. At each turn, the VLM receives: (1) the rendering of the current canvas, (2) an overall short caption of the object it is drawing, (3) description of the next part, (4) descriptions of previously drawn parts of the sketch along with their corresponding vector paths, and (5) the number of parts left to sketch after the current turn.  

The output is a sequence of paths (strokes) each coded as a curve. 
Since all strokes share the same set of SVG attributes (width, opacity, etc.) we instruct the model to output only the eight coordinates that define a cubic Bézier curve along with the SVG command letters \texttt{M} and \texttt{C}. The paths are separated by newline \texttt{\textbackslash n}. For example, a sequence of two paths will be presented as
\begin{equation}\vspace{-.5em}
\text{\small
\begin{tt}
  M 212 146 C 6 89 303 88 322 14\textbackslash nM 213 17 C 213 269 18 157 218 32\textbackslash n
\end{tt}
}
\end{equation}

Our method consists of two training stages: (1) supervised fine-tuning stage in which the model learns the correct output format and sketching policy for a single turn, and (2) multi-turn process-reward GRPO training to improve the visual quality of output. 

\subsection{Stage 1: Supervised Fine-Tuning}

We conduct SFT on the VLM agent using the standard cross-entropy loss (next token prediction) on input/output pairs. We augment our dataset by randomly sampling a maximum of 20 part permutations per sketch, yielding for each permutation the corresponding sequence of part descriptions, incomplete sketches (as strokes) and the associated incomplete renderings. For instance, suppose a sketch has parts A,B,C,D and E. A permutation of these might be C,B,D,E,A. The corresponding set of input/output pairs will include empty canvas + description of C (with the output being the ground truth strokes for C); canvas with C rendered + description of B (with the output consisting of the strokes for B); canvas with C+B rendered + description of D (with the output consisting of D), etc. See Fig.~\ref{fig:pipeline} for visualization. All permutation for a given sketch share the same ``global'' caption. This approach provides the agent with example of completing a sketch with arbitrary ordering of parts/turns.

The main purpose of the SFT stage is to train the agent to produce valid paths, and to train the model to generate a single part to extend an existing ground truth partial sketch (which will prepare it to the second stage in which it learns multi-turn generation).

\subsection{Stage 2: RL with Multi-turn Process-Reward GRPO}

After the SFT stage, the agent is capable of progressive generation when applied autoregressively (generate the first part, then generate the second part conditioned on observing the just generated first part, etc.) However, this creates a gap between the SFT training regime, in which the agent has only seen ``oracle'' intermediate states sampled from ground truth during SFT, and the inference time when it is given its own generations from previous steps. Indeed we observe a resulting deterioration in visual quality as the generation progresses.

To bridge this gap, we further train our agent with a reinforcement learning algorithm, Group Relative Policy Optimization (GRPO)~\cite{deepseekmath}.  GRPO computes the mean reward over multiple sampled trajectories (a group) as the baseline, replacing the need for an additional value function approximation model, which is usually of comparable size as the policy model~\cite{deepseekmath}. This makes GRPO more efficient than its predecessors like~\cite{ppo, ouyang2022traininglanguagemodelsfollow}. 

\noindent\textbf{GRPO preliminary}
We call a \emph{trajectory} a sampled sequence of responses $o^1, \ldots, o^T$ for a given input $q\sim P(Q)$. In our case, the trajectory is a sequence of sketch parts each adding to the previously generated parts.
Assuming the group size (number of sampled trajectories for a given problem) is $G$ and the length of steps for trajectory $g$ is $T_g$, then the collection of all the rewards for a group $\left\{\{o_g^t\}_{t=1}^{T_g}\right\}_{g=1}^G$ can be expressed as:
\begin{equation}
    \mathbf{R}=\left\{\, \{r_g^{t}\}_{t=1}^{T_g} \,\right\}_{g=1}^{G}.
    \label{eq:Roriginal}
\end{equation}

The standard GRPO normalizes the rewards with the mean and standard deviation of the \textit{entire} \(\mathbf{R}\), i.e.,
$
    \Tilde{r}_g^{t} = \left(r_g^{t}-\operatorname{mean}(\mathbf{R})\right)/\operatorname{std}(\mathbf{R}).
$
The advantage \(\Hat{A_g^{t}}\) of the current step is calculated as the sum of the normalized
rewards from the current and the following steps,
\begin{equation}
    \Hat{A_g^t} = \sum_{t'\geq t}^{T_g} \Tilde{r}_g^{t'}.
\end{equation}

\noindent\textbf{Process-reward calculation}
In iterative sketch generation the number of steps (parts) is fixed for a given sketch. Moreover, the ground truth of any intermediate state in a trajectory is also available to the reward model by simply assembling the ground truth paths for each previous parts together. Therefore, we can estimate intermediate rewards more precisely. Since all trajectories in a group have identical lengths (the number of parts), let us denote it by \( T = T_1 = T_2= \cdots= T_G\). Thus, the reward collection in~\eqref{eq:Roriginal} becomes
\begin{equation}
    \mathbf{R}=\left\{\, \{r_g^{t}\}_{t=1}^{T} \,\right\}_{g=1}^{G}.
    \label{eq:Rnew}
\end{equation}

Instead of estimating a unified baseline with all rewards in \(\mathbf{R}\), we compute normalized rewards and advantages within each step. Let \(\mathbf{R}^t = \left\{ r_{g'}^{t}\right\}_{g=1}^G\), then 
\begin{align}
    \Tilde{r}_g^{t} &= \frac{r_g^{t}-\operatorname{mean}(\mathbf{R}^t)}{\operatorname{std}(\mathbf{R}^t)},\\
\text{and}\quad 
    \Hat{A_g^t} &= \Tilde{r}_g^{t}.
\end{align}

We use two rewards to supervise GRPO training: DreamSim reward intended to capture visual quality, and path count reward encouraging appropriate brevity.

\noindent\textbf{DreamSim reward} In each step, we render the current canvas with CairoSVG \cite{cairosvg}, a lightweight rendering engine, and measure its (image-to-image) similarity to ground truth rendering at the same step. For this we use DreamSim~\cite{dreamsim} pre-trained ensemble model to compute cosine similarity between two images in embedded space. 

DreamSim is a learned perceptual similarity metric for images that aligns better with how humans judge visual similarity compared to CLIP~\cite{clip}, DINO~\cite{dino} and LPIPS~\cite{lpips}.  
Let $\operatorname{dreamsim}(I)$ be the embedding of an image $I$. The \textit{DreamSim reward} is 
\begin{equation}
     r_{\text{dreamsim}} = \cos\left(\operatorname{dreamsim}(I_{\text{gen}}),\operatorname{dreamsim}(I_\text{gt})\right),
\label{equation:dreamsim_reward}
\end{equation}
where \(I_{\text{gen}}\) is the current generated rendering and \(I_\text{gt}\) is the current ground truth rendering for the same set of parts.

\noindent\textbf{Path count reward} As identified by Liu et al. \cite{dr.grpo}, GRPO objective induces a bias towards longer trajectories. To keep the response length close to the distribution of the training data, we introduce the \textit {path count reward}:
\begin{equation}
    r_\text{pc} = \operatorname{max}(0, 1 - |N_\text{gt} - N| / N_{\text{gt}}),
\end{equation}
where \(N\) is the number of paths in the \textit{final} output and \(N_\text{gt}\)  is the number of paths in the \textit{final} ground truth. We only regularize the agent on the final number of paths rather than the number of paths for each individual part because empirically we find per-part path count signal to be too noisy. 

The combined reward is a weighted combination of the two rewards:
\begin{equation}
\label{combined_reward}
    r_g^t = r_\text{dreamsim} + \lambda r_\text{pc}.
\end{equation}
Before computing the rewards, we run the responses through a validity verifier. Any response that does not conform with the format will be assigned with \(\operatorname{min}(r_g^t)\) and its trajectory will be terminated at the current step; in such a case, \(N\) and \(N_\text{gt}\) are the cumulative path count up to the last successful step.

\noindent\textbf{Learning algorithm}
Our multi-turn process-reward GRPO learning objective builds on DeepSeekMath\cite{deepseekmath}.
Let
\(\rho_{g,k}^t(\theta) = \frac{\pi_\theta(o_{g,k}^t\mid q, o_{g,<k}^t)}{\pi_{\theta_{old}}(o_{g,k}^t\mid q, o_{g,<k}^t)}\) be the token level ratios where $\pi_{\theta}, \pi_{\theta_{old}}$ are the current and the old policy models during policy update (our VLM agent) and $q,o$ are questions and outputs sampled from the question dataset and the old policy $\pi_{\theta_{old}}$. $k$ indicates the token position of the response and $o_{g, <k}^t$ indicates the conditional generation process of $g$-th trajectory's $t$-th step response conditioned on its first $<k$ tokens. The learning objective, multi-turn and thus different from~\cite{deepseekmath}, is
\begin{equation}
\resizebox{0.9\linewidth}{!}{$
\begin{aligned}
    \mathcal{J}_{GRPO}(\theta) &= \mathbb{E}{\left[q \sim P(Q), \left\{\{o_g^t\}_{t=1}^{T}\right\}_{g=1}^G \sim \pi_{\theta_{old}}(O|q)\right]}  \\
    & \frac{1}{GT}\sum_{g=1}^G\sum_{t=1}^{T}\frac{1}{|o_g^t|} \sum_{k=1}^{|o_g^t|} \left\{ \min \left[ \rho_{g,k}^t(\theta)\, \hat{A}_{g,k}^t, \text{clip} \left( \rho_{g,k}^t(\theta), 1 - \epsilon, 1 + \epsilon \right)\, \hat{A}_{g,k}^t \right] - \beta \mathbb{D}_{KL}\left[\pi_{\theta} || \pi_{ref}\right]\right\} ,
\end{aligned}
$}
\label{eq:GRPO-obj}
\end{equation}

where $\epsilon$ and $\beta$ are hyper-parameters, and $\hat{A}_{g,k}^t$ is the $k$-th token level advantage. We estimate the KL divergence with the following unbiased estimator \cite{kl_approx}:
\begin{equation}
    \mathbb{D}_{KL}\left[\pi_{\theta} || \pi_{ref}\right] = \nu_{i,k}^t(\theta) - \log \nu_{i,k}^t(\theta) - 1,
\end{equation}
where
\(\nu_{g,k}^t(\theta) = \frac{\pi_{ref}(o_{g,k}^t\mid q, o_{g,<k}^t)}{\pi_\theta(o_{g,k}^t\mid q, o_{g,<k}^t)}\),
where $\pi_{\theta_{ref}}$ is the reference model. 
We present the pseudocode in the supplementary materials.

\section{Experiments} \label{experiments}

\begin{figure}[t]
\begin{minipage}[c]{.6\textwidth}
  \centering
  \includegraphics[width=0.99\linewidth]{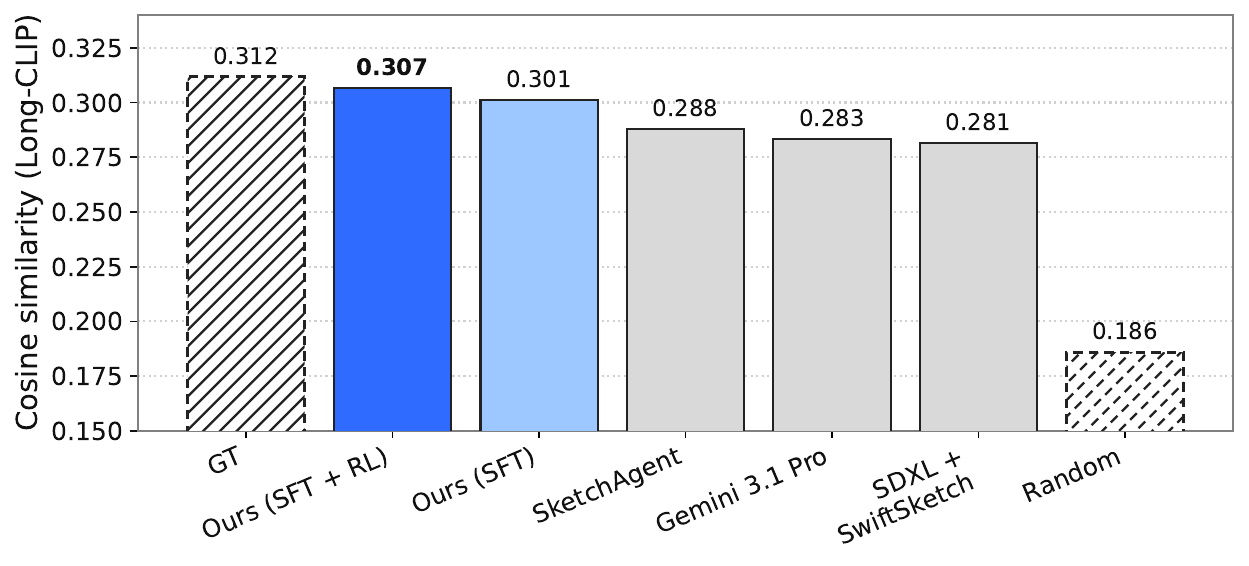}
\end{minipage}%
\begin{minipage}[c]{.4\textwidth}
\caption{The Long-CLIP cosine similarity across all tested models. The Ground Truth (GT) value and the Random value are the cosine similarity scores of text to the ground truth sketches from ControlSketch-Part and sketches of randomly sampled paths.}
  \label{fig:long-clip-bar-chart}
\end{minipage}
\vspace{-20pt}
\end{figure}

We experimentally assess generation quality across both the step-by-step sketching procedure and the final output, comparing against state-of-the-art methods. We further validate the contribution of our multi-turn process-reward GRPO training through ablation studies. Evaluation is conducted using both automatic metrics and user studies.

\subsection{Experimental Setup}

\noindent\textbf{Training data} We follow an established practice in two-stage LLM training pipelines~\cite{wang2025kiminaproverpreviewlargeformal,chen2025sftrlearlyinvestigation} that uses separate data for SFT and RL to prevent imitation bias, which has been found~\cite{kang2025quagmiressftrlposttraininghigh} to reduce exploration potential at the RL stage.
We reserve the high-quality (and relatively costly to create) ControlSketch-Part dataset for RL, and prepare an alternative dataset for the SFT stage. The SFT training dataset is obtained with the same pipeline described in Sec.~\ref{sec:datapipe} but annotated with Gemini 2.5 Flash, a model 6.7$\times$ cheaper than Gemini 3.0 Pro.

\noindent\textbf{Implementation details} We fine-tune Qwen3-VL-30B-A3B \cite{Qwen3VL} as the backbone of our sketching agent. For both stages, LoRA \cite{lora} with rank 64 is used for fine-tuning. We run SFT training with a learning rate of 2e-4 and a batch size of 128 for 5400 steps. RL training takes an additional 1000 steps with a batch size of 8, a group size of 8 and a learning rate of 3e-6. We use the reward proposed in \cref{combined_reward} with $\lambda=1.0$ and turn off KL divergence loss for RL training. Adam with $\beta_1=0.9, \beta_2=0.95$ and $\epsilon=1e-8$ is used throughout the entire training. We use Thinking Machines Lab's Tinker \cite{tinker} for both training stages. Bézier curve coordinates are rounded to the nearest ten for SFT training, while the original integer coordinates are retained for RL training.

\noindent\textbf{Baseline methods} We benchmark our method against three methods: SketchAgent~\cite{sketchagent}, Gemini 3.1 Pro and SDXL~\cite{sdxl}+SwiftSketch~\cite{swiftsketch}. SketchAgent is a Claude Sonnet-based sequential sketch generation method through zero-shot prompting. The original paper uses Claude Sonnet 3.5, which is no longer available, so we switch to the more recent Claude Sonnet 4.5. We also compare against Gemini 3.1 Pro, one of the latest general-purpose VLMs at the time of writing, used as a direct whole sketch generator. SwiftSketch is an image-to-sketch diffusion model. Since it requires an image as input, we first use SDXL, a text-to-image diffusion model, to generate images from text, and then apply SwiftSketch to convert them into sketches. For methods that require a single text caption, we concatenate all part descriptions.

\noindent\textbf{Evaluation metrics} Both automatic metrics and user studies are used to assess the visual quality of the generated sketches. Because the lengths of the concatenated text captions often exceed the maximum input length of CLIP~\cite{clip}, we use Long-CLIP~\cite{zhang2024longclip} with a maximum input length of 248 tokens, to evaluate how faithful the final sketch is to the text caption. For each sketch rendering, we compute the cosine similarity of image embedding to the embedding of the concatenated part descriptions, as a measurement of faithfulness to text input. Note that this uses different embeddings, and thus is a different metric, than DreamSim used in our reward mechanism (\cref{equation:dreamsim_reward}). 

We also conducted double-blind forced choice user preference studies between our method and baselines. These include two questions. In the first question, we ask the user to pick one from a pair of (whole) sketches based on the overall visual quality according to the associated text caption. In the second question, we present a looping animation showing part-by-part generation of a pair of sketches, and ask users to choose the sketch whose \textit{generation procedure} better matches the part descriptions. We ask the first question for comparison with all methods and ask both questions for comparison with SketchAgent, the only baseline capable of part-by-part generation.

\subsection{Experiment Results and Analysis}

\noindent\textbf{Long-CLIP cosine similarity}
\cref{fig:long-clip-bar-chart} reports the Long-CLIP cosine similarity scores of all methods and reference baselines including ground truth (GT) and randomly generated sketches (Random). The GT value is the mean Long-CLIP cosine similarity between concatenated part descriptions and the corresponding GT sketch in ControlSketch-Part, which can be viewed as the upper bound of the performance. The Random baseline is the mean Long-CLIP cosine similarity between concatenated part descriptions and sketches with randomly sampled strokes, where the number of cubic Bézier curves is sampled uniformly from [0, 32] and curve coordinates are sampled uniformly from [0, 512]. It establishes a lower bound on the metric.

Our full model (SFT + RL) achieves the best performance across all methods, surpassing the SFT-only variant, validating the contribution of both training stages. Among prior methods, SketchAgent performs best, suggesting that progressive, part-by-part generation holds a meaningful advantage over holistic approaches. Gemini 3.1 Pro, despite its general strength, falls short of specialist agents, highlighting that text-to-vector sketch generation remains a challenging domain where task-specific training still matters. SDXL + SwiftSketch trails all baselines, as errors from the text-to-image stage compound in the subsequent image-to-sketch generation. While the numerical range for all the non-random methods is fairly compressed, our user studies and qualitative results confirm that the metric differences correspond to meaningful visual quality distinctions.

\begin{figure}[t]
  \centering
  \includegraphics[width=0.95\linewidth]{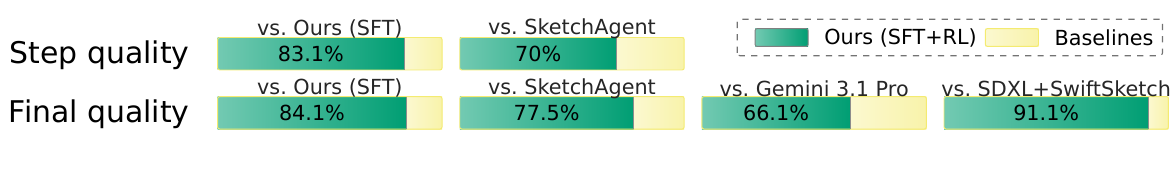}
  \vspace{-0.5cm}
  \caption{Pairwise preference studies conducted between our final model (SFT + RL) and the baselines. The title for each bar describes the baseline method that we are comparing against. The first column is the ablation between our final method (SFT + RL) versus the SFT only variant, which demonstrates the effectiveness of RL.}\vspace{-1em}
  \label{fig:user_study}
\end{figure}

\noindent\textbf{User studies}
We conduct user studies via Prolific, an online crowdsourcing platform. For step quality preference, we collect 426 responses per baseline comparison from 142 participants, and for final output quality preference, 560 responses per baseline comparison from 146 participants, totalling 3,092 pairwise comparisons. \cref{fig:user_study} reports the percentage of comparisons in which our method was preferred. Across both evaluation settings and all baselines, participants consistently favored our results.

\begin{figure}[t]
\centering
\tiny
\setlength{\tabcolsep}{0pt}
\renewcommand{\arraystretch}{0.0}
\newcommand{\lightrule}{\arrayrulecolor{black!20}\cmidrule[\lightrulewidth](l{0em}r{0em}){1-12}\arrayrulecolor{black}}
\begin{tabular}{
ccccccccccccccc
}
 & {\scalebox{0.8}{\tiny \shortstack{Ours\\(SFT+RL)}}} & {\scalebox{0.8}{\tiny \shortstack{Sketch\\-Agent}}} & {\scalebox{0.8}{\tiny \shortstack{Gemini\\3.1 pro}}} & {\scalebox{0.8}{\tiny \shortstack{SDXL+\\Swift\\Sketch}}} & & {\scalebox{0.8}{\tiny \shortstack{Ours\\(SFT+RL)}}} & {\scalebox{0.8}{\tiny \shortstack{Sketch\\-Agent}}} & {\scalebox{0.8}{\tiny \shortstack{Gemini\\3.1 pro}}} & {\scalebox{0.8}{\tiny \shortstack{SDXL+\\Swift\\Sketch}}} & & {\scalebox{0.8}{\tiny \shortstack{Ours\\(SFT+RL)}}} & {\scalebox{0.8}{\tiny \shortstack{Sketch\\-Agent}}} & {\scalebox{0.8}{\tiny \shortstack{Gemini\\3.1 pro}}} & {\scalebox{0.8}{\tiny \shortstack{SDXL+\\Swift\\Sketch}}} \\
\rotatebox{90}{\parbox{0.075\textwidth}{\centering\tiny Angel}} & \includegraphics[width=0.075\linewidth]{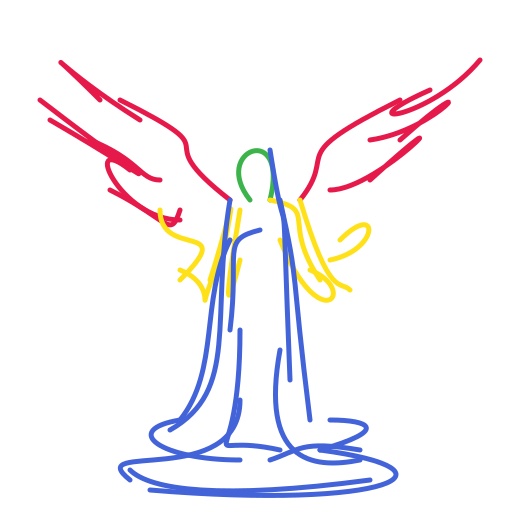} & \includegraphics[width=0.075\linewidth]{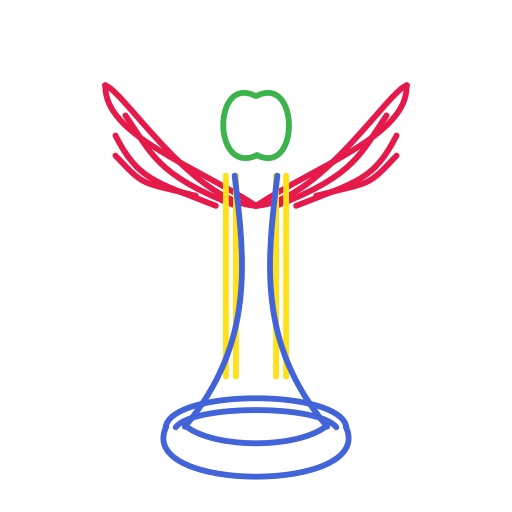} & \includegraphics[width=0.075\linewidth]{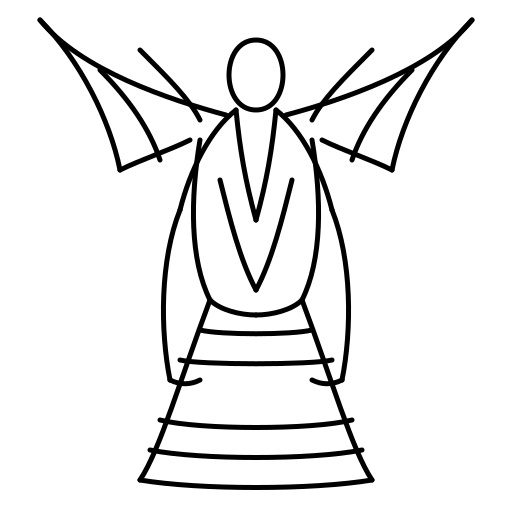} & \includegraphics[width=0.075\linewidth]{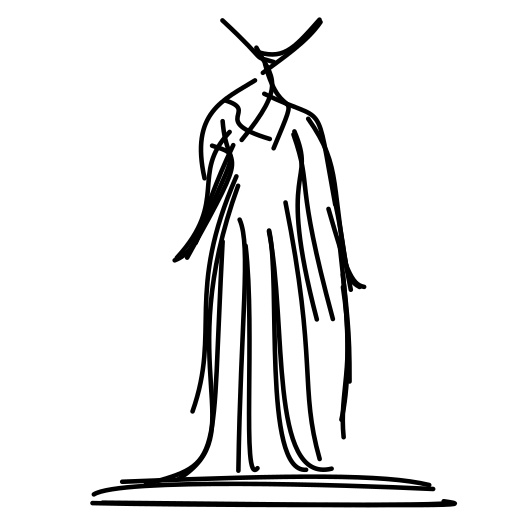} &\rotatebox{90}{\scalebox{0.82}{\parbox{0.075\linewidth}{\centering\tiny Astronaut}}} & \includegraphics[width=0.075\linewidth]{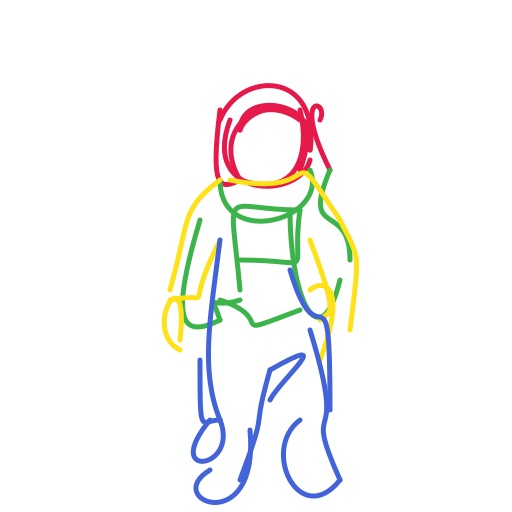} & \includegraphics[width=0.075\linewidth]{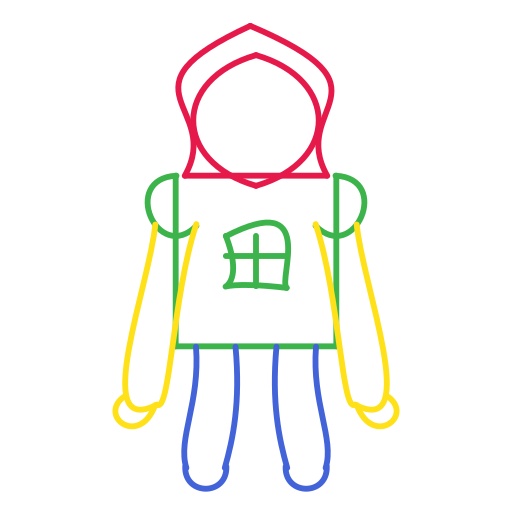} & \includegraphics[width=0.075\linewidth]{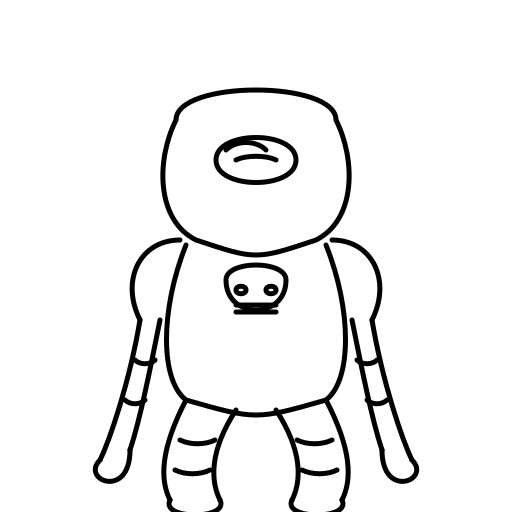} &\includegraphics[width=0.075\linewidth]{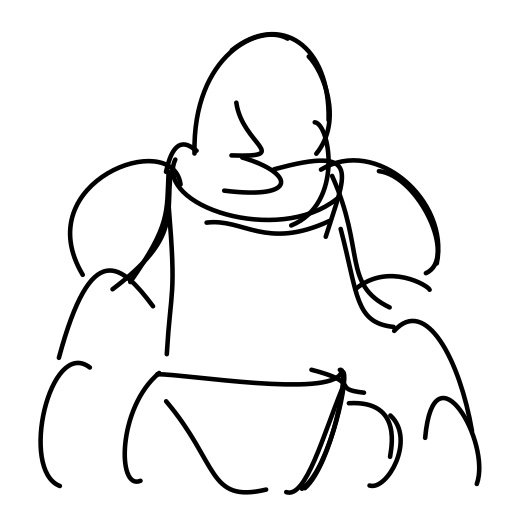} & \rotatebox{90}{\parbox{0.075\linewidth}{\centering\tiny Bear}} & \includegraphics[width=0.075\linewidth]{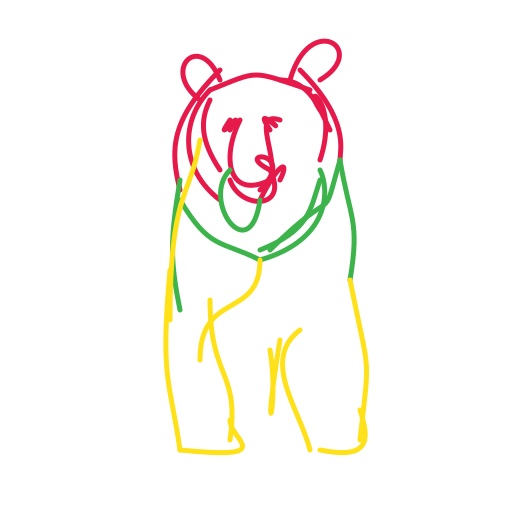} & \includegraphics[width=0.075\linewidth]{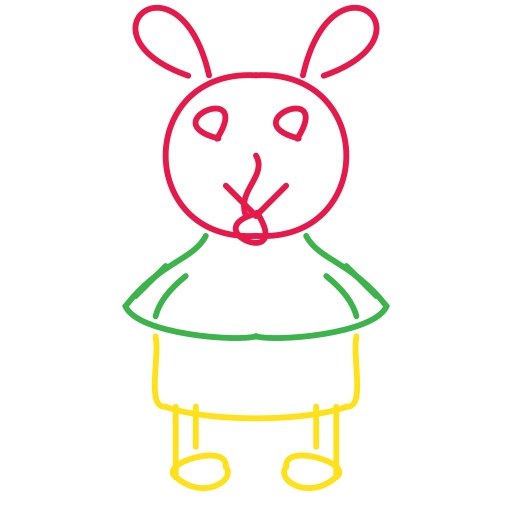} & \includegraphics[width=0.075\linewidth]{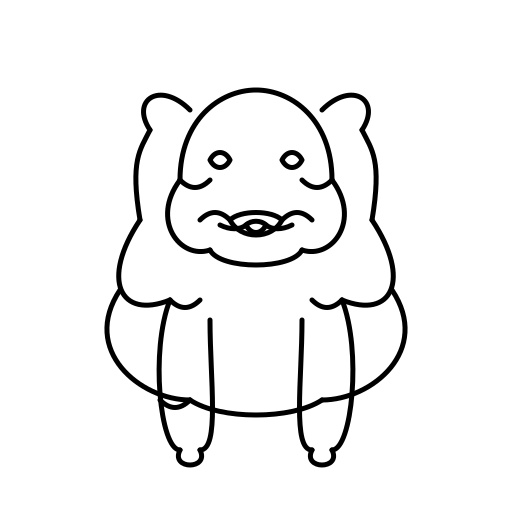} & \includegraphics[width=0.075\linewidth]{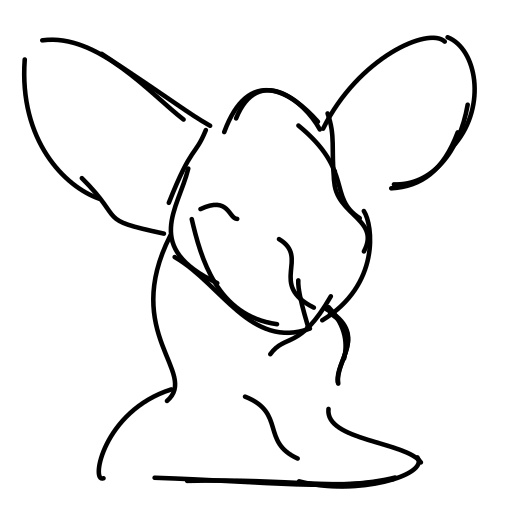} \\

\rotatebox{90}{\parbox{0.075\linewidth}{\centering\tiny Bicycle}} & \includegraphics[width=0.075\linewidth]{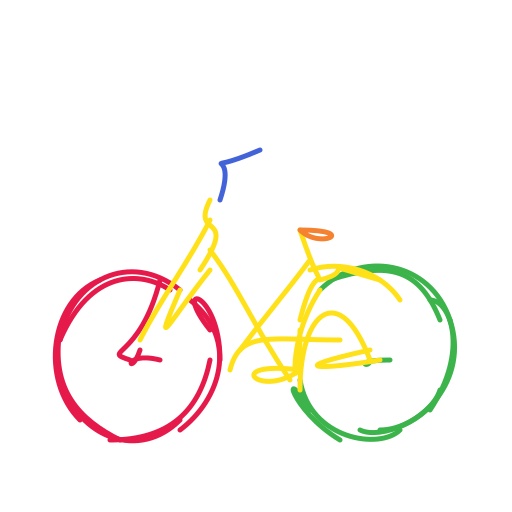} & \includegraphics[width=0.075\linewidth]{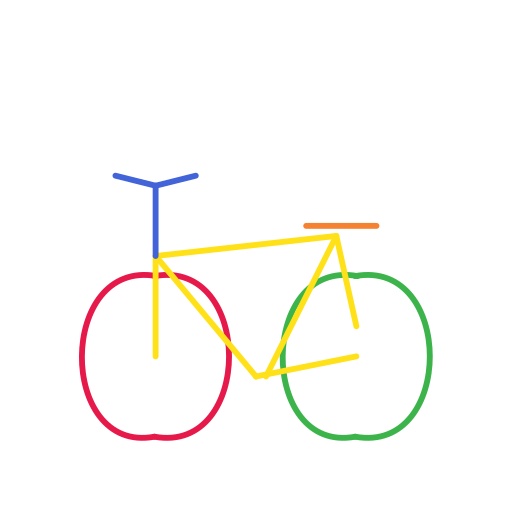} & \includegraphics[width=0.075\linewidth]{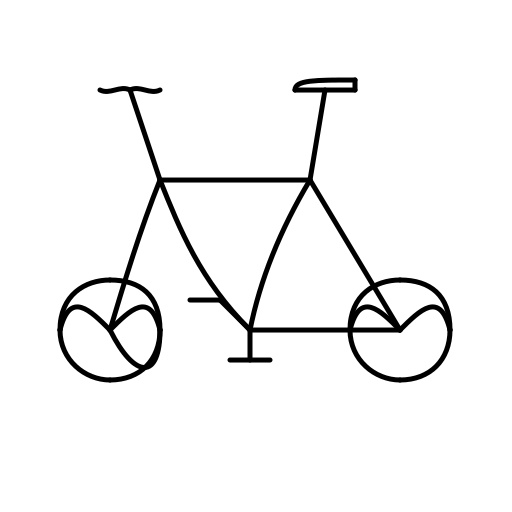} &\includegraphics[width=0.075\linewidth]{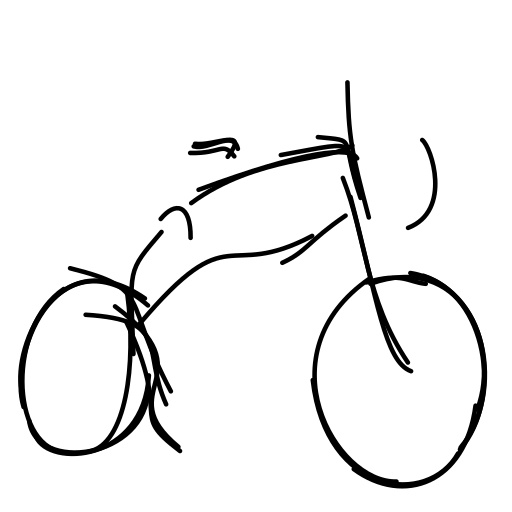} & \rotatebox{90}{\parbox{0.075\linewidth}{\centering\tiny Car}} & \includegraphics[width=0.075\linewidth]{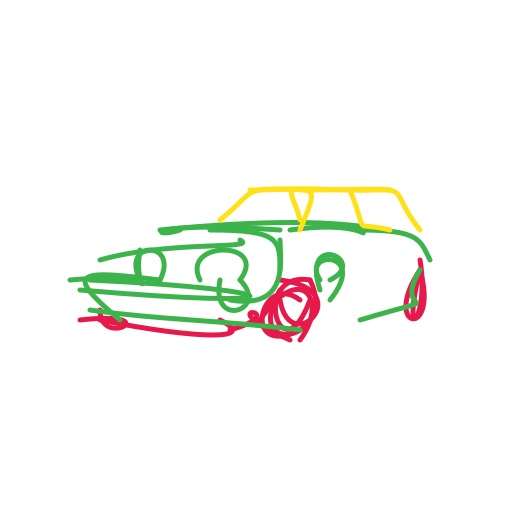} & \includegraphics[width=0.075\linewidth]{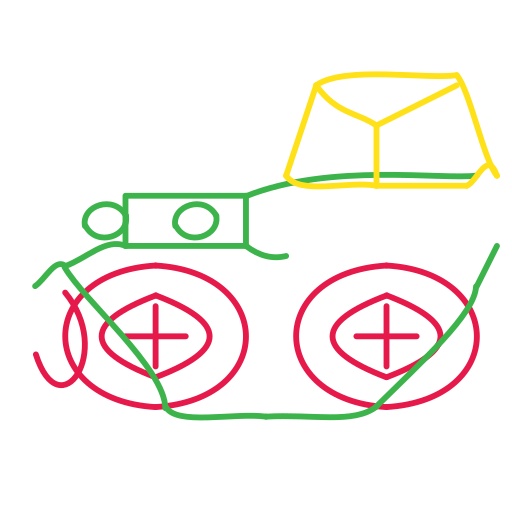} & \includegraphics[width=0.075\linewidth]{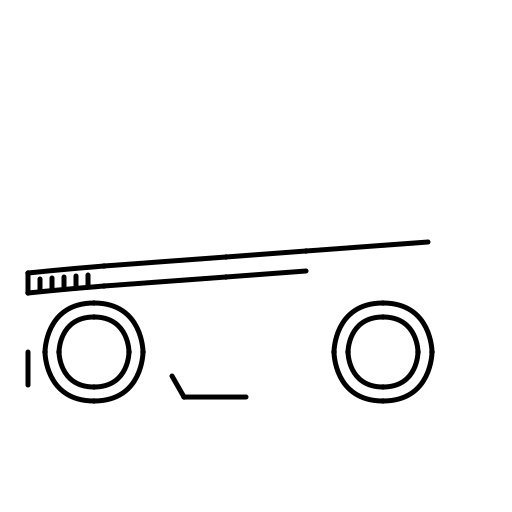} & \includegraphics[width=0.075\linewidth]{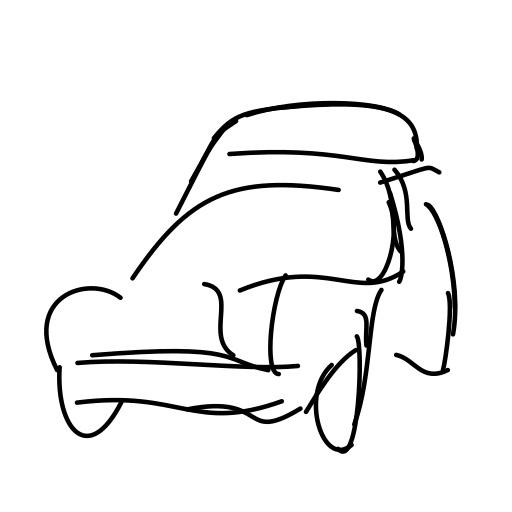} & \rotatebox{90}{\parbox{0.075\linewidth}{\centering\tiny Cat}} & \includegraphics[width=0.075\linewidth]{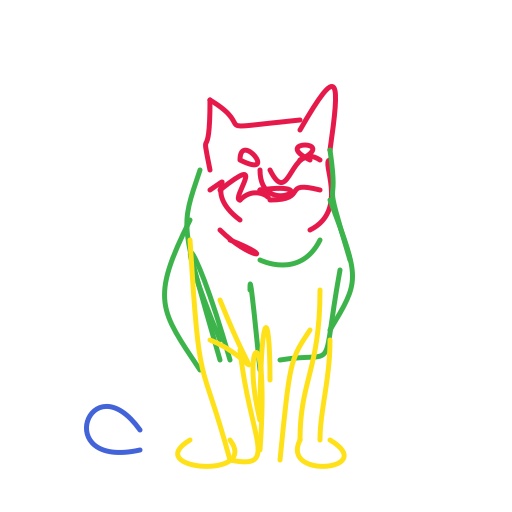} & \includegraphics[width=0.075\linewidth]{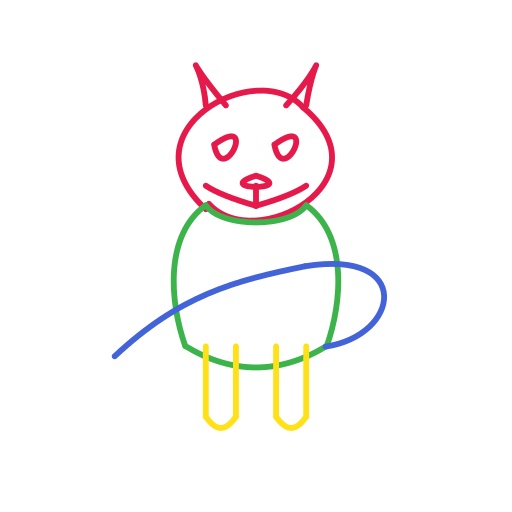} & \includegraphics[width=0.075\linewidth]{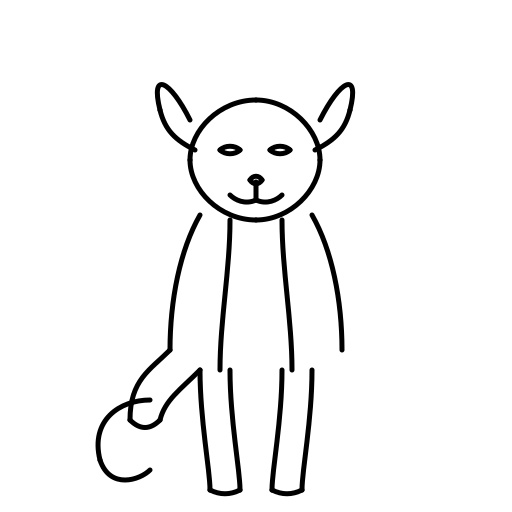} & \includegraphics[width=0.075\linewidth]{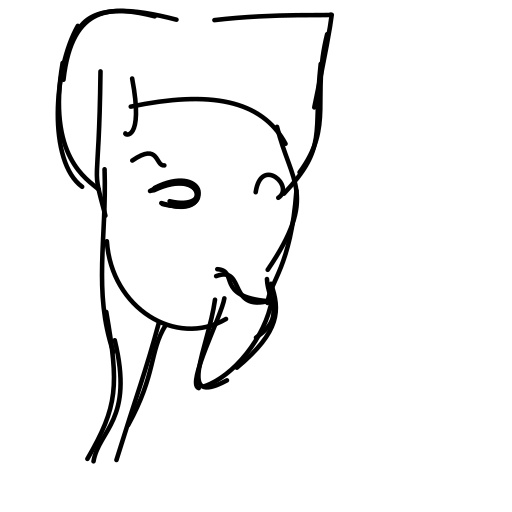} \\

\rotatebox{90}{\parbox{0.075\linewidth}{\centering\tiny Chair}} & \includegraphics[width=0.075\linewidth]{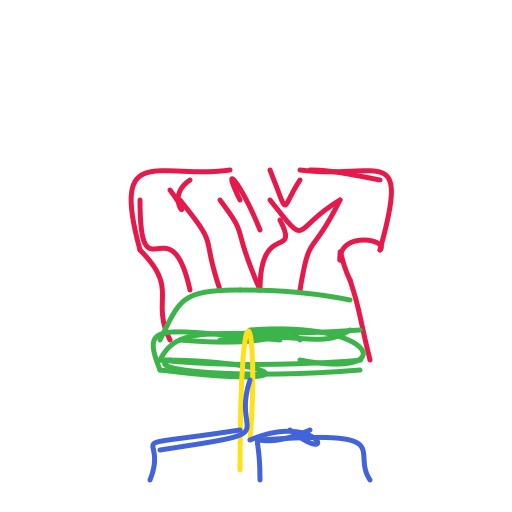} & \includegraphics[width=0.075\linewidth]{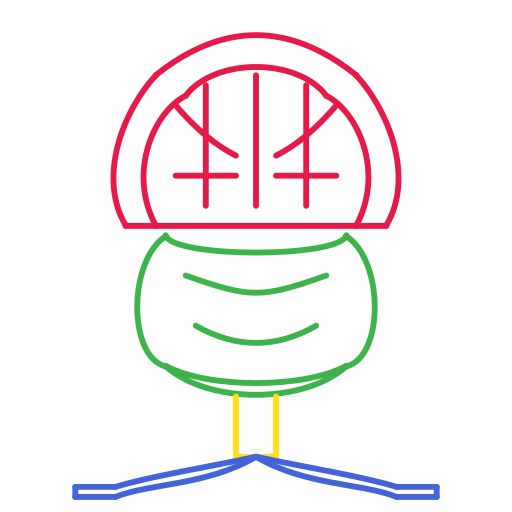} & \includegraphics[width=0.075\linewidth]{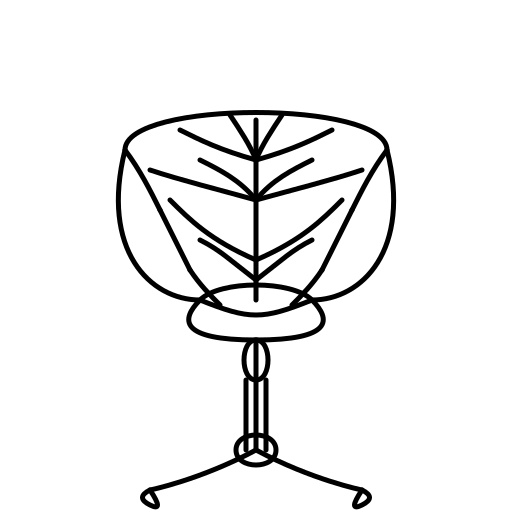} & \includegraphics[width=0.075\linewidth]{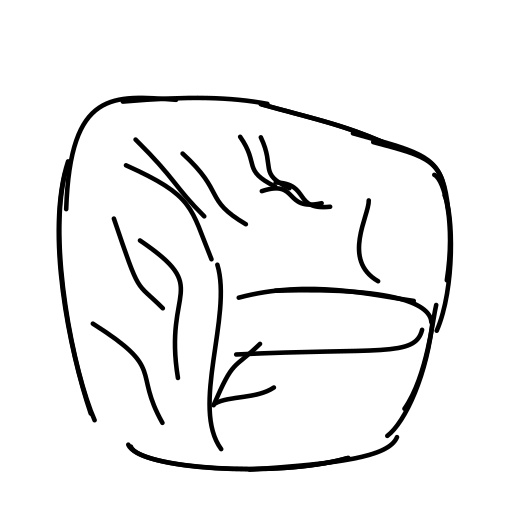} & \rotatebox{90}{\parbox{0.075\linewidth}{\centering\tiny Crab}} & \includegraphics[width=0.075\linewidth]{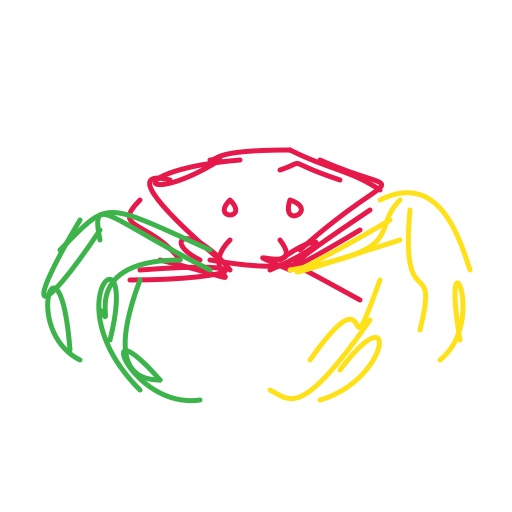} & \includegraphics[width=0.075\linewidth]{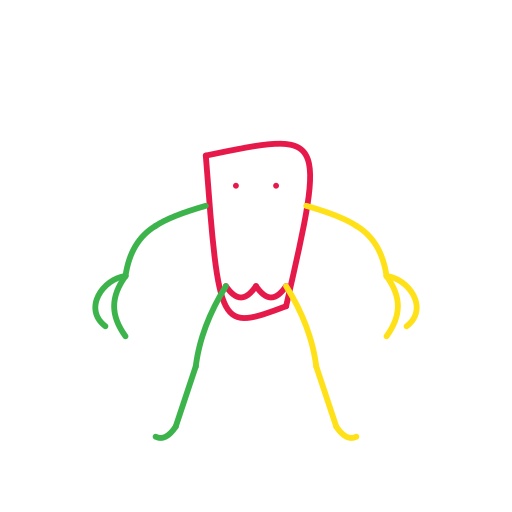} & \includegraphics[width=0.075\linewidth]{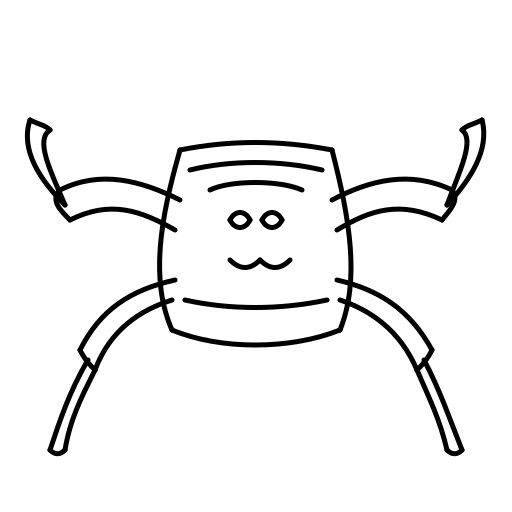} & \includegraphics[width=0.075\linewidth]{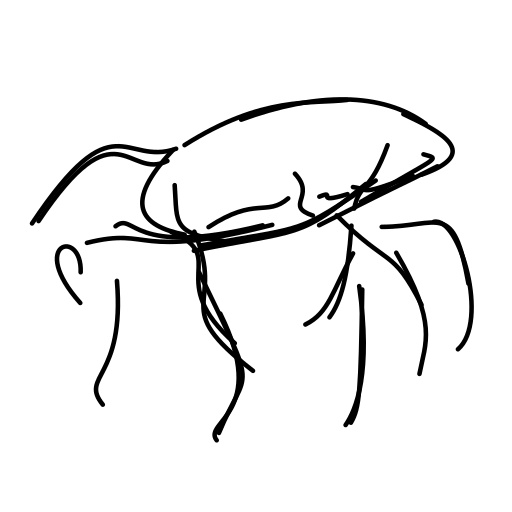} & \rotatebox{90}{\parbox{0.075\linewidth}{\centering\tiny Dog}} & \includegraphics[width=0.075\linewidth]{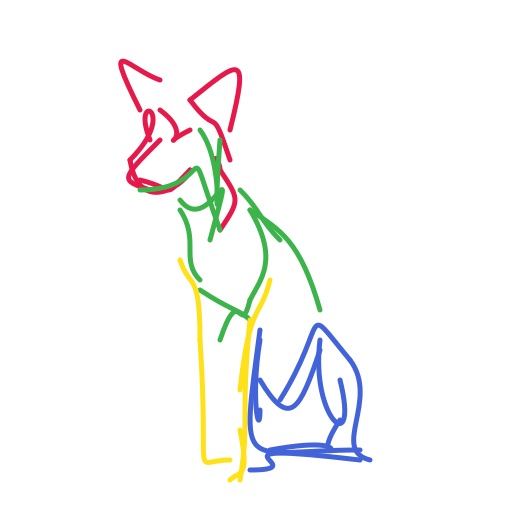} & \includegraphics[width=0.075\linewidth]{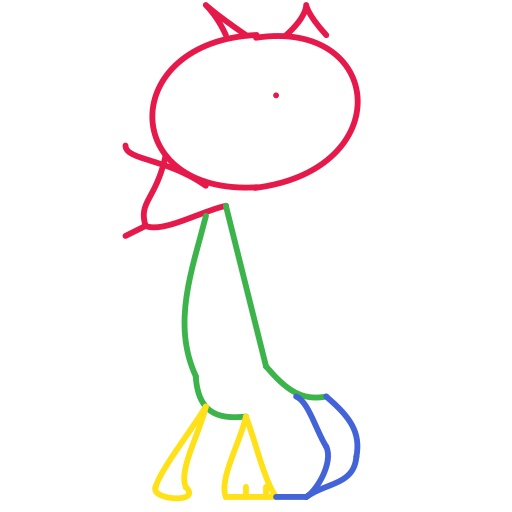} & \includegraphics[width=0.075\linewidth]{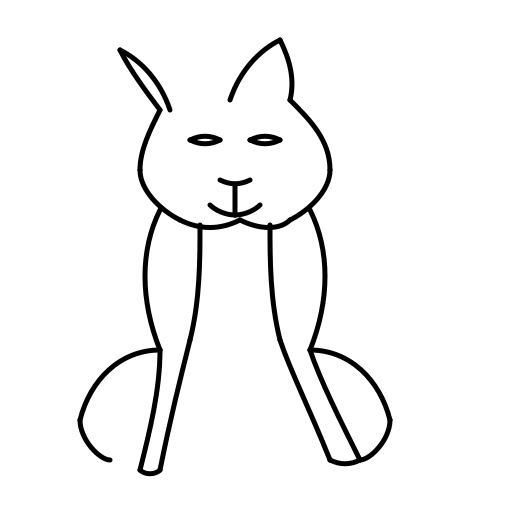} & \includegraphics[width=0.075\linewidth]{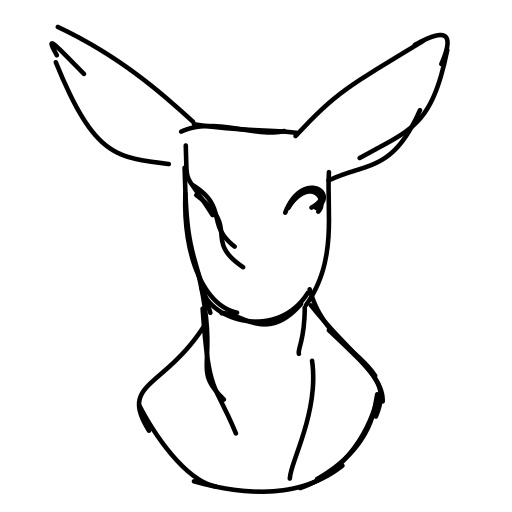} \\

\rotatebox{90}{\parbox{0.075\linewidth}{\centering\tiny Fish}} & \includegraphics[width=0.075\linewidth]{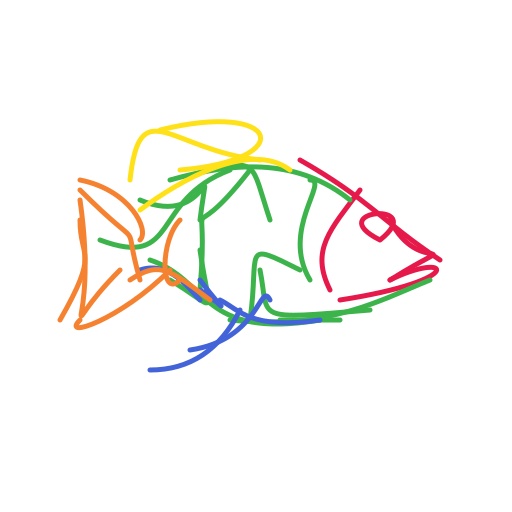} & \includegraphics[width=0.075\linewidth]{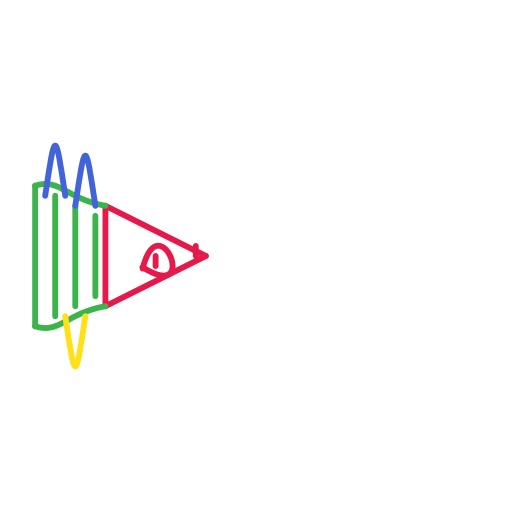} & \includegraphics[width=0.075\linewidth]{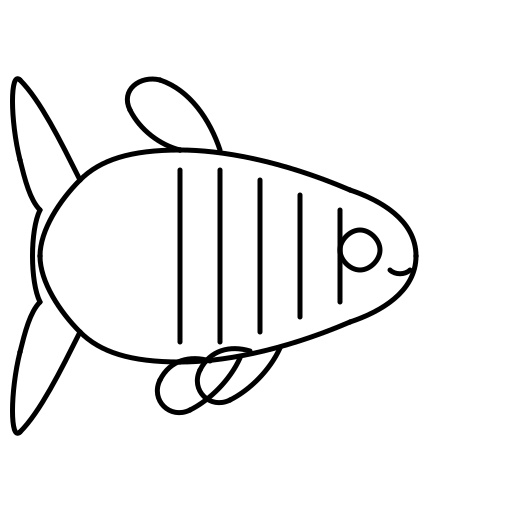} & \includegraphics[width=0.075\linewidth]{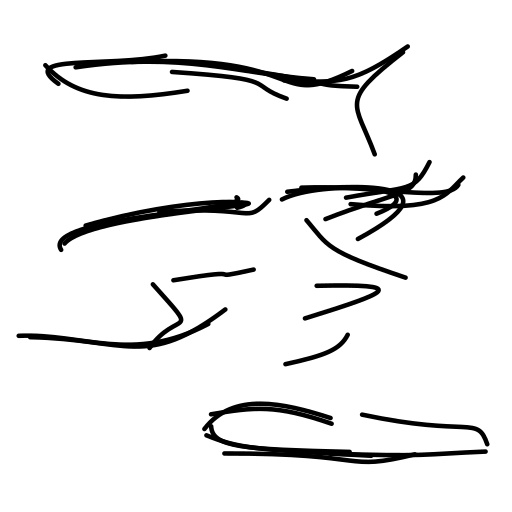} & \rotatebox{90}{\parbox{0.075\linewidth}{\centering\tiny Horse}} & \includegraphics[width=0.075\linewidth]{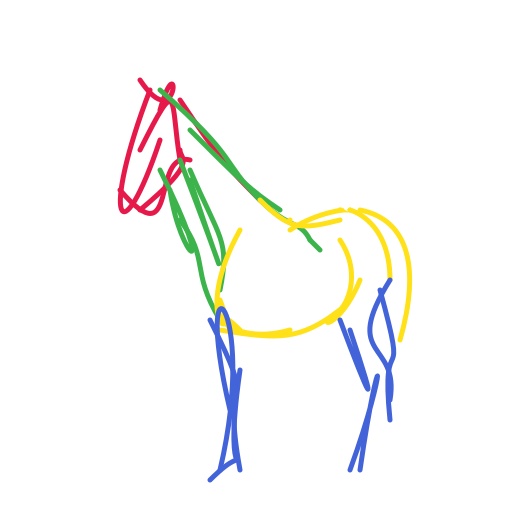} & \includegraphics[width=0.075\linewidth]{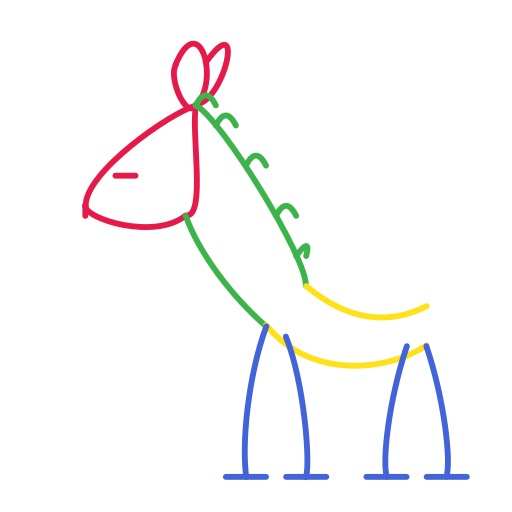} & \includegraphics[width=0.075\linewidth]{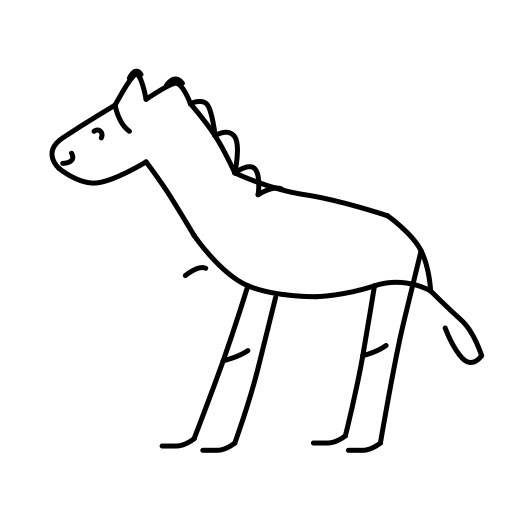} & \includegraphics[width=0.075\linewidth]{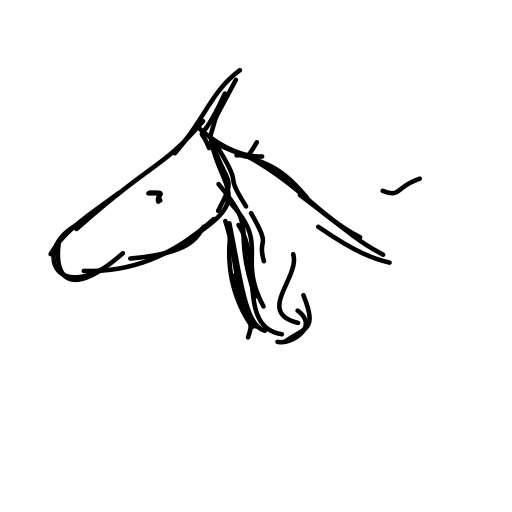} & \rotatebox{90}{\parbox{0.075\linewidth}{\centering\tiny Rabbit}} & \includegraphics[width=0.075\linewidth]{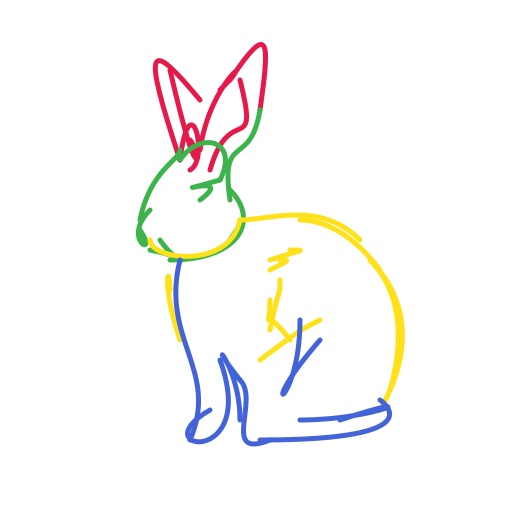} & \includegraphics[width=0.075\linewidth]{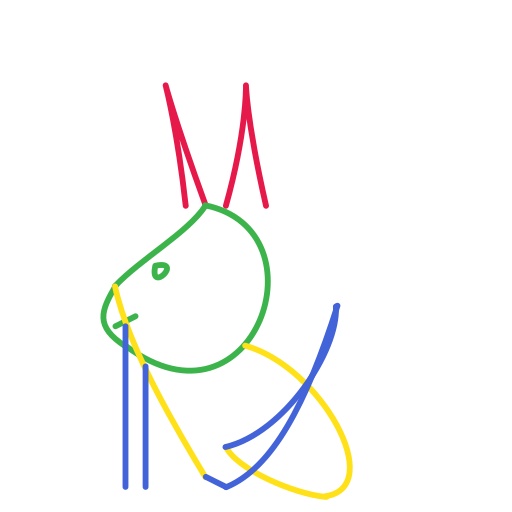} & \includegraphics[width=0.075\linewidth]{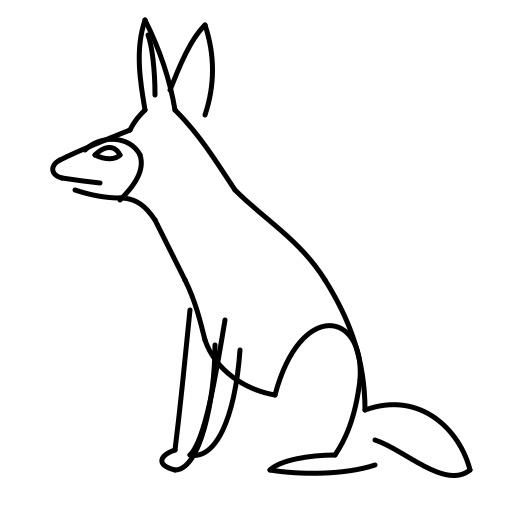} & \includegraphics[width=0.075\linewidth]{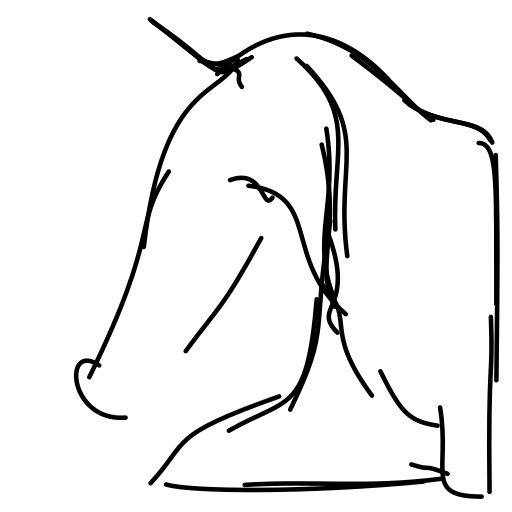} \\

\rotatebox{90}{\parbox{0.075\linewidth}{\centering\tiny Robot}} & \includegraphics[width=0.075\linewidth]{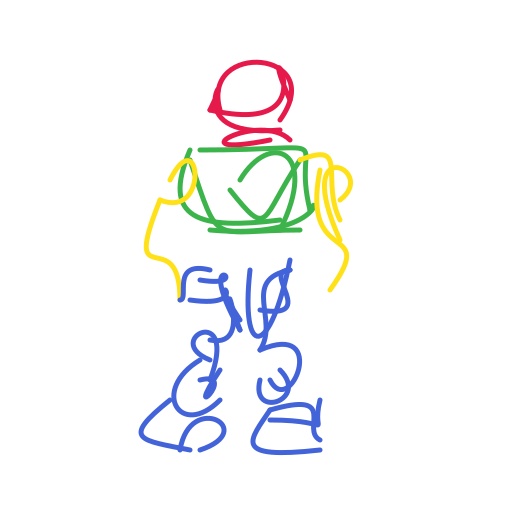} & \includegraphics[width=0.075\linewidth]{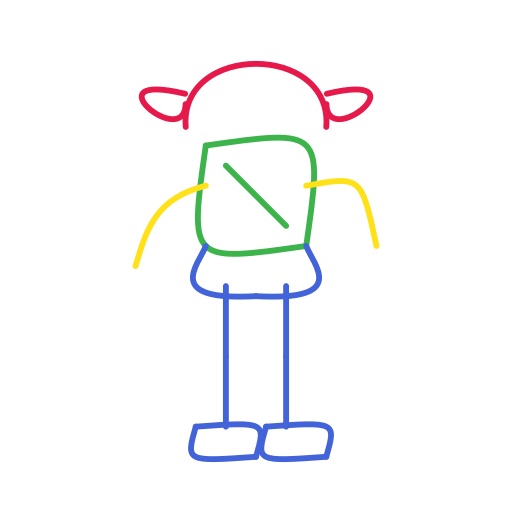} & \includegraphics[width=0.075\linewidth]{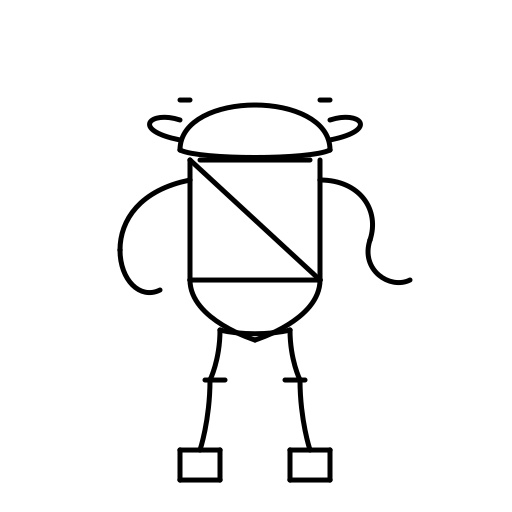} & \includegraphics[width=0.075\linewidth]{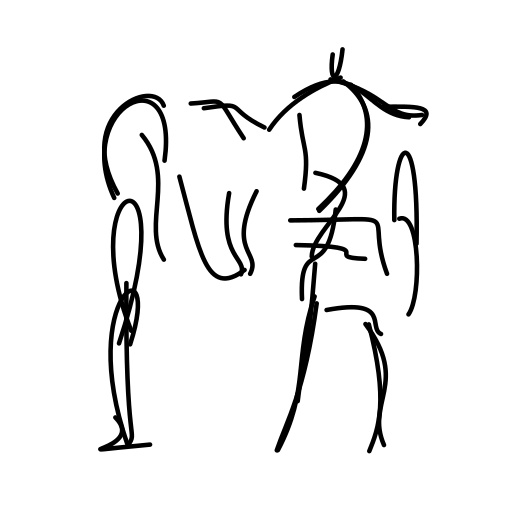} & \rotatebox{90}{\scalebox{0.82}{\parbox{0.075\linewidth}{\centering\tiny Sculpture}}} & \includegraphics[width=0.075\linewidth]{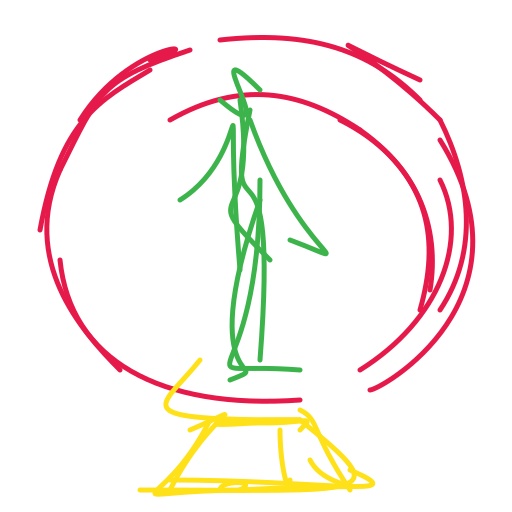} & \includegraphics[width=0.075\linewidth]{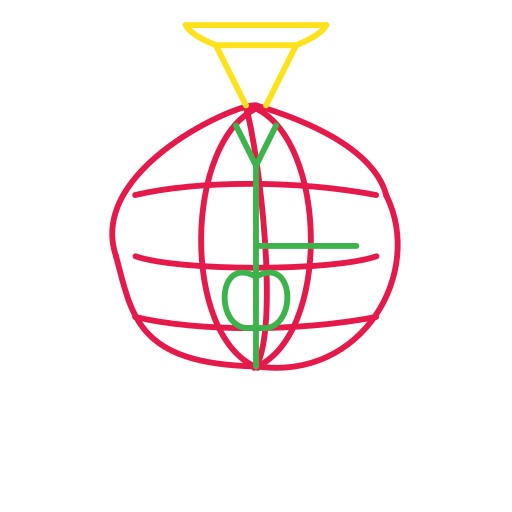} & \includegraphics[width=0.075\linewidth]{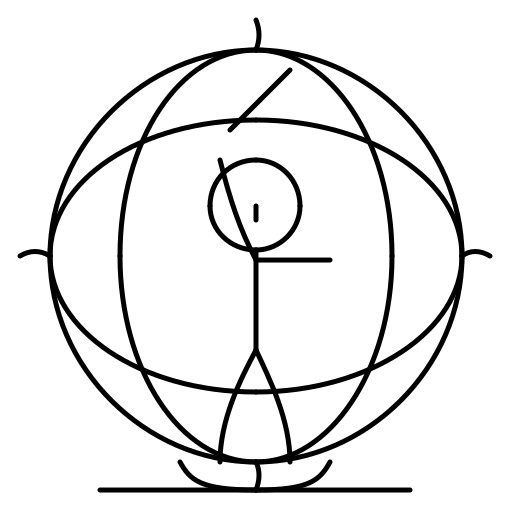} & \includegraphics[width=0.075\linewidth]{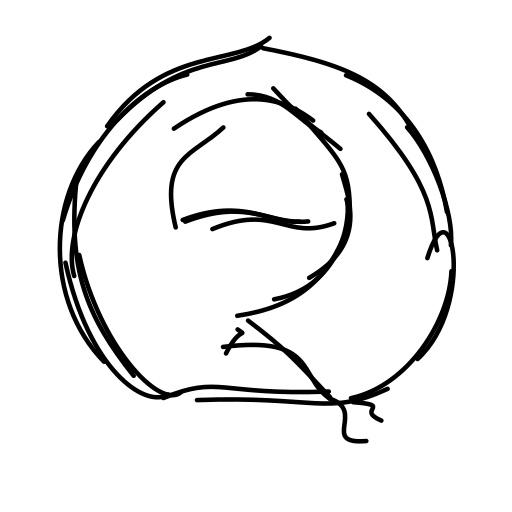} & \rotatebox{90}{\parbox{0.075\linewidth}{\centering\tiny Woman}} & \includegraphics[width=0.075\linewidth]{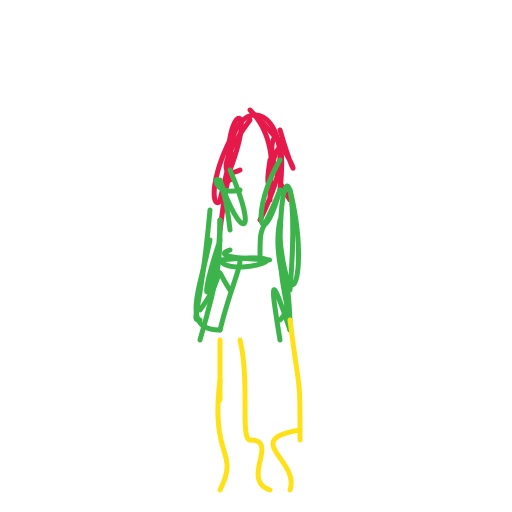} & \includegraphics[width=0.075\linewidth]{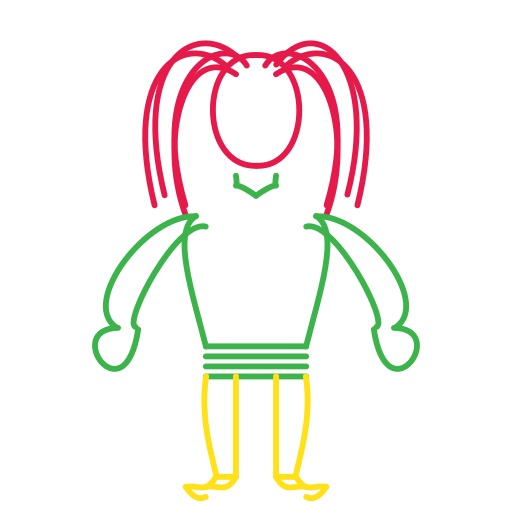} & \includegraphics[width=0.075\linewidth]{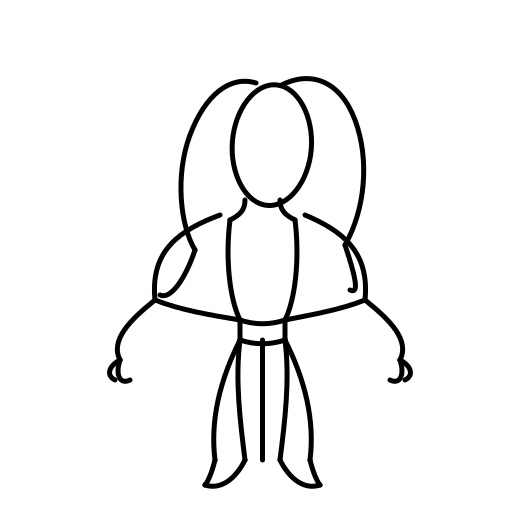} & \includegraphics[width=0.075\linewidth]{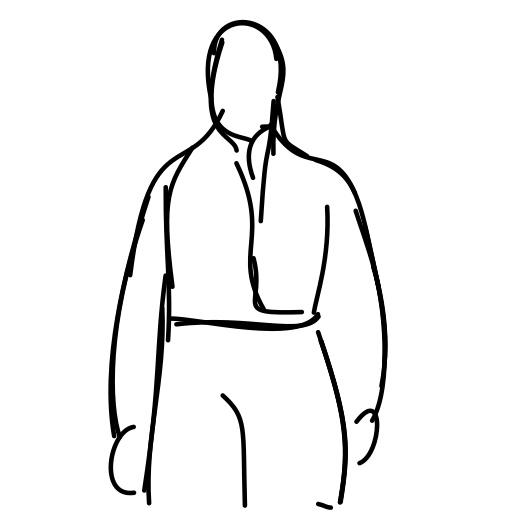}  \\
\end{tabular}
\caption{Qualitative comparison. Part-by-part generated samples are color-coded to illustrate different parts. One-shot generations are rendered in black. Samples in each group are generated with the same text input. Our model and training process do not rely on the class labels in any way, and we only show these for reference.}
\label{fig:example_comparison}

\end{figure}

\begin{figure}[!tbh]
\centering
\tiny
\setlength{\tabcolsep}{2pt}
\renewcommand{\arraystretch}{0.5}
\newcommand{\lightrule}{\arrayrulecolor{black!20}\cmidrule[\lightrulewidth](l{0em}r{0em}){1-12}\arrayrulecolor{black}}
\begin{tabular}{
cccccccccccc
}
\rotatebox{90}{\parbox{0.09\textwidth}{\centering\tiny Angel}} & \includegraphics[width=0.09\linewidth]{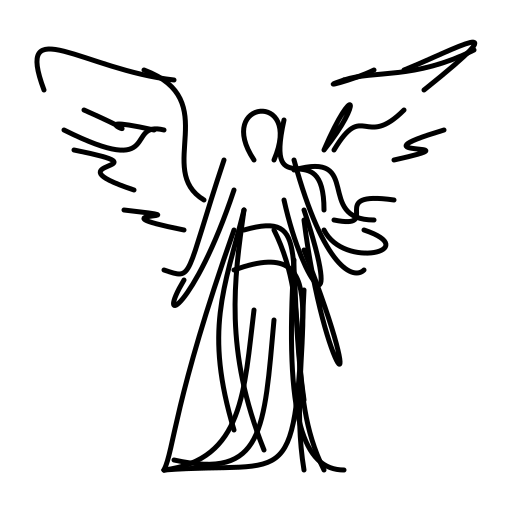} & \includegraphics[width=0.09\linewidth]{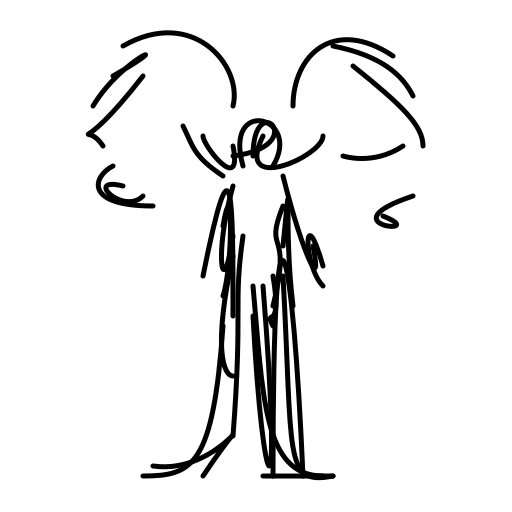} & \includegraphics[width=0.09\linewidth]{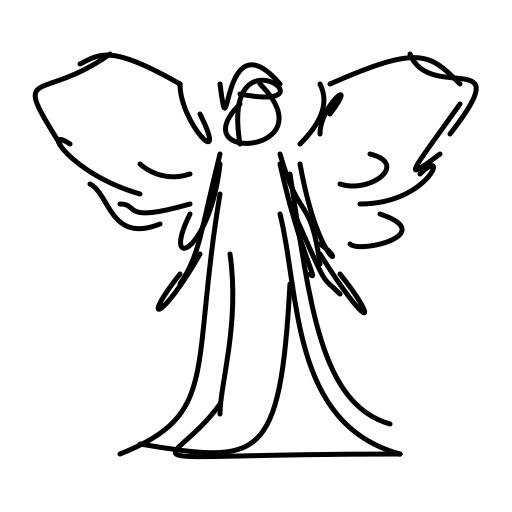} & \rotatebox{90}{\parbox{0.09\linewidth}{\centering\tiny Astronaut}} & \includegraphics[width=0.09\linewidth]{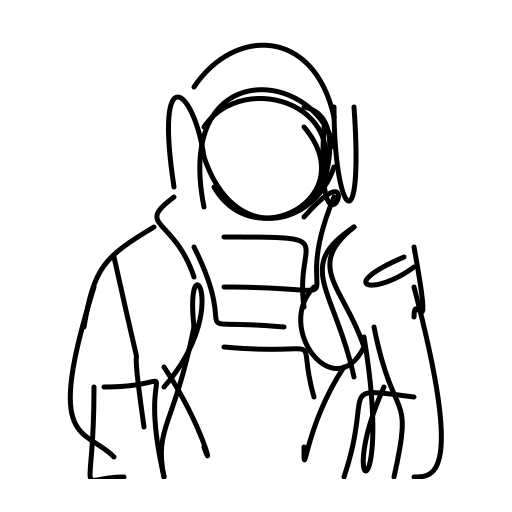} & \includegraphics[width=0.09\linewidth]{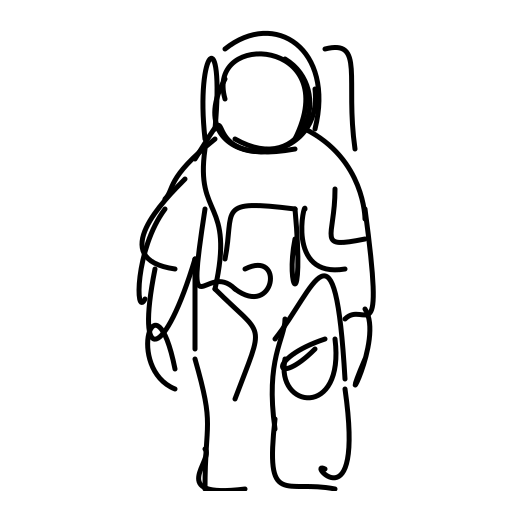} & \includegraphics[width=0.09\linewidth]{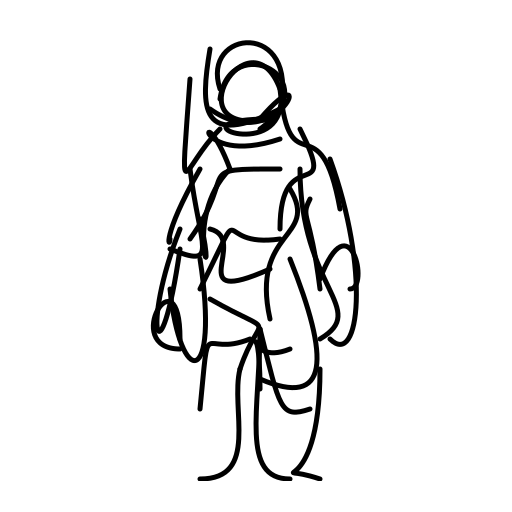} & \rotatebox{90}{\parbox{0.09\linewidth}{\centering\tiny Bear}} & \includegraphics[width=0.09\linewidth]{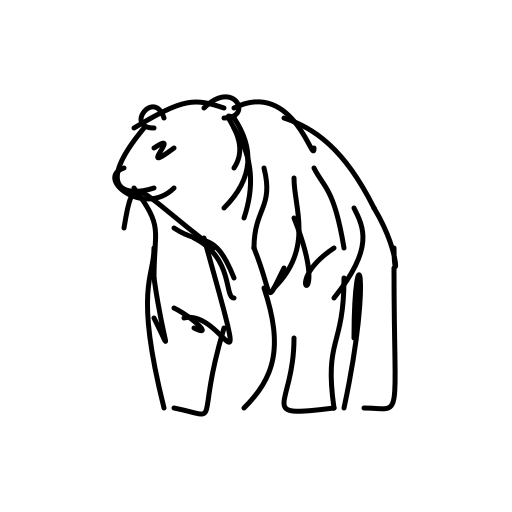} & \includegraphics[width=0.09\linewidth]{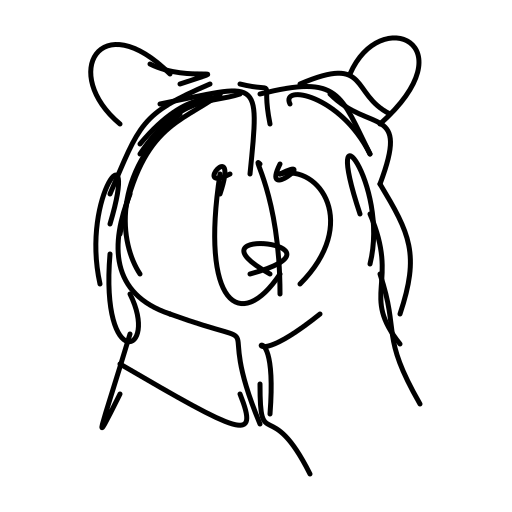} & \includegraphics[width=0.09\linewidth]{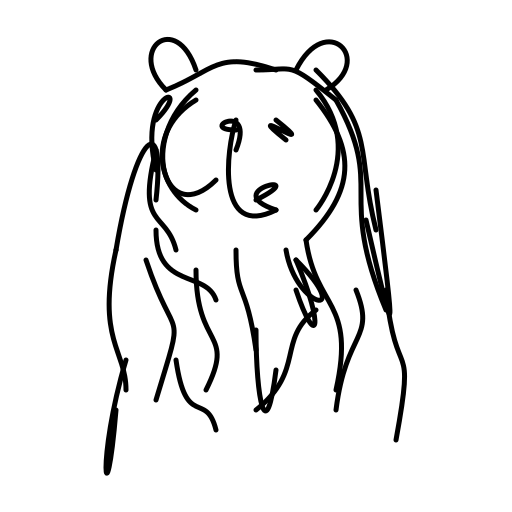} \\

\rotatebox{90}{\parbox{0.09\linewidth}{\centering\tiny Bicycle}} & \includegraphics[width=0.09\linewidth]{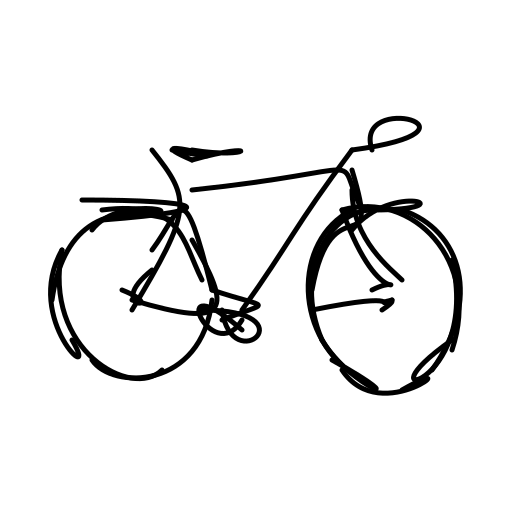} & \includegraphics[width=0.09\linewidth]{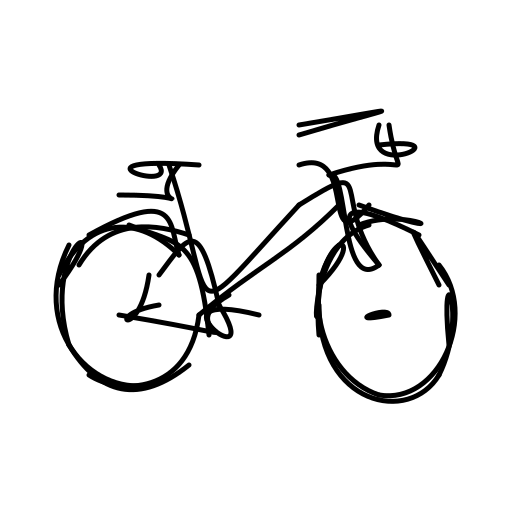} & \includegraphics[width=0.09\linewidth]{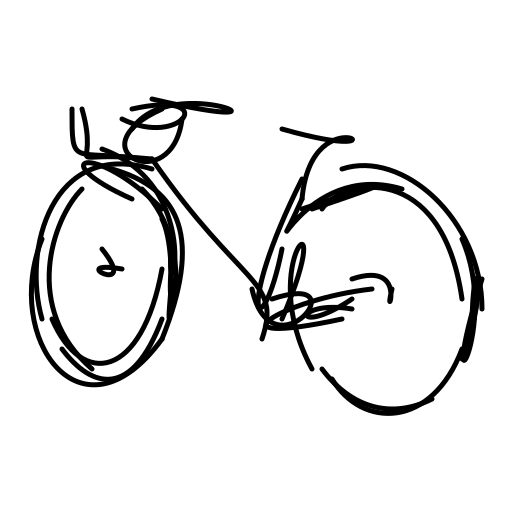} & \rotatebox{90}{\parbox{0.09\linewidth}{\centering\tiny Car}} & \includegraphics[width=0.09\linewidth]{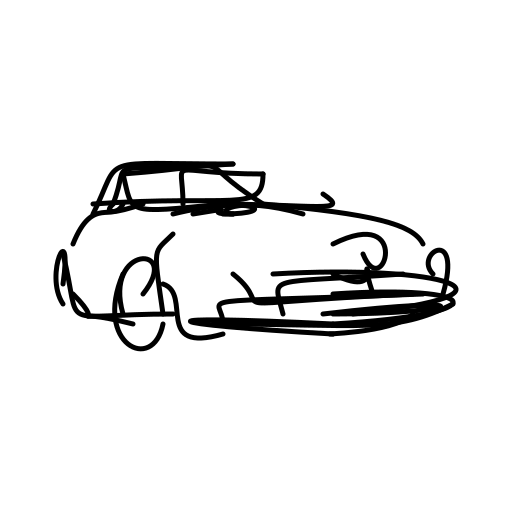} & \includegraphics[width=0.09\linewidth]{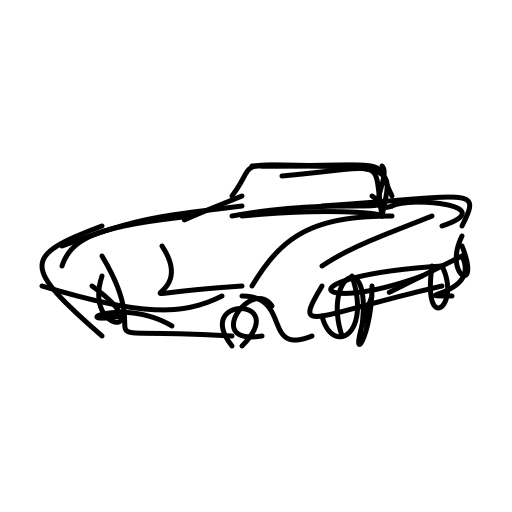} & \includegraphics[width=0.09\linewidth]{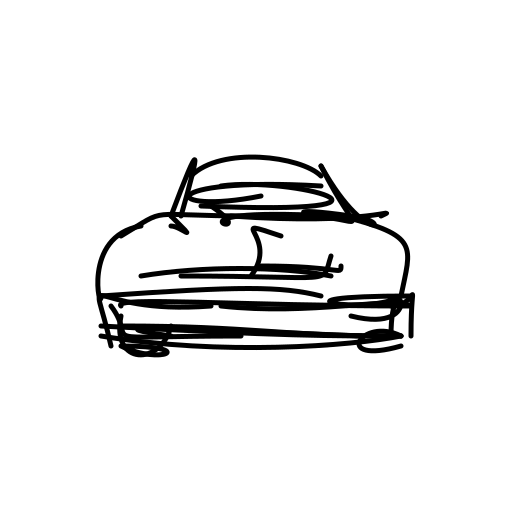} & \rotatebox{90}{\parbox{0.09\linewidth}{\centering\tiny Cat}} & \includegraphics[width=0.09\linewidth]{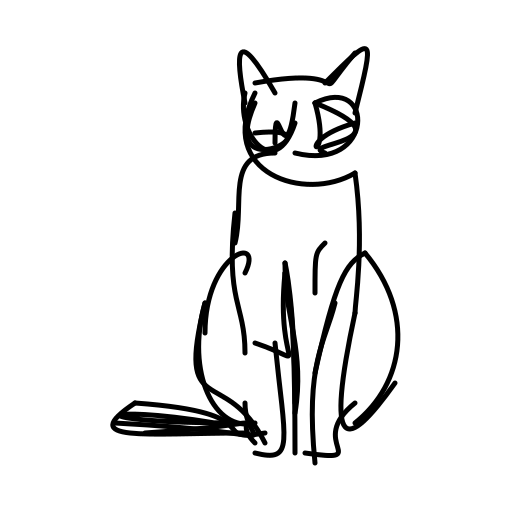} & \includegraphics[width=0.09\linewidth]{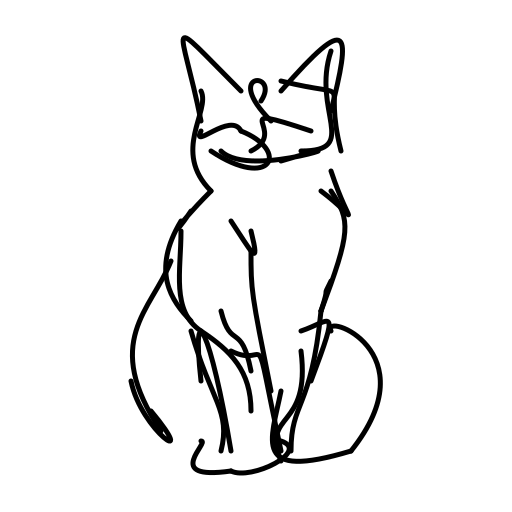} & \includegraphics[width=0.09\linewidth]{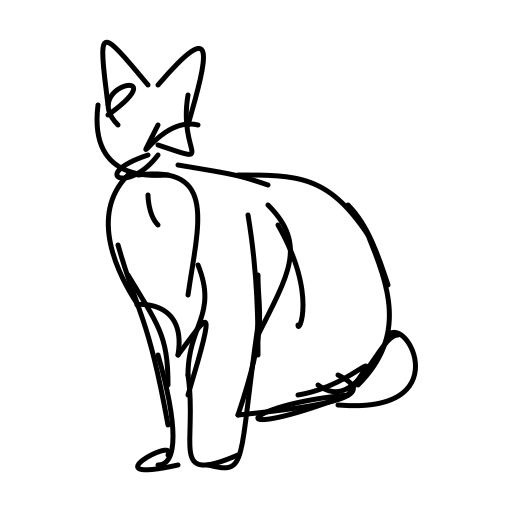} \\

\rotatebox{90}{\parbox{0.09\linewidth}{\centering\tiny Chair}} & \includegraphics[width=0.09\linewidth]{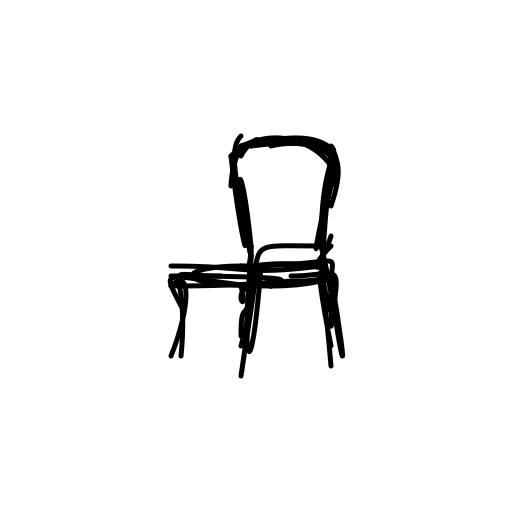} & \includegraphics[width=0.09\linewidth]{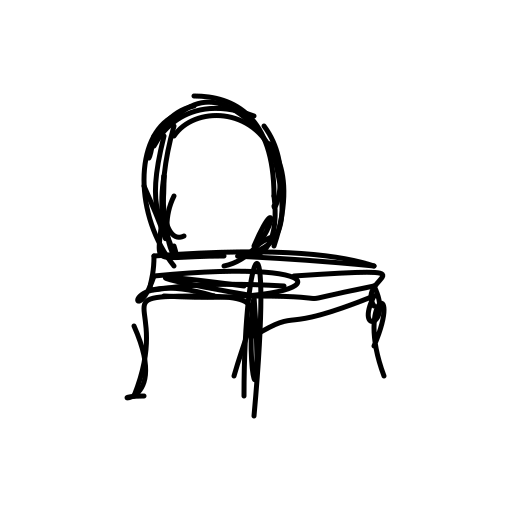} & \includegraphics[width=0.09\linewidth]{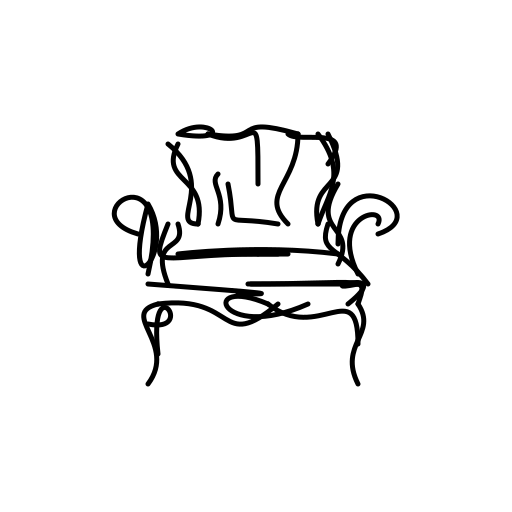} & \rotatebox{90}{\parbox{0.09\linewidth}{\centering\tiny Crab}} & \includegraphics[width=0.09\linewidth]{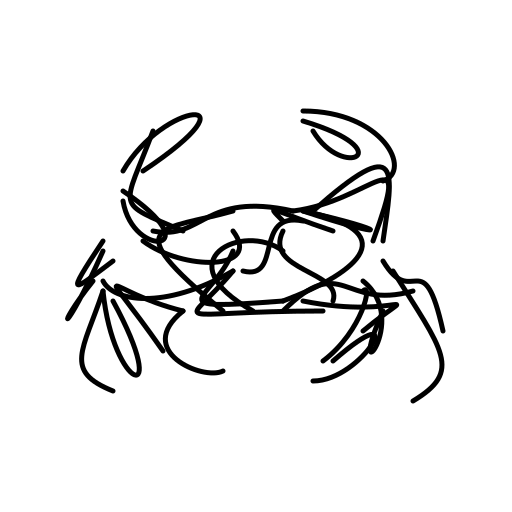} & \includegraphics[width=0.09\linewidth]{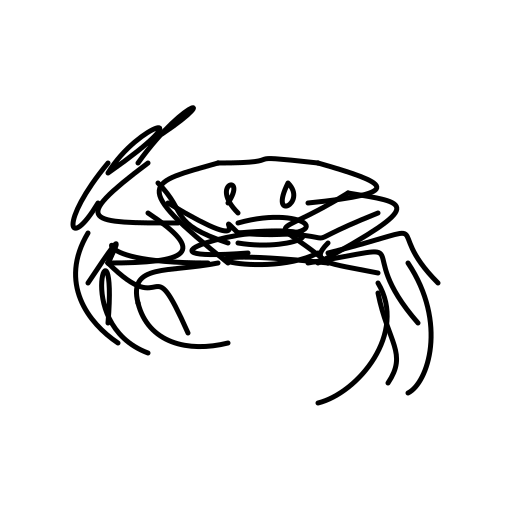} & \includegraphics[width=0.09\linewidth]{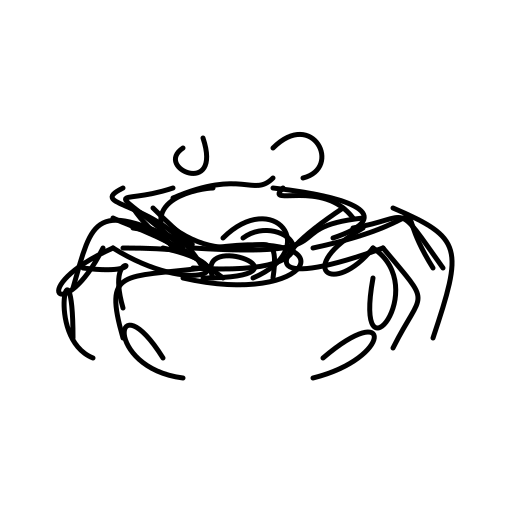} & \rotatebox{90}{\parbox{0.09\linewidth}{\centering\tiny Dog}} & \includegraphics[width=0.09\linewidth]{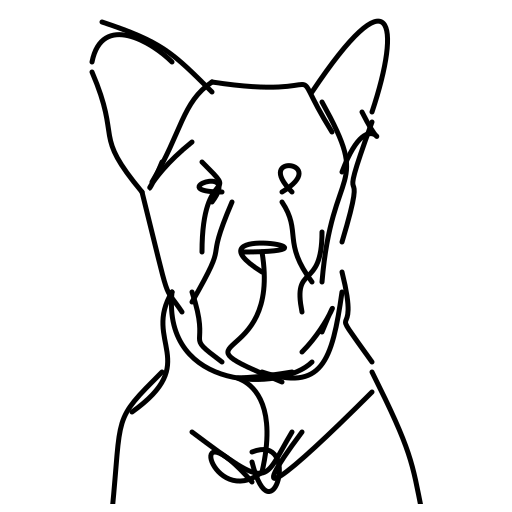} & \includegraphics[width=0.09\linewidth]{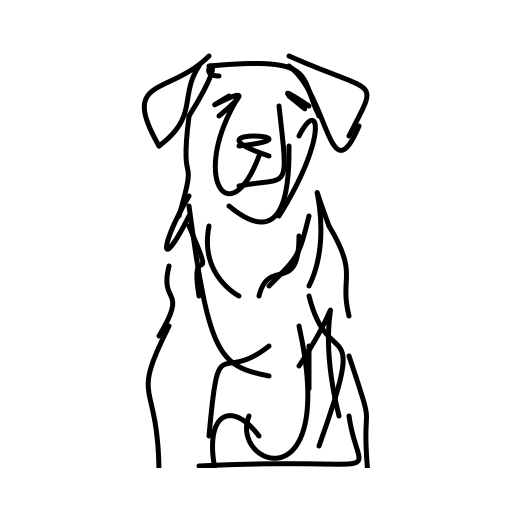} & \includegraphics[width=0.09\linewidth]{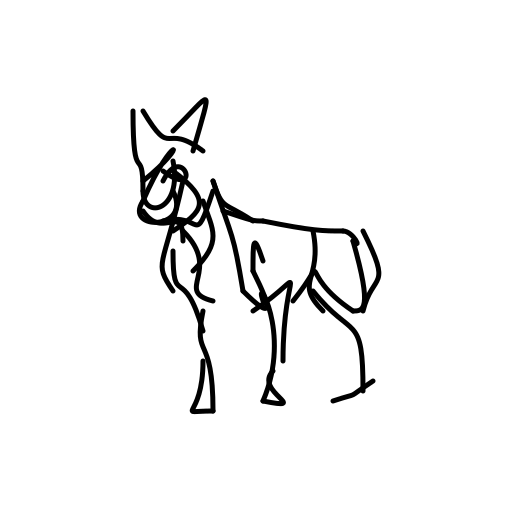} \\

\rotatebox{90}{\parbox{0.09\linewidth}{\centering\tiny Fish}} & \includegraphics[width=0.09\linewidth]{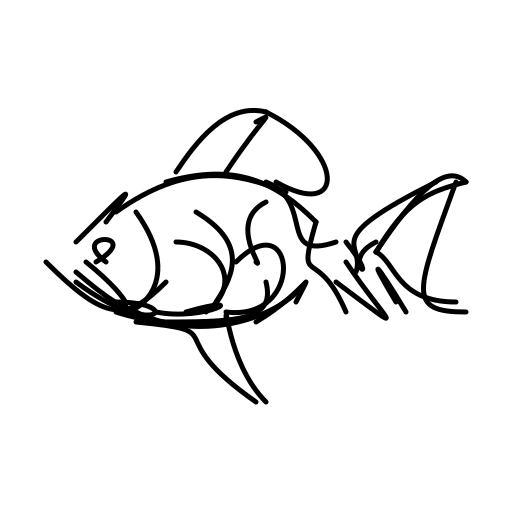} & \includegraphics[width=0.09\linewidth]{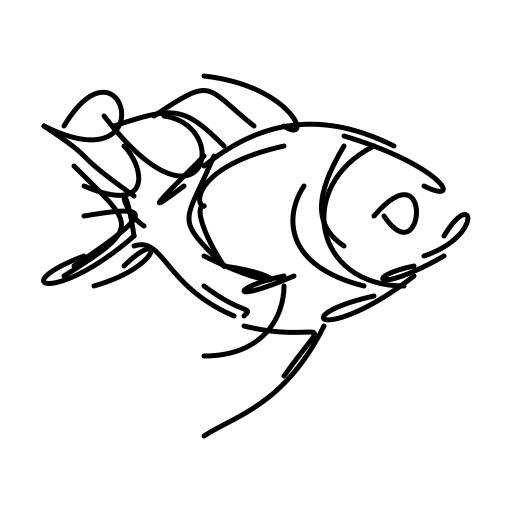} & \includegraphics[width=0.09\linewidth]{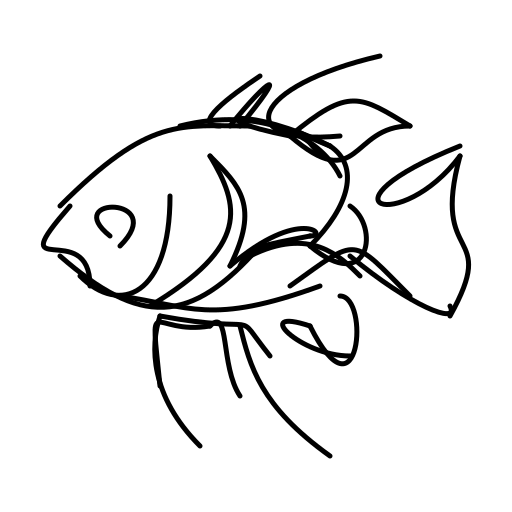} & \rotatebox{90}{\parbox{0.09\linewidth}{\centering\tiny Horse}} & \includegraphics[width=0.09\linewidth]{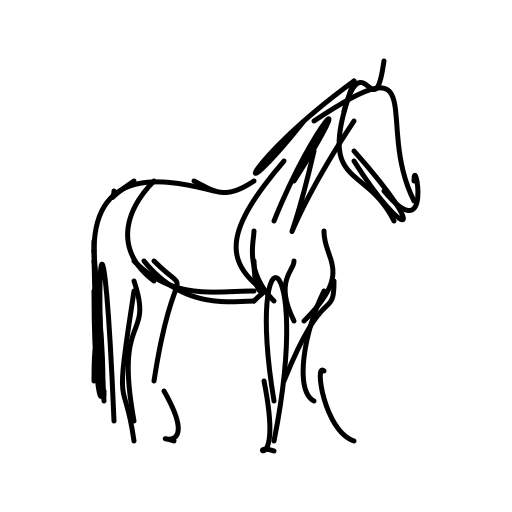} & \includegraphics[width=0.09\linewidth]{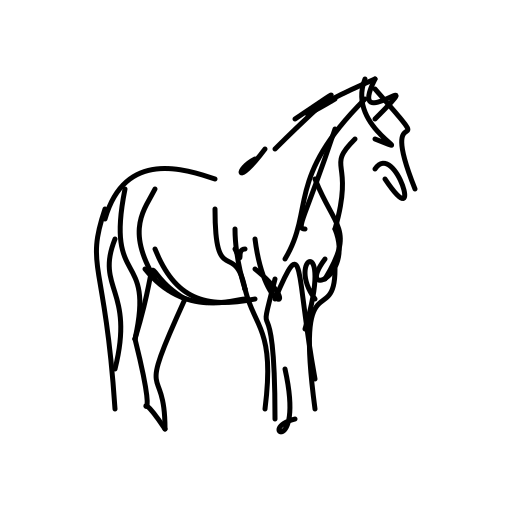} & \includegraphics[width=0.09\linewidth]{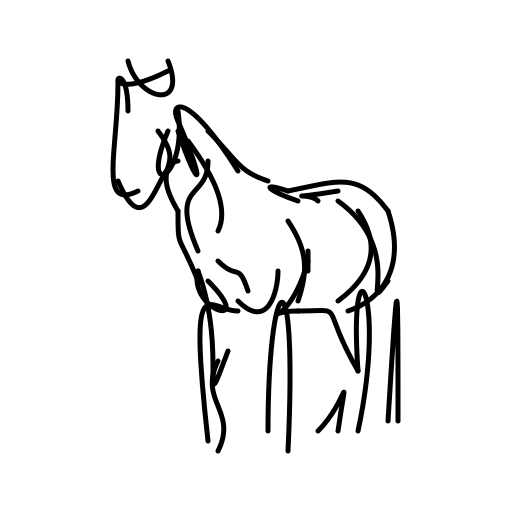} & \rotatebox{90}{\parbox{0.09\linewidth}{\centering\tiny Rabbit}} & \includegraphics[width=0.09\linewidth]{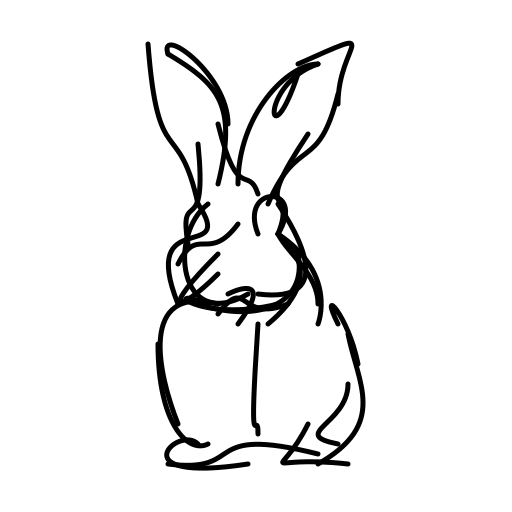} & \includegraphics[width=0.09\linewidth]{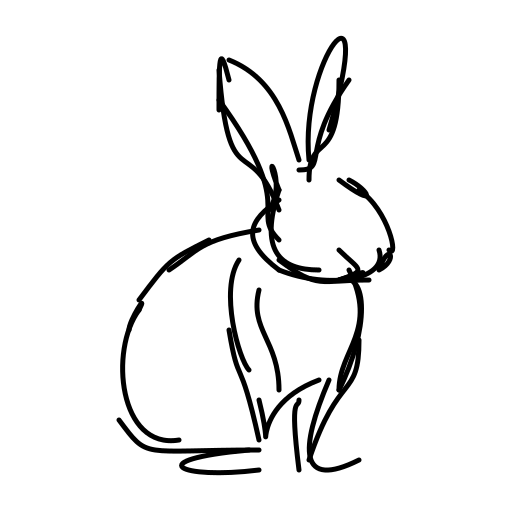} & \includegraphics[width=0.09\linewidth]{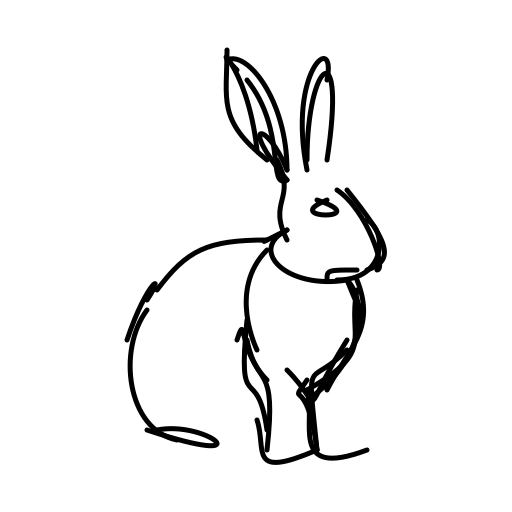} \\

\rotatebox{90}{\parbox{0.09\linewidth}{\centering\tiny Robot}} & \includegraphics[width=0.09\linewidth]{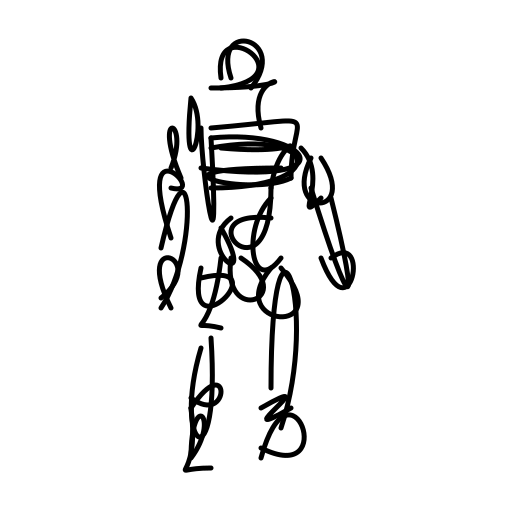} & \includegraphics[width=0.09\linewidth]{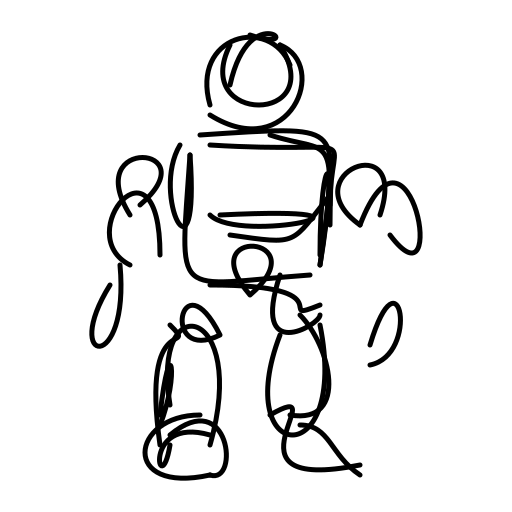} & \includegraphics[width=0.09\linewidth]{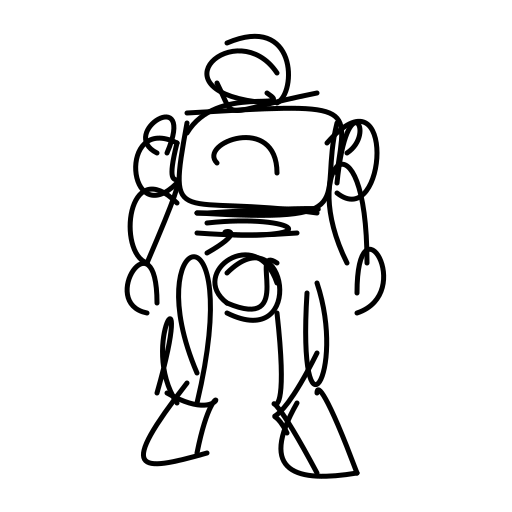} & \rotatebox{90}{\parbox{0.09\linewidth}{\centering\tiny Sculpture}} & \includegraphics[width=0.09\linewidth]{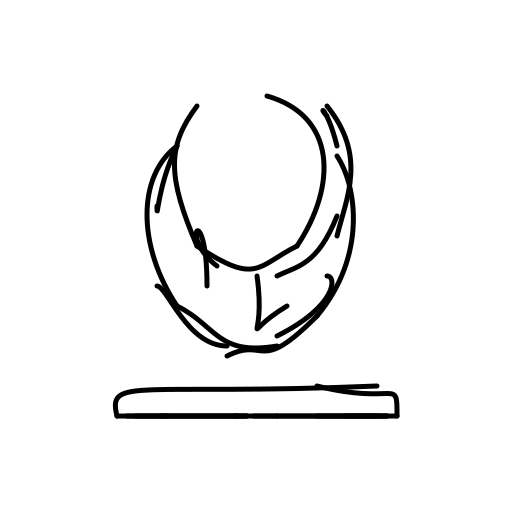} & \includegraphics[width=0.09\linewidth]{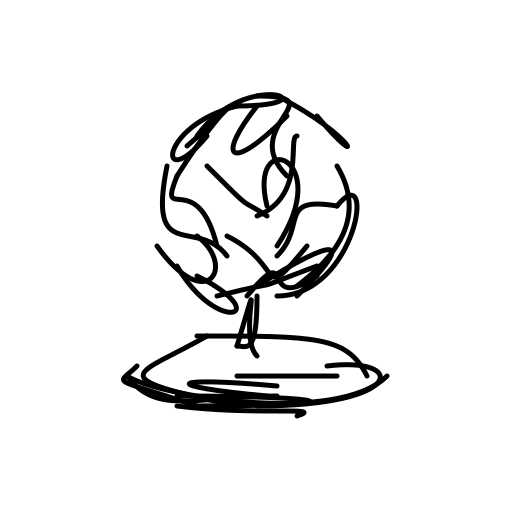} & \includegraphics[width=0.09\linewidth]{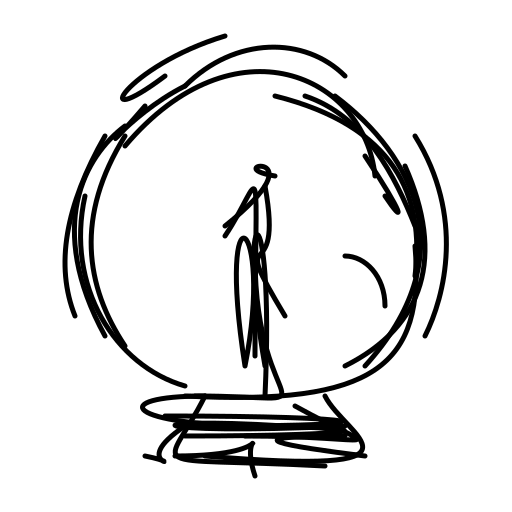} & \rotatebox{90}{\parbox{0.09\linewidth}{\centering\tiny Woman}} & \includegraphics[width=0.09\linewidth]{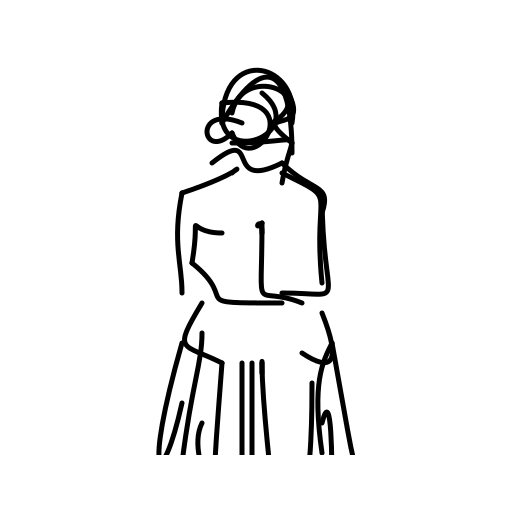} & \includegraphics[width=0.09\linewidth]{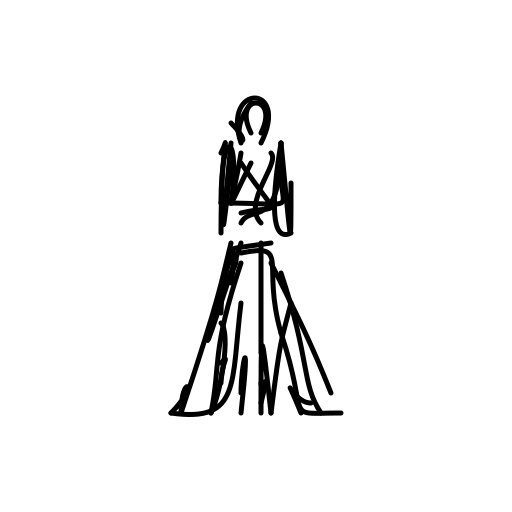} & \includegraphics[width=0.09\linewidth]{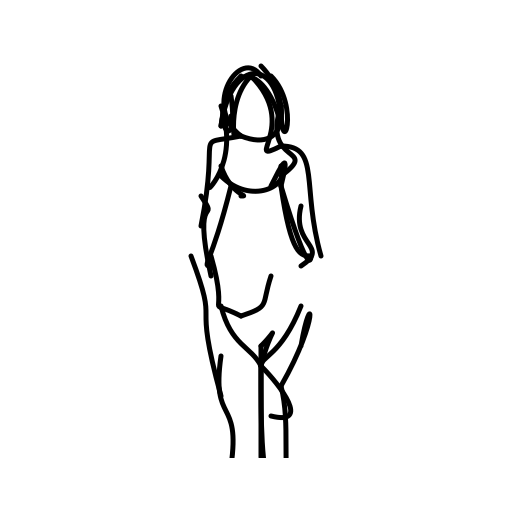} \\
\end{tabular}
\caption{Example outputs from Ours (SFT + RL) trained on ControlSketch-Part. Our model and training process do not rely on the class labels in any way, and we only show these for reference.}
\label{fig:ours_example_outputs}
\vspace{-10pt}
\end{figure}

\begin{figure}[t]
  \centering
  \includegraphics[width=0.99\linewidth]{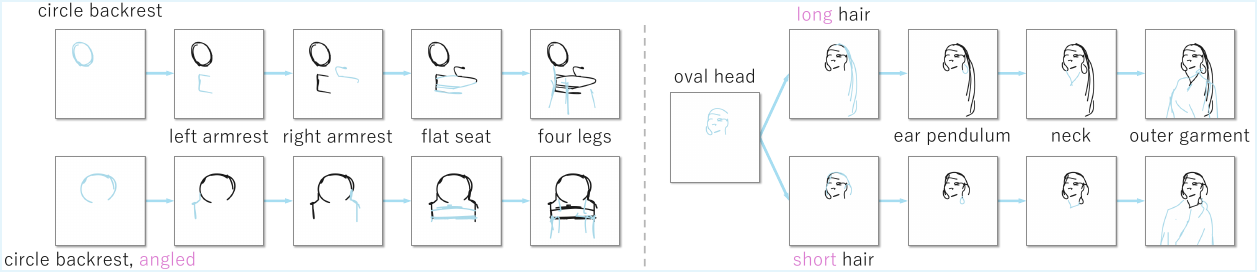}
  \caption{Additional progressive editing examples. Left: identical part descriptions with different initial canvasses lead to different outputs. Right: changing the description for an early part but keeping the subsequent part descriptions same produces two sketches with significant differences localized to the affected part.
}
  \label{fig:extra_edit}
\end{figure}

\noindent\textbf{Qualitative results} \cref{fig:example_comparison} shows our generated sketches alongside those of the baselines. Our sketches tend to contain smooth paths, and have a natural style with identifiable, meaningful parts. SketchAgent produces relatively clean part structure but favors simple geometric primitives and symmetric layouts, which limits visual quality. In a portion of outputs it produces misplaced or distorted components (e.g., car, dog, rabbit). Gemini 3.1 Pro shares a similar preference for simple geometries and symmetric layouts, and occasionally fails to produce a complete object (e.g., car). It also struggles to capture the distinguishing features of certain animals (e.g., bear, cat, and dog). SDXL + SwiftSketch can produce smooth, naturalistic sketches (e.g., bike and car), but is hindered by SDXL's lack of ability to adhere to long text inputs, which often causes it to miss details or misinterpret compositional relationships. SwiftSketch further degrades when the image generated by SDXL is low quality or lacks a clear foreground subject. Note that class labels are shown purely for reference and are not used by our model or training process in any way. We include more sketch examples of Ours(SFT + RL) in~\cref{fig:ours_example_outputs}. More progressive editing examples can be found in~\cref{fig:extra_edit}

\subsection{Ablation Study}
The performance gap (\cref{fig:long-clip-bar-chart}, \cref{fig:user_study}) between Ours (SFT + RL) and Ours (SFT) shows the benefits of RL training. We further investigate the usefulness of certain training strategies of our RL training. We ablate our multi-turn process-reward GRPO formulation against a single-turn GRPO baseline and a multi-turn outcome-reward GRPO baseline. The single-turn baseline treats the entire sketching process as a single completion with only a terminal reward on the final rendering. The multi-turn outcome-reward GRPO baseline uses the reward of the final rendering to compute advantages for all the steps, whereas our multi-turn process-reward GRPO setup uses intermediate-state rewards at each step, enabling dense credit assignment over the evolving canvas. 
We run the controlled ablation on Qwen2.5-VL-3B~\cite{bai2025qwen25vltechnicalreport}, and observed that multi-turn GRPO setups outperform single-turn GRPO while process-reward outperforms outcome-reward, as shown in Table~\ref{tab:one_turn_vs_multi_turn_rl}. The ablation study suggests that both multi-turn formulation and dense process-level rewards are important contributors to our model's final performance, each providing complementary benefits.

\begin{table}[t]
\begin{minipage}[c]{.3\textwidth}
\caption{Average Long-CLIP scores across different ablation configurations, using Qwen2.5-VL-3B.}
\label{tab:one_turn_vs_multi_turn_rl}
\vspace{-20pt}
\end{minipage}%
\begin{minipage}[c]{.7\textwidth}
\centering
\small
\setlength{\tabcolsep}{6pt}
\begin{tabular}{l c}
\toprule
Method & Long-CLIP $\uparrow$ \\
\midrule
Single-turn RL  & 0.281 \\
Multi-turn outcome-reward & 0.286 \\
Multi-turn process-reward (ours) & \textbf{0.298} \\
\bottomrule
\end{tabular}
\end{minipage}
\end{table}

\section{Conclusion}
\label{sec:conclusion}

We argue that part-level semantic annotation is a missing but critical ingredient for learning text-to-vector sketch generation. To close this gap, we developed a scalable automatic annotation pipeline and applied it to produce ControlSketch-Part, a dataset that enriches vector sketches with semantic part decompositions, per-part text descriptions, and path-to-part assignments. With this data in hand, we trained a VLM agent using a two-stage SFT+RL framework: SFT grounds the agent in output format and initializes the sketching policy for a single turn, while a novel multi-turn process-reward GRPO stage optimizes visual quality via intermediate visual rewards, closing the distribution gap between oracle training states and free-form inference. The resulting agent can generate structured sketches one part at a time. It outperforms prior methods across automatic metrics and user studies, and naturally supports localized editing operations (such as removal and replacement of strokes) with high visual quality. We expect that ControlSketch-Part and the proposed training framework will serve as useful resources for future research on structured multi-turn processes that benefit from visual feedback.

\FloatBarrier
\newpage
{
    \small
    \bibliographystyle{splncs04}
    \bibliography{main,rl_citations}

\begin{thebibliography}{10}
\providecommand{\url}[1]{\texttt{#1}}
\providecommand{\urlprefix}{URL }
\providecommand{\doi}[1]{https://doi.org/#1}

\bibitem{swiftsketch}
Arar, E., Frenkel, Y., Cohen-Or, D., Shamir, A., Vinker, Y.: Swiftsketch: A diffusion model for image-to-vector sketch generation. In: Proceedings of the Special Interest Group on Computer Graphics and Interactive Techniques Conference Conference Papers. SIGGRAPH Conference Papers '25, Association for Computing Machinery, New York, NY, USA (2025). \doi{10.1145/3721238.3730612}, \url{https://doi.org/10.1145/3721238.3730612}

\bibitem{Qwen3VL}
Bai, S., Cai, Y., Chen, R., Chen, K., Chen, X.H., Cheng, Z., Deng, L., Ding, W., Fang, R., Gao, C., Ge, C., Ge, W., Guo, Z., Huang, Q., Huang, Q., Huang, F., Hui, B., Jiang, S., Li, Z., Li, M., Li, M., Li, K., Lin, Z., Lin, J., Liu, X., Liu, J., Liu, C., Liu, Y., Liu, D., Liu, S., Lu, D., Luo, R., Lv, C., Men, R., Meng, L.Y., Ren, X., yi~Ren, X., Song, S., Sun, Y.C., Tang, J., Tu, J., Wan, J., Wang, P., Wang, P., Wang, Q., Wang, Y., Xie, T., Xu, Y., Xu, H., Xu, J., Yang, Z., Yang, M., Yang, J., Yang, A., Yu, B., Zhang, F., Zhang, H., Zhang, X., Zheng, B., Zhong, H., Zhou, J., Zhou, F., Zhou, J., Zhu, Y., Zhu, K.: Qwen3-vl technical report. ArXiv  \textbf{abs/2511.21631} (2025), \url{https://api.semanticscholar.org/CorpusID:283262018}

\bibitem{bai2025qwen25vltechnicalreport}
Bai, S., Chen, K., Liu, X., Wang, J., Ge, W., Song, S., Dang, K., Wang, P., Wang, S., Tang, J., Zhong, H., Zhu, Y., Yang, M., Li, Z., Wan, J., Wang, P., Ding, W., Fu, Z., Xu, Y., Ye, J., Zhang, X., Xie, T., Cheng, Z., Zhang, H., Yang, Z., Xu, H., Lin, J.: Qwen2.5-vl technical report (2025), \url{https://arxiv.org/abs/2502.13923}

\bibitem{dino}
Caron, M., Touvron, H., Misra, I., J\'egou, H., Mairal, J., Bojanowski, P., Joulin, A.: Emerging properties in self-supervised vision transformers. In: Proceedings of the International Conference on Computer Vision (ICCV) (2021)

\bibitem{chen2025sftrlearlyinvestigation}
Chen, H., Tu, H., Wang, F., Liu, H., Tang, X., Du, X., Zhou, Y., Xie, C.: Sft or rl? an early investigation into training r1-like reasoning large vision-language models (2025), \url{https://arxiv.org/abs/2504.11468}

\bibitem{autosketch}
Chin, H.Y., Shen, I.C., Chiu, Y.T., Shamir, A., Chen, B.Y.: Autosketch: Vlm-assisted style-aware vector sketch completion. In: Proceedings of the SIGGRAPH Asia 2025 Conference Papers. pp. 1--11 (2025)

\bibitem{beziersketch}
Das, A., Yang, Y., Hospedales, T., Xiang, T., Song, Y.Z.: B{\'e}ziersketch: A generative model for scalable vector sketches. In: European conference on computer vision. pp. 632--647. Springer (2020)

\bibitem{sketchode}
Das, A., Yang, Y., Hospedales, T., Xiang, T., Song, Y.Z.: Sketch{ODE}: Learning neural sketch representation in continuous time. In: International Conference on Learning Representations (2022), \url{https://openreview.net/forum?id=c-4HSDAWua5}

\bibitem{chirodiff}
Das, A., Yang, Y., Hospedales, T., Xiang, T., Song, Y.Z.: Chirodiff: Modelling chirographic data with diffusion models. In: The Eleventh International Conference on Learning Representations (2023), \url{https://openreview.net/forum?id=1ROAstc9jv}

\bibitem{clipdraw}
Frans, K., Soros, L., Witkowski, O.: Clipdraw: Exploring text-to-drawing synthesis through language-image encoders. Advances in Neural Information Processing Systems  \textbf{35},  5207--5218 (2022)

\bibitem{dreamsim}
Fu, S., Tamir, N., Sundaram, S., Chai, L., Zhang, R., Dekel, T., Isola, P.: Dreamsim: Learning new dimensions of human visual similarity using synthetic data. In: Advances in Neural Information Processing Systems. vol.~36, pp. 50742--50768 (2023)

\bibitem{geng2025jsonschemabench}
Geng, S., Cooper, H., Moskal, M., Jenkins, S., Berman, J., Ranchin, N., West, R., Horvitz, E., Nori, H.: Jsonschemabench: A rigorous benchmark of structured outputs for language models. arXiv preprint arXiv:2501.10868  (2025)

\bibitem{sketchrnn}
Ha, D., Eck, D.: A neural representation of sketch drawings. In: International Conference on Learning Representations (2018)

\bibitem{harada2025curse}
Harada, K., Yamazaki, Y., Taniguchi, M., Kojima, T., Iwasawa, Y., Matsuo, Y.: Curse of instructions: Large language models cannot follow multiple instructions at once. OpenReview  (2024), \url{https://openreview.net/pdf?id=R6q67CDBCH}

\bibitem{ho2020denoising}
Ho, J., Jain, A., Abbeel, P.: Denoising diffusion probabilistic models. Advances in neural information processing systems  \textbf{33},  6840--6851 (2020)

\bibitem{hoffmann2022chinchilla}
Hoffmann, J., Borgeaud, S., Mensch, A., Buchatskaya, E., Cai, T., Rutherford, E., de~Las~Casas, D., Hendricks, L.A., Welbl, J., Clark, A., et~al.: Training compute-optimal large language models. In: Proceedings of the 36th International Conference on Neural Information Processing Systems. pp. 30016--30030 (2022)

\bibitem{lora}
Hu, E.J., Shen, Y., Wallis, P., Allen-Zhu, Z., Li, Y., Wang, S., Wang, L., Chen, W., et~al.: Lora: Low-rank adaptation of large language models. ICLR  \textbf{1}(2), ~3 (2022)

\bibitem{kang2025quagmiressftrlposttraininghigh}
Kang, F., Kuchnik, M., Padthe, K., Vlastelica, M., Jia, R., Wu, C.J., Ardalani, N.: Quagmires in sft-rl post-training: When high sft scores mislead and what to use instead (2025), \url{https://arxiv.org/abs/2510.01624}

\bibitem{cairosvg}
Kozea: Cairosvg: Convert your svg files to pdf and png (https://cairosvg.org/) (2025), \url{https://cairosvg.org/}

\bibitem{liu2024lost}
Liu, N.F., Lin, K., Hewitt, J., Paranjape, A., Bevilacqua, M., Petroni, F., Liang, P.: Lost in the middle: How language models use long contexts. Transactions of the association for computational linguistics  \textbf{12},  157--173 (2024)

\bibitem{dr.grpo}
Liu, Z., Chen, C., Li, W., Qi, P., Pang, T., Du, C., Lee, W.S., Lin, M.: Understanding r1-zero-like training: A critical perspective. arXiv preprint arXiv:2503.20783  (2025)

\bibitem{ouyang2022traininglanguagemodelsfollow}
Ouyang, L., Wu, J., Jiang, X., Almeida, D., Wainwright, C.L., Mishkin, P., Zhang, C., Agarwal, S., Slama, K., Ray, A., Schulman, J., Hilton, J., Kelton, F., Miller, L., Simens, M., Askell, A., Welinder, P., Christiano, P., Leike, J., Lowe, R.: Training language models to follow instructions with human feedback (2022), \url{https://arxiv.org/abs/2203.02155}

\bibitem{sdxl}
Podell, D., English, Z., Lacey, K., Blattmann, A., Dockhorn, T., M{\"u}ller, J., Penna, J., Rombach, R.: Sdxl: Improving latent diffusion models for high-resolution image synthesis. In: The Twelfth International Conference on Learning Representations (2024)

\bibitem{poole2022dreamfusion}
Poole, B., Jain, A., Barron, J.T., Mildenhall, B.: Dreamfusion: Text-to-3d using 2d diffusion. arXiv preprint arXiv:2209.14988  (2022)

\bibitem{sketchdreamer}
Qu, Z., Xiang, T., Song, Y.Z.: Sketchdreamer: Interactive text-augmented creative sketch ideation. arXiv preprint arXiv:2308.14191  (2023)

\bibitem{clip}
Radford, A., Kim, J.W., Hallacy, C., Ramesh, A., Goh, G., Agarwal, S., Sastry, G., Askell, A., Mishkin, P., Clark, J., et~al.: Learning transferable visual models from natural language supervision. In: International conference on machine learning. pp. 8748--8763. PmLR (2021)

\bibitem{rodriguez2025renderingawarereinforcementlearningvector}
Rodriguez, J.A., Zhang, H., Puri, A., Feizi, A., Pramanik, R., Wichmann, P., Mondal, A., Samsami, M.R., Awal, R., Taslakian, P., Gella, S., Rajeswar, S., Vazquez, D., Pal, C., Pedersoli, M.: Rendering-aware reinforcement learning for vector graphics generation (2025), \url{https://arxiv.org/abs/2505.20793}

\bibitem{kl_approx}
Schulman, J.: Approximating kl divergence (2020), \url{http://joschu.net/blog/kl-approx.html}

\bibitem{ppo}
Schulman, J., Wolski, F., Dhariwal, P., Radford, A., Klimov, O.: Proximal policy optimization algorithms. arXiv preprint arXiv:1707.06347  (2017)

\bibitem{deepseekmath}
Shao, Z., Wang, P., Zhu, Q., Xu, R., Song, J., Bi, X., Zhang, H., Zhang, M., Li, Y., Wu, Y., et~al.: Deepseekmath: Pushing the limits of mathematical reasoning in open language models. arXiv preprint arXiv:2402.03300  (2024)

\bibitem{song2020score}
Song, Y., Sohl-Dickstein, J., Kingma, D.P., Kumar, A., Ermon, S., Poole, B.: Score-based generative modeling through stochastic differential equations. arXiv preprint arXiv:2011.13456  (2020)

\bibitem{tinker}
{Thinking Machines Lab}: Tinker (2025), \url{https://thinkingmachines.ai/tinker/}

\bibitem{villalobos2024position}
Villalobos, P., Ho, A., Sevilla, J., Besiroglu, T., Heim, L., Hobbhahn, M.: Position: Will we run out of data? limits of llm scaling based on human-generated data. In: Forty-first International Conference on Machine Learning (2024)

\bibitem{Vinker_2023_ICCV}
Vinker, Y., Alaluf, Y., Cohen-Or, D., Shamir, A.: Clipascene: Scene sketching with different types and levels of abstraction. In: Proceedings of the IEEE/CVF International Conference on Computer Vision (ICCV). pp. 4146--4156 (October 2023)

\bibitem{vinker2022clipasso}
Vinker, Y., Pajouheshgar, E., Bo, J.Y., Bachmann, R.C., Bermano, A.H., Cohen-Or, D., Zamir, A., Shamir, A.: Clipasso: Semantically-aware object sketching. ACM Trans. Graph.  \textbf{41}(4) (2022), \url{https://doi.org/10.1145/3528223.3530068}

\bibitem{sketchagent}
Vinker, Y., Shaham, T.R., Zheng, K., Zhao, A., Fan, J.E., Torralba, A.: Sketchagent: Language-driven sequential sketch generation (2024), \url{https://arxiv.org/abs/2411.17673}

\bibitem{wang2025kiminaproverpreviewlargeformal}
Wang, H., Unsal, M., Lin, X., Baksys, M., Liu, J., Santos, M.D., Sung, F., Vinyes, M., Ying, Z., Zhu, Z., Lu, J., de~Saxc{\'e}, H., Bailey, B., Song, C., Xiao, C., Zhang, D., Zhang, E., Pu, F., Zhu, H., Liu, J., Bayer, J., Michel, J., Yu, L., Dreyfus-Schmidt, L., Tunstall, L., Pagani, L., Machado, M., Bourigault, P., Wang, R., Polu, S., Barroyer, T., Li, W.D., Niu, Y., Fleureau, Y., Hu, Y., Yu, Z., Wang, Z., Yang, Z., Liu, Z., Li, J.: Kimina-prover preview: Towards large formal reasoning models with reinforcement learning (2025), \url{https://arxiv.org/abs/2504.11354}

\bibitem{xing2025reasonsvghybridrewardrl}
Xing, X., Guan, Y., Zhang, J., Xu, D., Yu, Q.: Reason-svg: Hybrid reward rl for aha-moments in vector graphics generation (2025), \url{https://arxiv.org/abs/2505.24499}

\bibitem{diffsketcher}
Xing, X., Wang, C., Zhou, H., Zhang, J., Yu, Q., Xu, D.: Diffsketcher: Text guided vector sketch synthesis through latent diffusion models. In: Thirty-seventh Conference on Neural Information Processing Systems (2023), \url{https://openreview.net/forum?id=CY1xatvEQj}

\bibitem{zhang2024longclip}
Zhang, B., Zhang, P., Dong, X., Zang, Y., Wang, J.: Long-clip: Unlocking the long-text capability of clip. In: European conference on computer vision. pp. 310--325. Springer (2024)

\bibitem{lpips}
Zhang, R., Isola, P., Efros, A.A., Shechtman, E., Wang, O.: The unreasonable effectiveness of deep features as a perceptual metric. In: CVPR (2018)

\bibitem{zhou2025strokefusion}
Zhou, J., Zhou, Y., Yang, H., Xu, P., Huang, H.: Strokefusion: Vector sketch generation via joint stroke-udf encoding and latent sequence diffusion. arXiv preprint arXiv:2503.23752  (2025)

\end{thebibliography}
}
\clearpage
\setcounter{page}{1}
\maketitlesupplementary

\cref{sec:supp-pseudo-code} presents the pseudo code for our multi-turn process-reward GRPO training. \cref{sec:supp-prompt-templates} documents the prompt templates for our automatic data annotation pipeline. \cref{sec:supp-additional-results} shows additional part-by-part sketching results of our model. \cref{sec:supp-additional-dataset-examples} provides further examples from our ControlSketch-Part dataset. In~\cref{sec:supp-limitations}, we present failure cases and discuss limitations and future work.

\section{Pseudo Code}
\label{sec:supp-pseudo-code}
\begin{algorithm}[ht]
  \small
  \caption{Multi-turn Process-reward GRPO}
  \textbf{Input} initial policy model $\pi_{\theta_{\text{init}}}$; reward models $r_\phi$; task prompts $\mathcal{D}$; 
  hyperparameters $\epsilon$, $\beta$, $\mu$
  \begin{algorithmic}[1]
    \State policy model $\pi_\theta \leftarrow \pi_{\theta_{\text{init}}}$
    \For{iteration = 1, \dots, I}
       \State reference model $\pi_{ref} \leftarrow \pi_{\theta}$
      \For{step = 1, \dots, M}
      \State Sample a batch $\mathcal{D}_b$ from $\mathcal{D}$
      \State Update the old policy model $\pi_{\theta_{old}} \leftarrow \pi_{\theta}$ 
      \State Sample $G$ trajectories $\left\{\{o_g^t\}_{t=1}^T\right\}^G_{g=1} \sim \pi_{\theta_{old}} (\cdot \mid q) $ for each question $q \in \mathcal{D}_b$
      \State Compute rewards $\left\{\{r_g^t\}_{t=1}^T\right\}^G_{g=1}$ for each sampled output $o_g^t$ by running $r_{\phi}$ 
      \State Compute $\hat{A}_{g,k}^t$ for the $k$-th token of $o_g^t$ through group relative advantage estimation.
      \For{GRPO iteration = 1, \dots, $\mu$}
        \State Update the policy model $\pi_{\theta}$ by maximizing the GRPO objective (\cref{eq:GRPO-obj})
      \EndFor
    \EndFor 
    \EndFor 
  \end{algorithmic}
  \textbf{Output} $\pi_\theta$
  \label{alg:iter-grpo}
\end{algorithm}

\section{Prompt Templates for the Data Collection Pipeline}
\label{sec:supp-prompt-templates}

This section documents the seven prompt templates used in the automatic annotation pipeline described in the main paper. 

\promptmeta{Placeholders}{
\begin{itemize}[label=\textbullet]
\item \placeholder{min\_parts} and \placeholder{max\_parts} are the minimum and maximum number of parts the agent can return, respectively.
\item \placeholder{rendering} represents the rendered image of the sketch.
\item \placeholder{diagnostic\_vis} represents the rendered image of the diagnostic visualization.
\item \placeholder{svg\_text} denotes the raw SVG source code,
\item \placeholder{joined\_parts} expands to one line per part in the form \texttt{Part1: ...}, \texttt{Part2: ...},
\item \placeholder{old\_parts\_json} is the previous part decomposition results.
\item \placeholder{old\_assignments\_json} is the previous path assignment results.
\item \placeholder{critique\_json} is the critique output from the previous step.
\item \placeholder{stepx\_instruction} is the complete instruction of step x.
\end{itemize}
}

\subsection*{Step 1: Initial Part Decomposition}

\begin{promptlisting}{Prompt}
<rendering>
You are given a black-and-white sketch image.
By examining the inputs, propose a set of parts that can effectively decompose the object into meaningful components.
1. Describe all visible details exhaustively in each part, with concise language. The set of parts must be collectively exhaustive and complementary.
2. When appropriate, prefer a finer-grained decomposition into meaningful parts, but do not split a single coherent part artificially.
3. Avoid high-level part such as "a dog" or "a woman".
4. Avoid semantically meaningless part such as "two lines" or "a curve".
5. You cannot have more than one parts describing the same or overlapping component of the object. I.e. **Information about the same part MUST be a single part**.
6. Do not explicitly mention strokes, lines, dots, or marks as such, but rather what they represent in the real world.
7. Do not describe drawing marks (e.g., "lines indicating legs" or "strokes forming a wheel"); name the actual object parts directly (e.g., "legs", "wheel").
8. Ignore isolated, clearly unintended marks or strokes that do not contribute to the main object structure.
9. Do not mention colors, medium, art/style/linework, lighting/composition/camera, emotions, intent, or subjective qualities.
10. Use specific, concrete part names (e.g., "expanded wings", "long tail") and avoid vague descriptions such as "expanded structure" or "long object".
11. Do not merge two clearly separate structures into one part for brevity.
12. Be specific about quantities when clearly visible; use exact numbers (e.g., "four legs") instead of generic terms like "legs".
13. Include details about the object's orientation, posture and motion when they are clearly depicted and visually distinctive. Note: "facing left/right" should mean facing the viewer's left/right, not the object's own left/right.
The number of parts should be between <min_parts> and <max_parts>, inclusive.
Provide your output as a JSON array of strings, each string being one part description.
Only return the JSON array, nothing else.
\end{promptlisting}

\begin{promptlisting}{Response Schema}
{
    "type": "array",
    "items": {"type": "string"},
    "minItems": <min_parts>,
    "maxItems": <max_parts>
}
\end{promptlisting}

\subsection*{Step 2: Part Critique}

\begin{promptlisting}{Prompt}
<rendering>
You are auditing a previous decomposition answer.

Original task instruction:
<step1_instruction>

Previous answer (JSON array of parts):
<old_parts_json>

You are also provided with the original sketch image.
Please closely read the original task instruction and check whether the previous answer follows each numbered requirement in that instruction, one by one.
For every violation, add an issue that explicitly references the violated requirement number(s), explains why it is violated, and suggests a concrete fix.
If a requirement is satisfied, do not add an issue for it.
If you believe any part is not correctly described, also add an issue for it.
Focus on strict requirement compliance rather than style preference.
Return ONLY JSON matching schema.
\end{promptlisting}

\begin{promptlisting}{Response Schema}
{
    "type": "object",
    "properties": {
        "issues": {
            "type": "array",
            "items": {
                "type": "object",
                "properties": {
                    "type": {"type": "string"},
                    "severity": {"type": "string", "enum": ["low", "medium", "high"]},
                    "reason": {"type": "string"},
                    "suggested_fix": {"type": "string"},
                },
                "required": ["type", "reason"],
            },
        },
        "summary": {"type": "string"},
        "should_revise": {"type": "boolean"},
    },
    "required": ["issues", "summary", "should_revise"]
}
\end{promptlisting}

\subsection*{Step 3: Part Refinement}

\begin{promptlisting}{Prompt}
<rendering>
Revise the previous part decomposition using the critique.

Previous answer:
<old_parts_json>

Critique JSON:
<critique_json>

You are also provided with the original sketch image.
Revision rules:
- If current parts are already good, keep them unchanged.
- Otherwise edit to fix errors.
- Output <min_parts> to <max_parts> non-overlapping semantic parts.
- Use concise but descriptive phrases.
Return ONLY JSON matching schema.
\end{promptlisting}

\begin{promptlisting}{Response Schema}
{
    "type": "array",
    "items": {"type": "string"},
    "minItems": <min_parts>,
    "maxItems": <max_parts>
}
\end{promptlisting}

\subsection*{Step 4: Initial Path Assignment}
Assume that the current sketch contains K paths.
\begin{promptlisting}{Prompt}
<rendering>
Here is the svg file of a sketch.
<svg_text>
You are also provided with the rendering image of this svg.
The image contains K paths. By examining this svg code and its corresponding rendered raster image, assign each path to one of the parts provided below.
<joined_parts>
1. Return your answer in JSON format with keys Path1, Path2, ..., PathK and values being the part label (e.g., Part1).
2. Use only the provided part labels.
3. Each path must be assigned to exactly one part.
4. All K paths must be assigned, and every provided part must be used at least once.
\end{promptlisting}

\begin{promptlisting}{Response Schema}
{
    "type": "object",
    "properties": {f"Path{i}": {"type": "string", "enum": [f"Part{i}" for i in range(1,num_parts+1)]} for i in range(1, K+1)},
    "required": [f"Path{i}" for i in range(1, K+1)],
}
\end{promptlisting}

\subsection*{Step 5: Path Assignment Critique with Diagnostic Visualization}

\begin{promptlisting}{Prompt}
<rendering>
<diagnostic_vis>
You are auditing a previous path-to-part assignment for an SVG sketch.

Original assignment task prompt:
<step4_instruction>

Previous assignment JSON:
<old_assignments_json>

Inputs provided:
- Original sketch rendering image.
- A color-coded paired image where left panel is part descriptions and right panel is the sketch with paths colored by assigned part.

Critique rules:
1. Check compliance against each numbered requirement in the task prompt.
2. Verify semantic correctness between part descriptions and assigned colored paths.
3. For each part, reason about whether there are any paths incorrectly assigned or missing.
4. For each issue, give concrete fix suggestions (what paths should move and why).
5. If no problems are found, return empty issues and should_revise=false.
Return ONLY JSON matching the schema.
\end{promptlisting}

\begin{promptlisting}{Response Schema}
{
    "type": "object",
    "properties": {
        "issues": {
            "type": "array",
            "items": {
                "type": "object",
                "properties": {
                    "type": {"type": "string"},
                    "severity": {"type": "string", "enum": ["low", "medium", "high"]},
                    "reason": {"type": "string"},
                    "suggested_fix": {"type": "string"},
                },
                "required": ["type", "reason"],
            },
        },
        "summary": {"type": "string"},
        "should_revise": {"type": "boolean"},
    },
    "required": ["issues", "summary", "should_revise"],
}
\end{promptlisting}

\subsection*{Step 6: Path Assignment Refinement}

\begin{promptlisting}{Prompt}
<rendering>
Revise the previous path-to-part assignment.

Original assignment task prompt:
<step4_instruction>

Previous assignment JSON:
<old_assignments_json>

Critique JSON:
<critique_json>

You are also provided with the original sketch image.
Revision rules:
- Follow every requirement in the original task prompt.
- If critique indicates no issue, keep assignment unchanged.
- Otherwise apply minimal but sufficient edits.
- Output only a JSON object with keys Path1..PathK and values among allowed Part labels.
Return ONLY the JSON object.
\end{promptlisting}

\begin{promptlisting}{Response Schema}
{
    "type": "object",
    "properties": {f"Path{i}": {"type": "string", "enum": [f"Part{i}" for i in range(1,num_parts+1)]} for i in range(1, K+1)},
    "required": [f"Path{i}" for i in range(1, K+1)],
}
\end{promptlisting}

\subsection*{Step 7: Caption Generation}

\begin{promptlisting}{Prompt}
You are given a black-and-white sketch image.
You are also provided with candidate object parts for your reference:
<joined_parts>
Write a short, strictly objective and literal caption describing the depicted objects.
1. Interpret visible marks, lines, and shapes as real-world object features, not as artistic or drawing elements.
2. Do not refer to the image as a sketch, drawing, or artwork.
3. Do not mention colors, materials, artistic style, linework, lighting, composition, camera, emotions, intent, or subjective qualities.
4. Do not add inferred, speculative, or imaginative details beyond what is directly visible.
5. Ignore isolated, clearly unintended marks or strokes that do not contribute to the main object structure.
6. Include only essential, clearly visible, and iconic information.
7. Focus exclusively on the visual content of the image.
8. Include details about the object's orientation, posture and motion when they are clearly depicted and visually distinctive. Note: "facing left/right" should mean facing the viewer's left/right, not the object's own left/right.
9. Limit the caption to 25 words or fewer.
\end{promptlisting}

\clearpage

\section{Additional Part-by-Part Results}
\vspace{-20pt}
\label{sec:supp-additional-results}
\FloatBarrier
\begin{figure}[H]
\centering
\begin{tabular}{c}
    \includegraphics[width=1.0\textwidth]{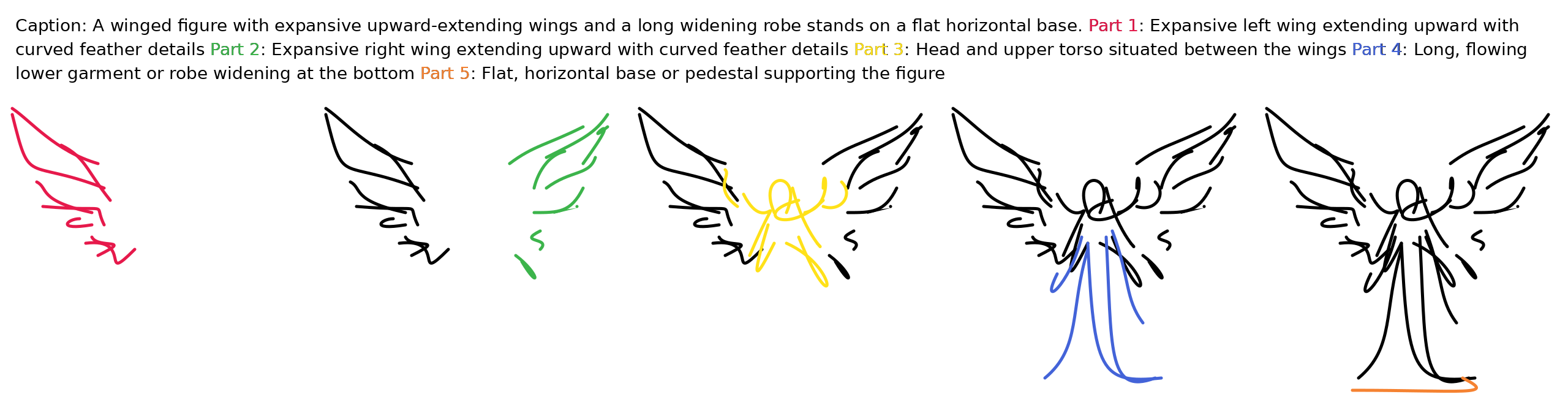} \\
    \includegraphics[width=1.0\textwidth]{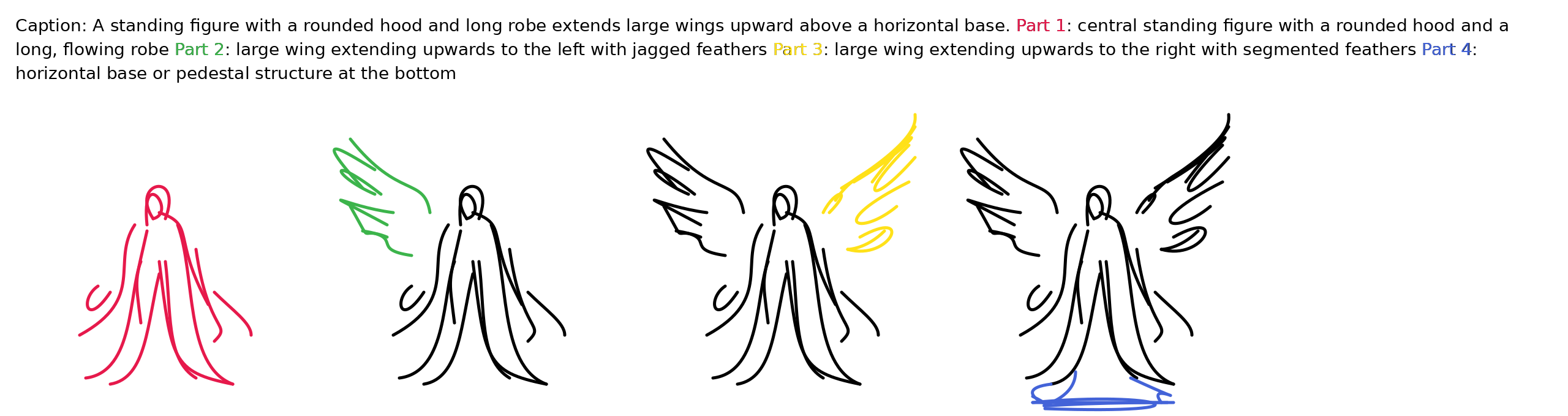} \\
    \includegraphics[width=1.0\textwidth]{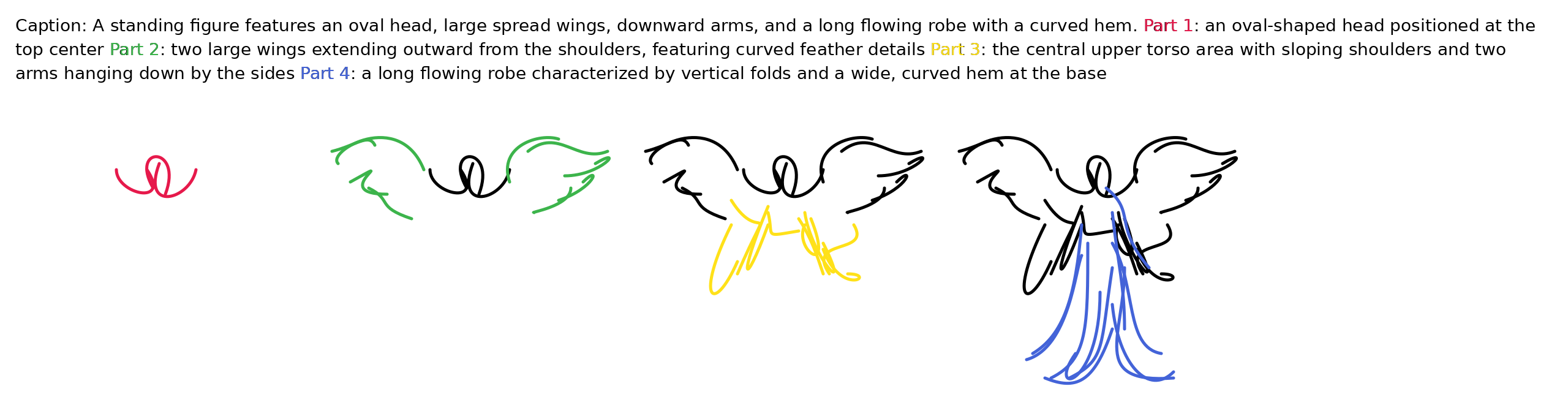} \\
    \includegraphics[width=1.0\textwidth]{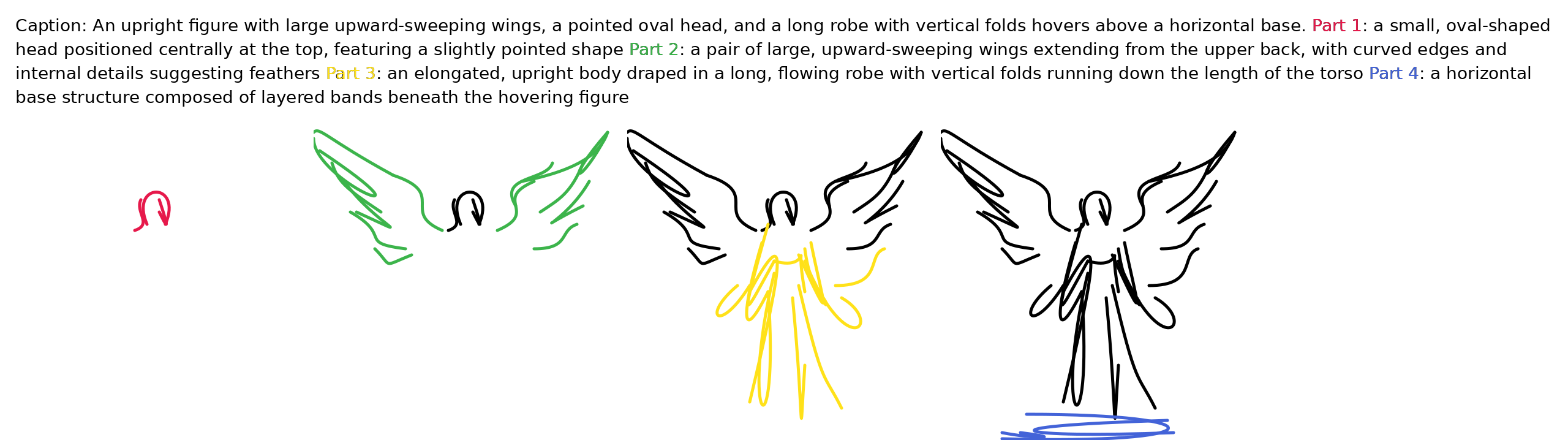} \\
    \includegraphics[width=1.0\textwidth]{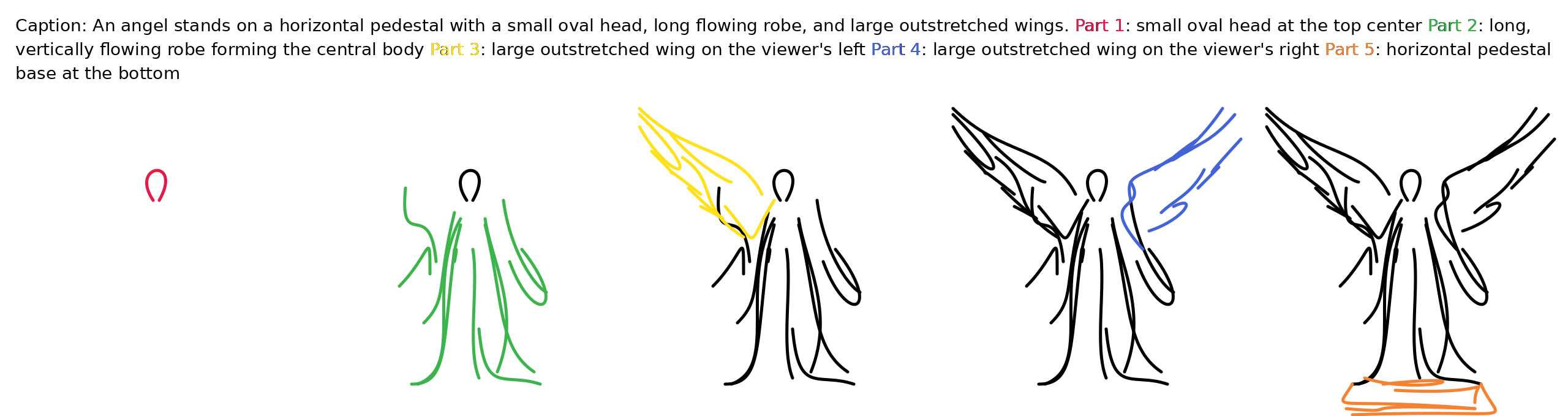}
\end{tabular}
\caption{Additional part-by-part results of our model. Part descriptions and caption appear above each sketch's cumulative frames, with newly added parts color-coded to match corresponding part labels. }
\end{figure}

\begin{figure}[H]
\centering
\begin{tabular}{c}
    \includegraphics[width=1.0\textwidth]{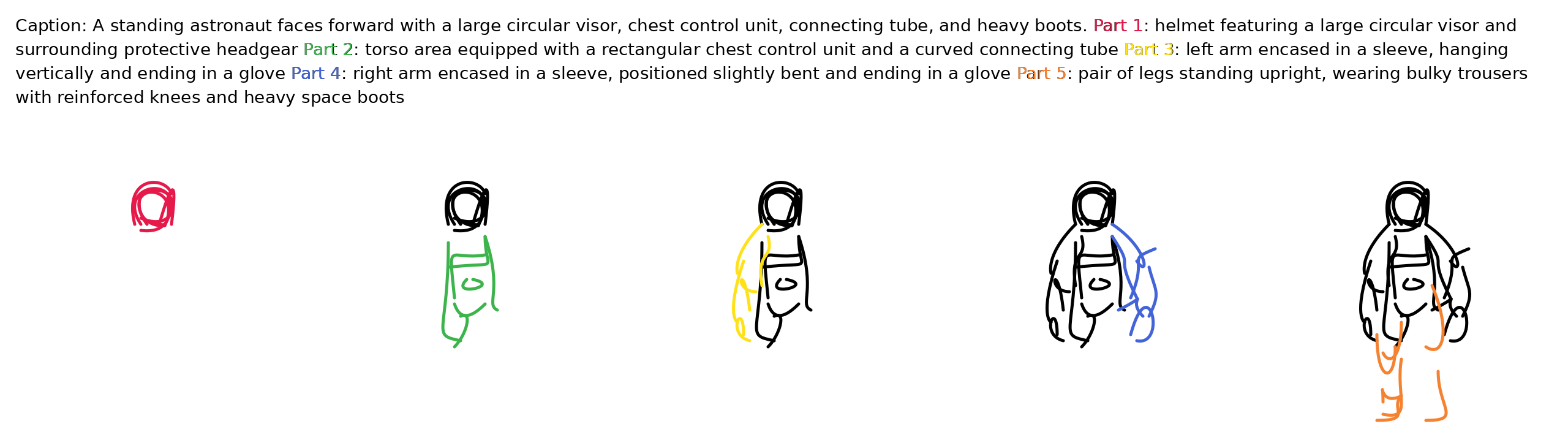} \\
    \includegraphics[width=1.0\textwidth]{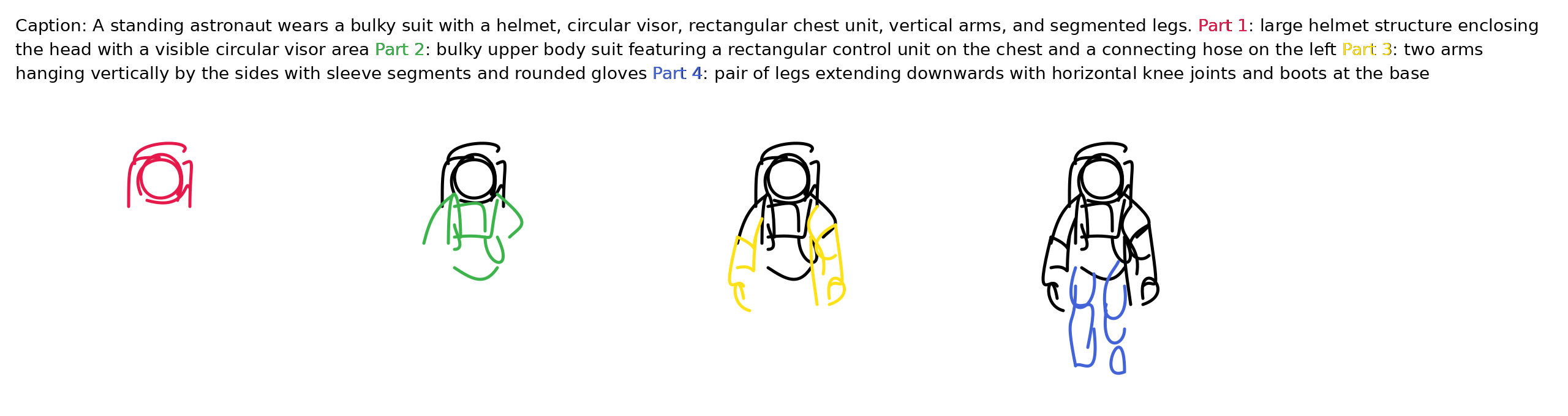} \\
    \includegraphics[width=1.0\textwidth]{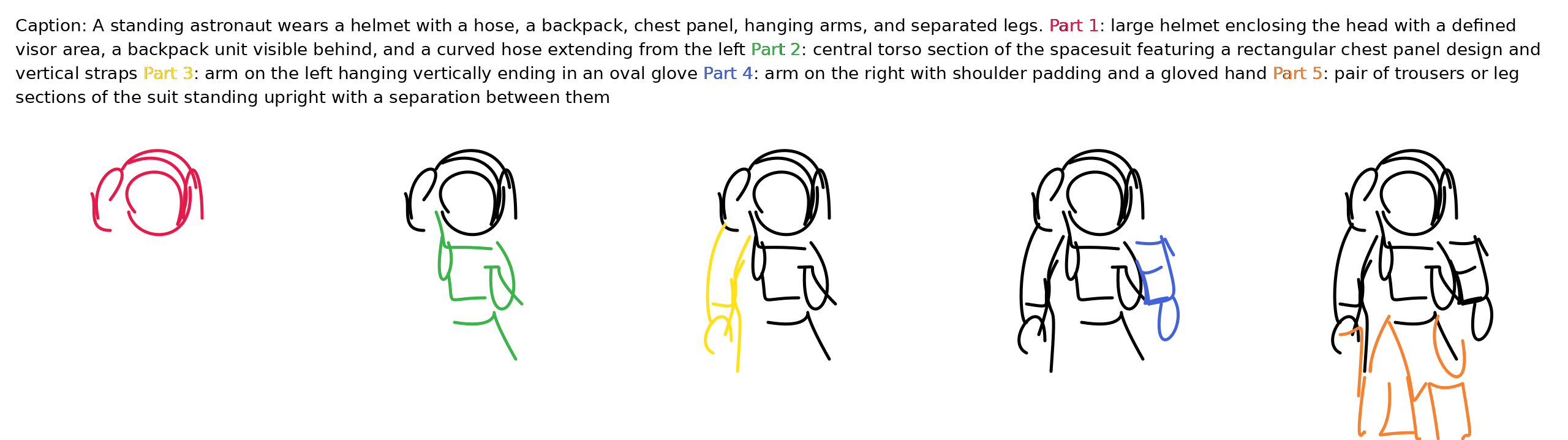} \\
    \includegraphics[width=1.0\textwidth]{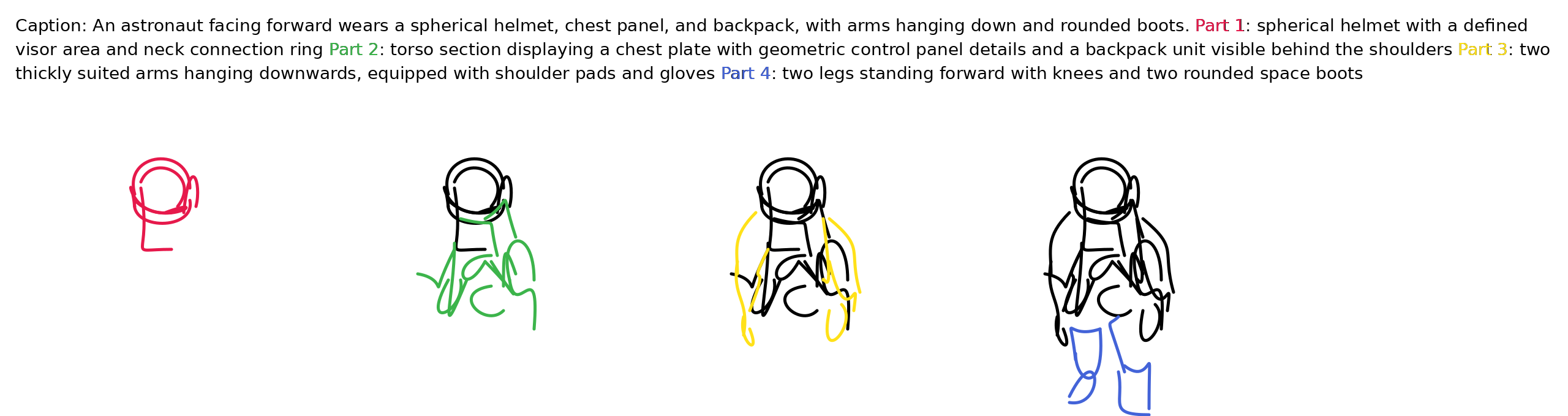} \\
    \includegraphics[width=1.0\textwidth]{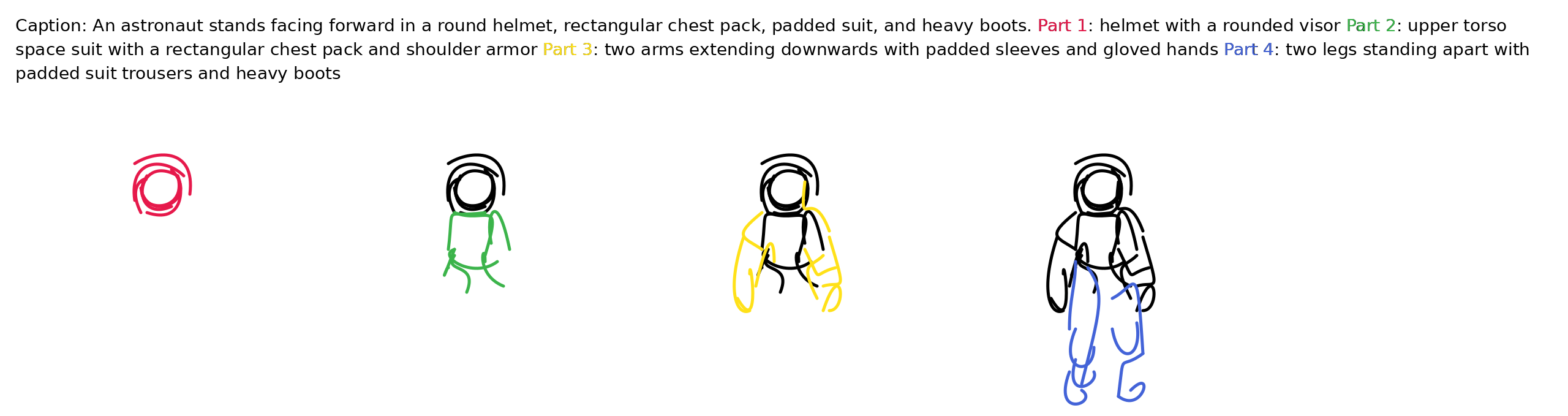}
\end{tabular}
\caption{Additional part-by-part results of our model (continued).}
\end{figure}

\begin{figure}[H]
\centering
\begin{tabular}{c}
    \includegraphics[width=1.0\textwidth]{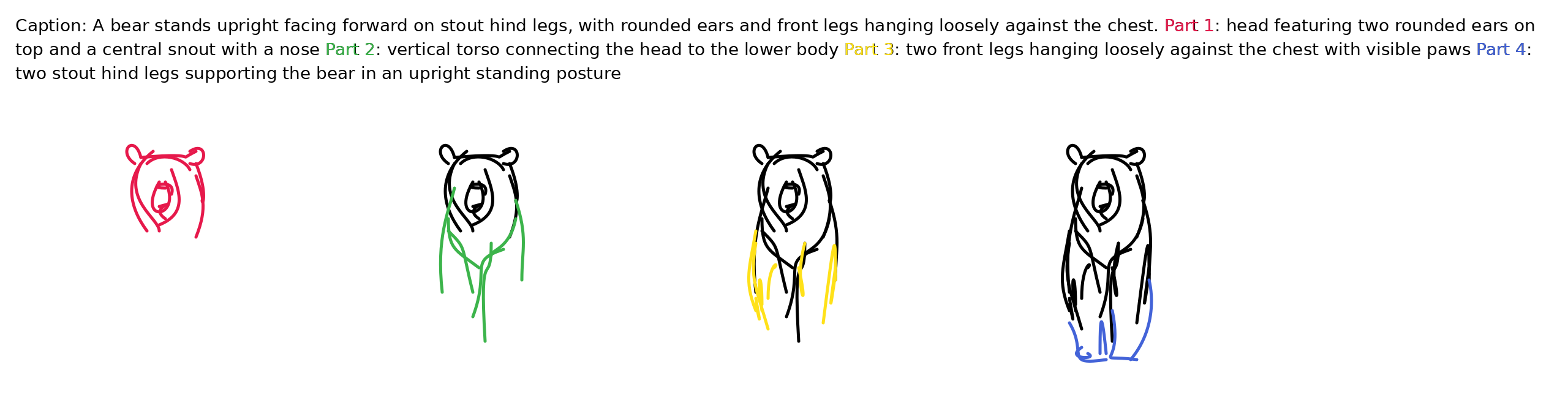} \\
    \includegraphics[width=1.0\textwidth]{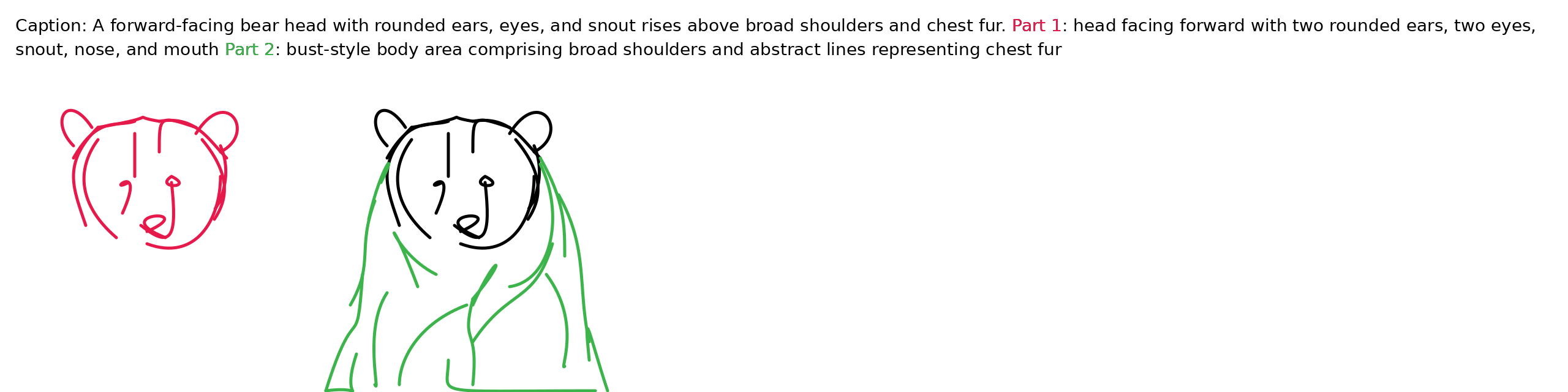} \\
    \includegraphics[width=1.0\textwidth]{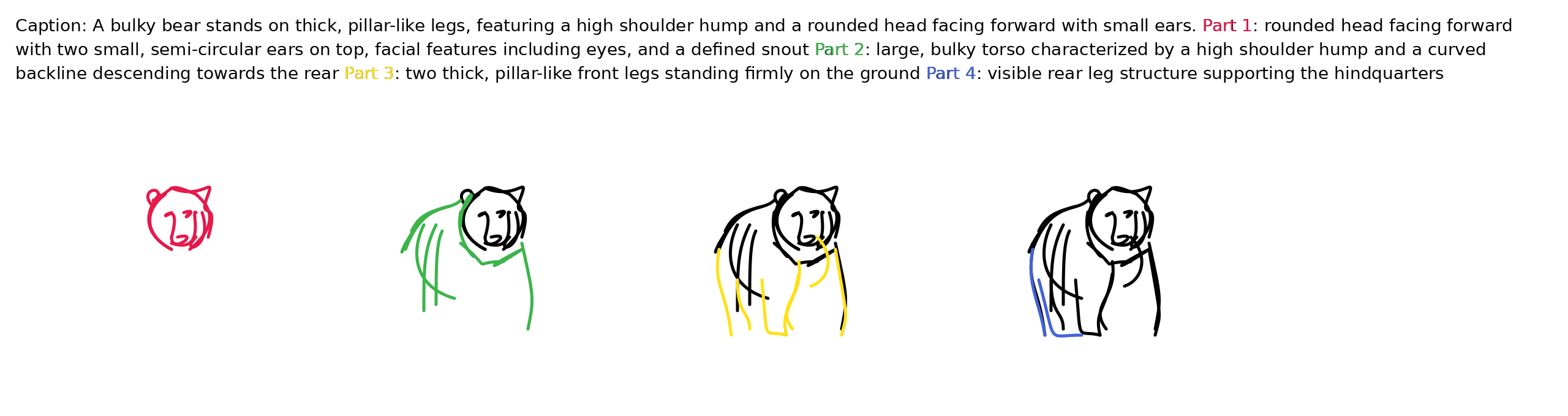} \\
    \includegraphics[width=1.0\textwidth]{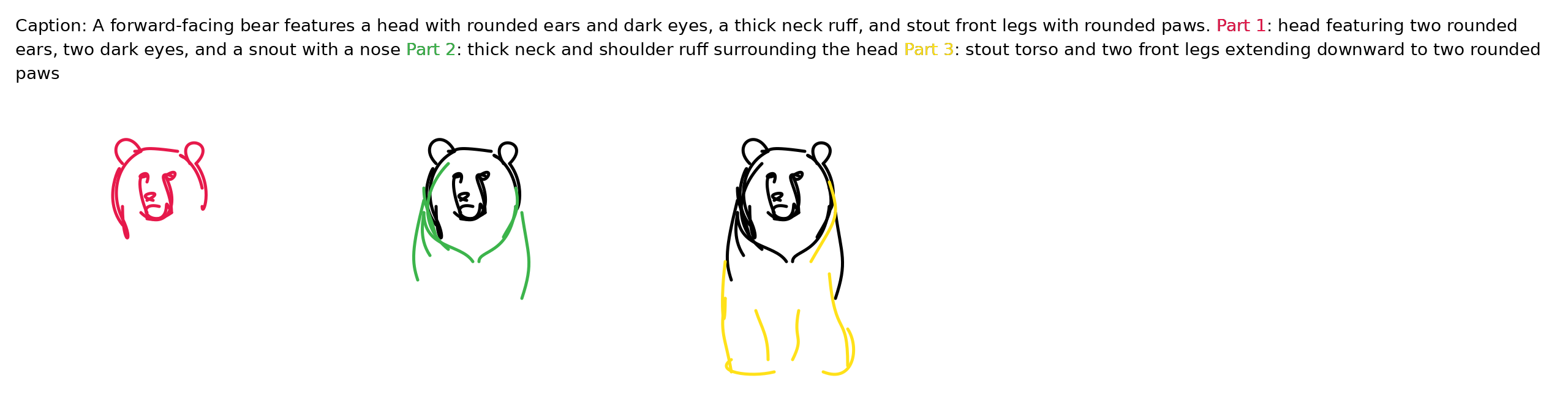} \\
    \includegraphics[width=1.0\textwidth]{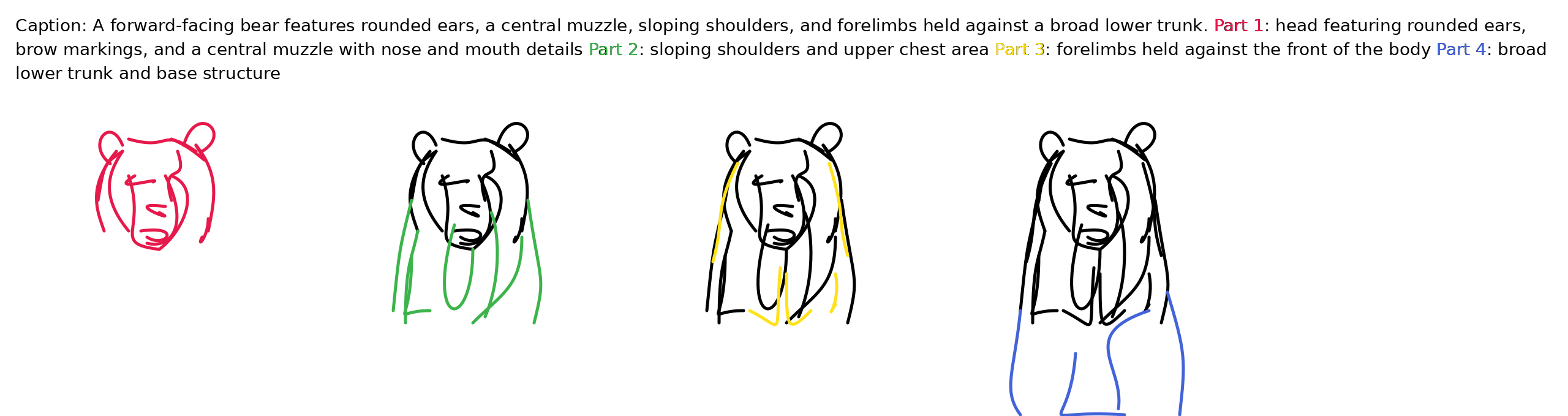}
\end{tabular}
\caption{Additional part-by-part results of our model (continued).}
\end{figure}

\begin{figure}[H]
\centering
\begin{tabular}{c}
    \includegraphics[width=1.0\textwidth]{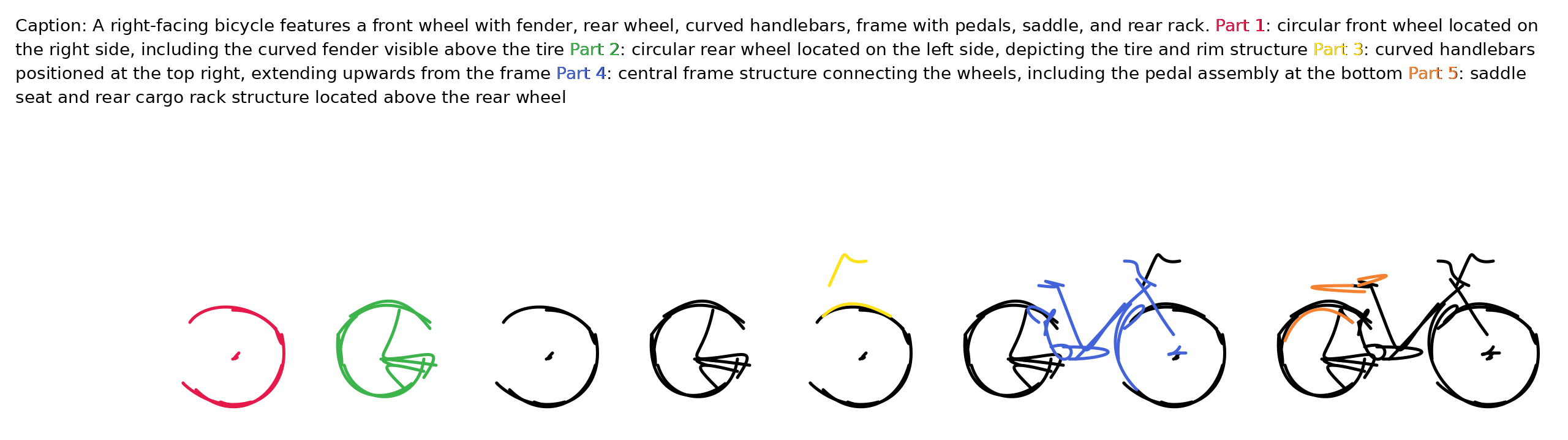} \\
    \includegraphics[width=1.0\textwidth]{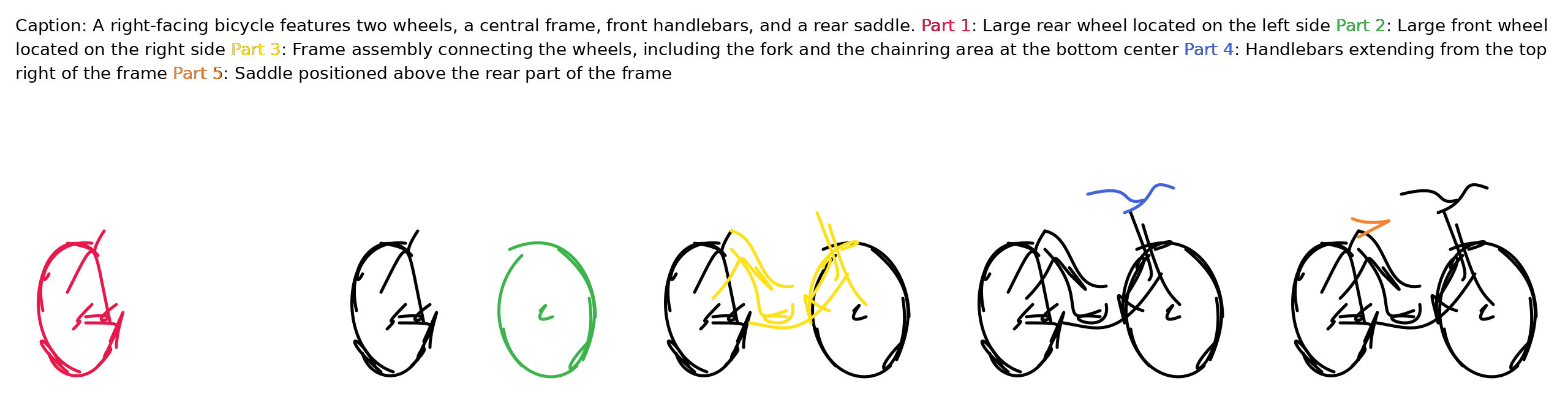} \\
    \includegraphics[width=1.0\textwidth]{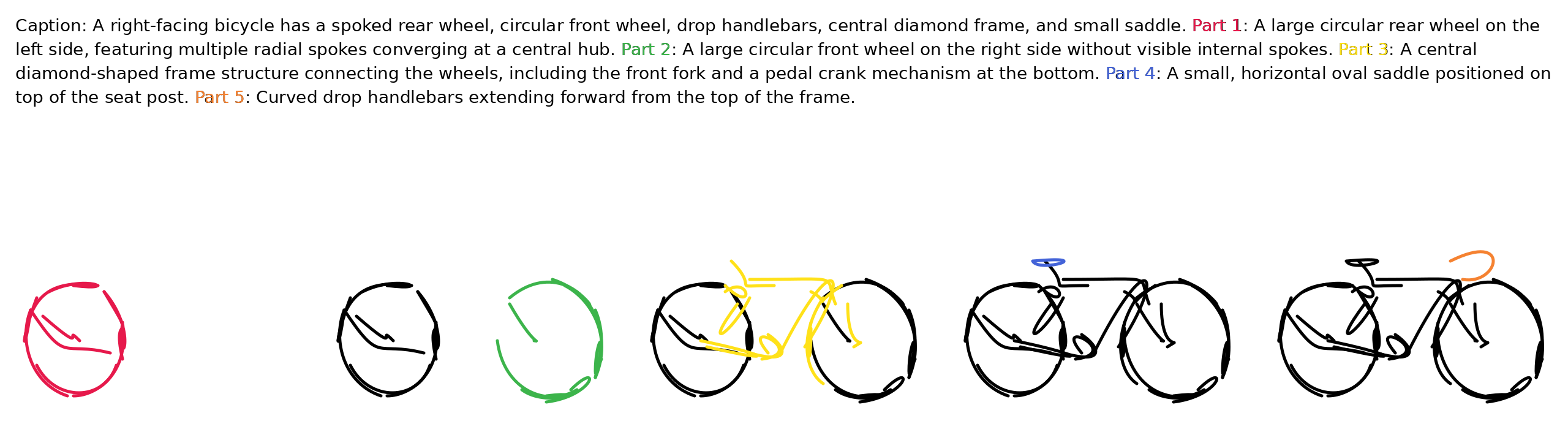} \\
    \includegraphics[width=1.0\textwidth]{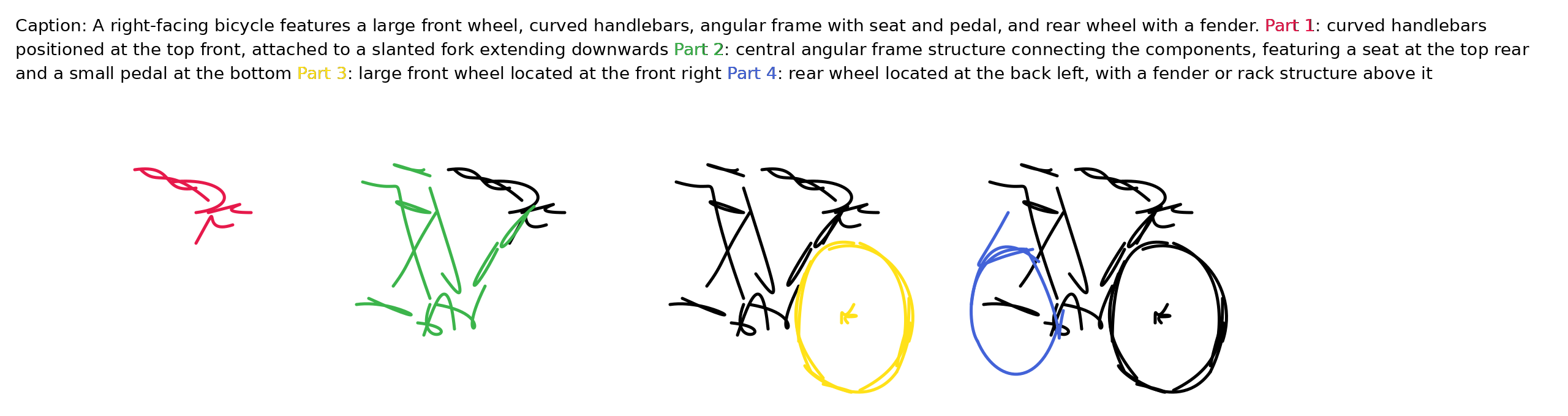} \\
    \includegraphics[width=1.0\textwidth]{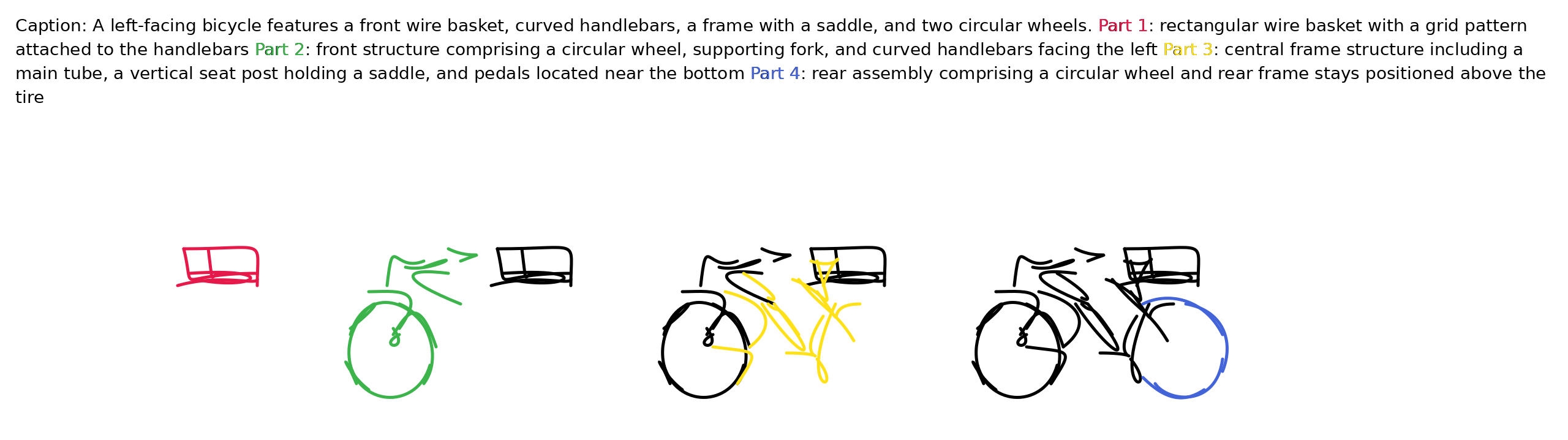}
\end{tabular}
\caption{Additional part-by-part results of our model (continued).}
\end{figure}

\begin{figure}[H]
\centering
\begin{tabular}{c}
    \includegraphics[width=1.0\textwidth]{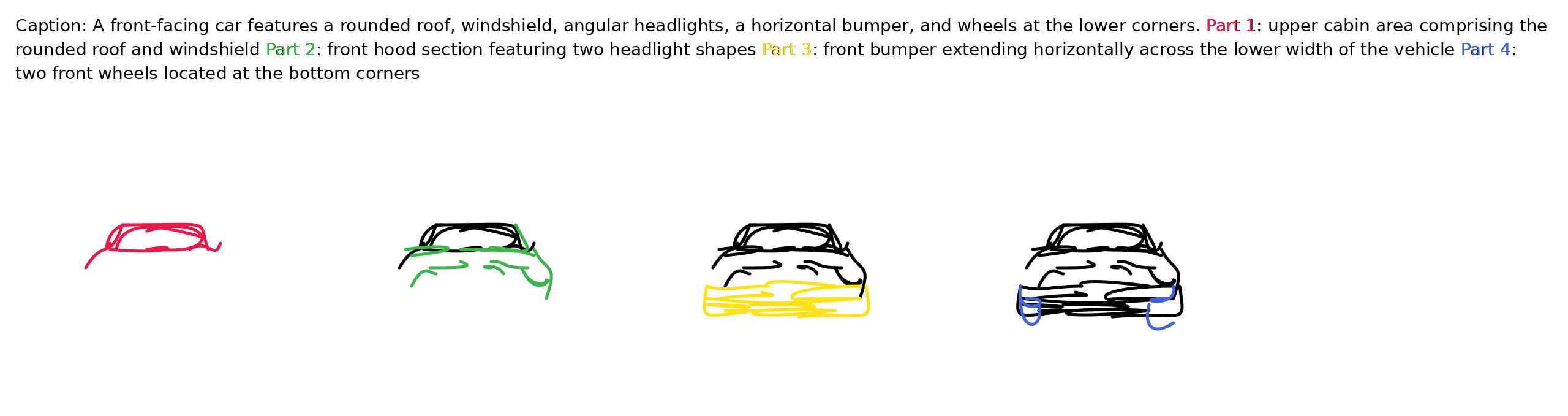} \\
    \includegraphics[width=1.0\textwidth]{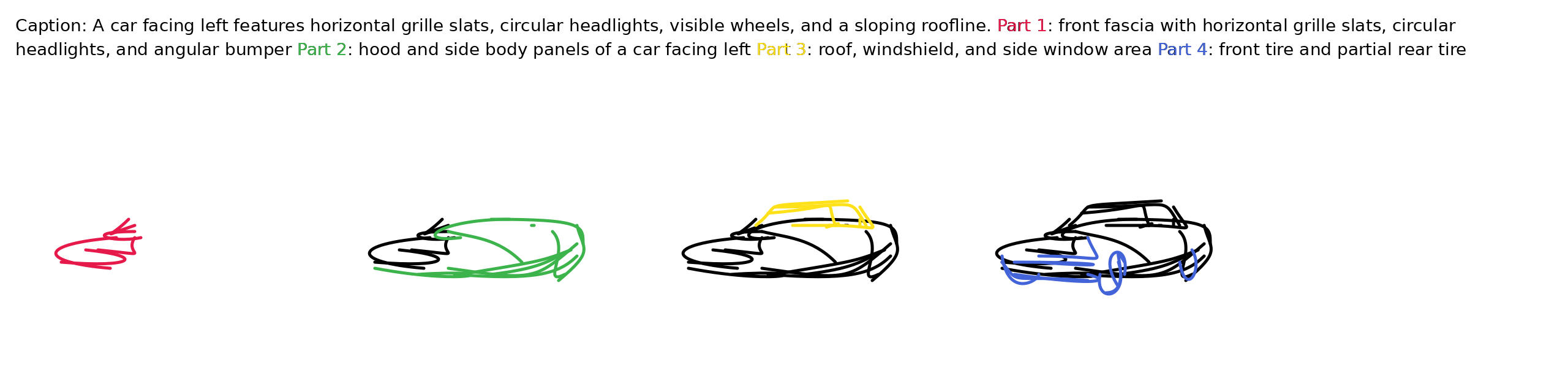} \\
    \includegraphics[width=1.0\textwidth]{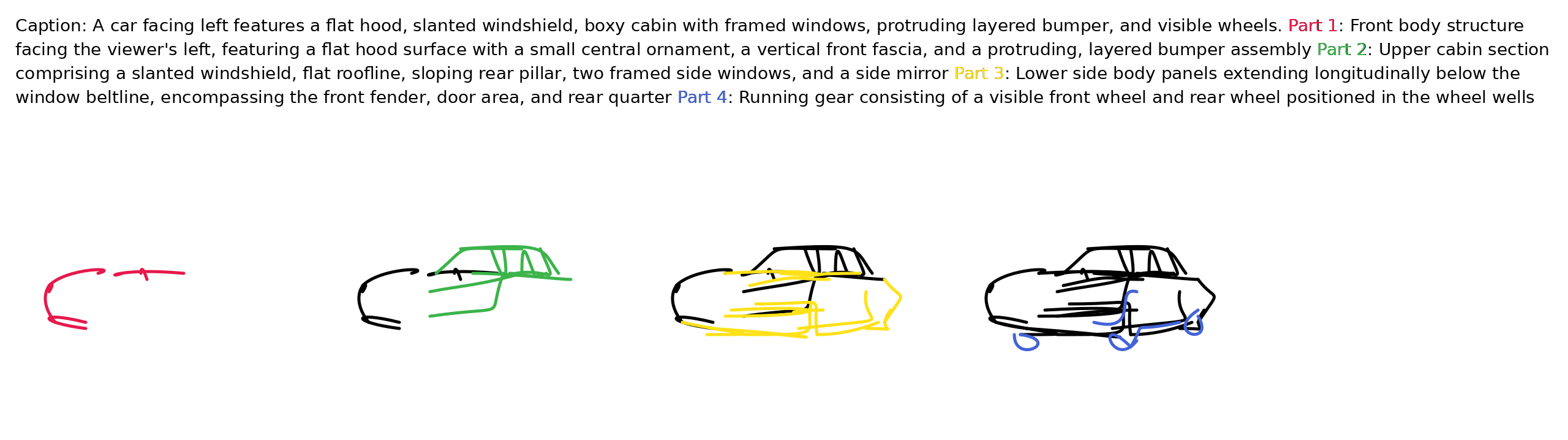} \\
    \includegraphics[width=1.0\textwidth]{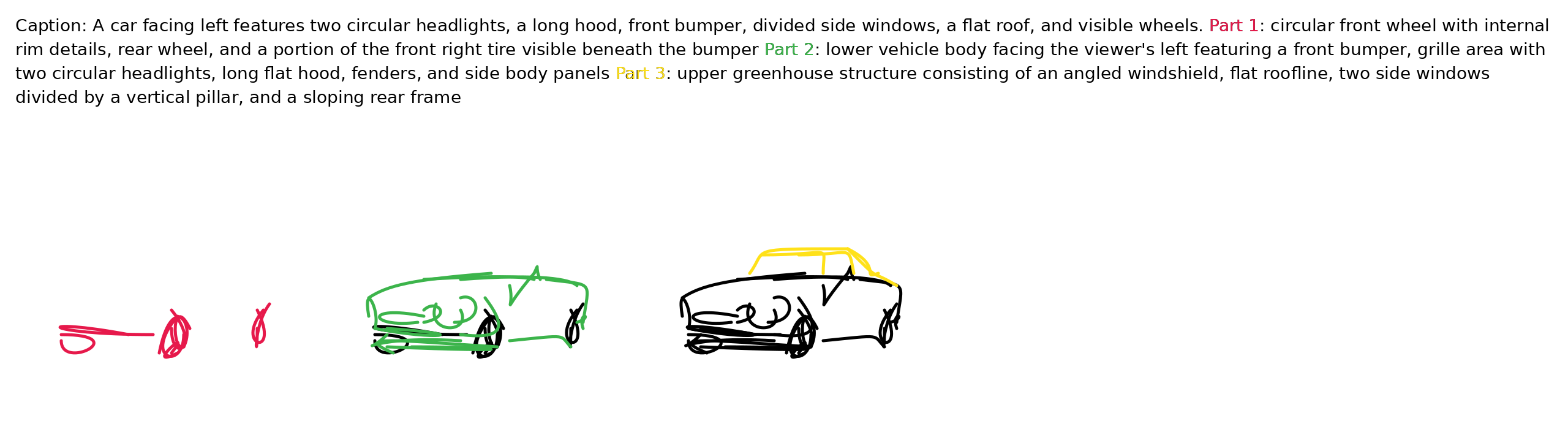} \\
    \includegraphics[width=1.0\textwidth]{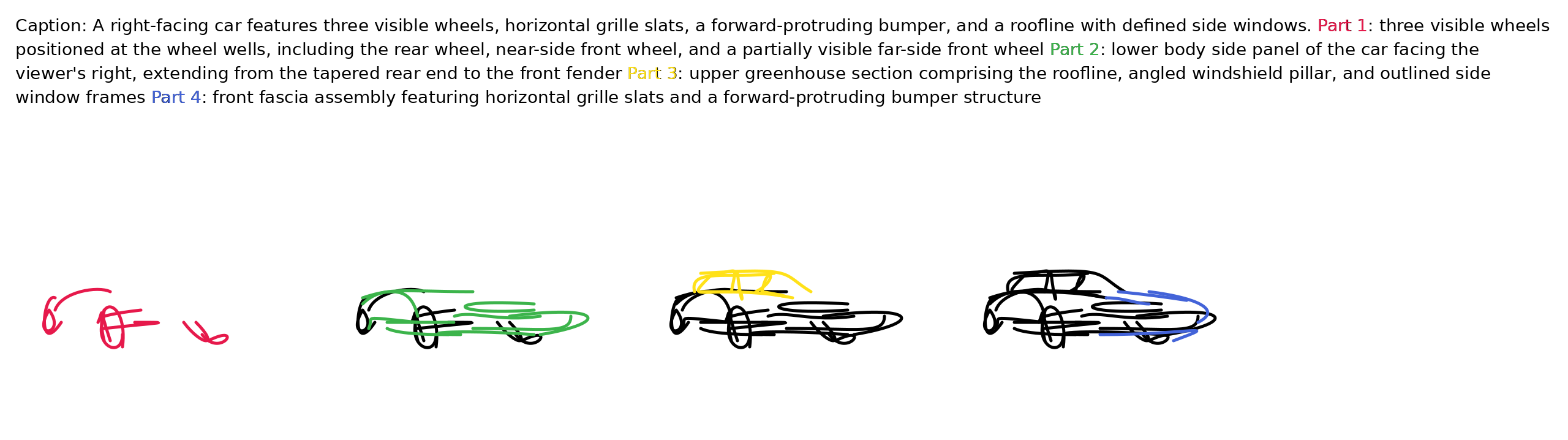}
\end{tabular}
\caption{Additional part-by-part results of our model (continued).}
\end{figure}

\begin{figure}[H]
\centering
\begin{tabular}{c}
    \includegraphics[width=1.0\textwidth]{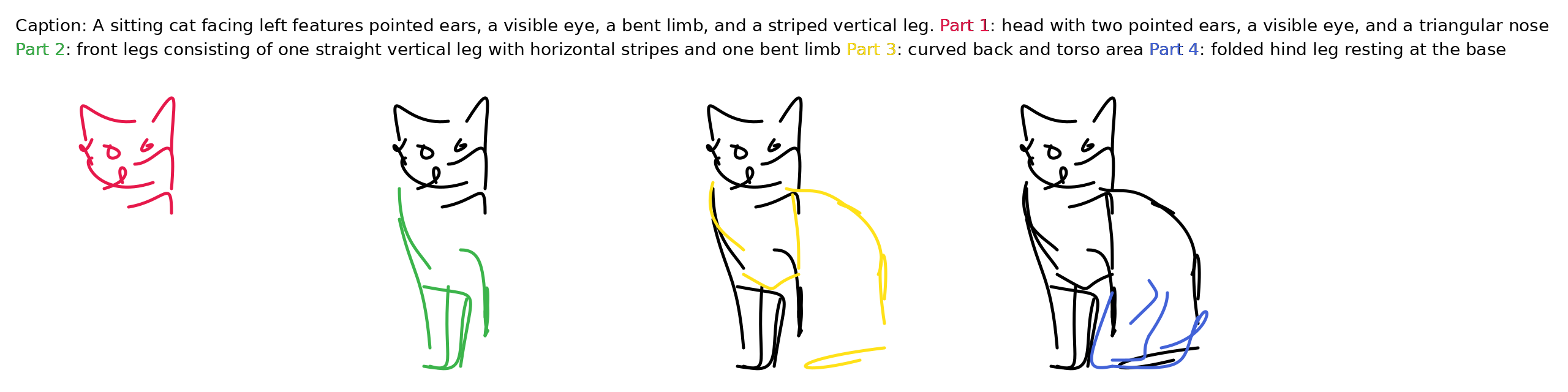} \\
    \includegraphics[width=1.0\textwidth]{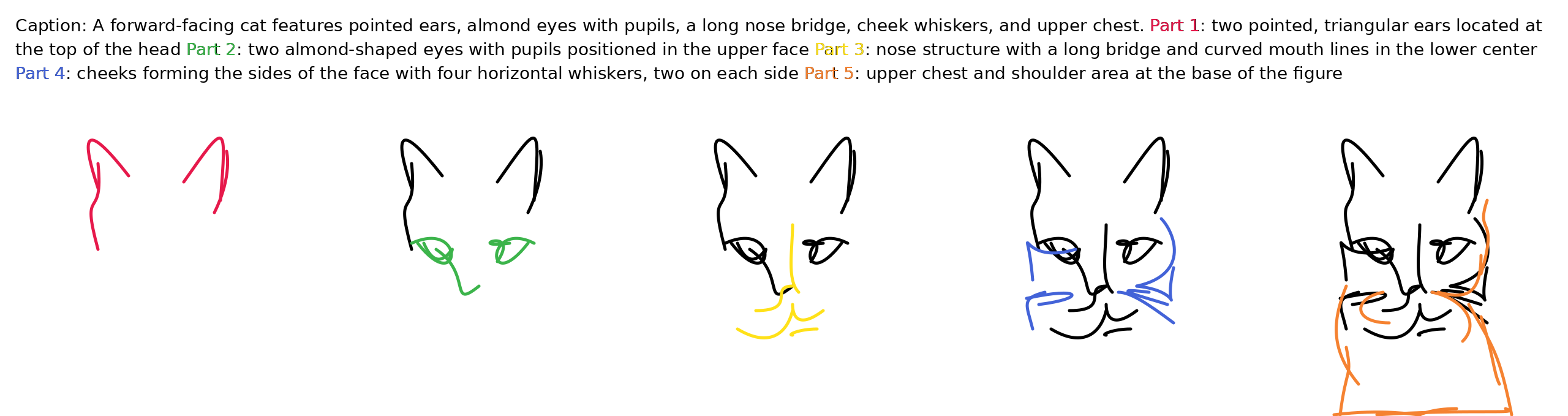} \\
    \includegraphics[width=1.0\textwidth]{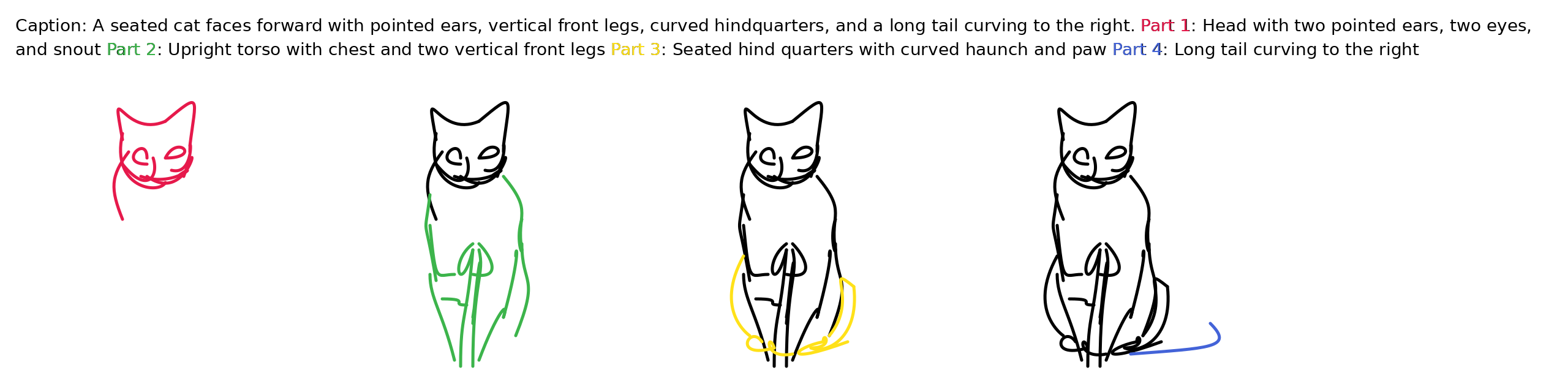} \\
    \includegraphics[width=1.0\textwidth]{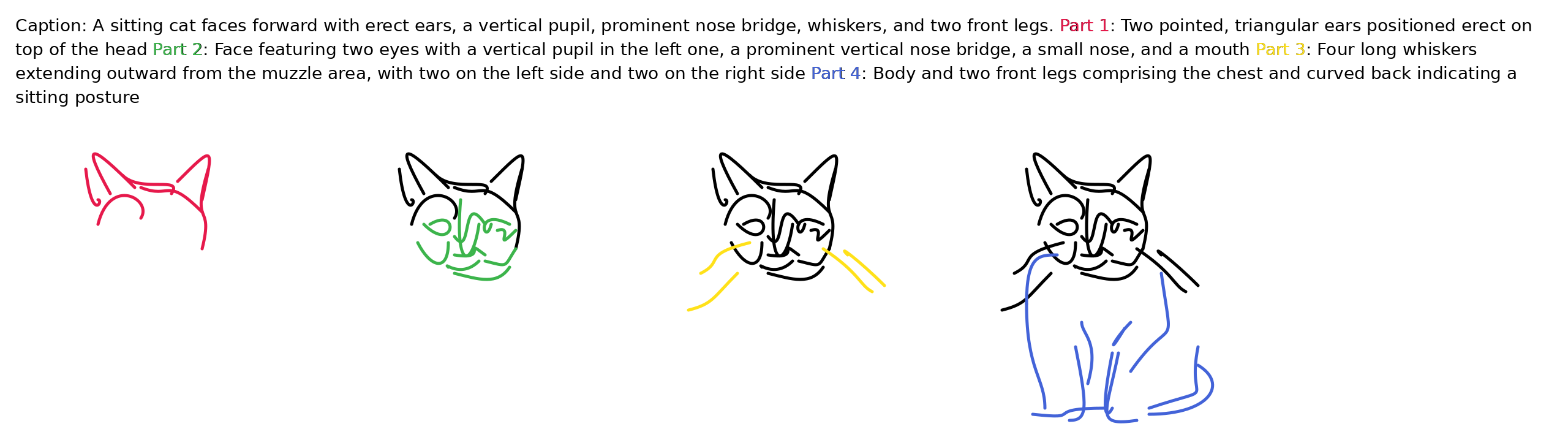} \\
    \includegraphics[width=1.0\textwidth]{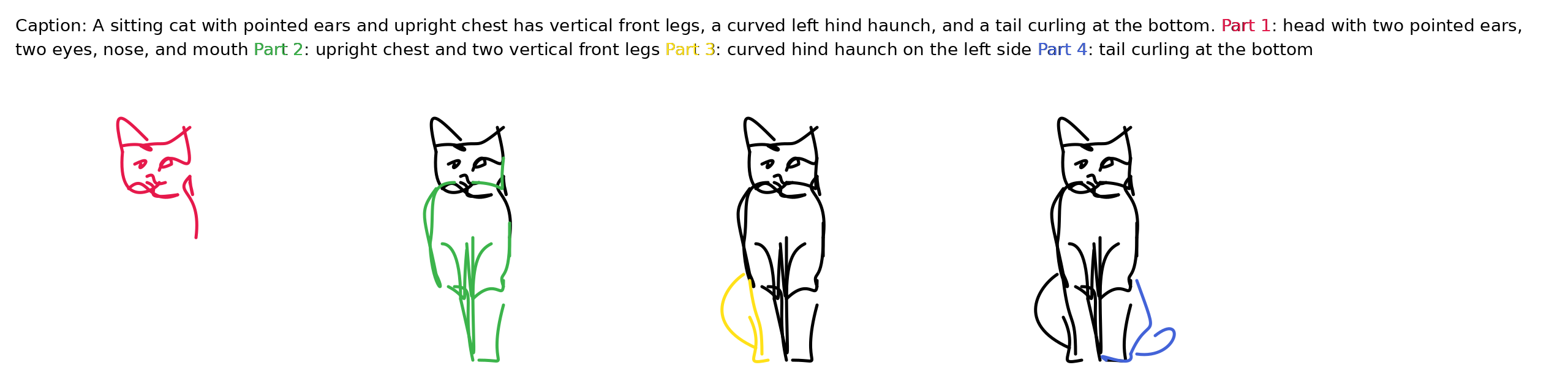}
\end{tabular}
\caption{Additional part-by-part results of our model (continued).}
\end{figure}

\begin{figure}[H]
\centering
\begin{tabular}{c}
    \includegraphics[width=1.0\textwidth]{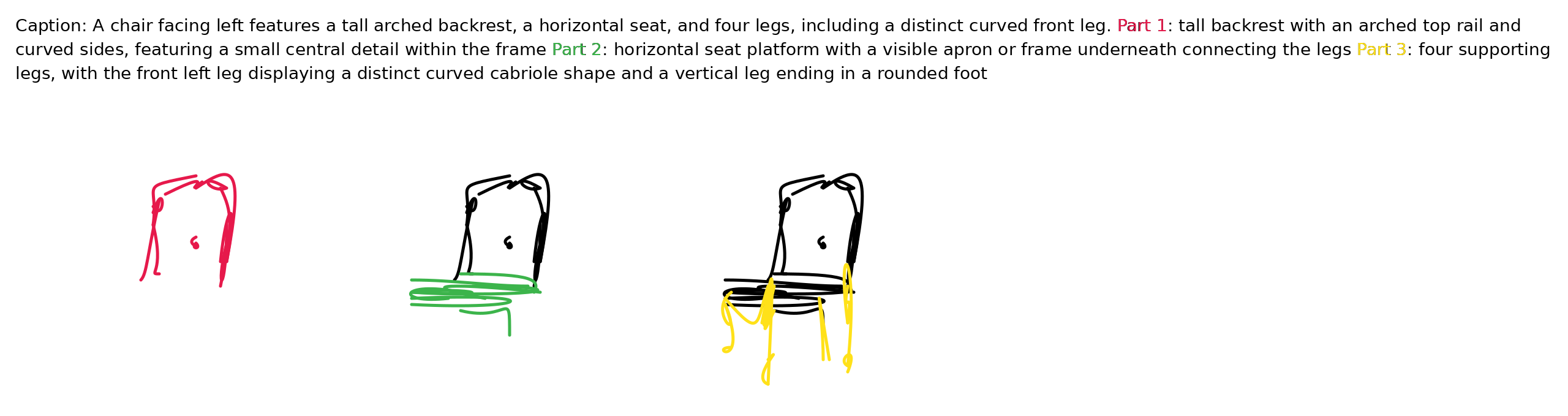} \\
    \includegraphics[width=1.0\textwidth]{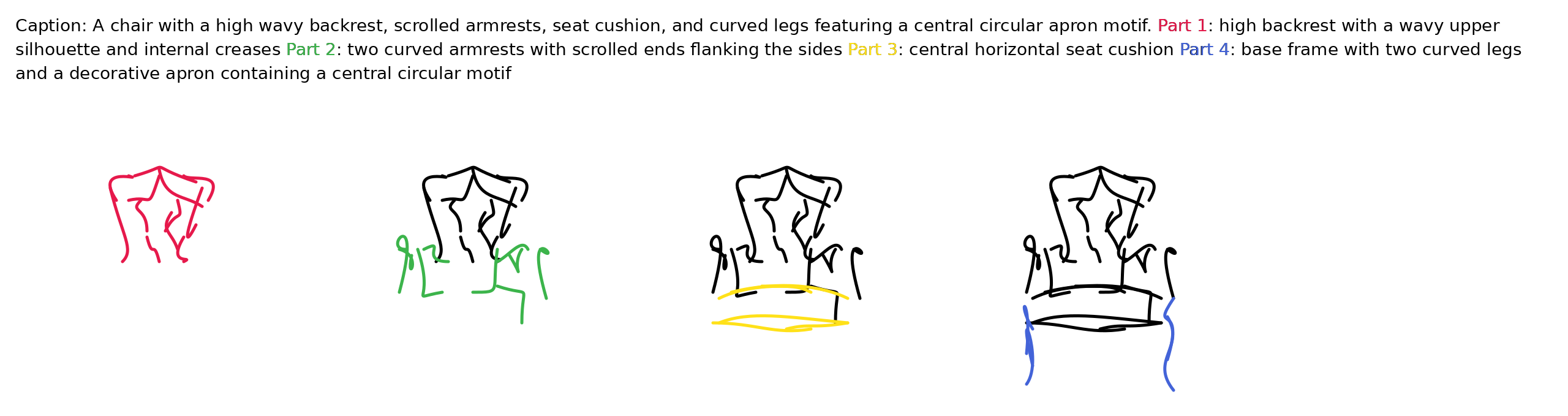} \\
    \includegraphics[width=1.0\textwidth]{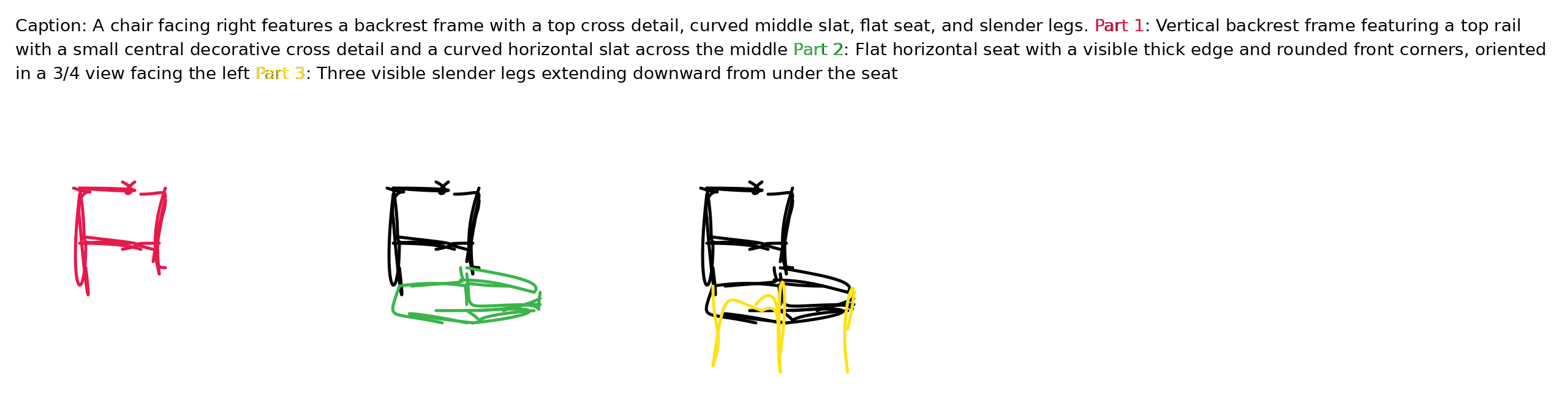} \\
    \includegraphics[width=1.0\textwidth]{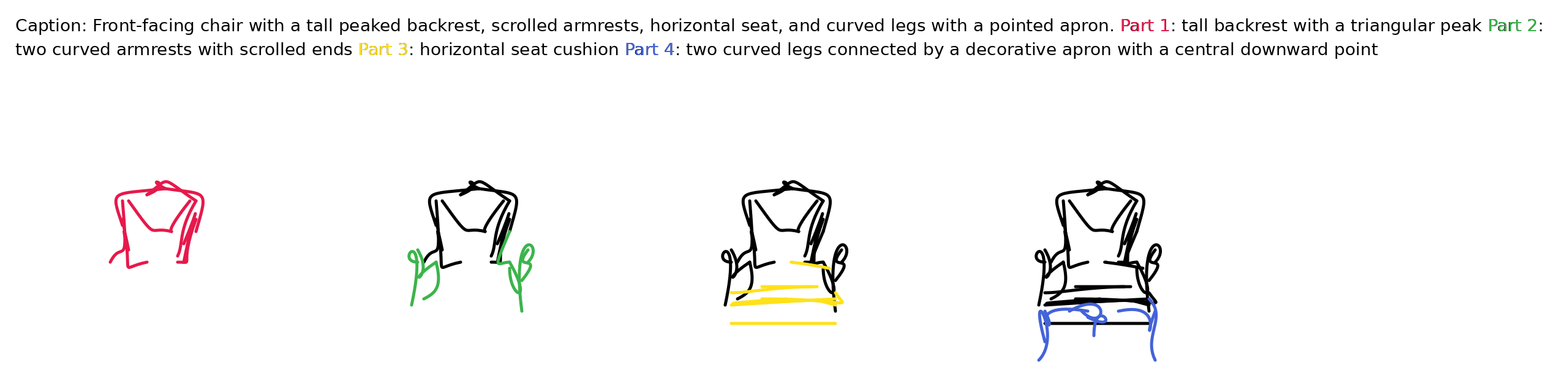} \\
    \includegraphics[width=1.0\textwidth]{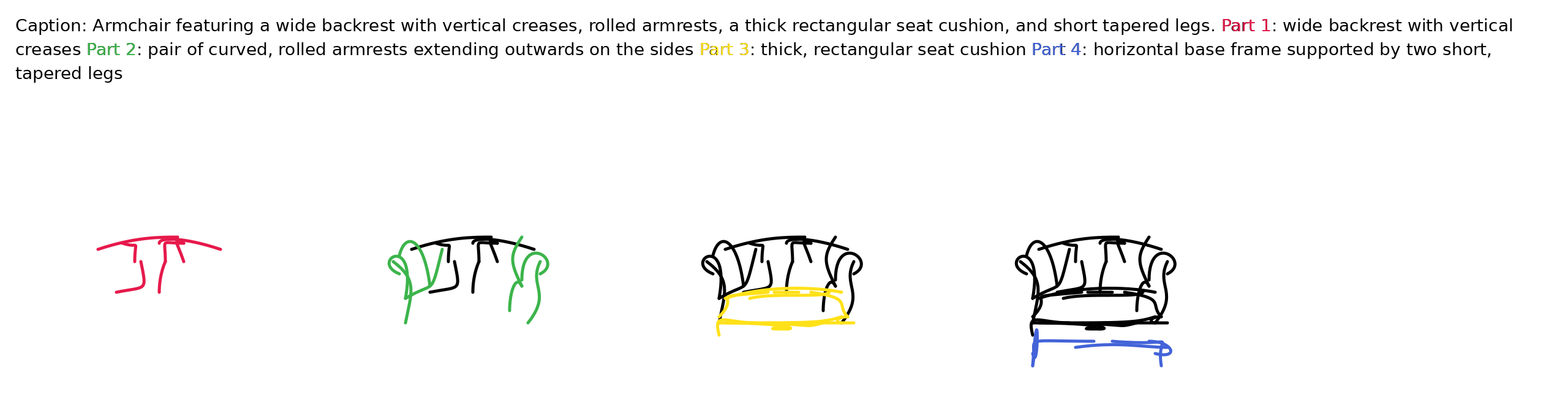}
\end{tabular}
\caption{Additional part-by-part results of our model (continued).}
\end{figure}

\begin{figure}[H]
\centering
\begin{tabular}{c}
    \includegraphics[width=1.0\textwidth]{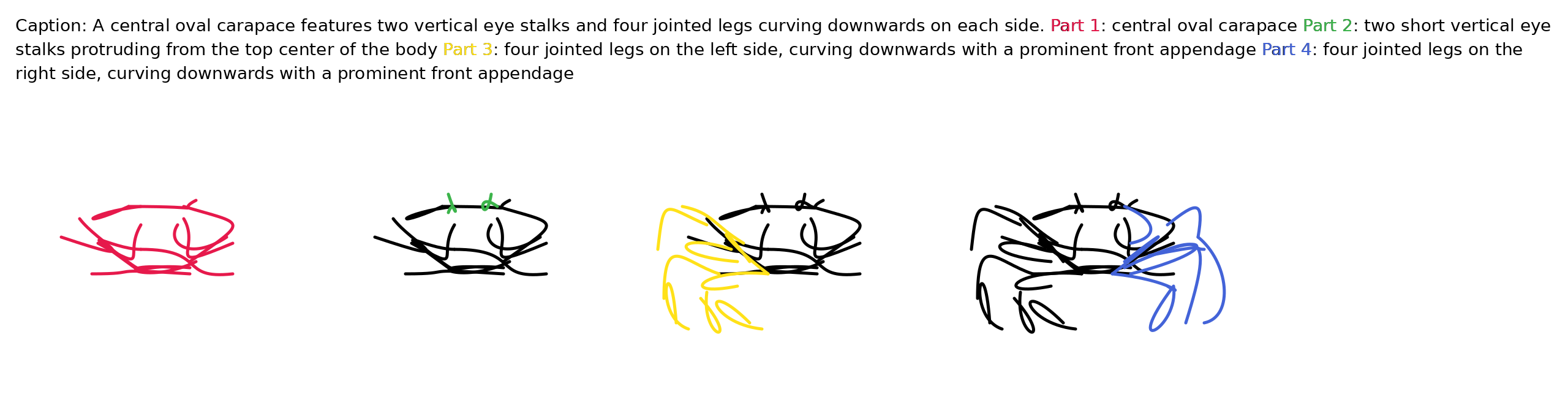} \\
    \includegraphics[width=1.0\textwidth]{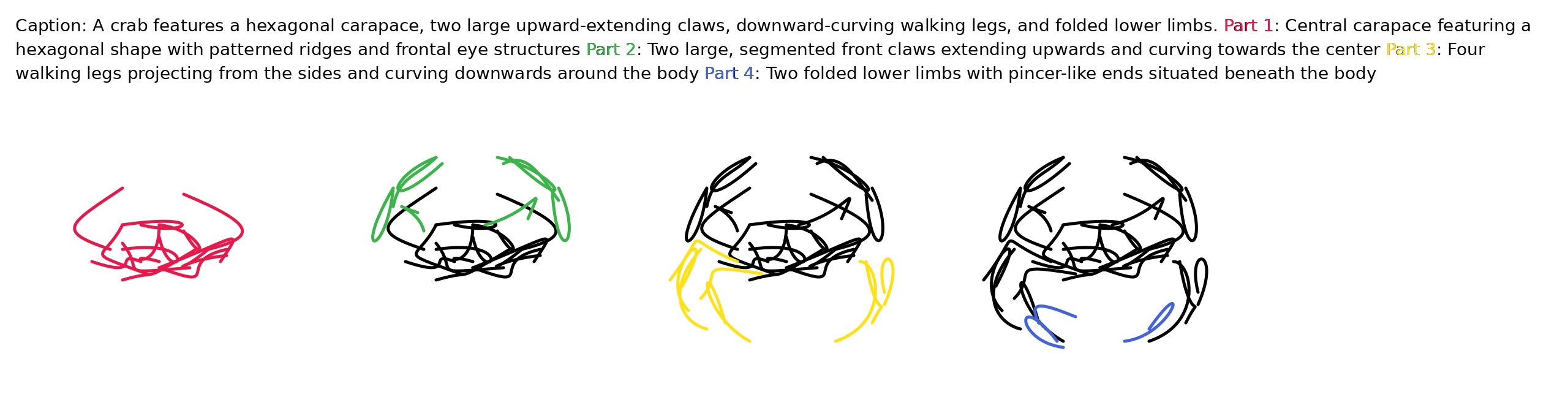} \\
    \includegraphics[width=1.0\textwidth]{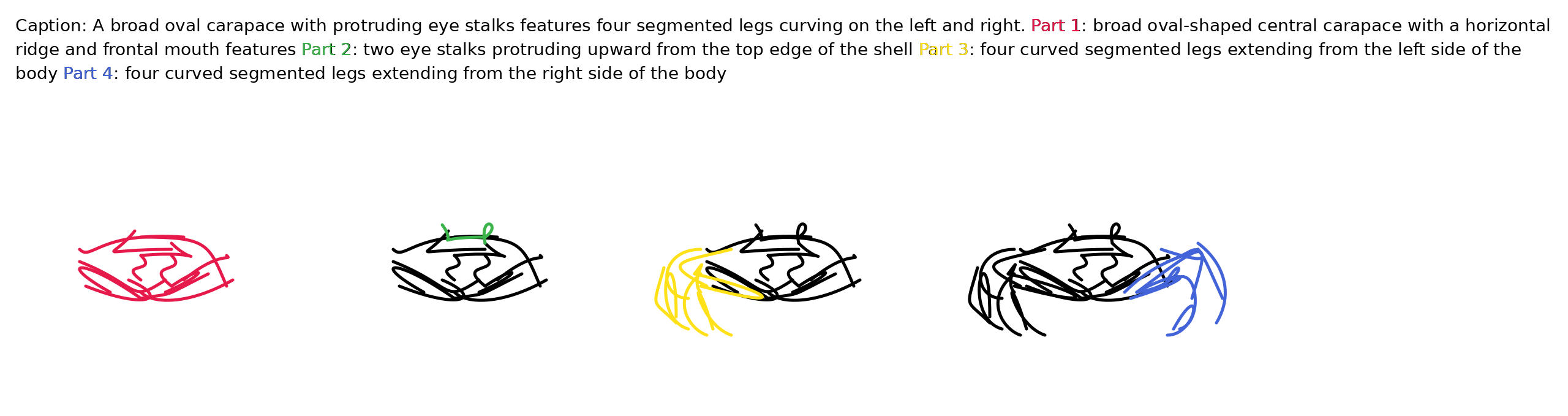} \\
    \includegraphics[width=1.0\textwidth]{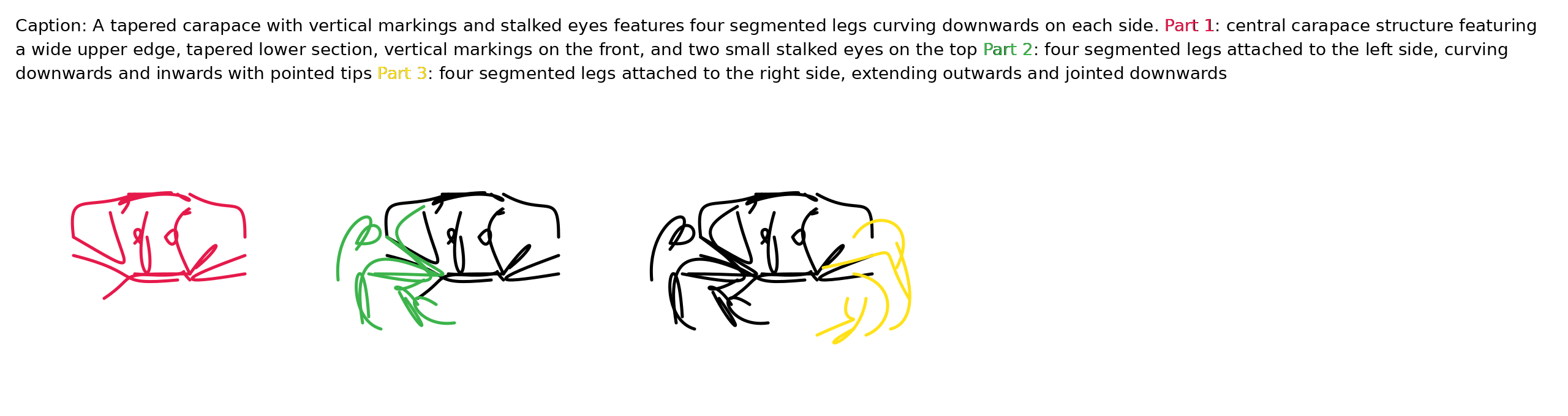} \\
    \includegraphics[width=1.0\textwidth]{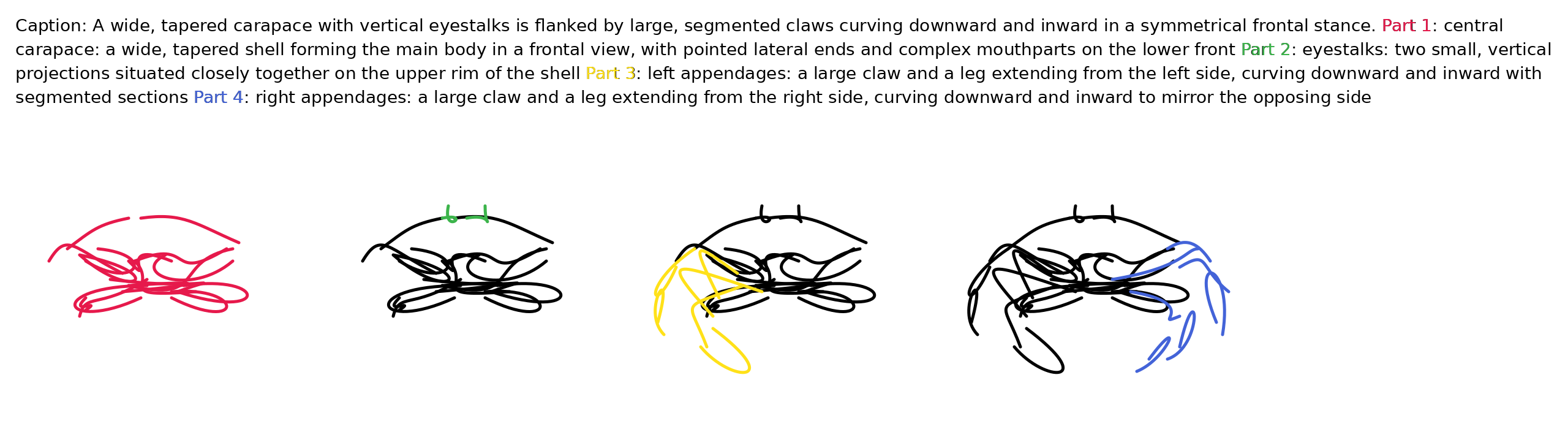}
\end{tabular}
\caption{Additional part-by-part results of our model (continued).}
\end{figure}

\begin{figure}[H]
\centering
\begin{tabular}{c}
    \includegraphics[width=1.0\textwidth]{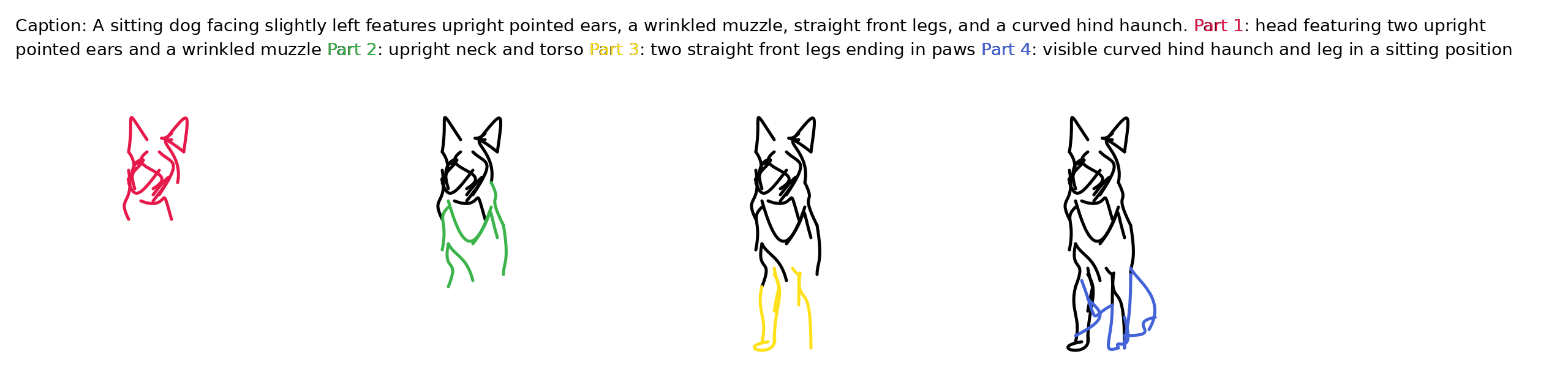} \\
    \includegraphics[width=1.0\textwidth]{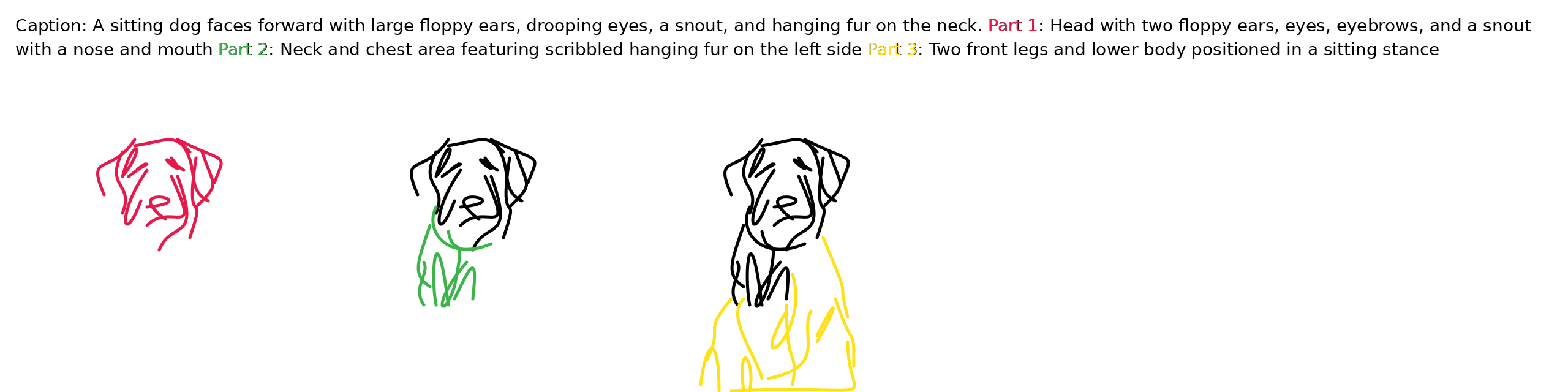} \\
    \includegraphics[width=1.0\textwidth]{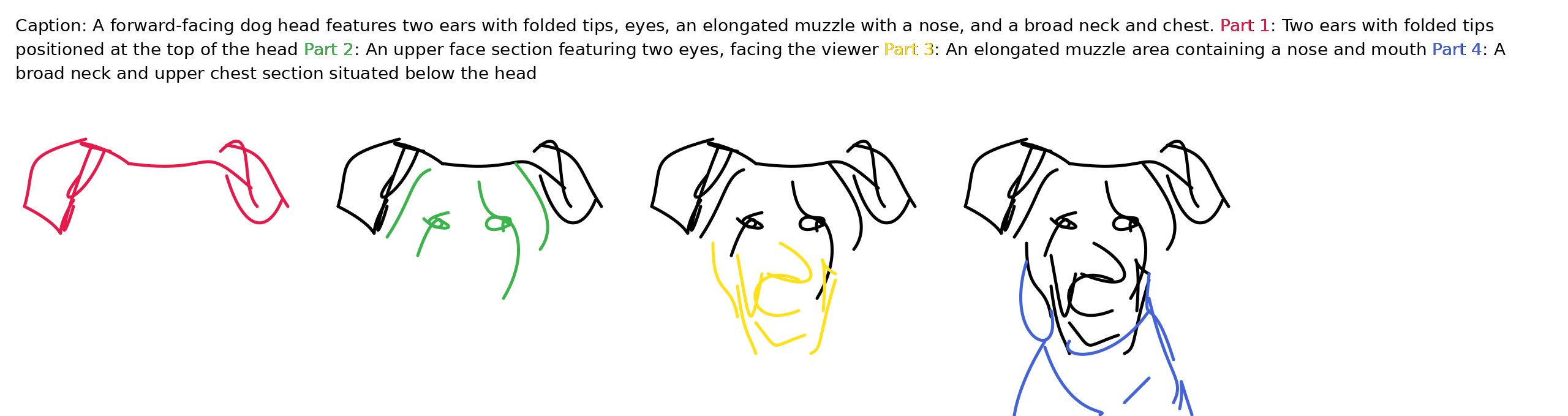} \\
    \includegraphics[width=1.0\textwidth]{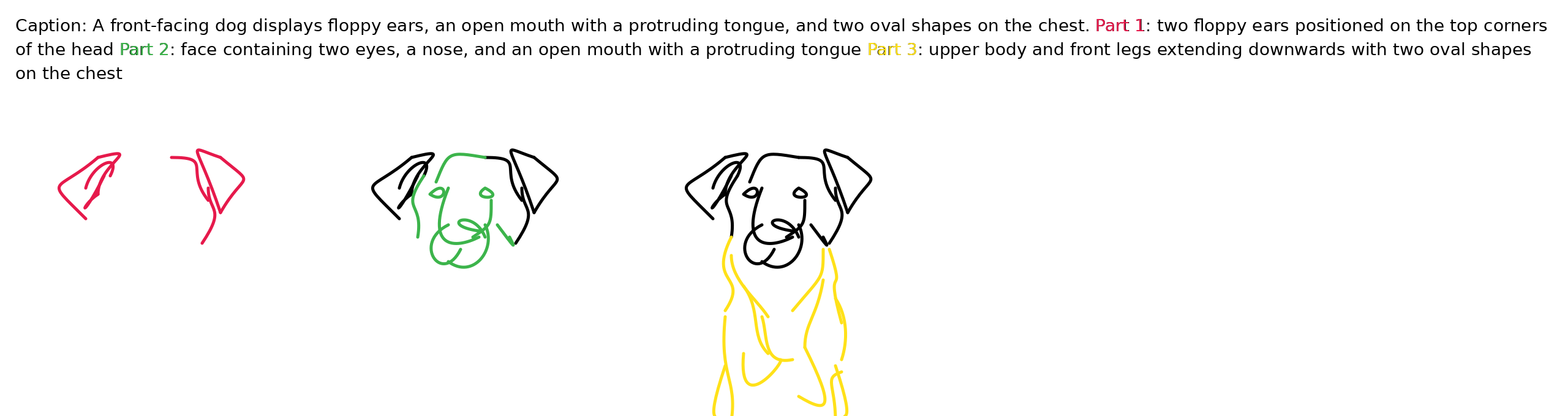} \\
    \includegraphics[width=1.0\textwidth]{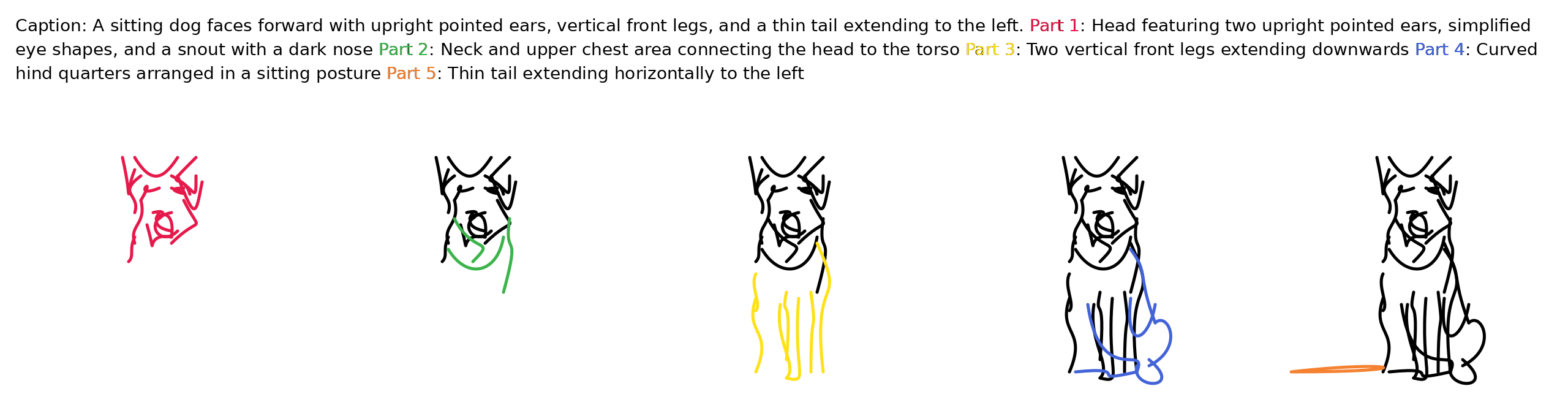}
\end{tabular}
\caption{Additional part-by-part results of our model (continued).}
\end{figure}

\begin{figure}[H]
\centering
\begin{tabular}{c}
    \includegraphics[width=1.0\textwidth]{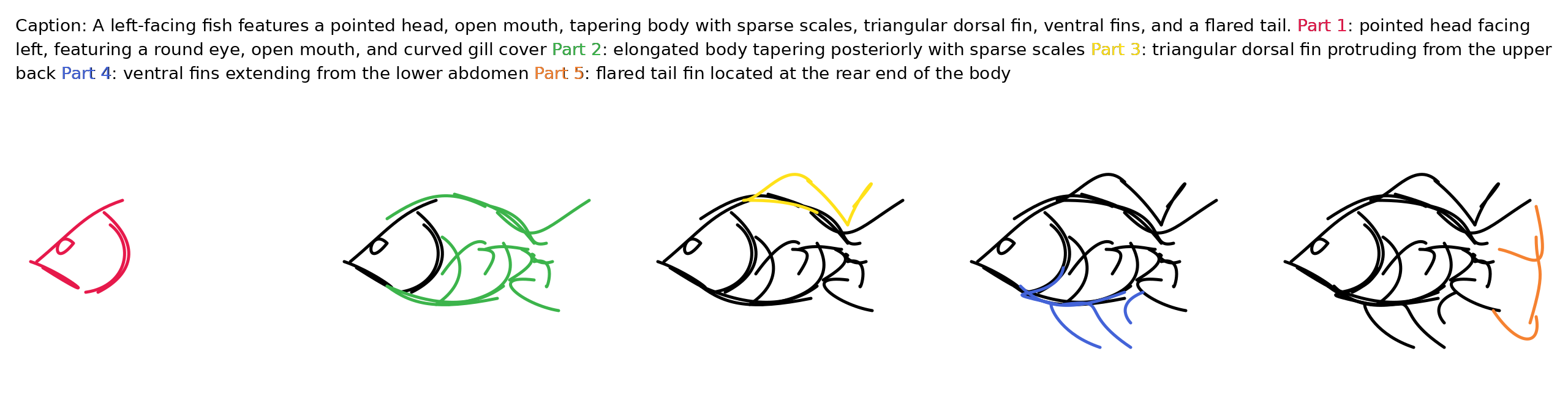} \\
    \includegraphics[width=1.0\textwidth]{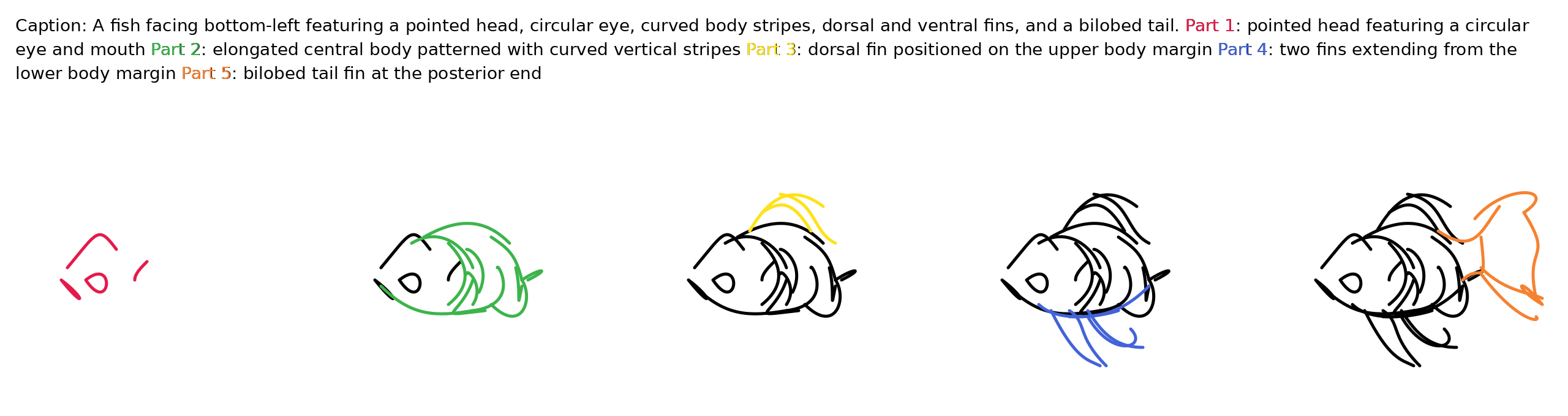} \\
    \includegraphics[width=1.0\textwidth]{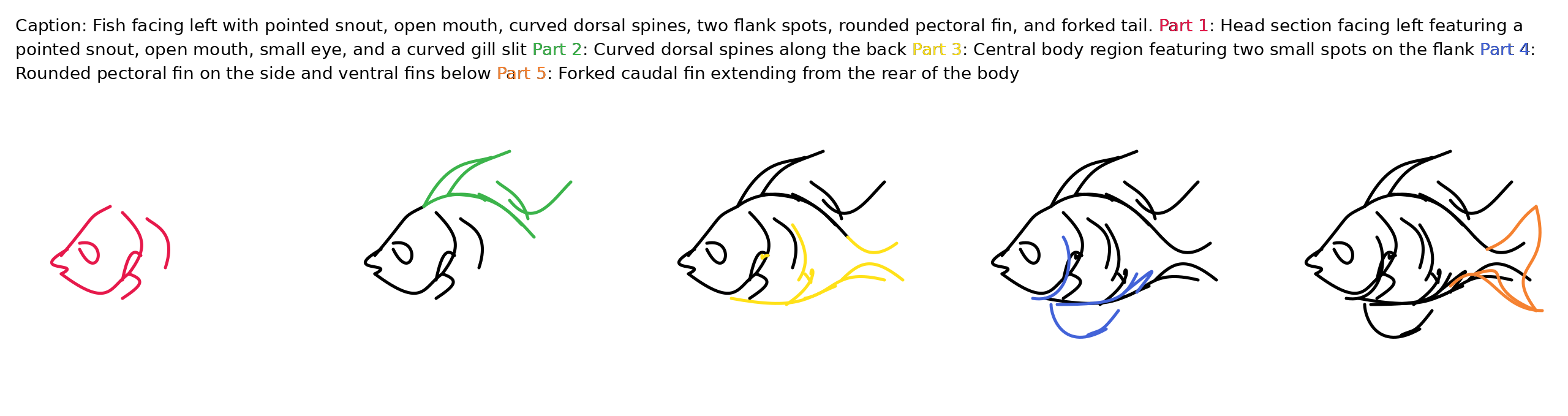} \\
    \includegraphics[width=1.0\textwidth]{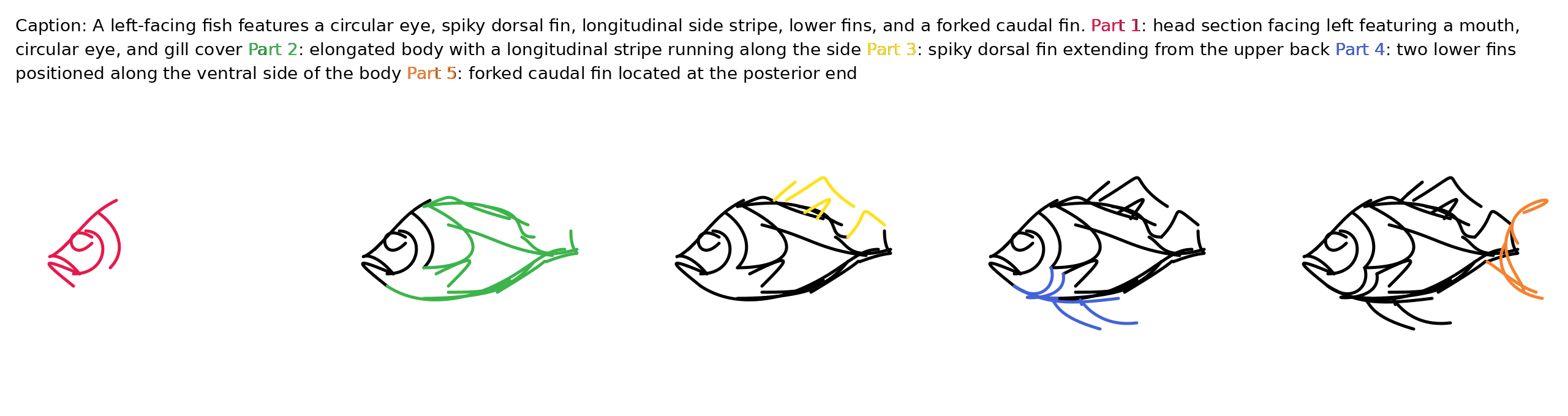} \\
    \includegraphics[width=1.0\textwidth]{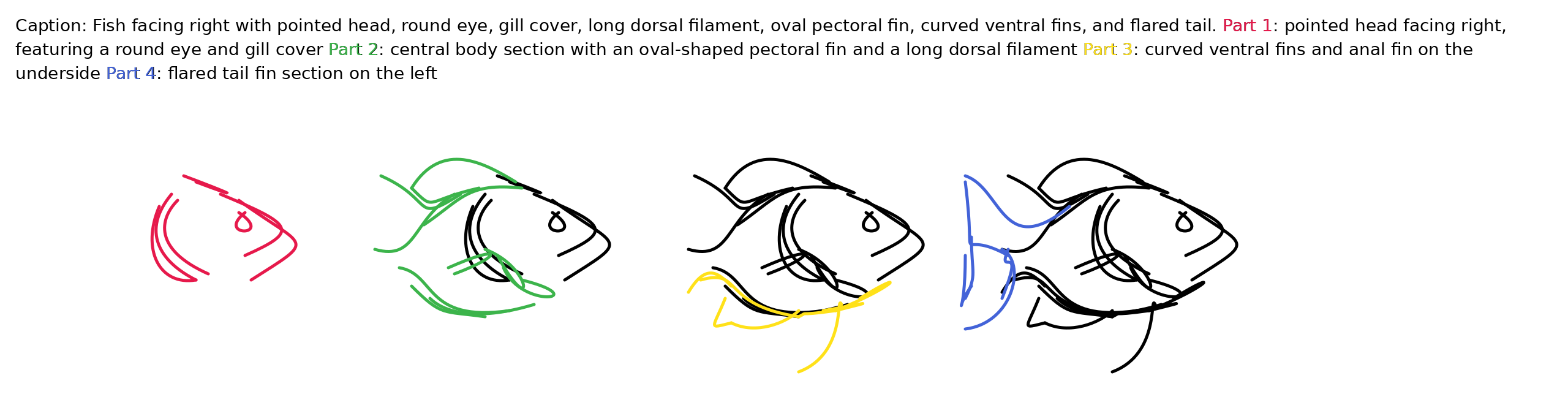}
\end{tabular}
\caption{Additional part-by-part results of our model (continued).}
\end{figure}

\begin{figure}[H]
\centering
\begin{tabular}{c}
    \includegraphics[width=1.0\textwidth]{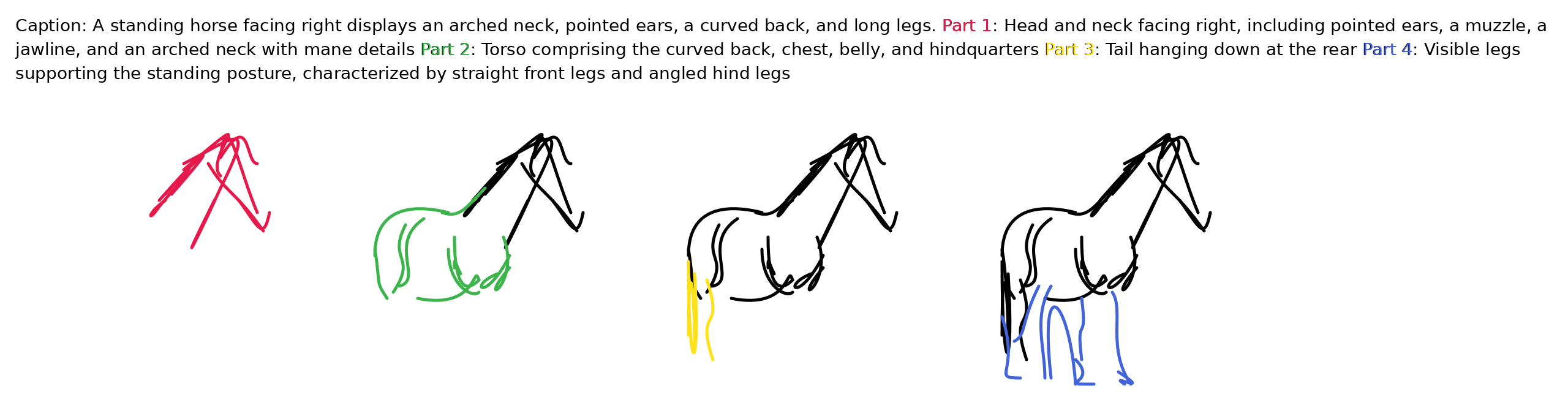} \\
    \includegraphics[width=1.0\textwidth]{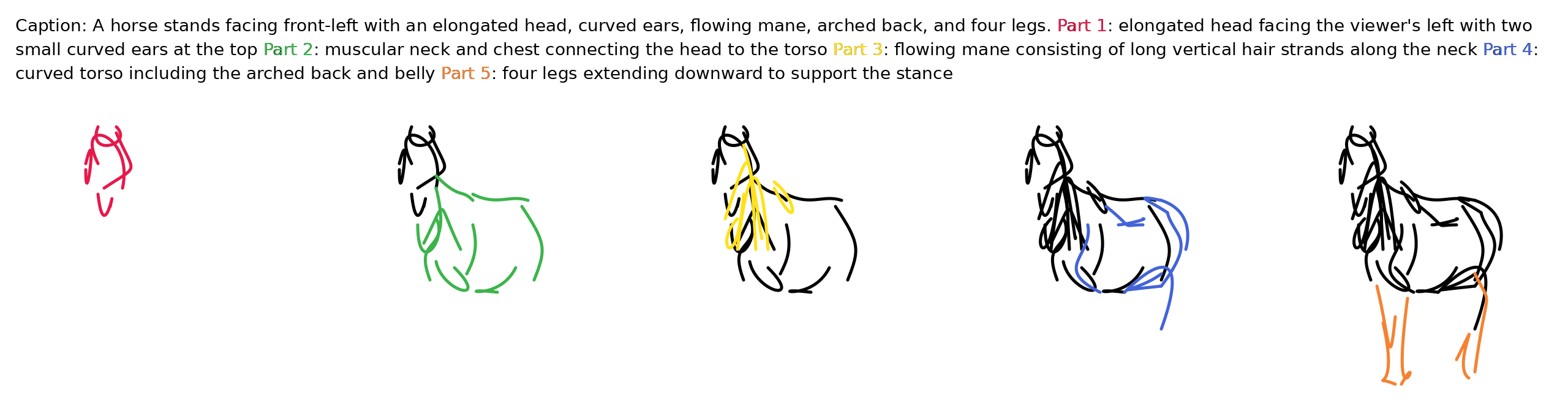} \\
    \includegraphics[width=1.0\textwidth]{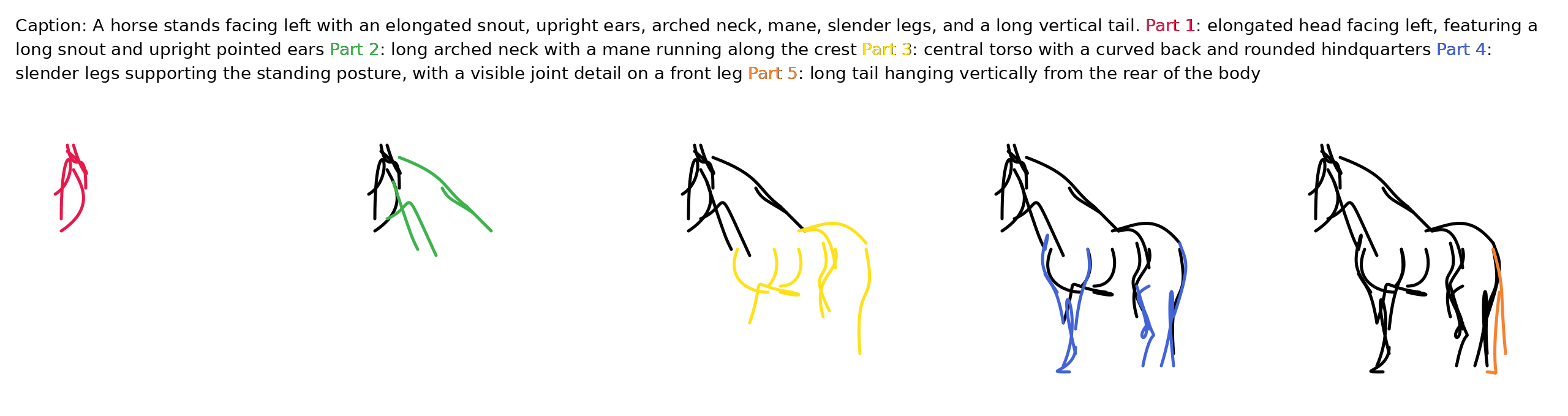} \\
    \includegraphics[width=1.0\textwidth]{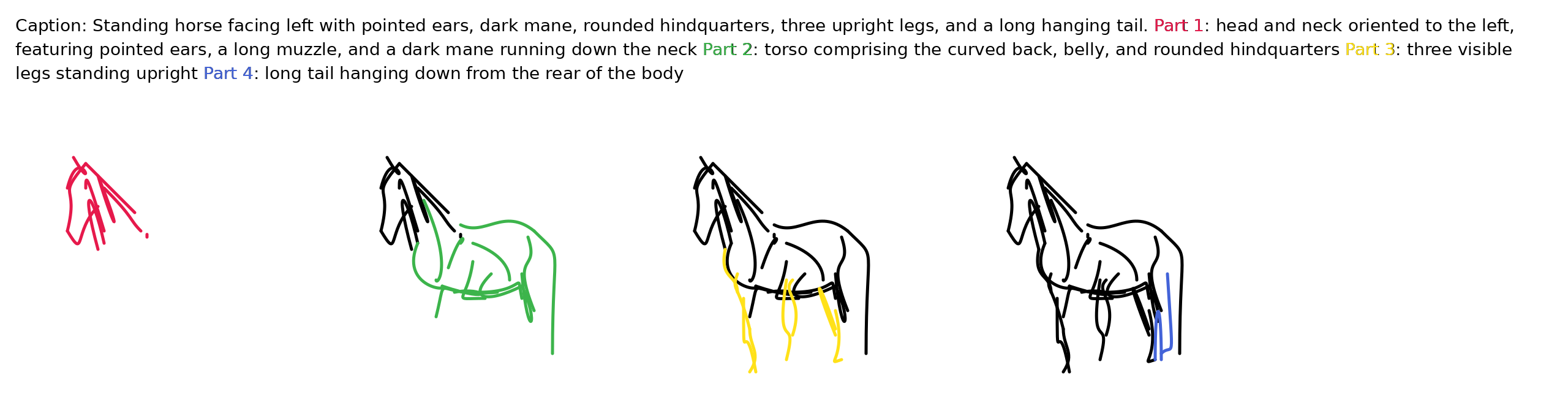} \\
    \includegraphics[width=1.0\textwidth]{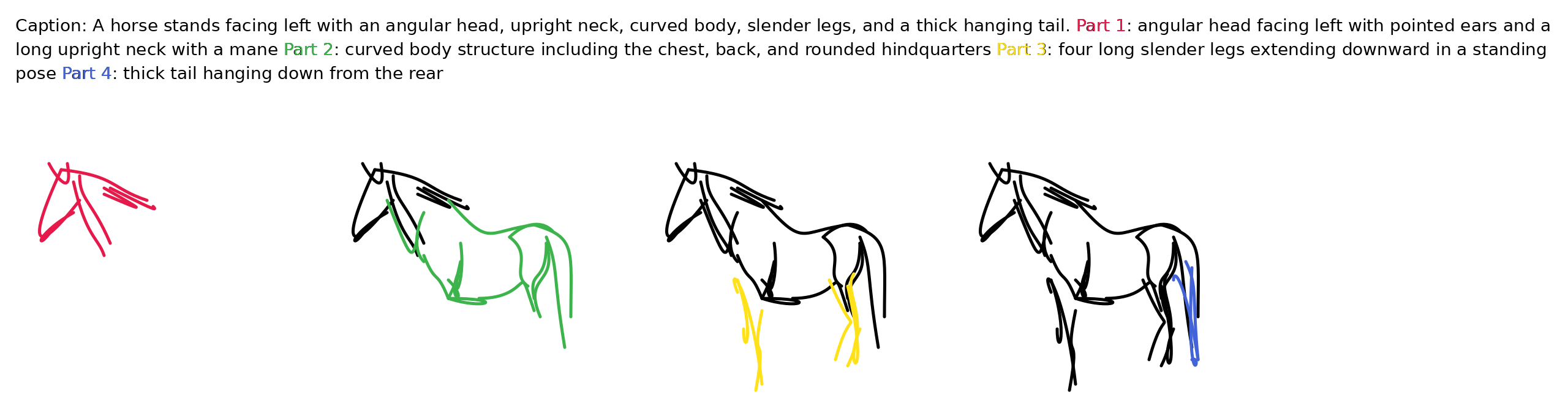}
\end{tabular}
\caption{Additional part-by-part results of our model (continued).}
\end{figure}

\begin{figure}[H]
\centering
\begin{tabular}{c}
    \includegraphics[width=1.0\textwidth]{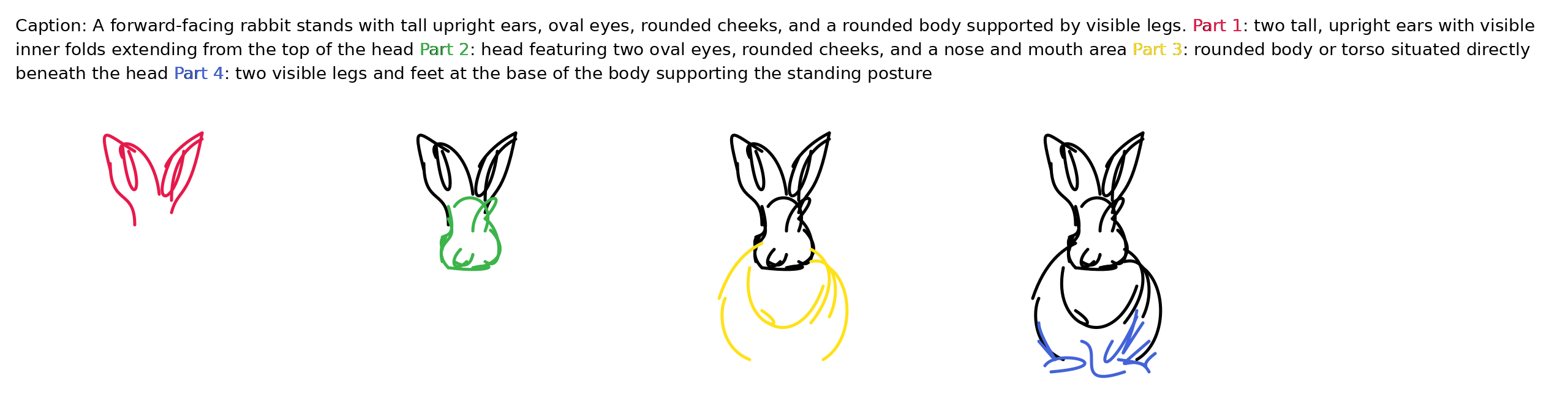} \\
    \includegraphics[width=1.0\textwidth]{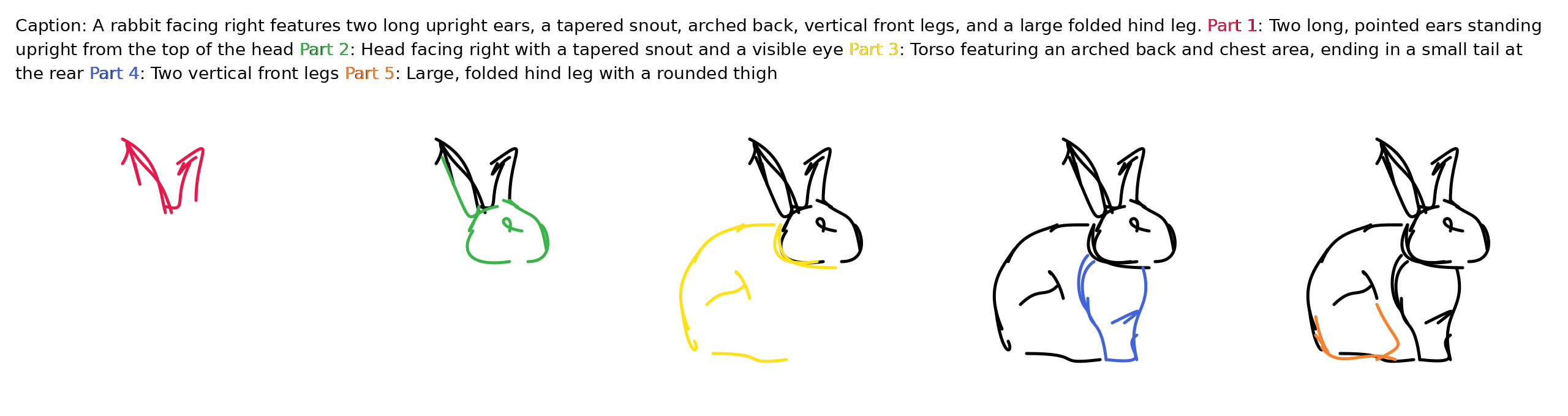} \\
    \includegraphics[width=1.0\textwidth]{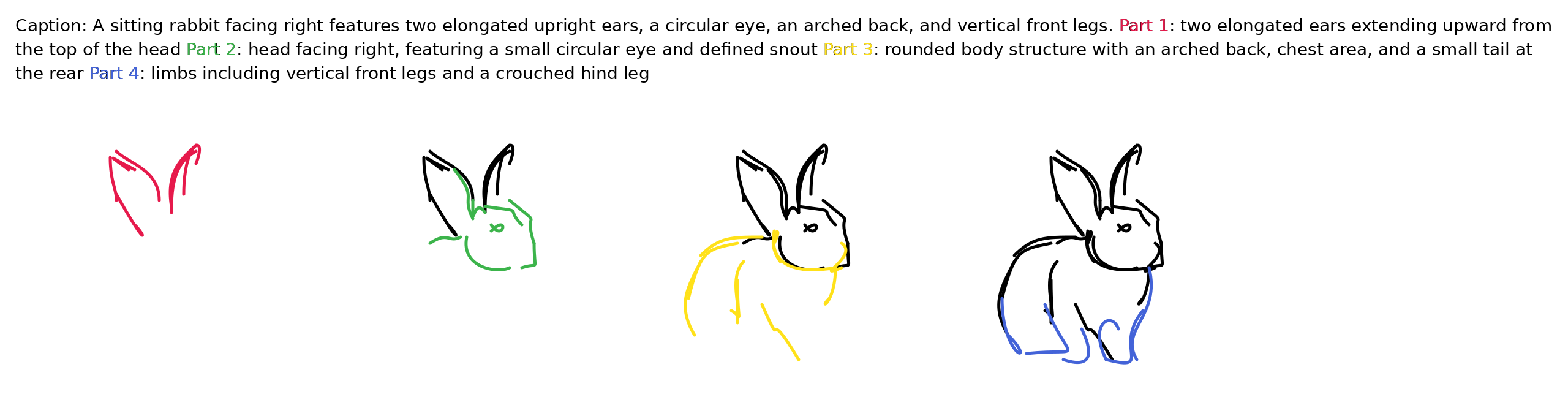} \\
    \includegraphics[width=1.0\textwidth]{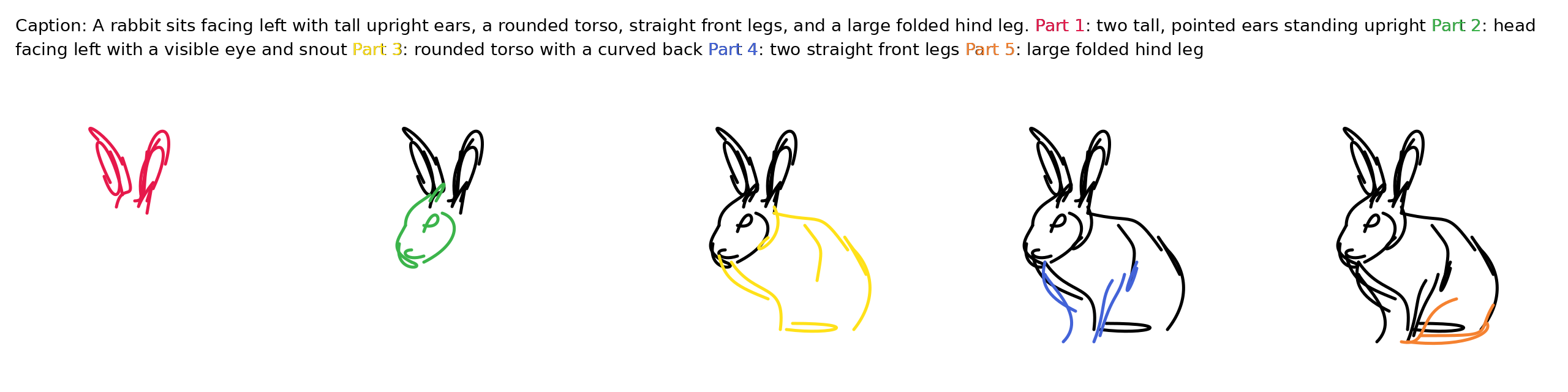} \\
    \includegraphics[width=1.0\textwidth]{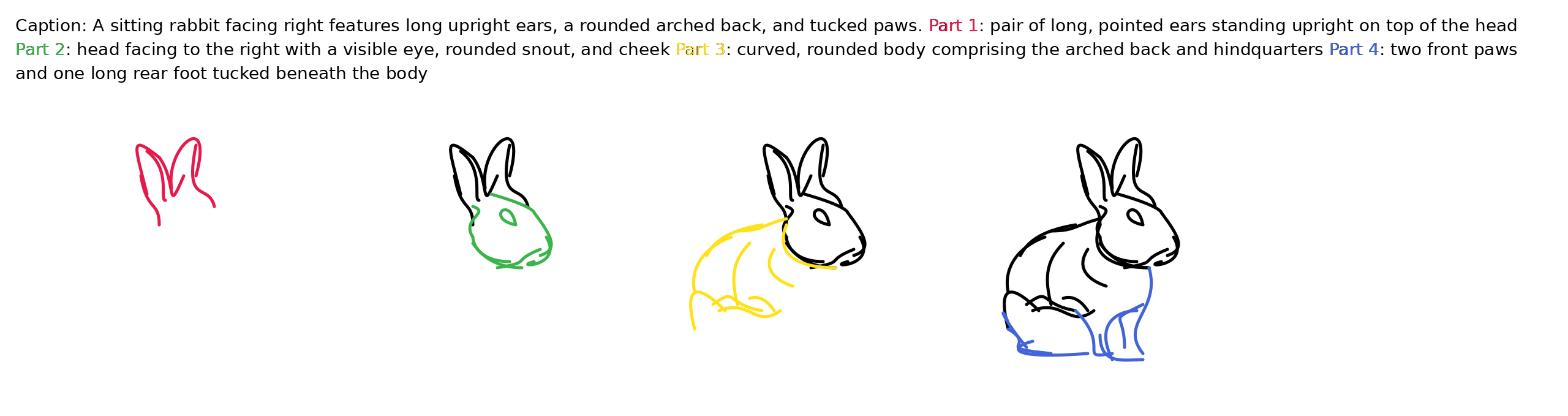}
\end{tabular}
\caption{Additional part-by-part results of our model (continued).}
\end{figure}

\begin{figure}[H]
\centering
\begin{tabular}{c}
    \includegraphics[width=1.0\textwidth]{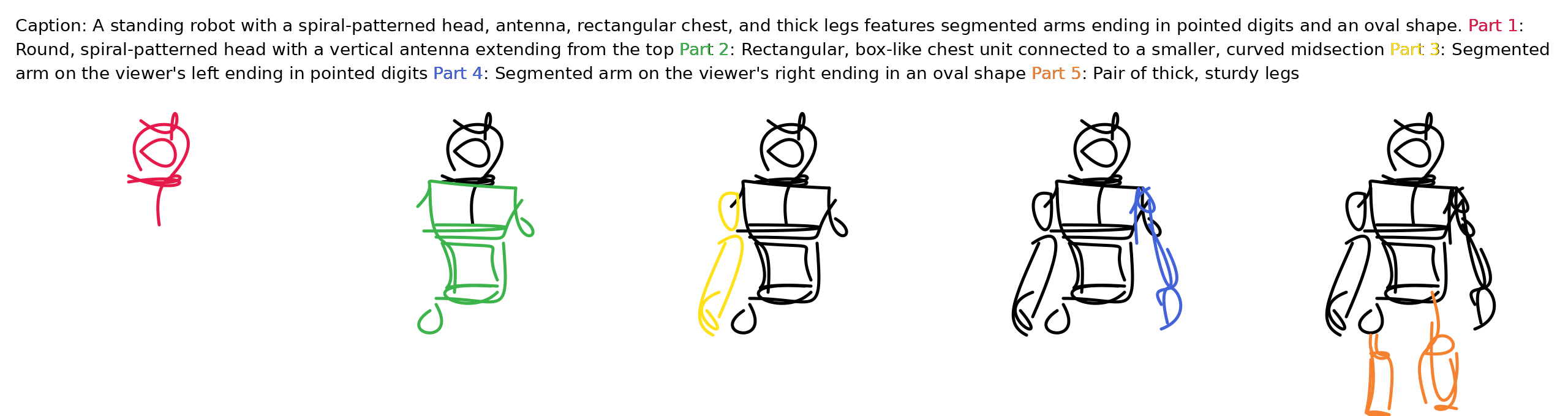} \\
    \includegraphics[width=1.0\textwidth]{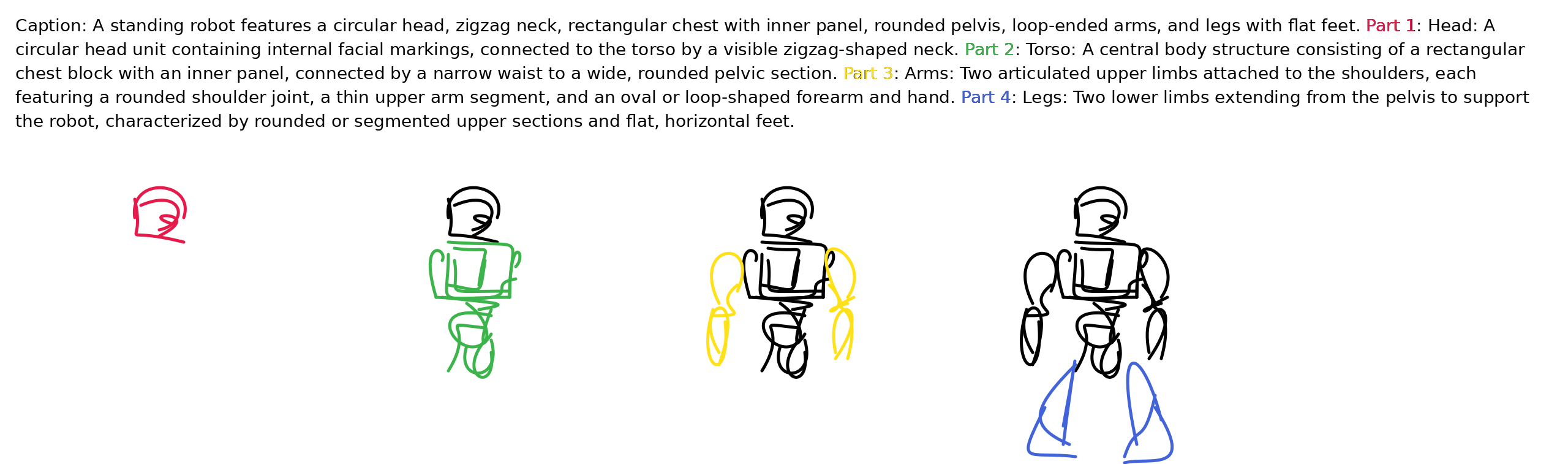} \\
    \includegraphics[width=1.0\textwidth]{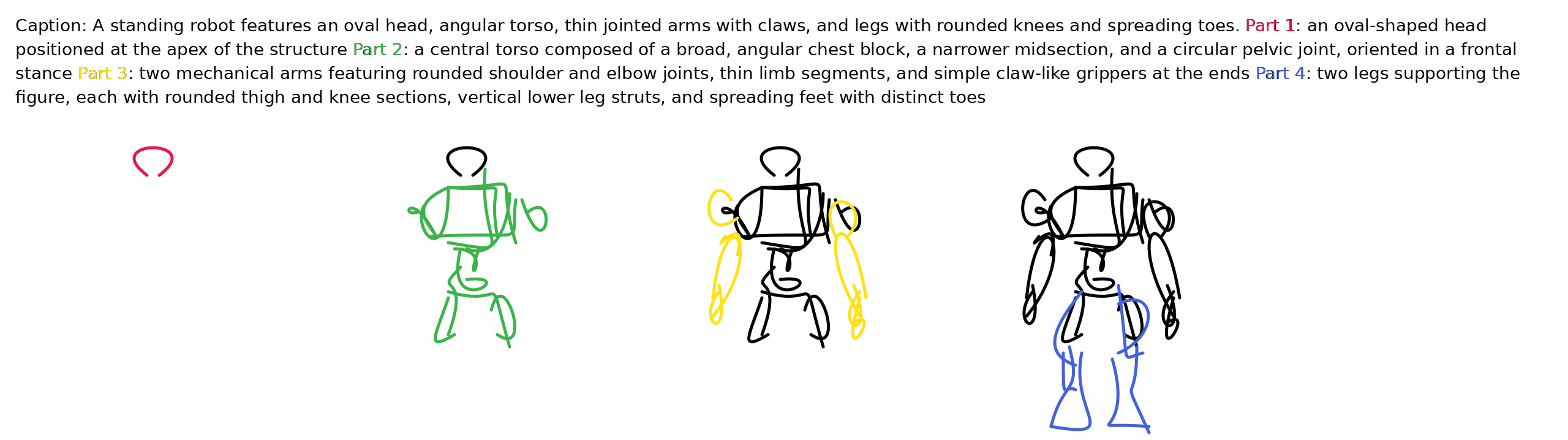} \\
    \includegraphics[width=1.0\textwidth]{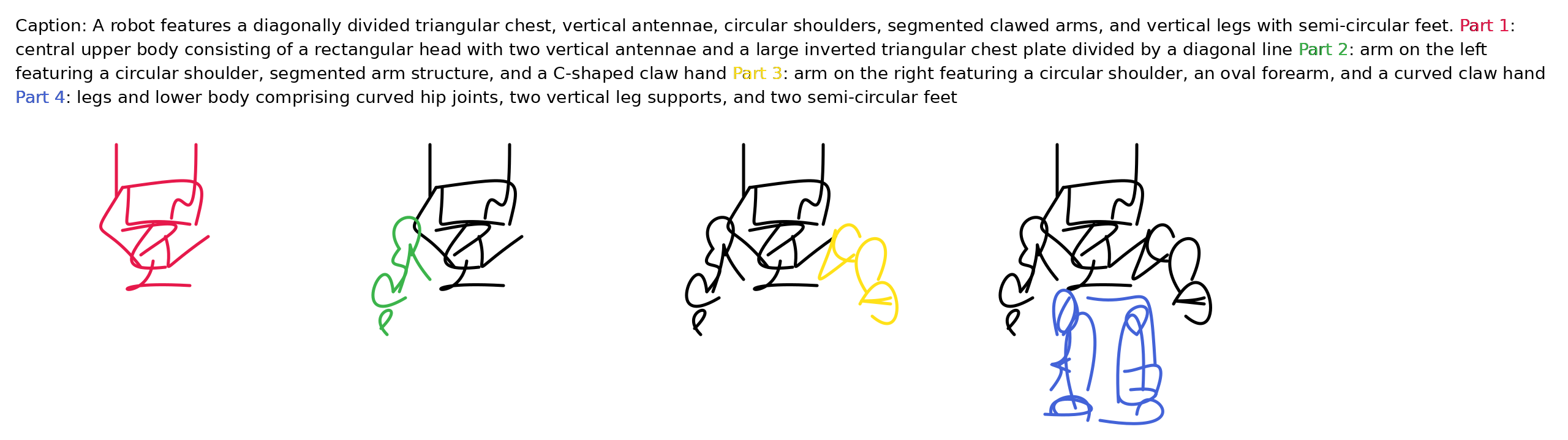} \\
    \includegraphics[width=1.0\textwidth]{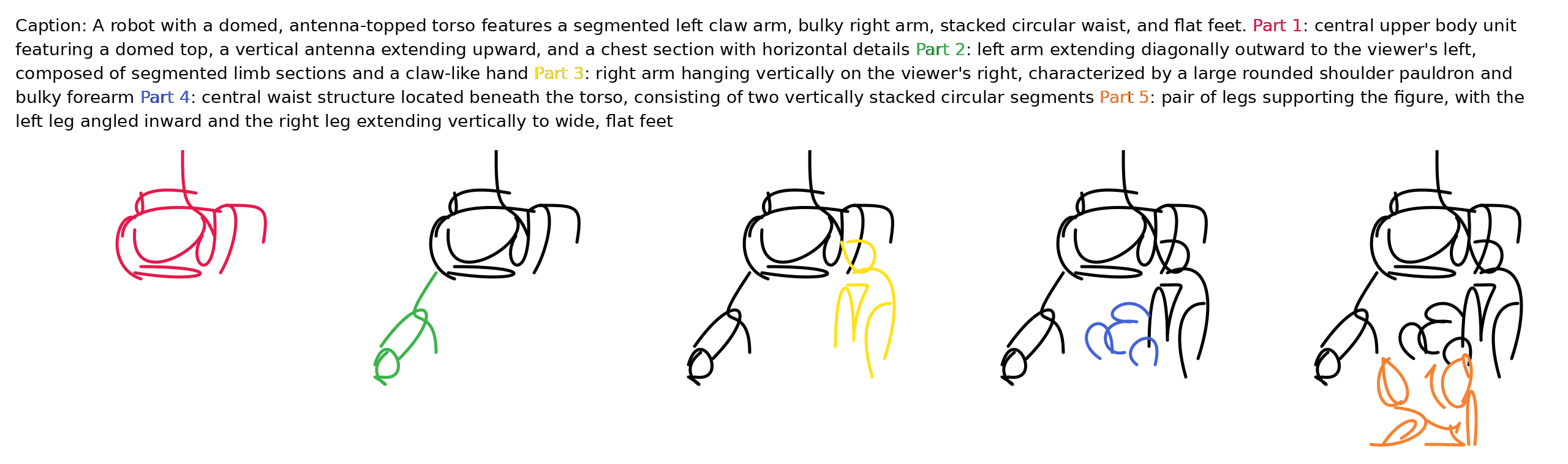}
\end{tabular}
\caption{Additional part-by-part results of our model (continued).}
\end{figure}

\begin{figure}[H]
\centering
\begin{tabular}{c}
    \includegraphics[width=1.0\textwidth]{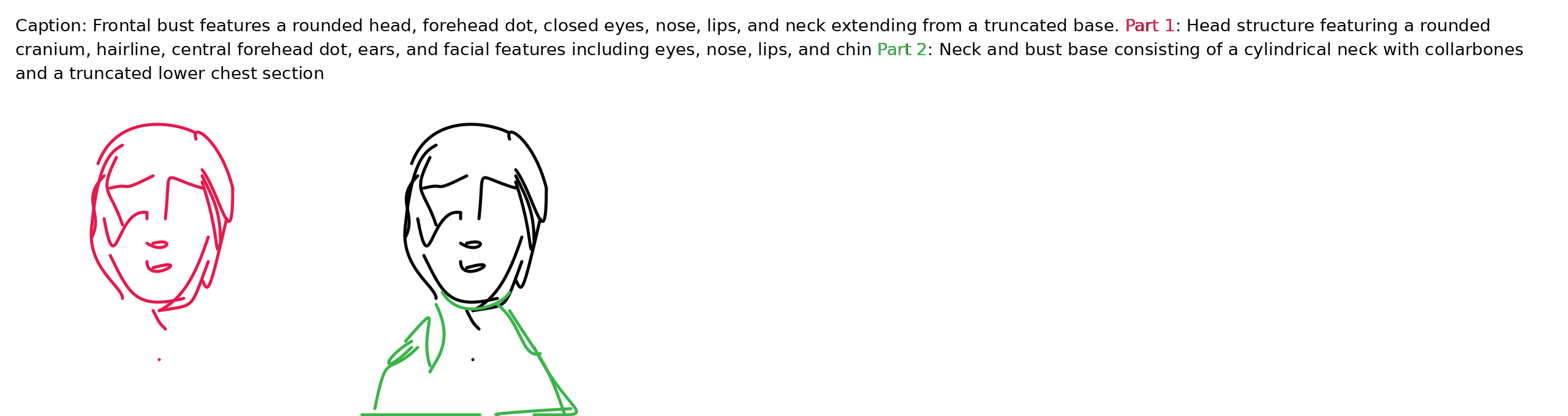} \\
    \includegraphics[width=1.0\textwidth]{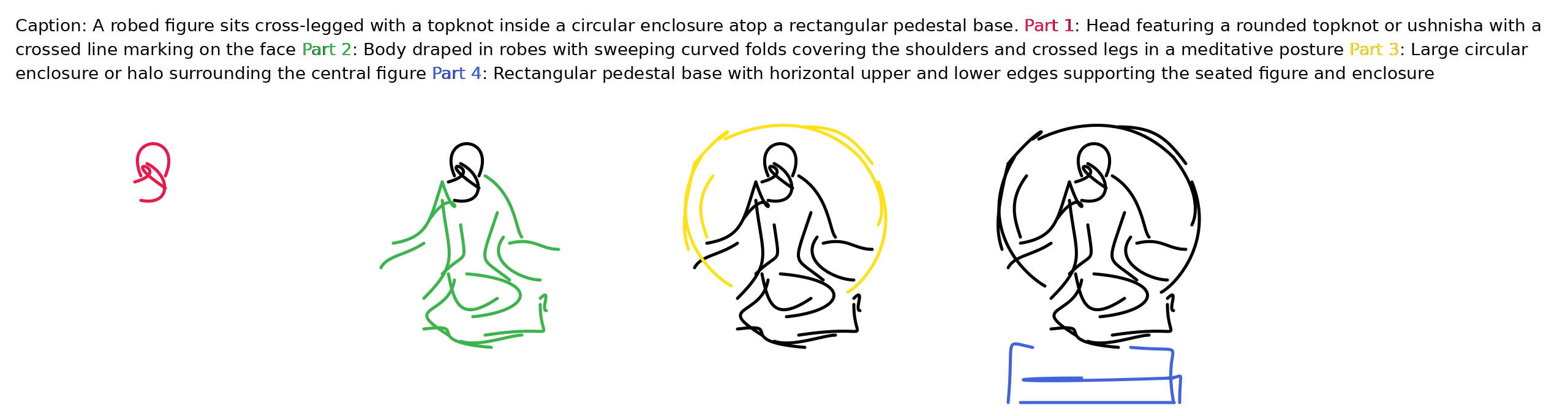} \\
    \includegraphics[width=1.0\textwidth]{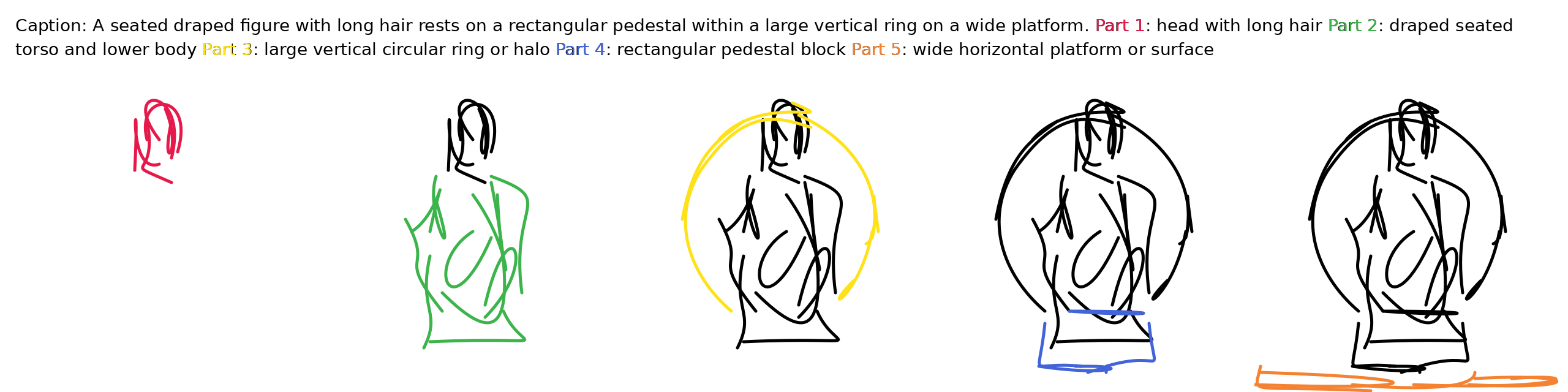} \\
    \includegraphics[width=1.0\textwidth]{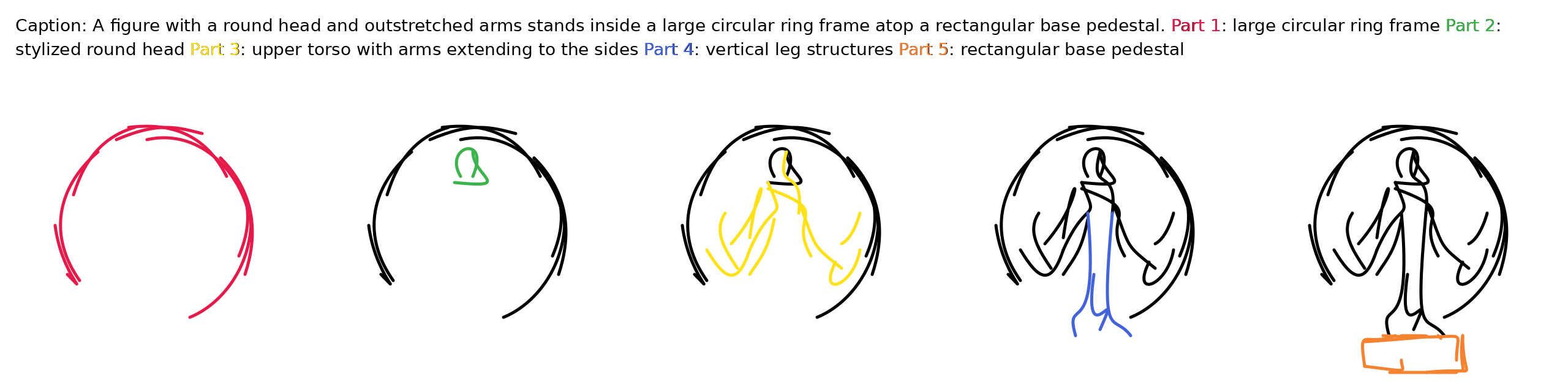} \\
    \includegraphics[width=1.0\textwidth]{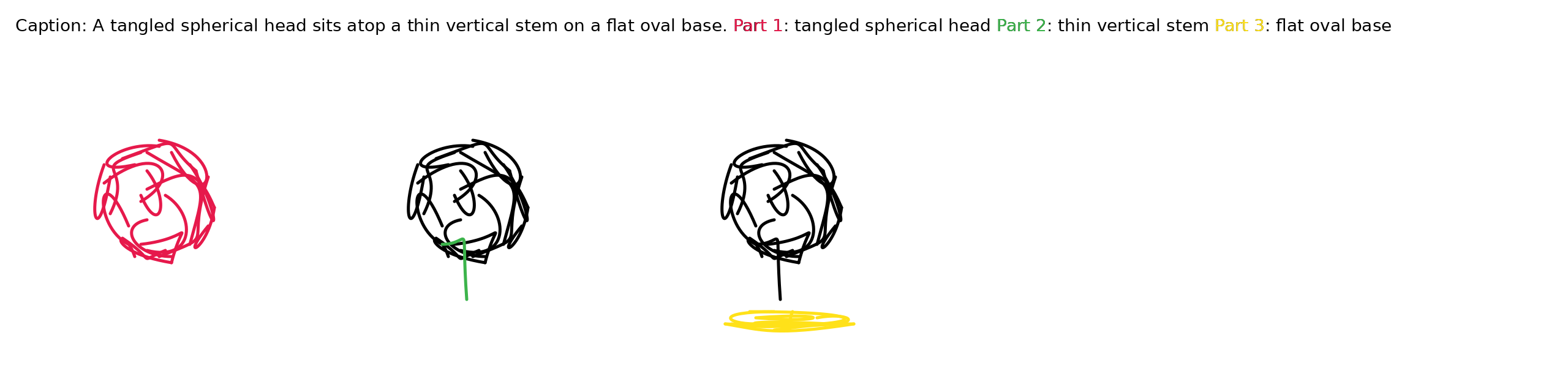}
\end{tabular}
\caption{Additional part-by-part results of our model (continued).}
\end{figure}

\begin{figure}[H]
\centering
\begin{tabular}{c}
    \includegraphics[width=1.0\textwidth]{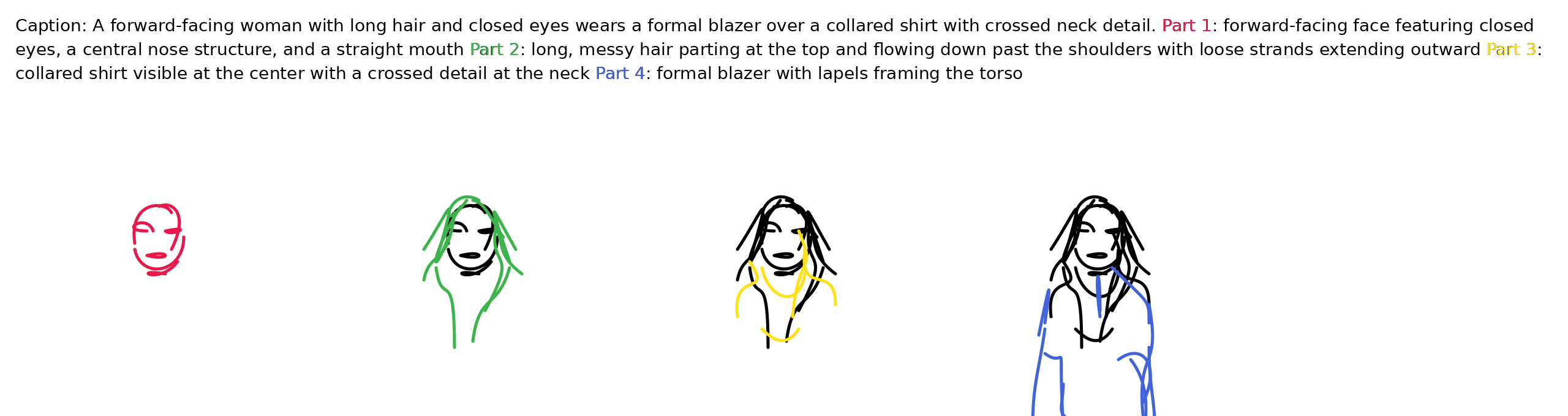} \\
    \includegraphics[width=1.0\textwidth]{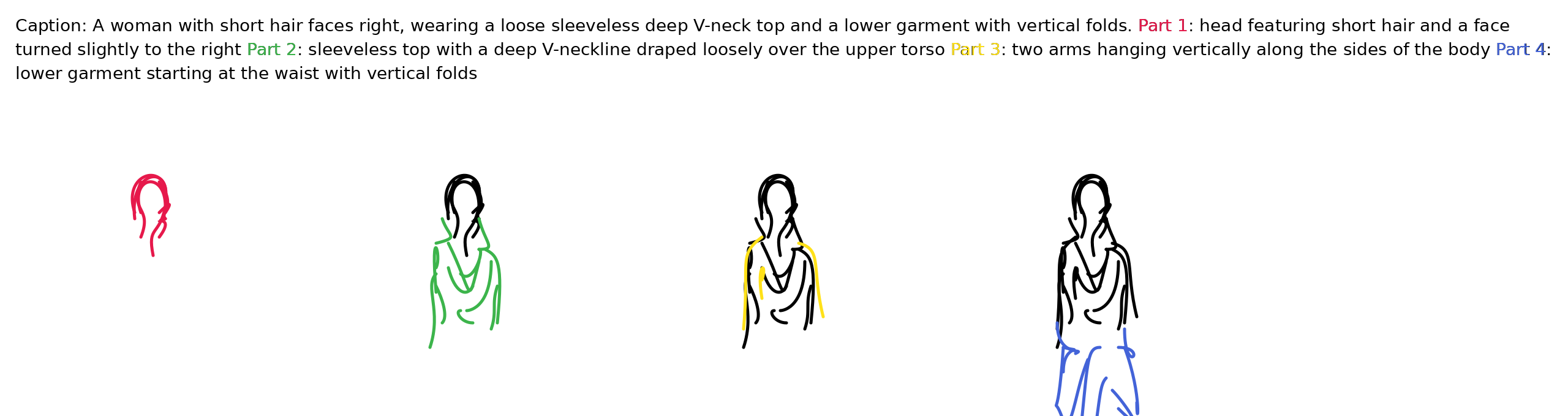} \\
    \includegraphics[width=1.0\textwidth]{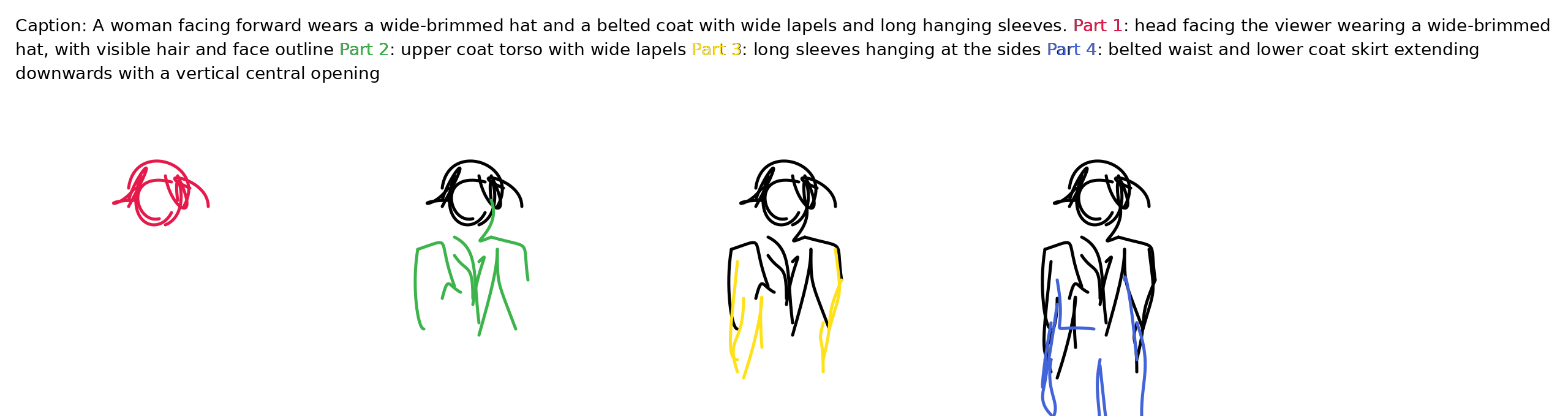} \\
    \includegraphics[width=1.0\textwidth]{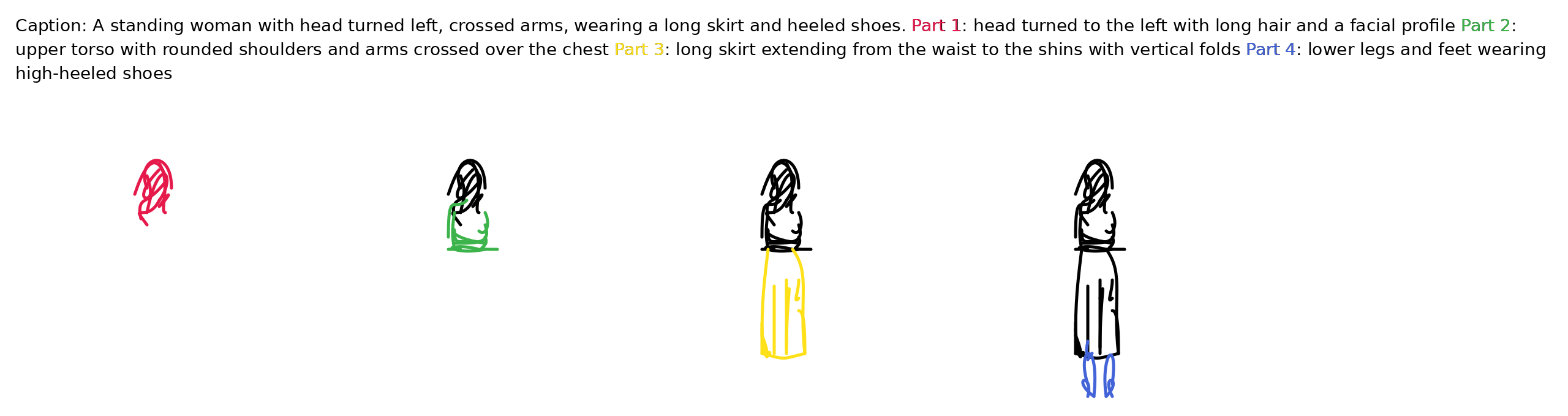} \\
    \includegraphics[width=1.0\textwidth]{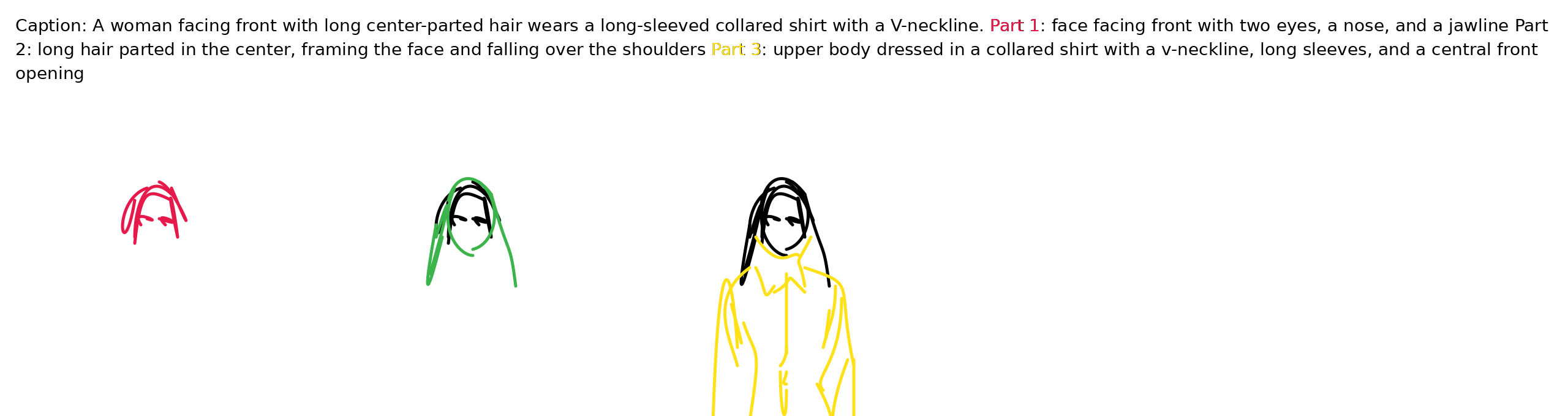}
\end{tabular}
\caption{Additional part-by-part results of our model (continued).}
\end{figure}

\clearpage

\section{Additional ControlSketch-Part Dataset Examples}
\label{sec:supp-additional-dataset-examples}
\FloatBarrier
\begin{table}[H]
\centering
\begin{tabular}{cc}
    \includegraphics[width=0.5\textwidth]{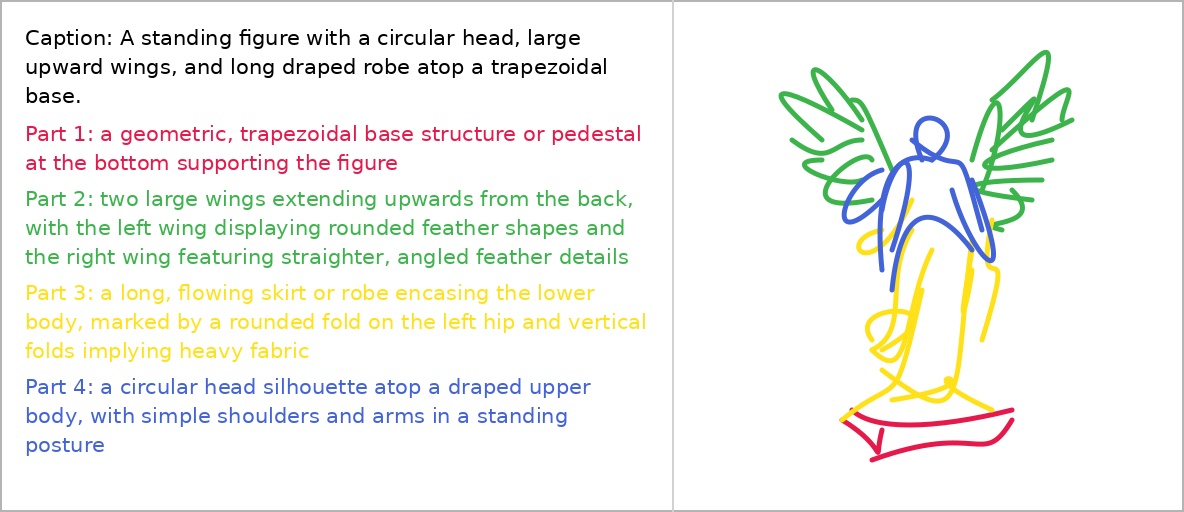} &
    \includegraphics[width=0.5\textwidth]{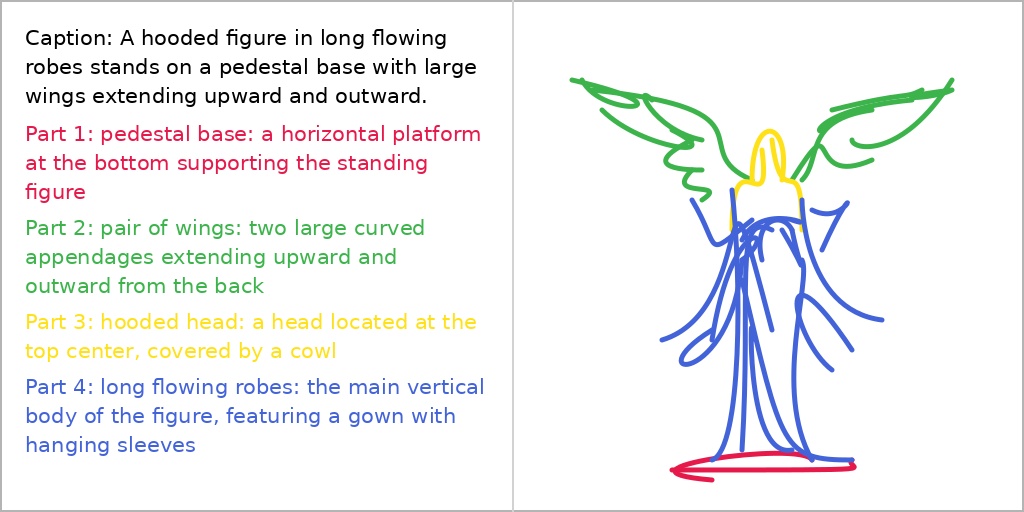} \\
    \includegraphics[width=0.5\textwidth]{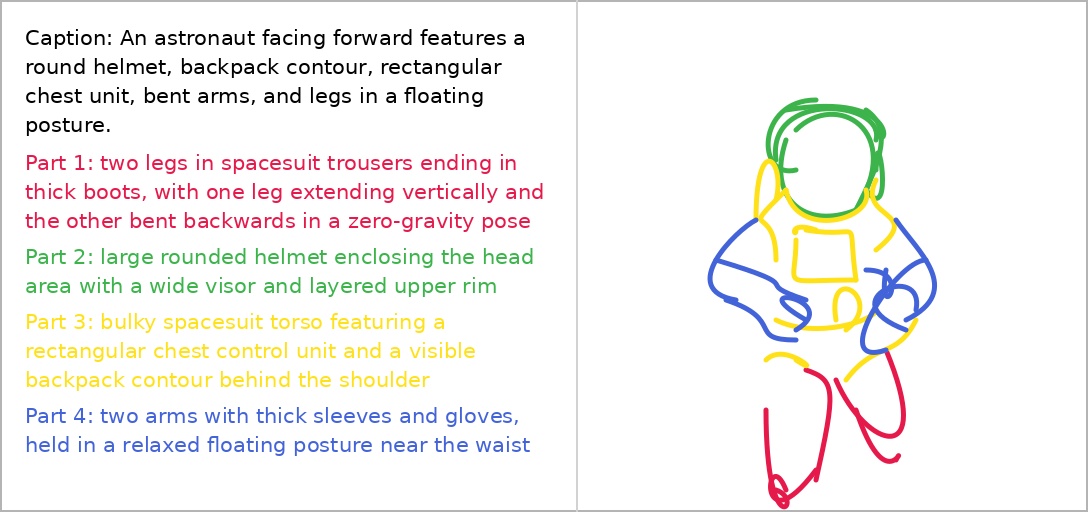} &
    \includegraphics[width=0.5\textwidth]{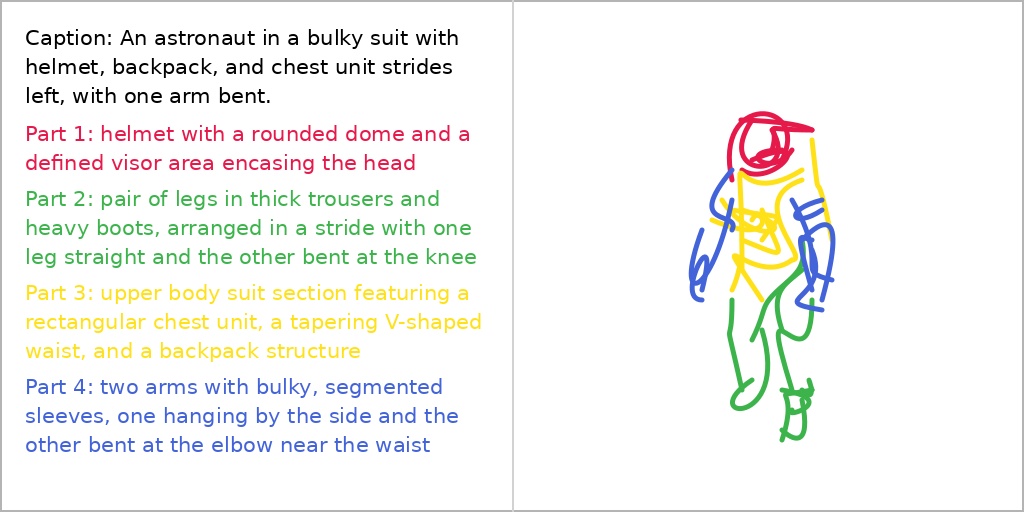} \\
    \includegraphics[width=0.5\textwidth]{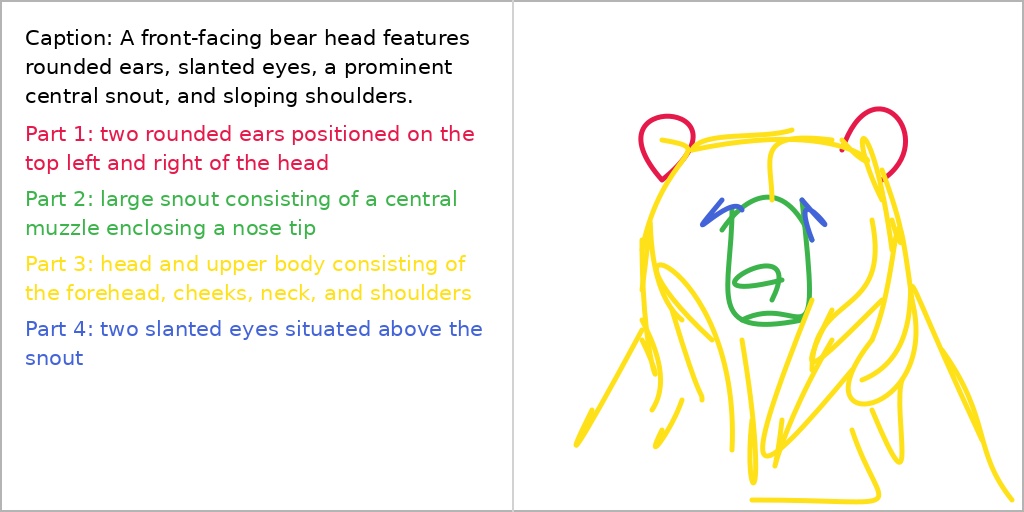} &
    \includegraphics[width=0.5\textwidth]{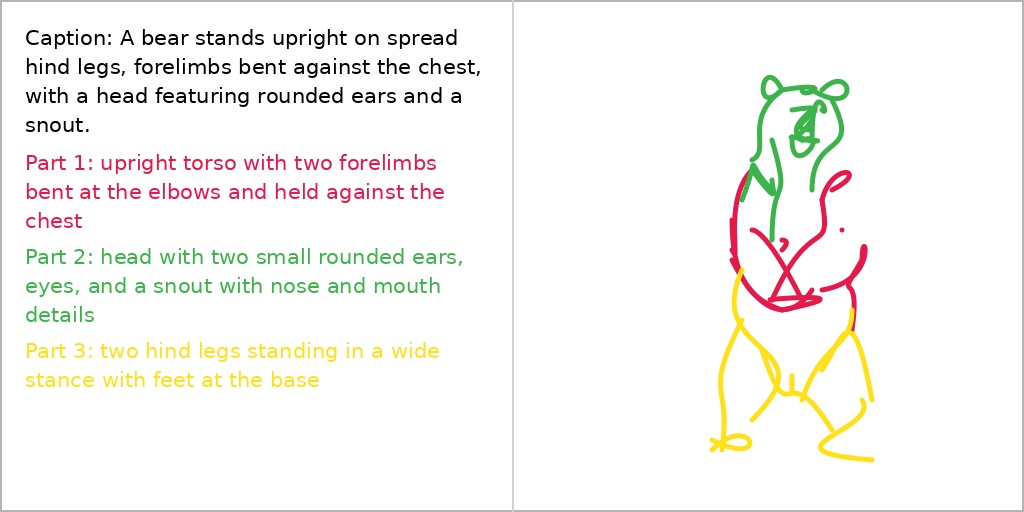} \\
    \includegraphics[width=0.5\textwidth]{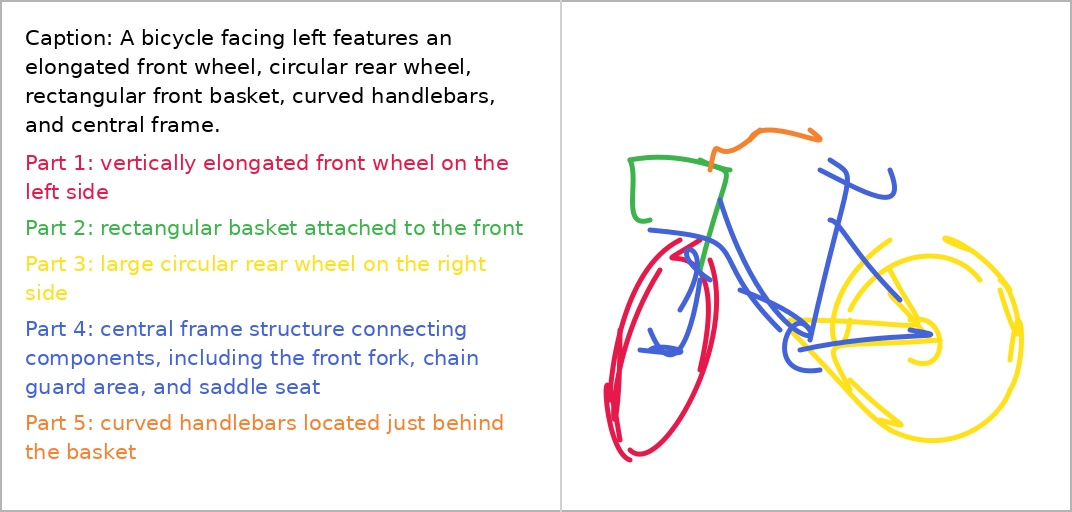} &
    \includegraphics[width=0.5\textwidth]{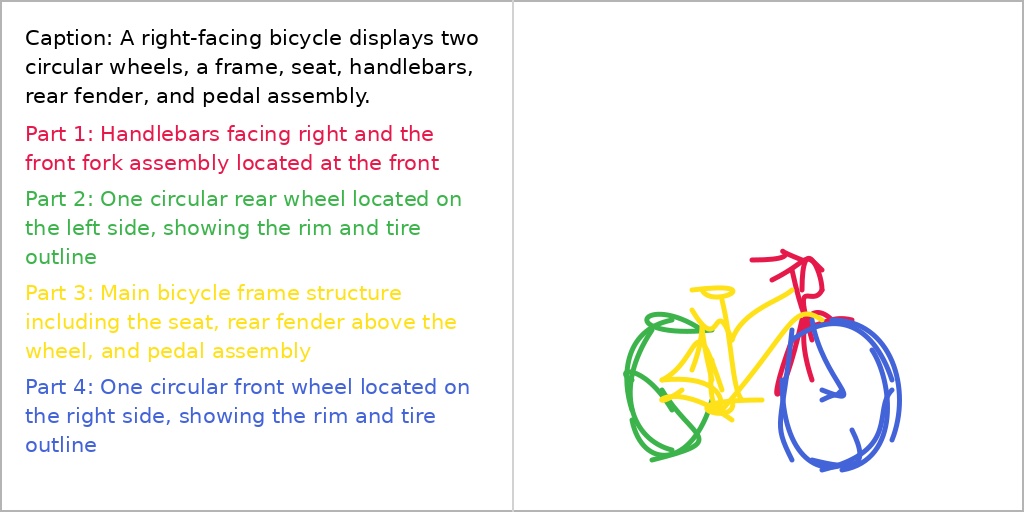} \\
    \includegraphics[width=0.5\textwidth]{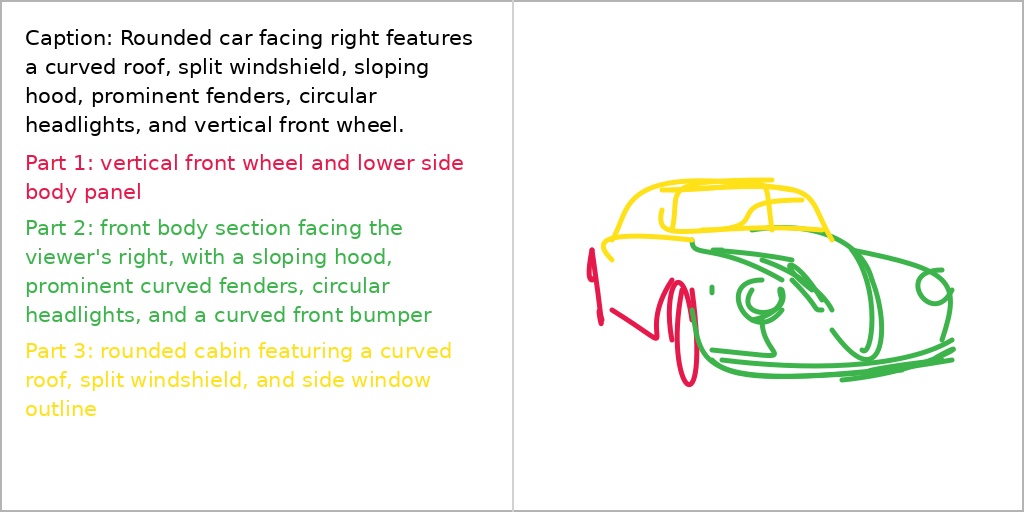} &
    \includegraphics[width=0.5\textwidth]{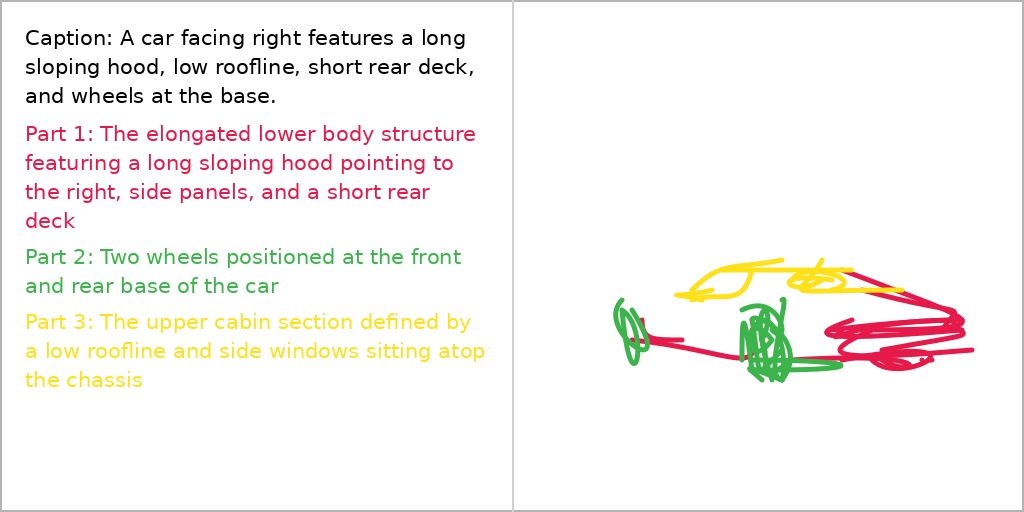} \\
\end{tabular}
\caption{Additional Examples of the ControlSketch-Part Dataset.}
\end{table}

\begin{table}[H]
\centering
\begin{tabular}{cc}
    \includegraphics[width=0.5\textwidth]{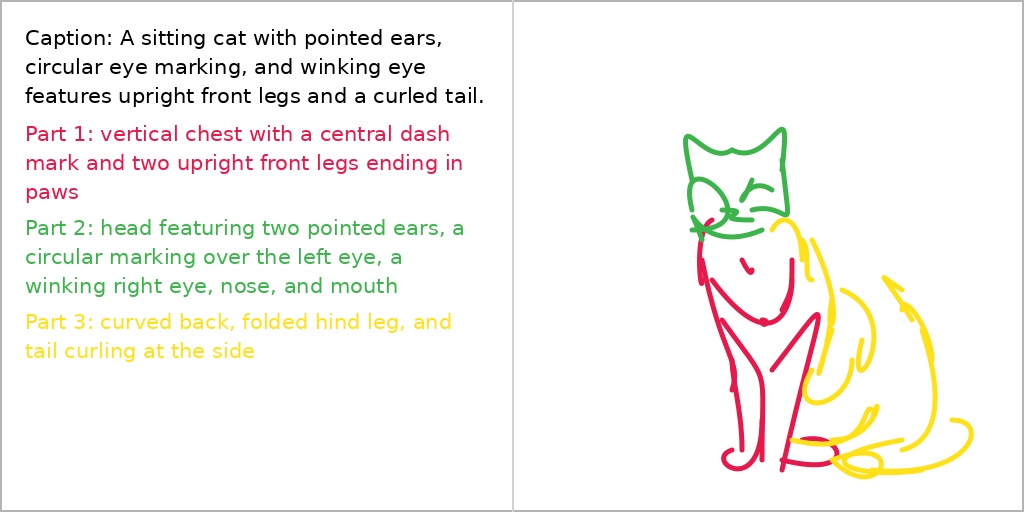} &
    \includegraphics[width=0.5\textwidth]{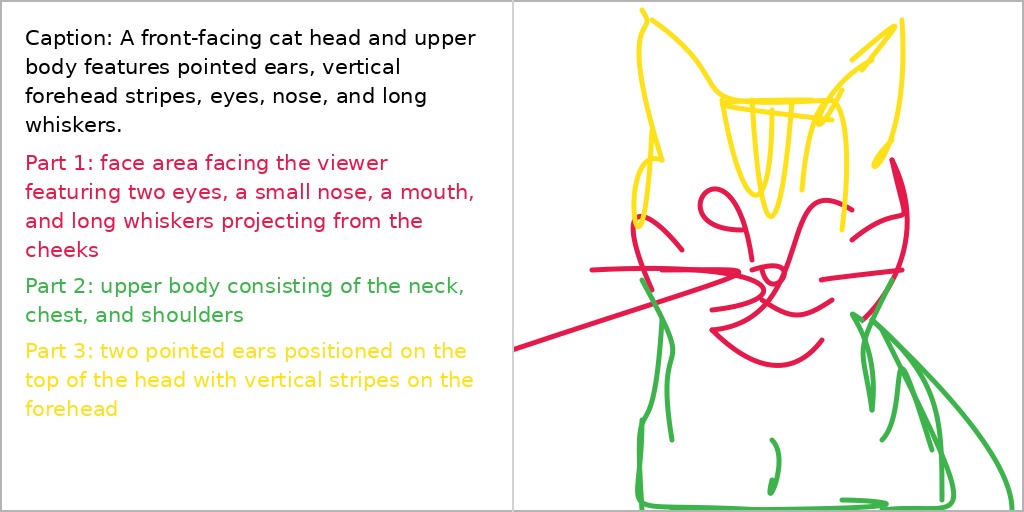} \\
    \includegraphics[width=0.5\textwidth]{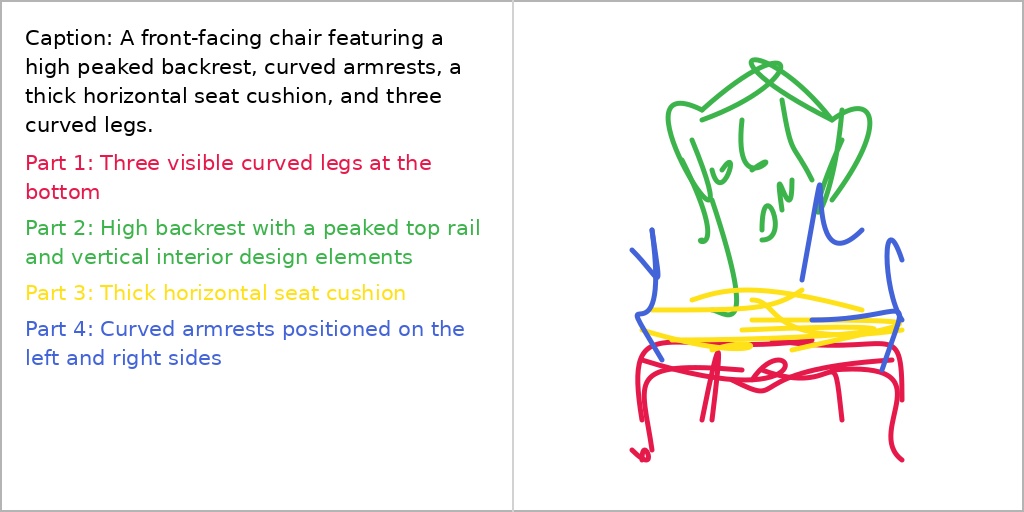} &
    \includegraphics[width=0.5\textwidth]{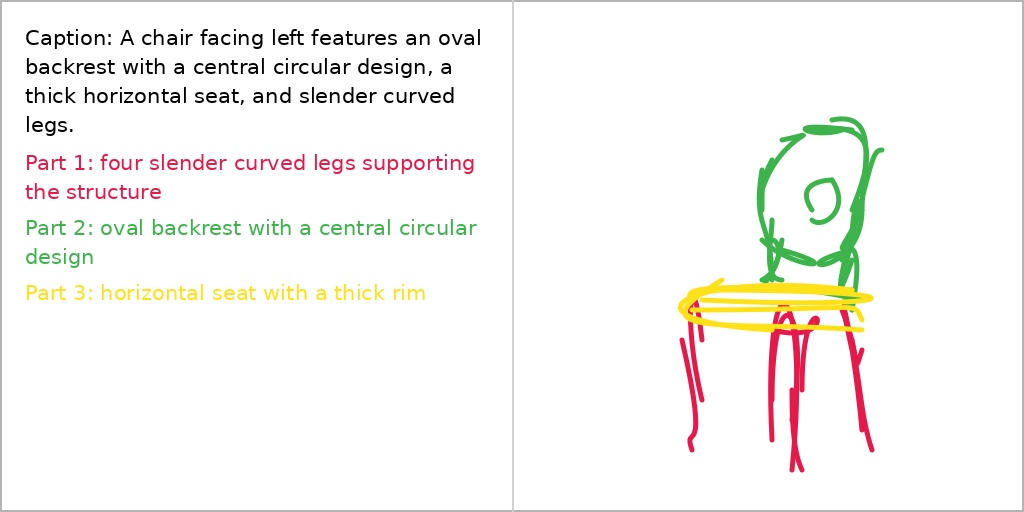} \\
    \includegraphics[width=0.5\textwidth]{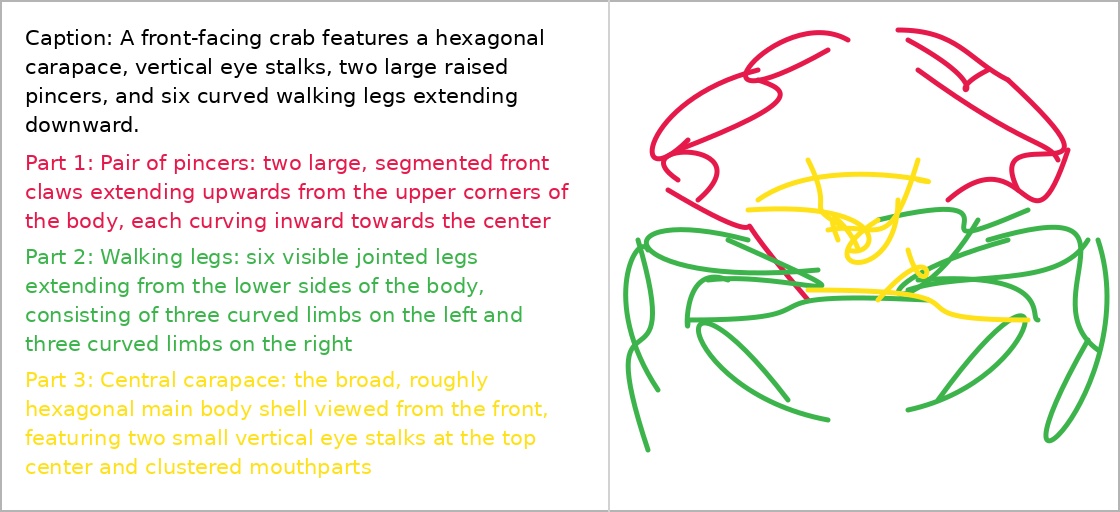} &
    \includegraphics[width=0.5\textwidth]{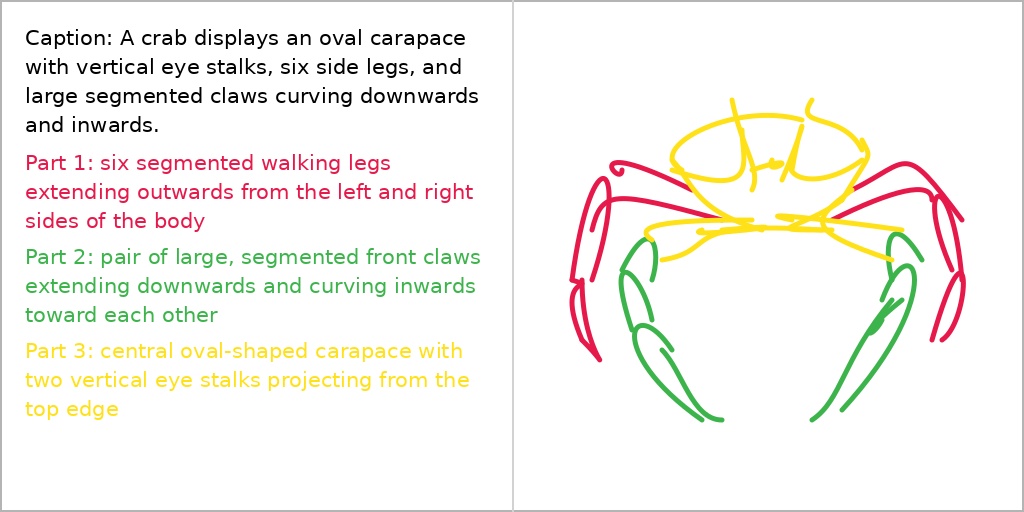} \\
    \includegraphics[width=0.5\textwidth]{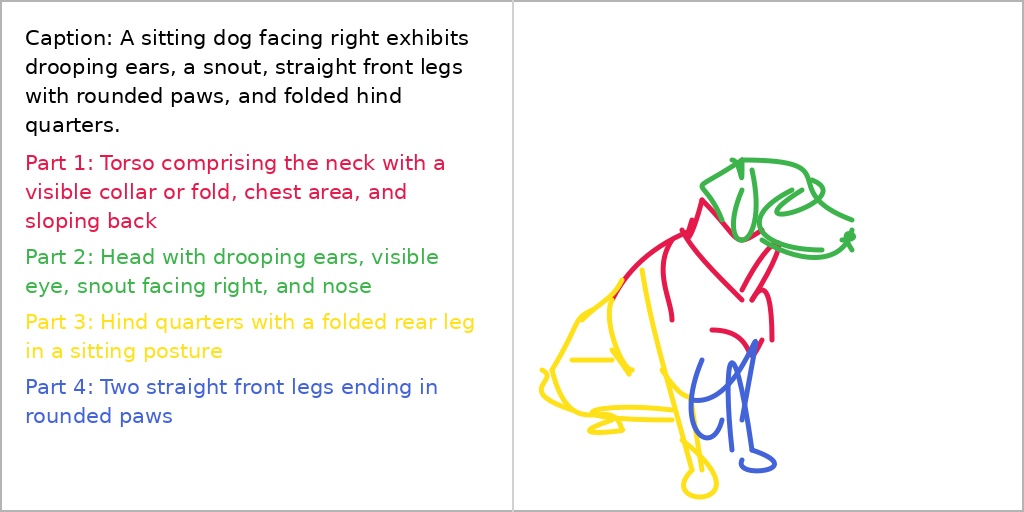} &
    \includegraphics[width=0.5\textwidth]{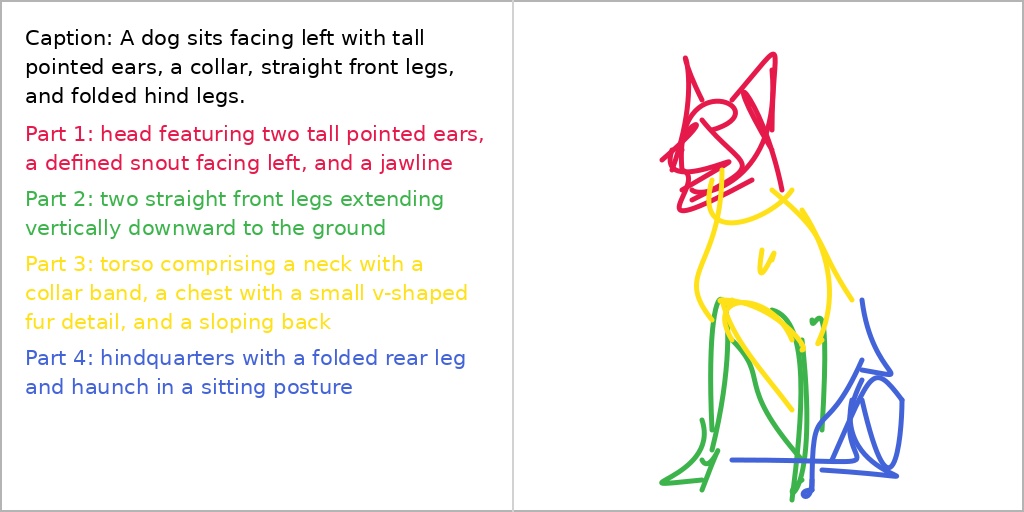} \\
    \includegraphics[width=0.5\textwidth]{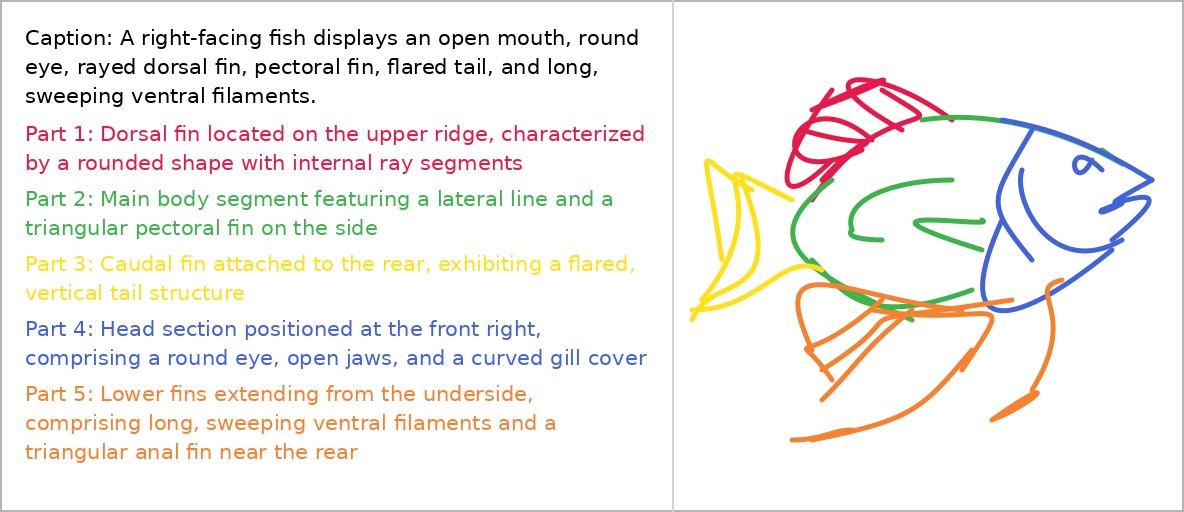} &
    \includegraphics[width=0.5\textwidth]{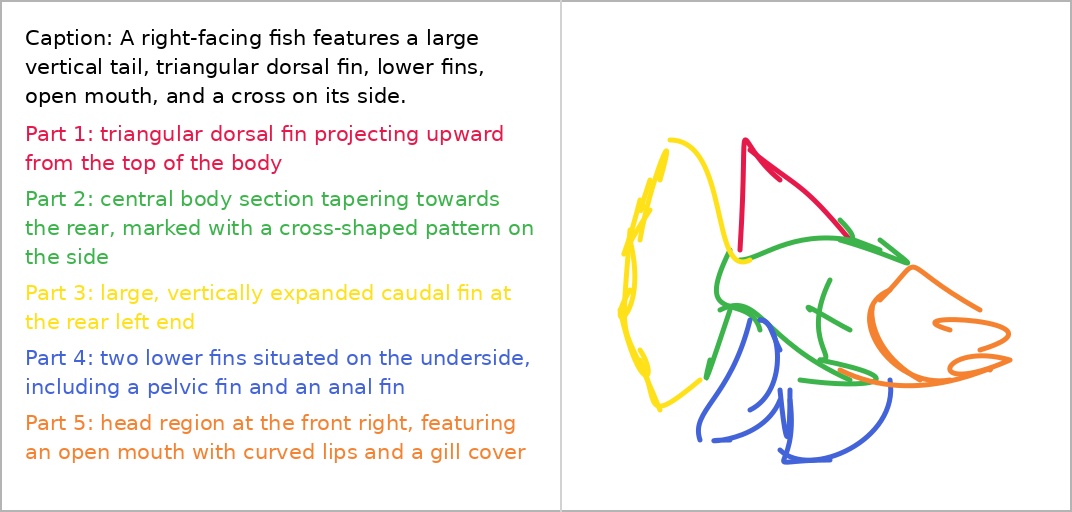} \\
\end{tabular}
\caption{Additional Examples of the ControlSketch-Part Dataset (continued).}
\end{table}

\begin{table}[H]
\centering
\begin{tabular}{cc}
    \includegraphics[width=0.5\textwidth]{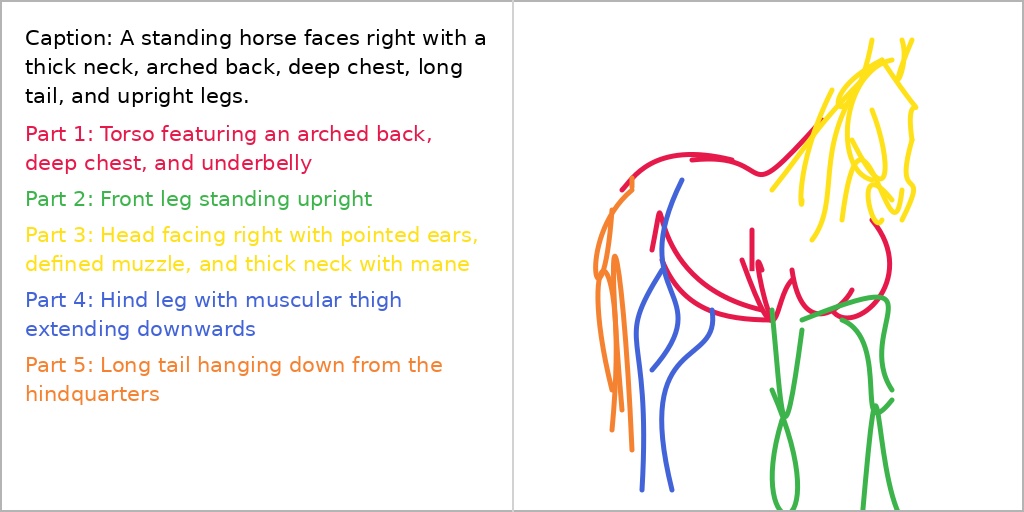} &
    \includegraphics[width=0.5\textwidth]{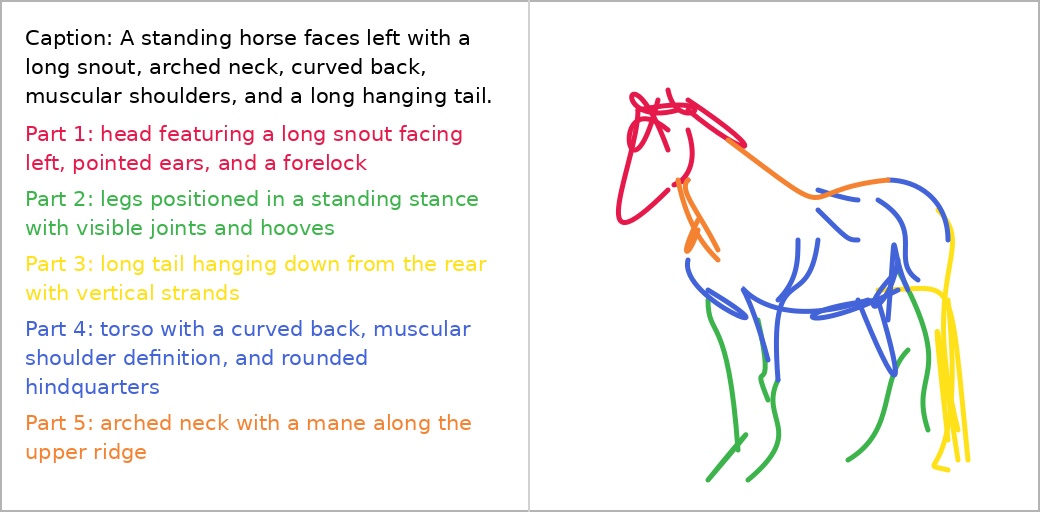} \\
    \includegraphics[width=0.5\textwidth]{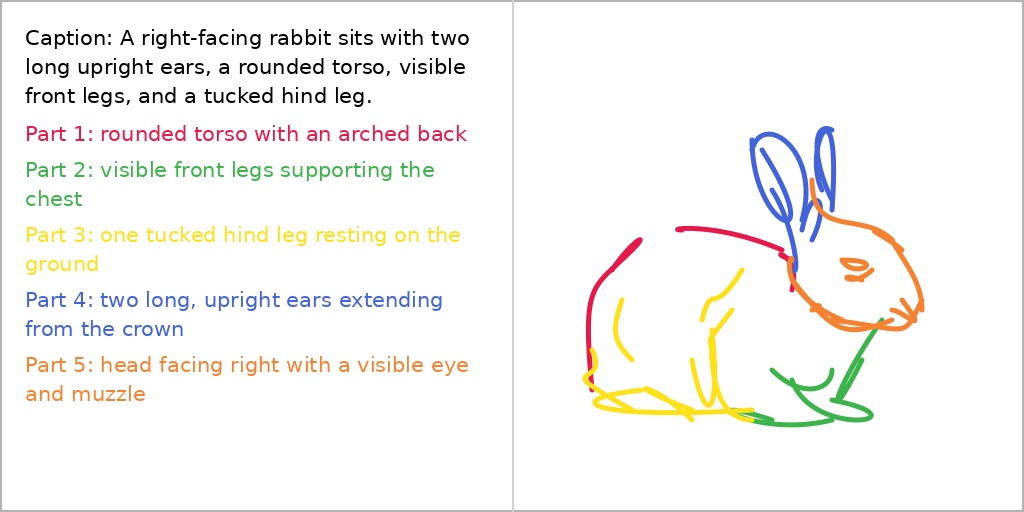} &
    \includegraphics[width=0.5\textwidth]{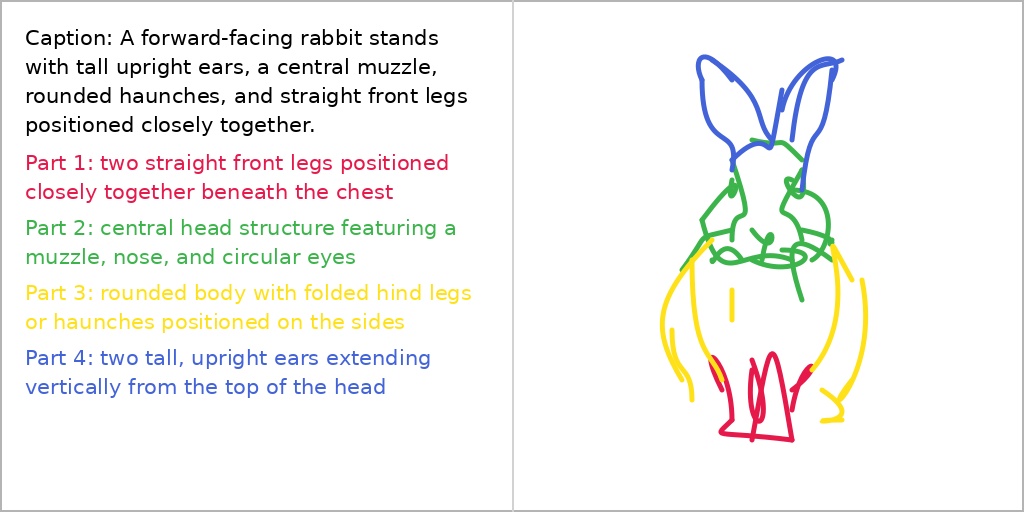} \\
    \includegraphics[width=0.5\textwidth]{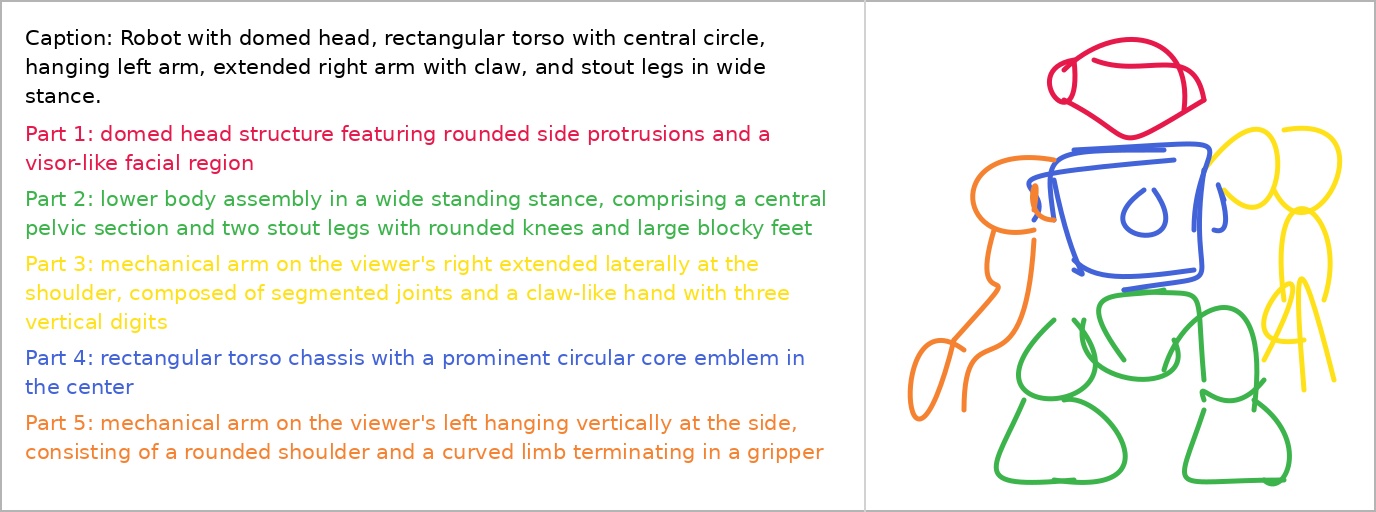} &
    \includegraphics[width=0.5\textwidth]{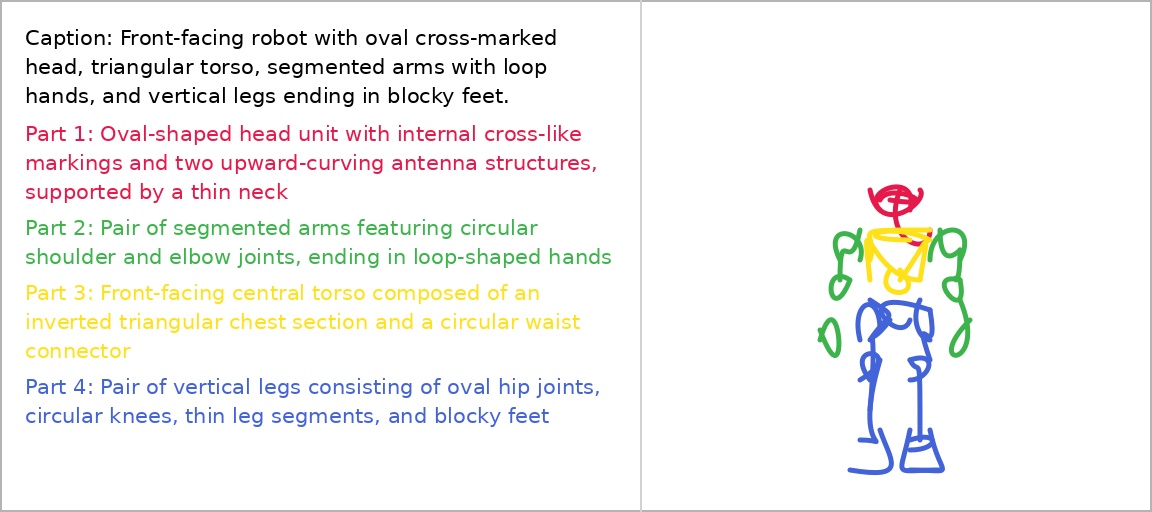} \\
    \includegraphics[width=0.5\textwidth]{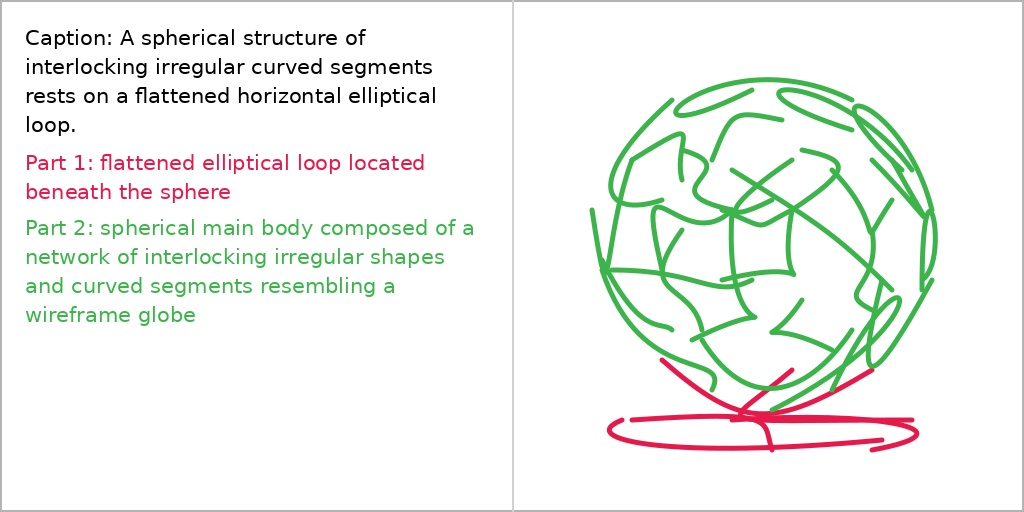} &
    \includegraphics[width=0.5\textwidth]{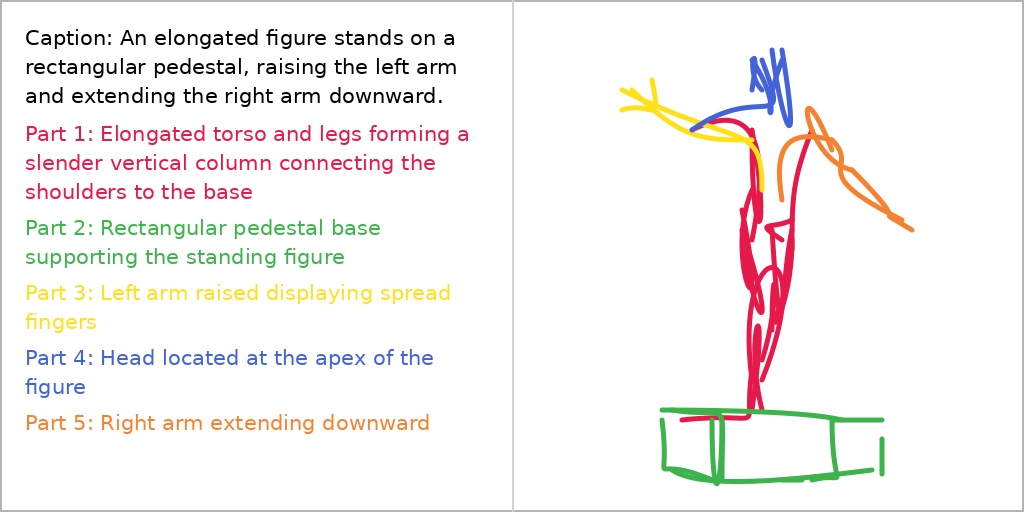} \\
    \includegraphics[width=0.5\textwidth]{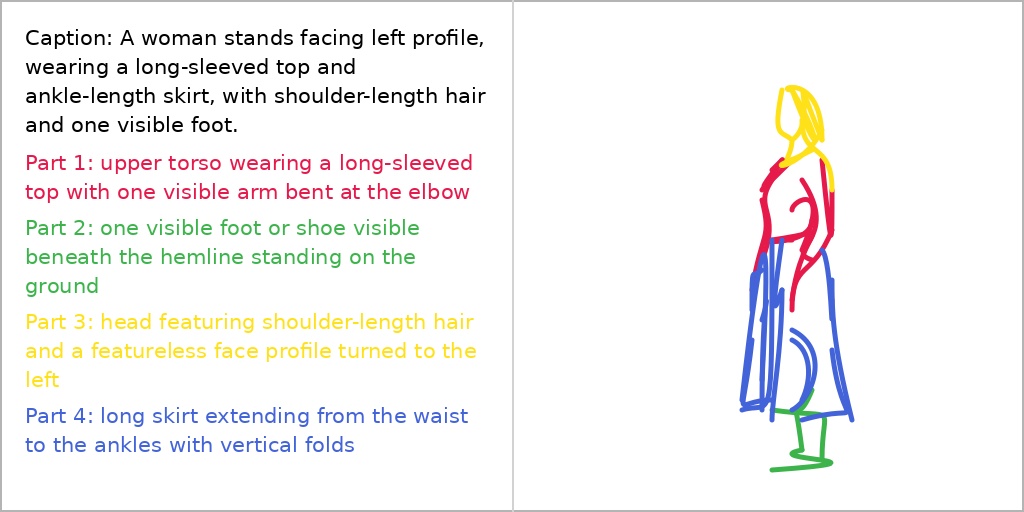} &
    \includegraphics[width=0.5\textwidth]{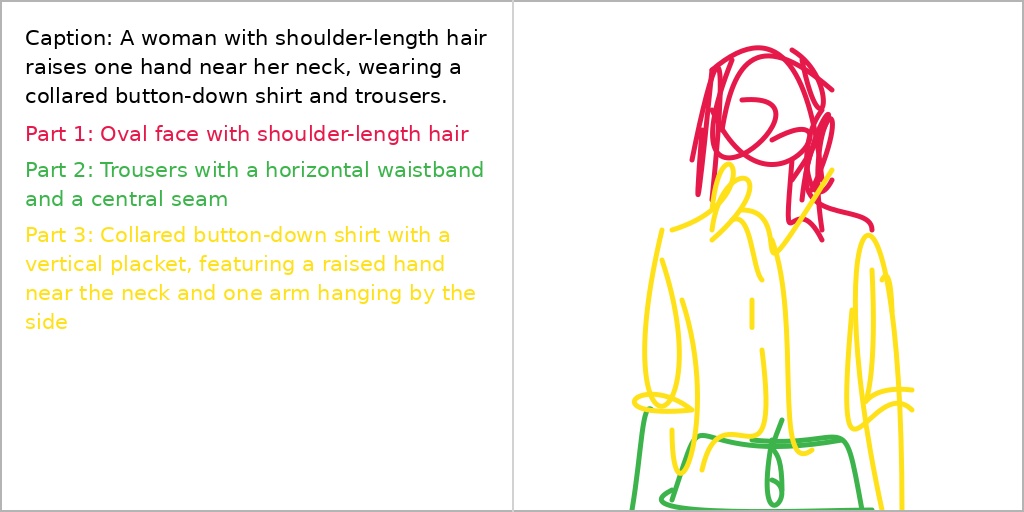} \\
\end{tabular}
\caption{Additional Examples of the ControlSketch-Part Dataset (continued).}
\end{table}

\clearpage
\section{Failure cases, Limitations and Future Work}
\label{sec:supp-limitations}
\subsection{Failure cases}
\begin{figure}[H]
\centering

\begin{subfigure}{\textwidth}
    \centering
    \includegraphics[width=0.8\textwidth]{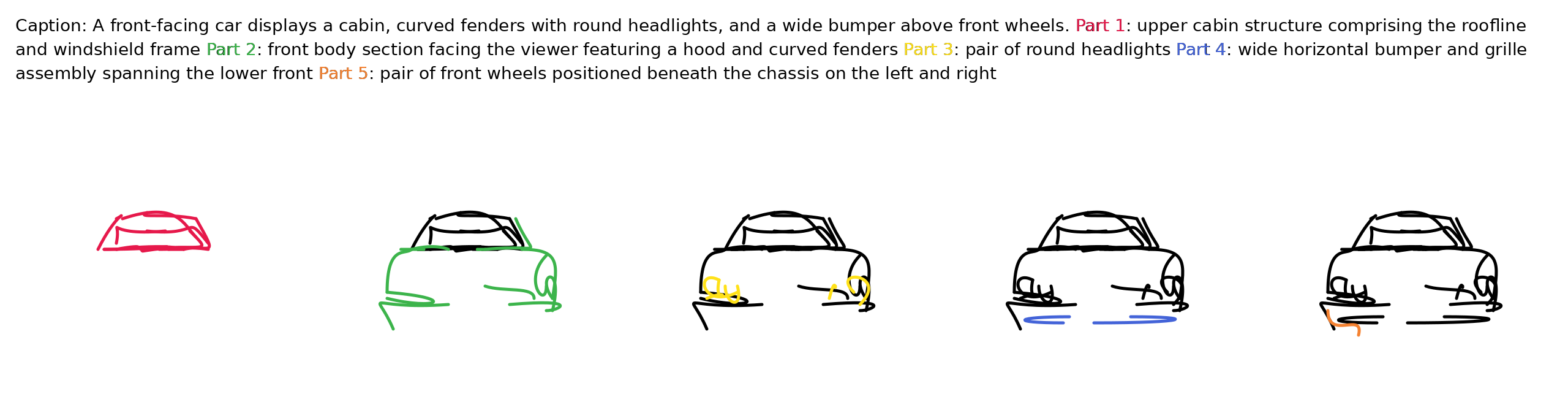}
    \caption{}
    \label{fig:supp-failure-a}
\end{subfigure}

\begin{subfigure}{\textwidth}
    \centering
    \includegraphics[width=0.8\textwidth]{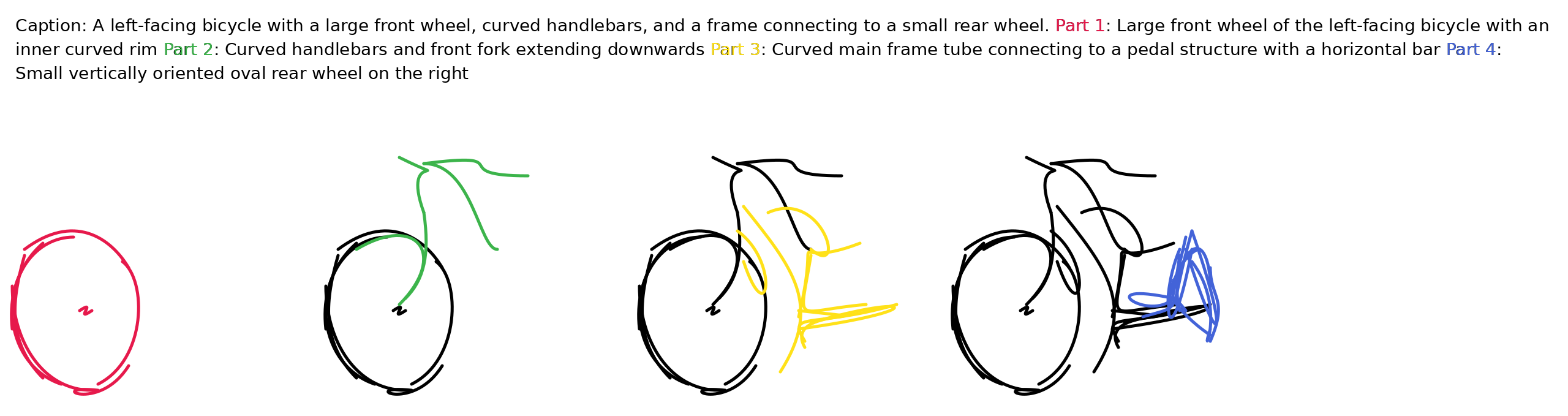}
    \caption{}
    \label{fig:supp-failure-b}
\end{subfigure}

\begin{subfigure}{\textwidth}
    \centering
    \includegraphics[width=0.8\textwidth]{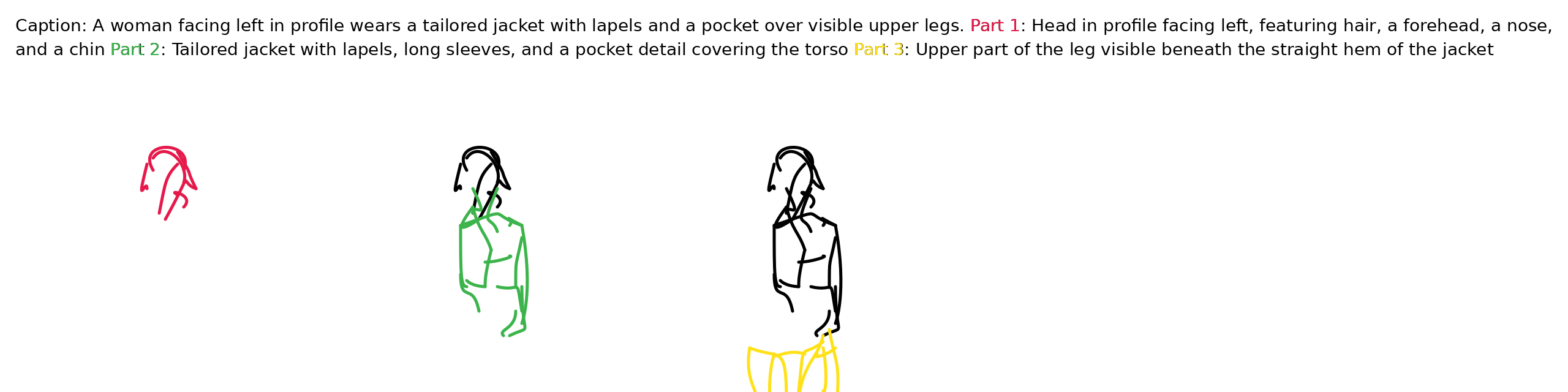}
    \caption{}
    \label{fig:supp-failure-c}
\end{subfigure}

\caption{Failure cases.}
\end{figure}

Sketches in the ControlSketch dataset all contain a fixed number of paths. As a result, the path count reward incentivizes the agent to match the ``ground-truth'' path count, which may lead to premature stopping once this count is reached, even if the corresponding part is not fully drawn. This behavior can be found in the omitted right wheel in~\cref{fig:supp-failure-a}. A second failure mode is erroneous topologies for unfamiliar semantic concepts. For example, the agent fails to correctly depict the ``vertically oriented oval rear wheel'' in~\cref{fig:supp-failure-b}, a structure that is relatively rare in the dataset. In addition, while RL training substantially mitigates part-misalignment errors, occasional misplacements remain. In~\cref{fig:supp-failure-c}, for example, the jacket is positioned too far to the right, creating an unnatural gap between the jacket and the upper legs. 

\subsection{Limitations}
\begin{figure}
    \centering
    \begin{subfigure}[b]{0.1666\textwidth}
        \centering
        \includegraphics[width=\linewidth]{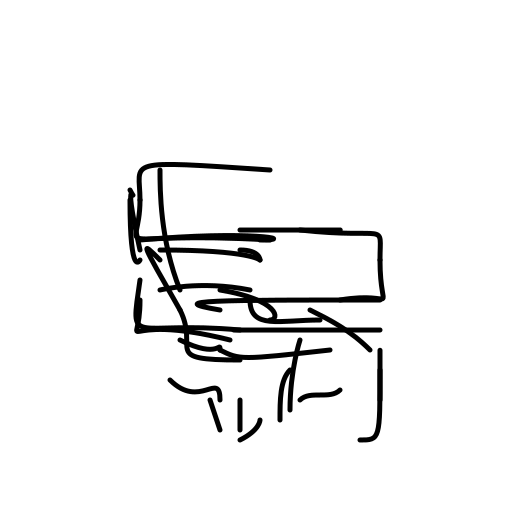}
        \caption{bed}
        \label{fig:limitation_bed}
    \end{subfigure}%
    \begin{subfigure}[b]{0.1666\textwidth}
        \centering
        \includegraphics[width=\linewidth]{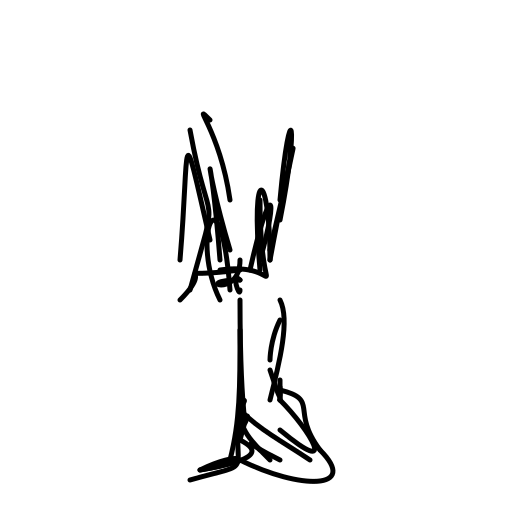}
        \caption{fork}
        \label{fig:limitation_fork}
        
    \end{subfigure}%
    \begin{subfigure}[b]{0.1666\textwidth}
        \centering
        \includegraphics[width=\linewidth]{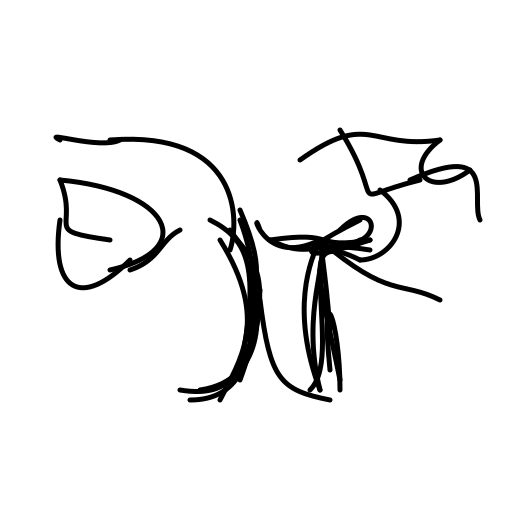}
        \caption{whale}
        \label{fig:limitation_whale}
        
    \end{subfigure}%
    \begin{subfigure}[b]{0.1666\textwidth}
        \centering
        \includegraphics[width=\linewidth]{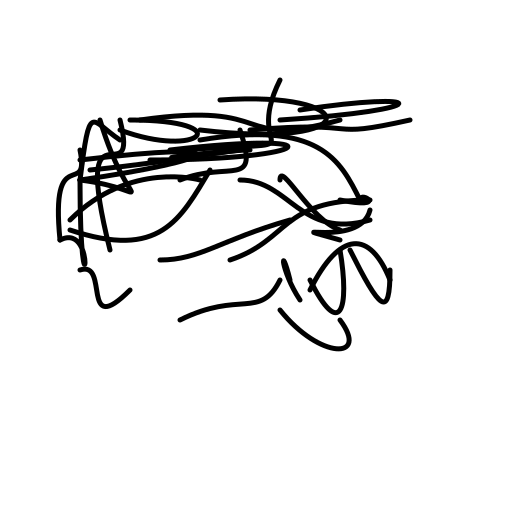}
        \caption{helicopter}
        \label{fig:limitation_heli}
        
    \end{subfigure}%
    \begin{subfigure}[b]{0.1666\textwidth}
        \centering
        \includegraphics[width=\linewidth]{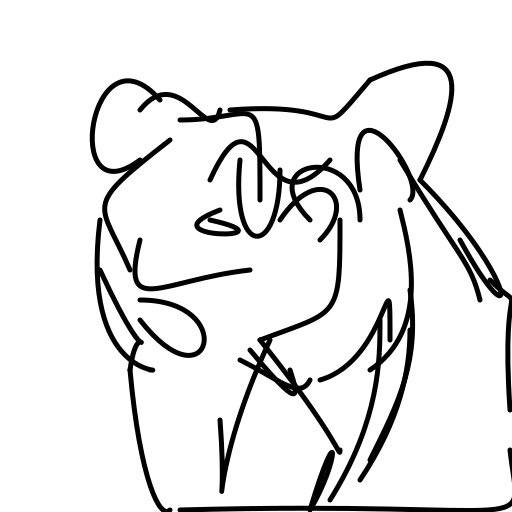}
        \caption{tiger}
        \label{fig:limitation_tiger}
        
    \end{subfigure}%
    \begin{subfigure}[b]{0.1666\textwidth}
        \centering
        \includegraphics[width=\linewidth]{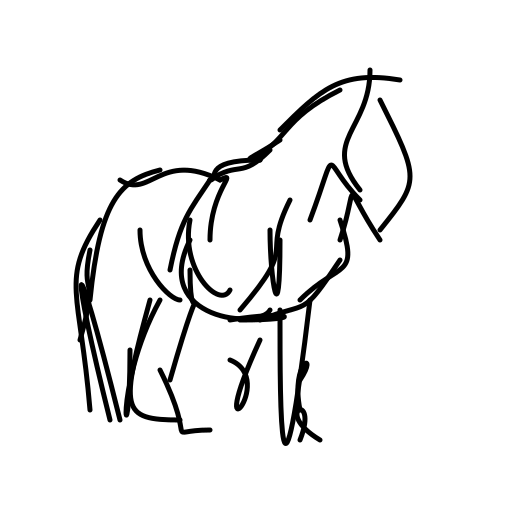}
        \caption{camel}
        \label{fig:limitation_camel}
        
    \end{subfigure}%
    \\
    \begin{subfigure}[b]{0.1666\textwidth}
        \centering
        \includegraphics[width=\linewidth]{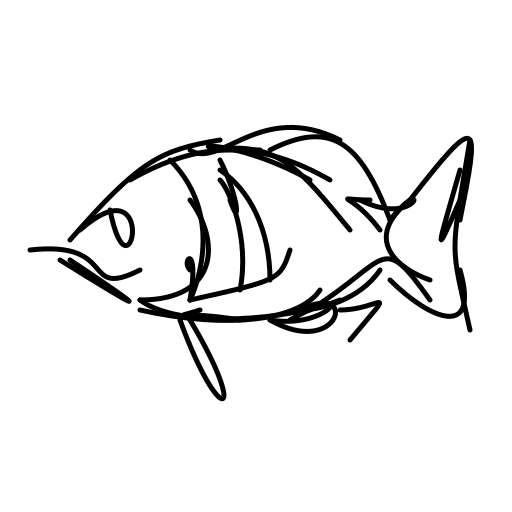}
        \caption{shark}
        \label{fig:limitation_shark}
        
    \end{subfigure}%
    \begin{subfigure}[b]{0.1666\textwidth}
        \centering
        \includegraphics[width=\linewidth]{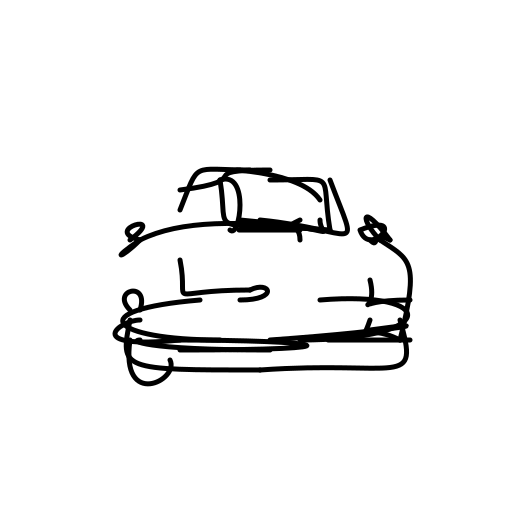}
        \caption{bus}
        \label{fig:limitation_bus}
        
    \end{subfigure}%
    \begin{subfigure}[b]{0.1666\textwidth}
        \centering
        \includegraphics[width=\linewidth]{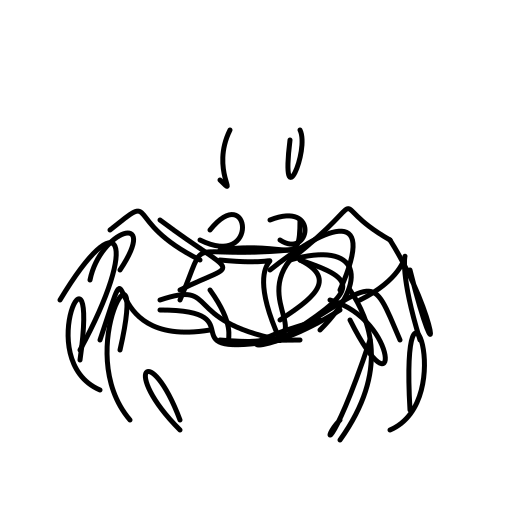}
        \caption{spider}
        \label{fig:limitation_spider}
        
    \end{subfigure}%
    \begin{subfigure}[b]{0.1666\textwidth}
        \centering
        \includegraphics[width=\linewidth]{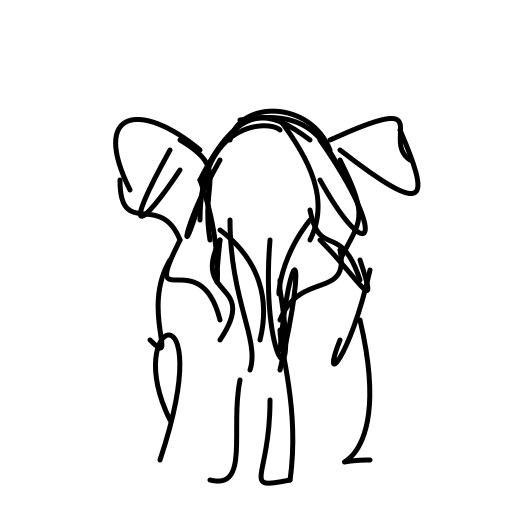}
        \caption{elephant}
        \label{fig:limitation_elephant}
    \end{subfigure}%
    \begin{subfigure}[b]{0.1666\textwidth}
        \centering
        \includegraphics[width=\linewidth]{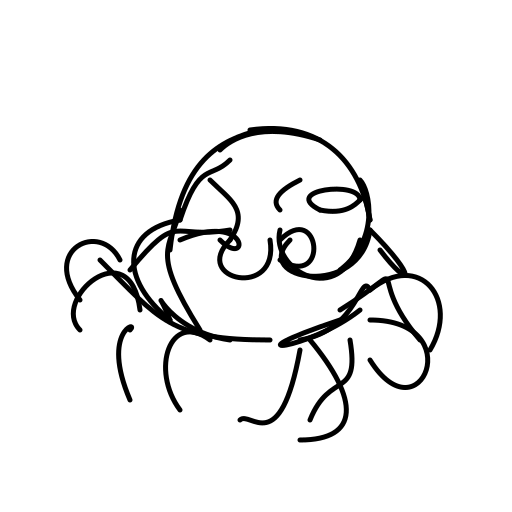}
        \caption{octopus}
        \label{fig:limitation_octopus}
    \end{subfigure}%
    \begin{subfigure}[b]{0.1666\textwidth}
        \centering
        \includegraphics[width=\linewidth]{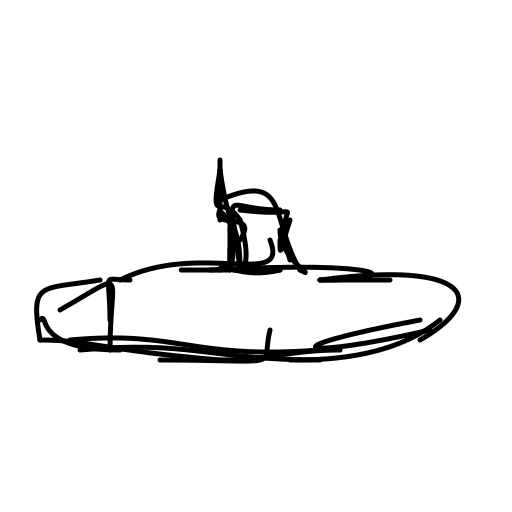}
        \caption{submarine}
        \label{fig:limitation_submarine}
    \end{subfigure}%
    \caption{Sample outputs on test data. The actual categories are below the sketches. These categories are unseen by the agent during training.}
    \label{fig:ood}
\end{figure}

The primary bottleneck toward a more general sketching agent is data. Our work is limited in its coverage of object categories beyond those present in the ControlSketch dataset. \cref{fig:ood} shows several sample outputs on object categories absent from the training data, i.e., out-of-distribution (OOD) inputs to the sketching agent.

We observe two distinct failure modes. First, for most unseen categories, the agent fails to produce meaningful sketches altogether. See~\cref{fig:limitation_bed,fig:limitation_fork,fig:limitation_whale,fig:limitation_heli} for representative examples. Second, when the input caption and part descriptions contain semantic concepts closely related to those seen during training, the agent tends to output sketches of ``nearest-neighbor'' categories rather than the intended objects. For instance, the agent generates sketches~\cref{fig:limitation_tiger,fig:limitation_camel,fig:limitation_shark,fig:limitation_bus,fig:limitation_spider} resembling the seen classes \textit{bear}, \textit{horse}, \textit{fish}, \textit{car}, and \textit{crab}, rather than their intended classes \textit{tiger}, \textit{camel}, \textit{shark}, \textit{bus}, and \textit{spider}, respectively.

Despite these limitations, we find that when OOD inputs consist of relatively general part descriptions, the agent occasionally produces sketches that reasonably approximate the intended object. See~\cref{fig:limitation_elephant,fig:limitation_octopus,fig:limitation_submarine} for such outputs which are from randomly selected OOD examples in the test partition. We believe that training on a significantly larger and more diverse dataset would substantially improve generalization, analogous to the ``emergent capabilities'' achieved by large-scale text-to-pixel image diffusion models.

\begin{figure}
    \centering
    \begin{subfigure}[t]{0.32\textwidth}
        \centering
        \includegraphics[width=0.49\linewidth]{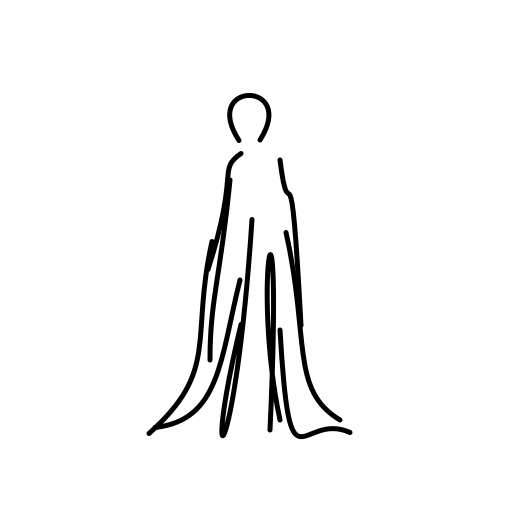}%
        \hfill
        \includegraphics[width=0.49\linewidth]{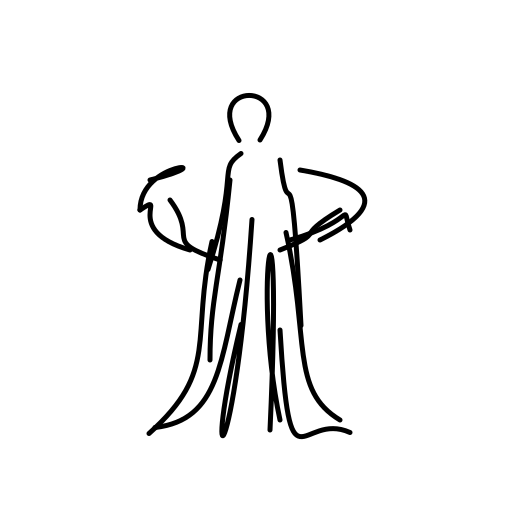}
        \caption{``two raised, segmented front claws with curved pincers''}
        \label{fig:pairA}
    \end{subfigure}%
    \hfill\vrule\hfill
    \begin{subfigure}[t]{0.32\textwidth}
        \centering
        \includegraphics[width=0.49\linewidth]{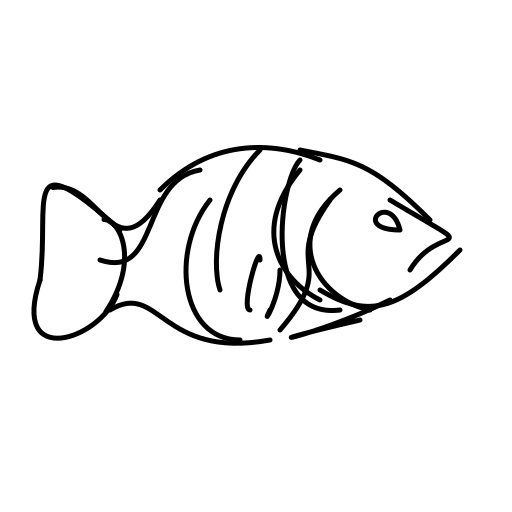}%
        \hfill
        \includegraphics[width=0.49\linewidth]{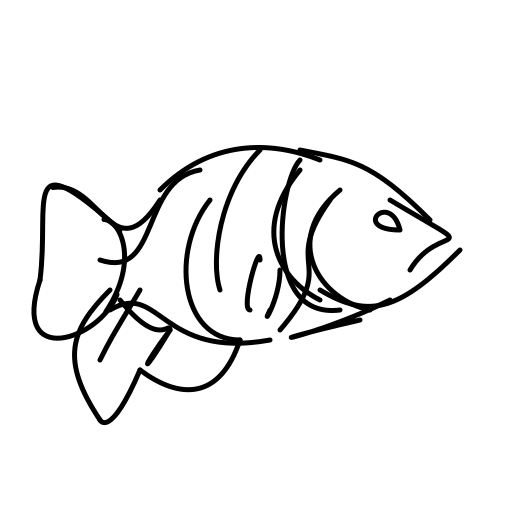}
        \caption{``walking legs''}
        \label{fig:pairB}
    \end{subfigure}%
    \hfill\vrule\hfill
    \begin{subfigure}[t]{0.32\textwidth}
        \centering
        \includegraphics[width=0.49\linewidth]{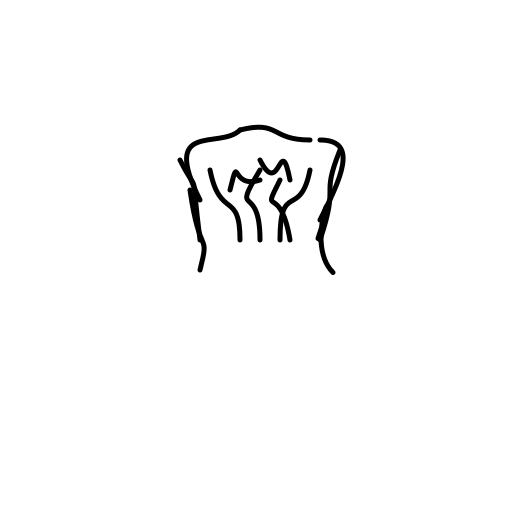}%
        \hfill
        \includegraphics[width=0.49\linewidth]{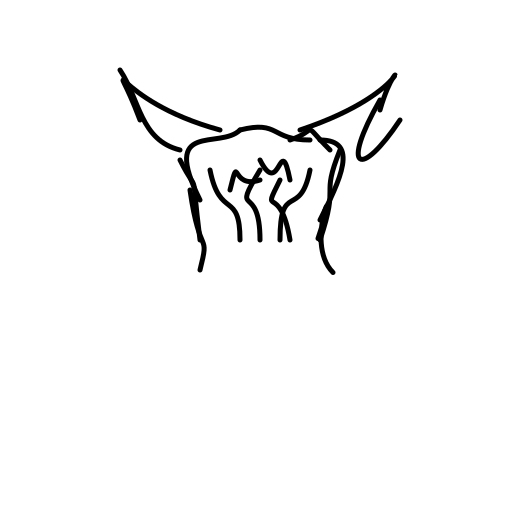}
        \caption{``a pair of feathered wings extending upwards from backrest''}
        \label{fig:pairC}
    \end{subfigure}%
    \caption{Adding out-of-domain parts. Each pair shows the original incomplete sketch (left) and the result after the agent attempts to add an out-of-domain part (right). The text below each pair is the part description provided as input to the agent.}
    \label{fig:ood_parts}
\end{figure}

Beyond object-level generalization, the agent also struggles to generalize to out-of-domain \textit{parts}. \cref{fig:ood_parts} shows several examples in which the agent is asked to attach parts from unrelated categories to familiar objects, e.g. segmented pincers on an angel, walking legs on a fish, or feathered wings on a chair. In each pair, the left image shows the original incomplete sketch and the right image shows the result after the agent attempts to add the specified part. Most of the time the agent fails to produce accurate depictions of these OOD parts. 

Finally, a higher-level limitation stems from the inherent difficulty of large language models in predicting precise numerical coordinates. This imprecision, compounded by the stochasticity introduced by temperature-based sampling, can occasionally produce sketches with drastically degraded stroke placement. \cref{fig:worst} presents the lowest-quality validation outputs we have observed across all in-distribution categories. In practice, outputs of this quality are very infrequent and represent only edge cases.

\begin{figure}[t]
    \centering
    \includegraphics[width=\linewidth]{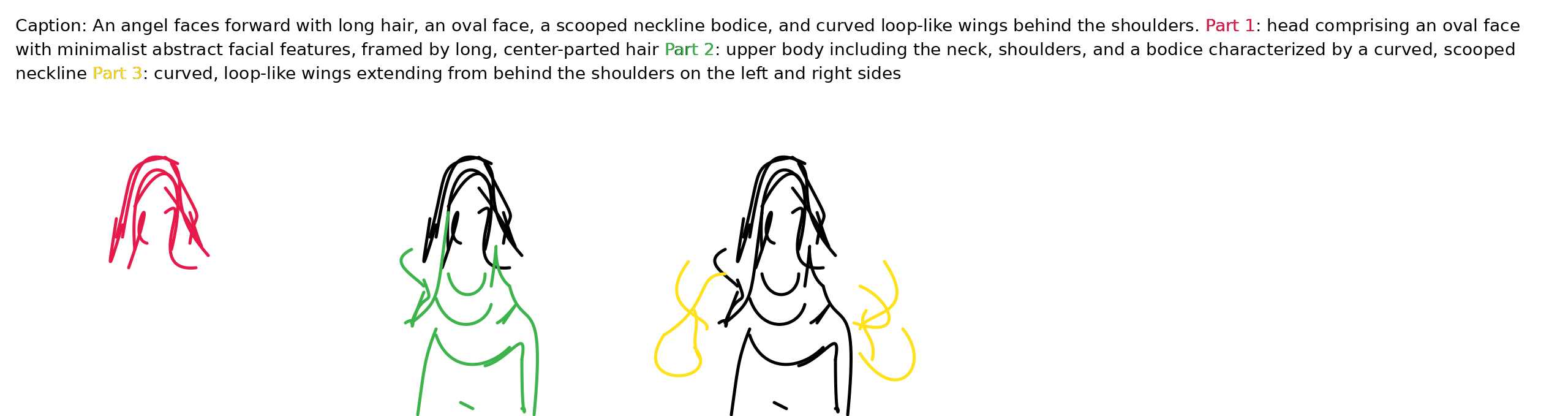}\\
    \includegraphics[width=\linewidth]{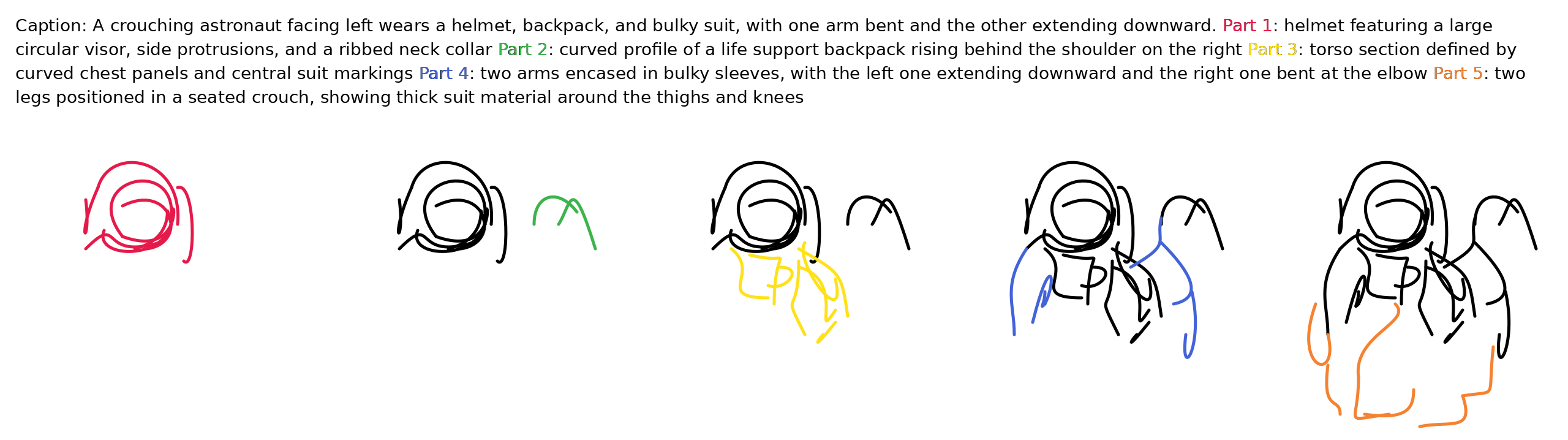}\\
    \includegraphics[width=\linewidth]{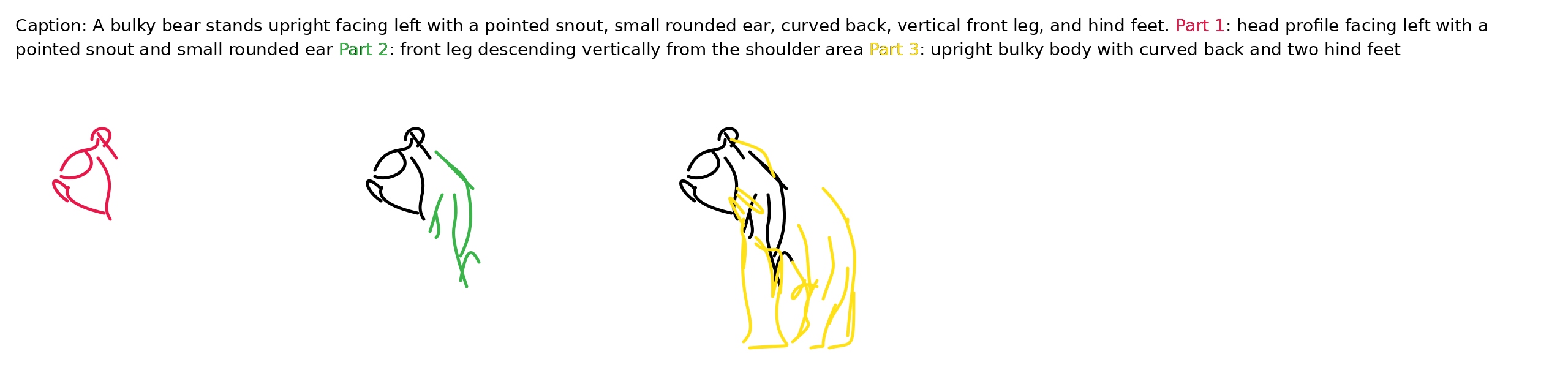}\\
    \includegraphics[width=\linewidth]{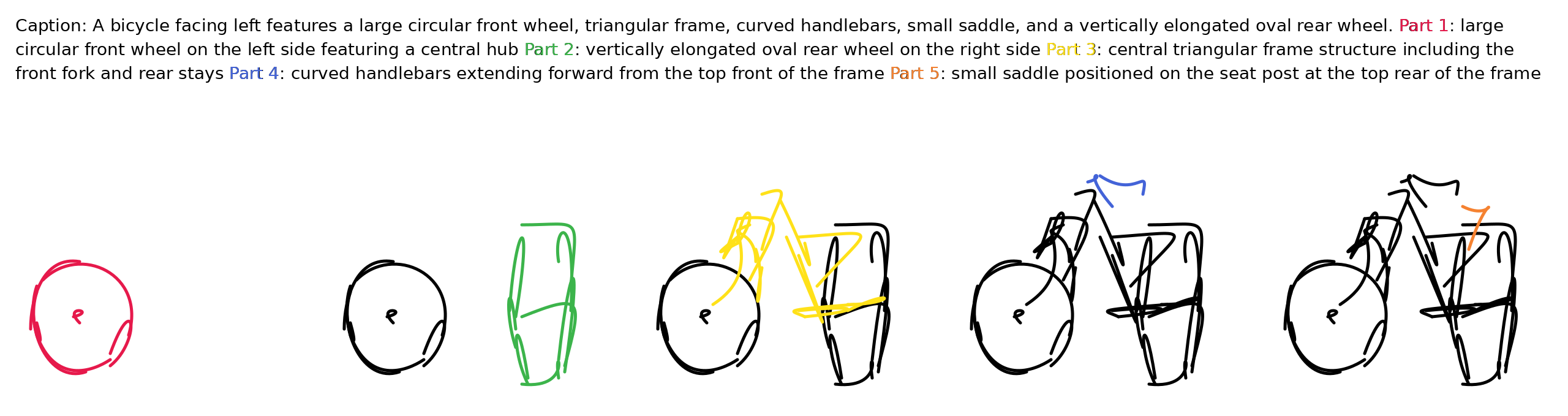}\\
    \includegraphics[width=\linewidth]{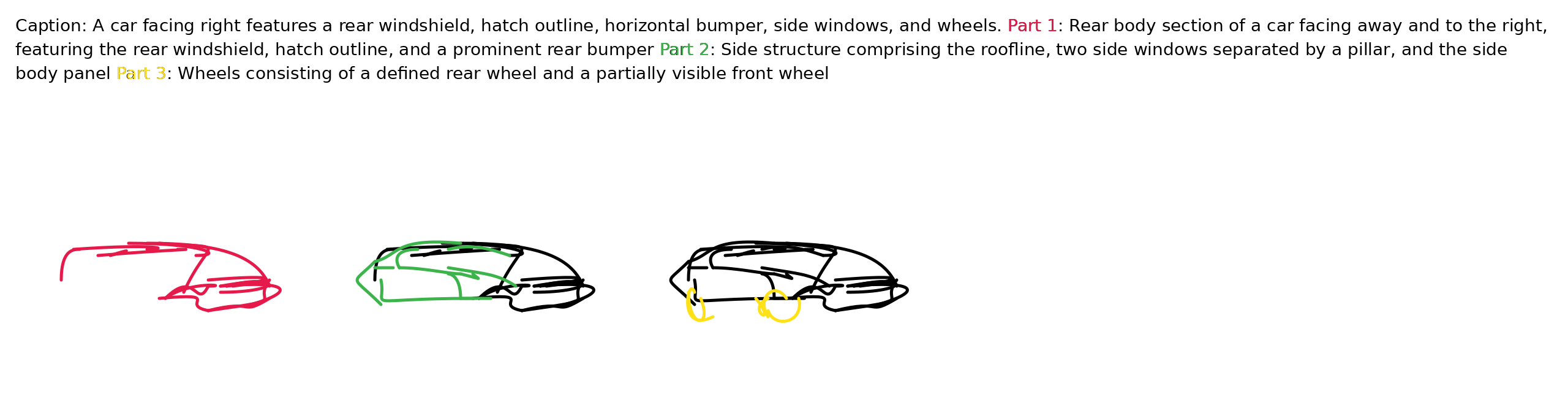}
    \caption{Worst-quality outputs across all in-distribution categories.}
    \label{fig:worst}
\end{figure}

\begin{figure}[t]
    \ContinuedFloat
    \centering
    \includegraphics[width=\linewidth]{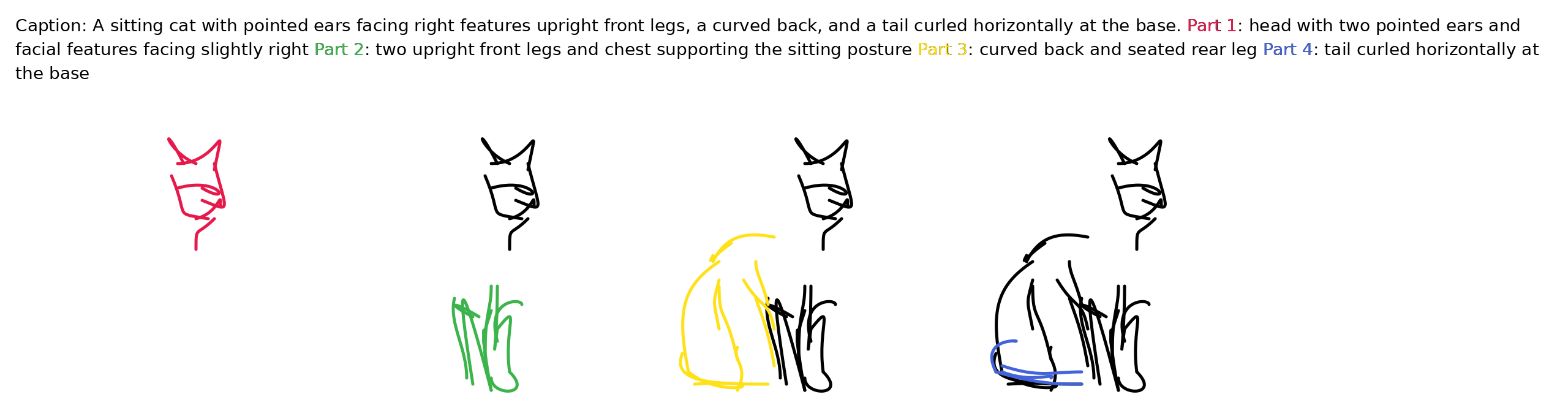}\\
    \includegraphics[width=\linewidth]{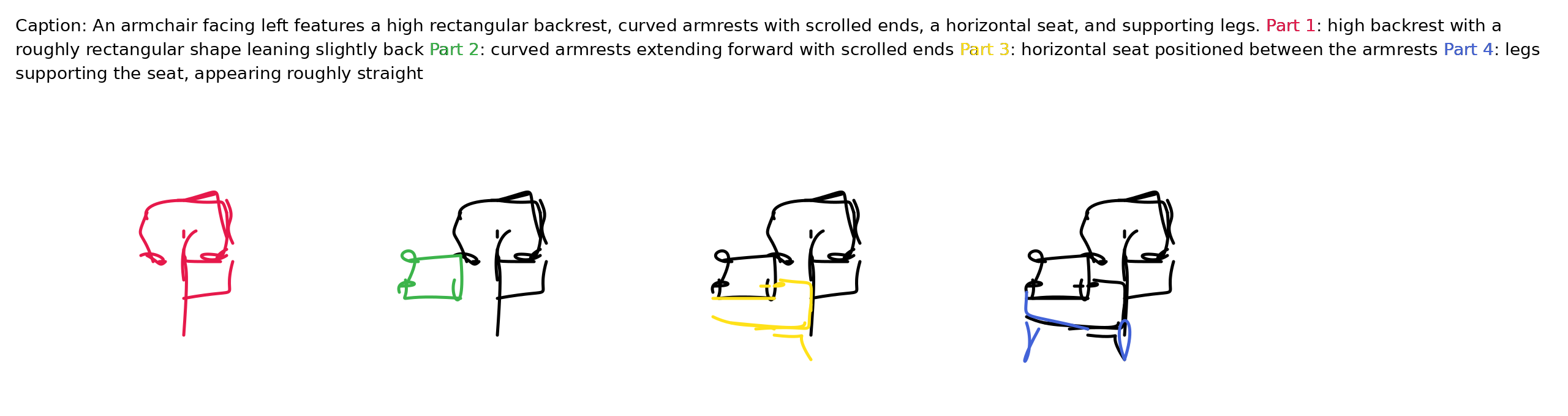}\\
    \includegraphics[width=\linewidth]{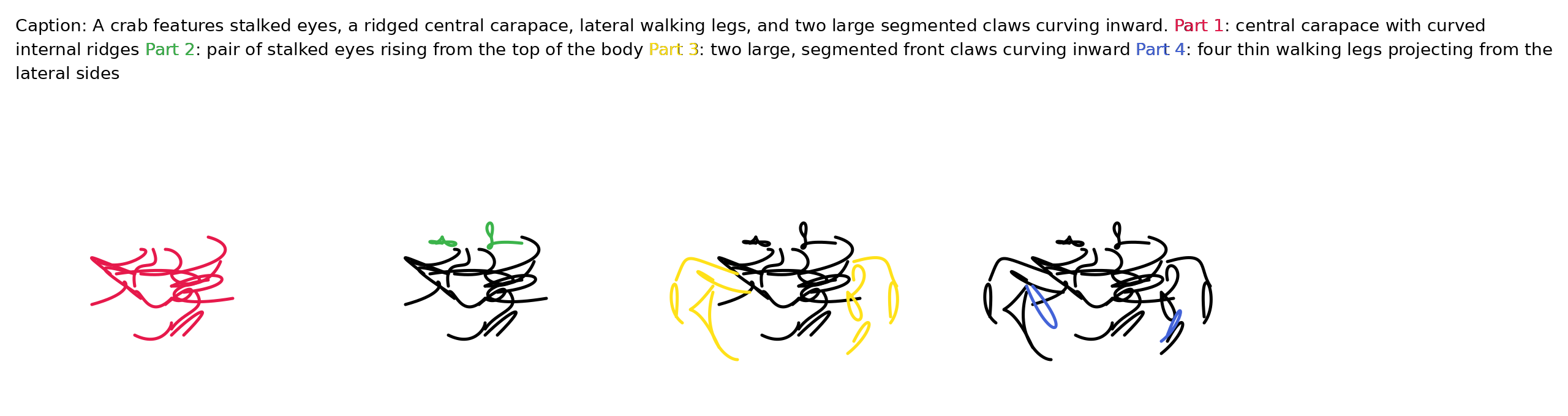}\\
    \includegraphics[width=\linewidth]{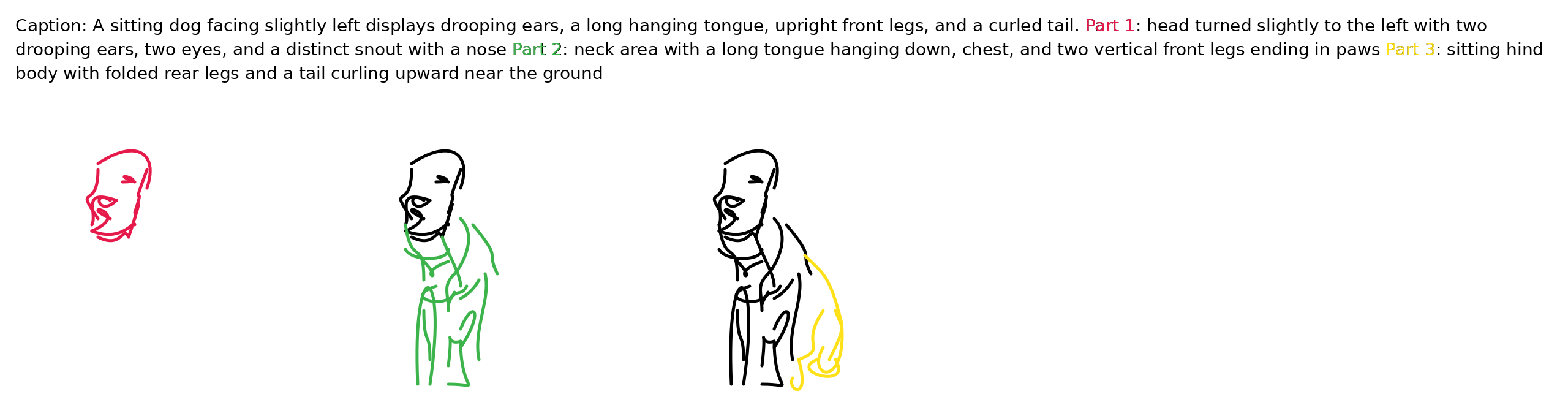}\\
    \includegraphics[width=\linewidth]{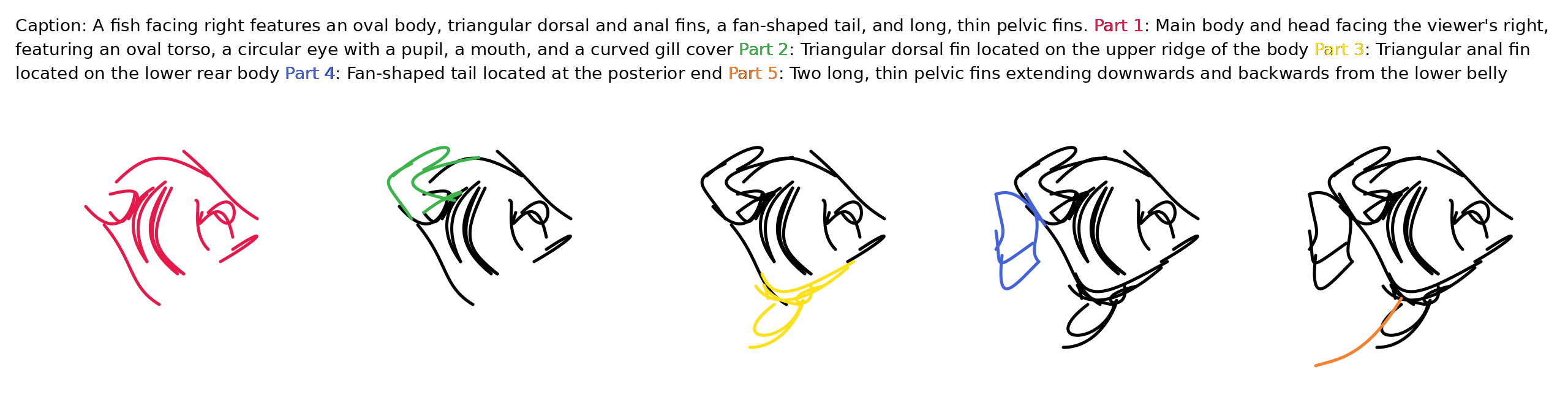}
    \caption[]{Worst-quality outputs (continued).}
\end{figure}

\begin{figure}[t]
    \ContinuedFloat
    \centering
    \includegraphics[width=\linewidth]{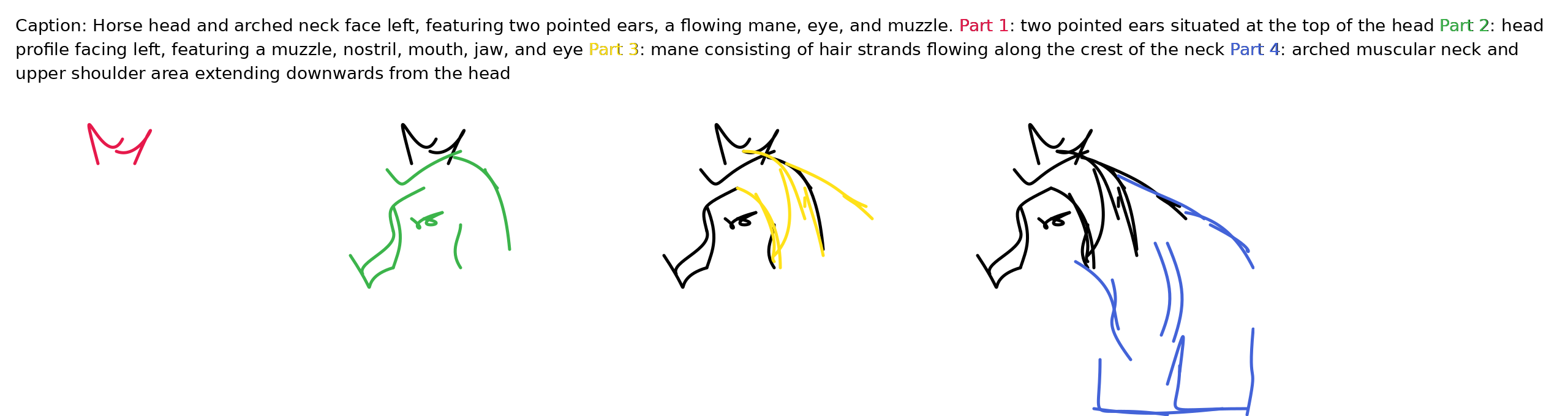}\\
    \includegraphics[width=\linewidth]{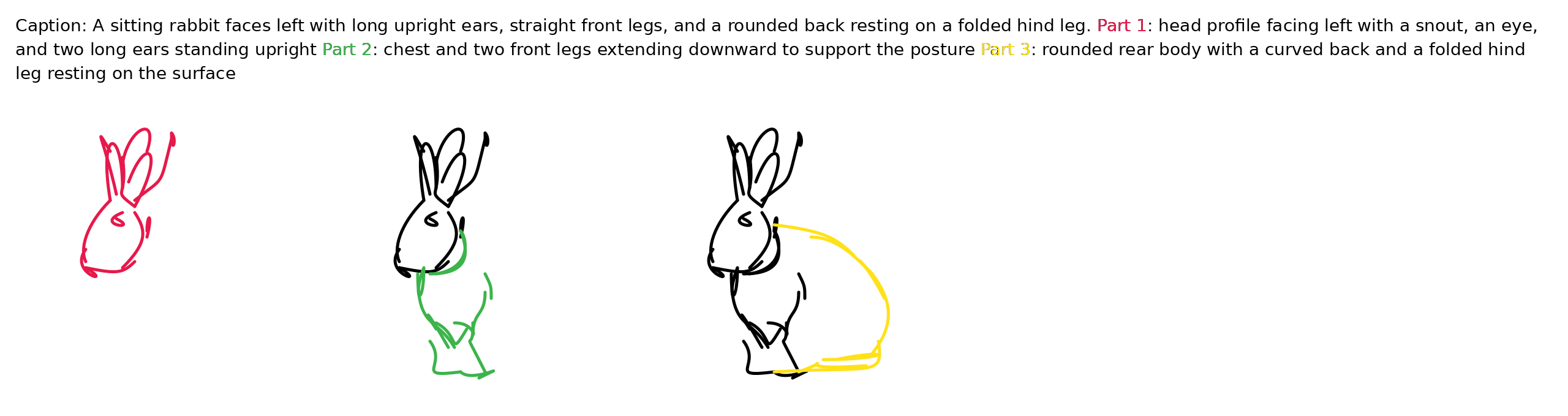}\\
    \includegraphics[width=\linewidth]{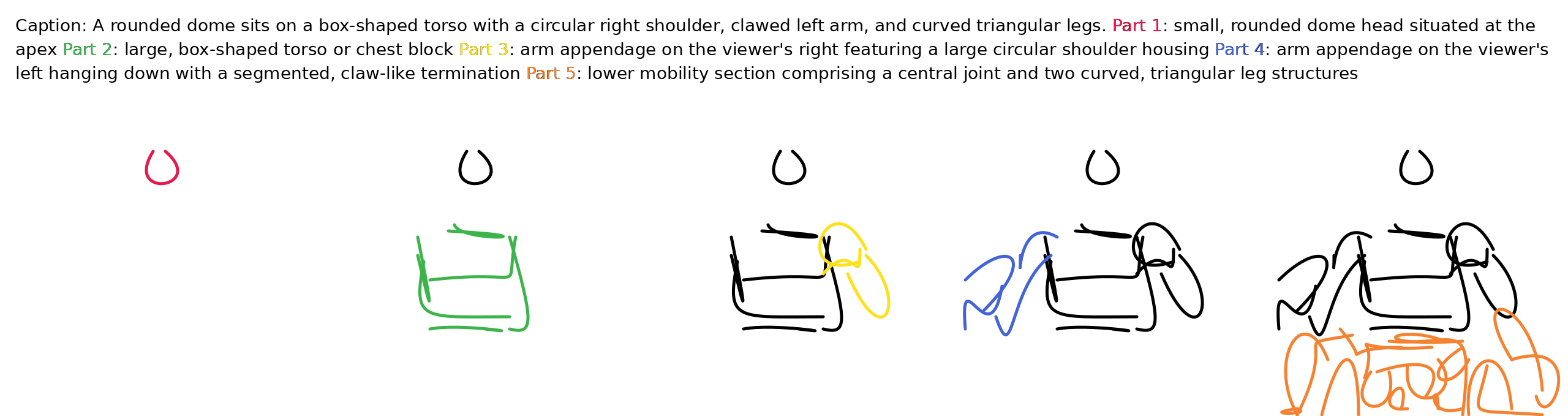}\\
    \includegraphics[width=\linewidth]{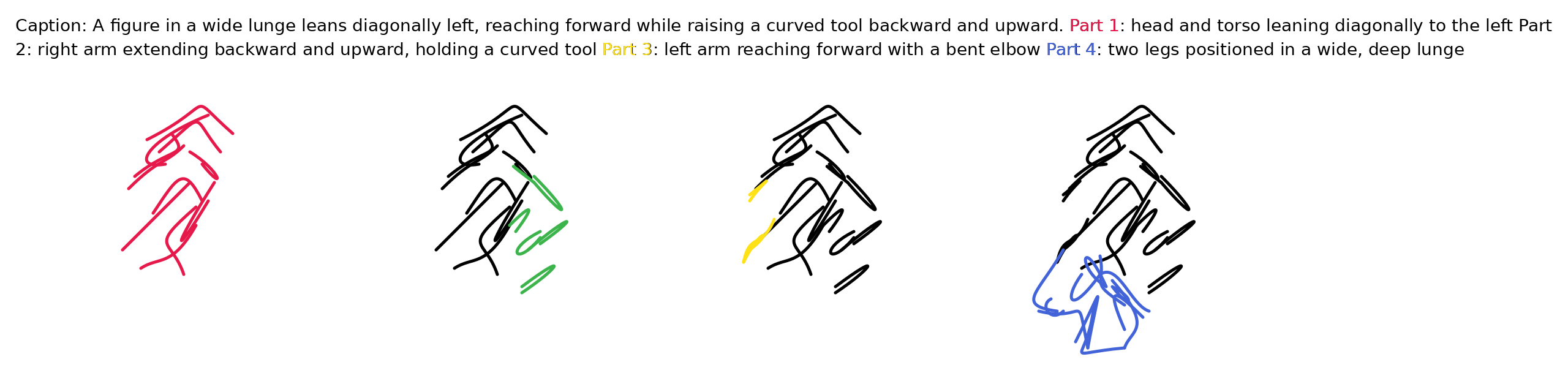}\\
    \includegraphics[width=\linewidth]{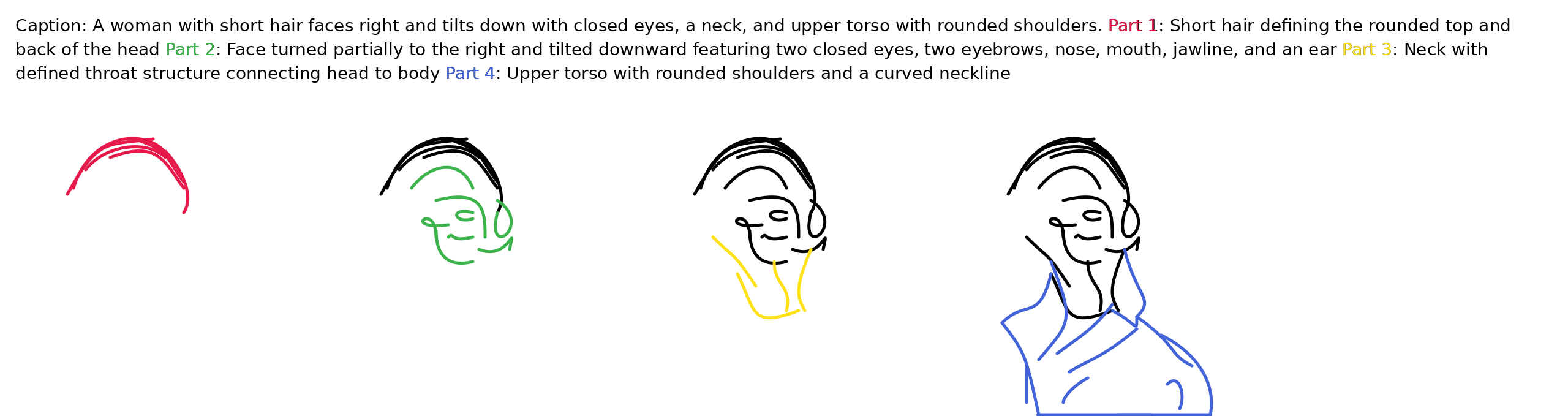}
    \caption[]{Worst-quality outputs (continued).}
\end{figure}

\subsection{Future Work}
The current pipeline is designed for generating one part at a time. In future work, a planning agent could coordinate multiple agents to generate different parts in parallel. Furthermore, as mentioned before, enabling the system to refine unsatisfactory intermediate outputs may further improve overall sketch quality. 

Another promising direction is to incorporate richer natural language reasoning into the generation process, for example by introducing chain-of-thought reasoning before generating each part.

The sketching capabilities of our agent could also be leveraged to support visual reasoning tasks. One possible direction is to extend the agent’s abilities to generate auxiliary figures for tasks such as geometry problems.

In addition, equipping the agent with a self-evaluation mechanism, where the agent assesses the quality of its own intermediate outputs and selectively regenerates unsatisfactory parts, could mitigate error accumulation and improve robustness over longer generation sequences.

\end{document}